%% file: main.tex
\documentclass[10pt,twocolumn,letterpaper]{article}

\usepackage[pagenumbers]{wacv}            %

\input{preamble}

\definecolor{wacvblue}{rgb}{0.21,0.49,0.74}
\usepackage[pagebackref,breaklinks,colorlinks,allcolors=wacvblue]{hyperref}

\def\wacvPaperID{107} %
\def\confName{WACV}
\def\confYear{2026}

\title{Locally Explaining Prediction Behavior via Gradual Interventions and Measuring Property Gradients}

\author{Niklas Penzel$^1$ \quad
Joachim Denzler$^1$ \\
$^1$Computer Vision Group, Friedrich Schiller University Jena, Germany\\
{\tt\small \{niklas.penzel, joachim.denzler\}@uni-jena.de}}

\begin{document}
\maketitle

\input{sec/0_abstract}

\input{sec/1_intro}

\input{sec/2_related_work}

\input{sec/3_method}
\input{sec/4_experiments}

\input{sec/5_limitations}
\input{sec/6_conclusions}

{
    \small
    \bibliographystyle{ieeenat_fullname}
    \bibliography{main}
}

\input{sec/X_suppl}

\FloatBarrier

\end{document}

%% file: preamble.tex
\usepackage[dvipsnames]{xcolor}

\usepackage{tikz}
\usetikzlibrary{automata, positioning, patterns, patterns.meta, calc}
\usepackage{bm}

\newcommand{\class}{\mathbb{F}}
\newcommand{\data}{\mathcal{D}_{\text{train}}}
\newcommand{\cweights}{\theta}
\newcommand{\siglevel}{\delta} 

\newcommand{\foi}{\mathsf{X}} 
\newcommand{\rfoi}{\mathsf{x}}
\newcommand{\fset}{\mathfrak{X}} 

\newcommand{\statistic}{\mathcal{S}}

\newcommand{\propgrad}{\xspace$\mathbb{E}[|\nabla_\foi|]$\xspace}

\usepackage{algorithm}
\usepackage{algpseudocode}
\usepackage{nicefrac}
\usepackage{todonotes}
\usepackage{aligned-overset}
\usepackage{svg}
\usepackage{multirow}
\usepackage{pifont}
\newcommand{\cmark}{\ding{51}}%
\newcommand{\xmark}{\ding{55}}%
\usepackage{titletoc}
\usepackage{placeins}
\usepackage{float}

%% file: sec/0_abstract.tex
\begin{abstract}
Deep learning models achieve high predictive performance but lack intrinsic interpretability, hindering our understanding of the learned prediction behavior.
Existing local explainability methods focus on associations, neglecting the causal drivers of model predictions. 
Other approaches adopt a causal perspective but primarily provide global, model-level explanations.
However, for specific inputs, it's unclear whether globally identified factors apply locally.
To address this limitation, we introduce a novel framework for local interventional explanations by leveraging recent advances in image-to-image editing models. 
Our approach performs gradual interventions on semantic properties to quantify the corresponding impact on a model's predictions using a novel score, the expected property gradient magnitude. 
We demonstrate the effectiveness of our approach through an extensive empirical evaluation on a wide range of architectures and tasks.
First, we validate it in a synthetic scenario and demonstrate its ability to locally identify biases.
Afterward, we apply our approach to investigate medical skin lesion classifiers, analyze network training dynamics, and study a pre-trained CLIP model with real-life interventional data.
Our results highlight the potential of interventional explanations on the property level to reveal new insights into the behavior of deep models.\footnote{Project page: \url{https://propgrad.github.io/}}
\end{abstract}

%% file: sec/1_intro.tex
\section{Introduction}
\label{sec:intro}

\begin{figure*}[t]
\centering
\begin{subfigure}{0.49\textwidth}
    \includegraphics[width=\linewidth, trim=1cm 0.5cm 0cm 0.6cm, clip]{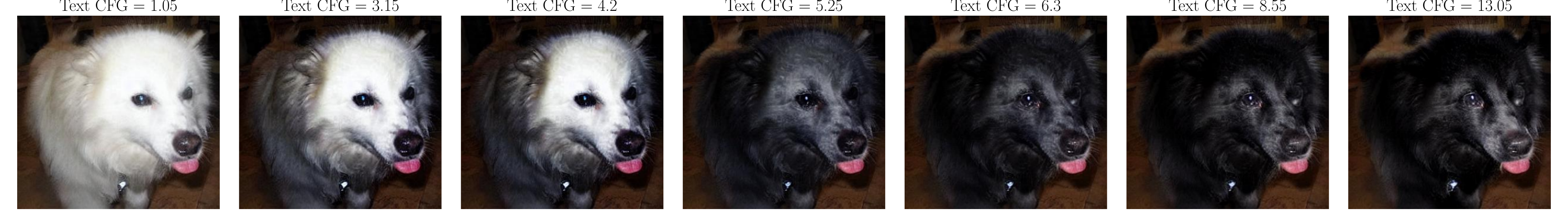}
    \caption{Fur color intervention on a light-furred dog. }
    \label{fig:fur-intervent}
\end{subfigure}
\begin{subfigure}{0.49\textwidth}
    \includegraphics[width=\linewidth, trim=0cm 0.5cm 1cm 0.6cm, clip]{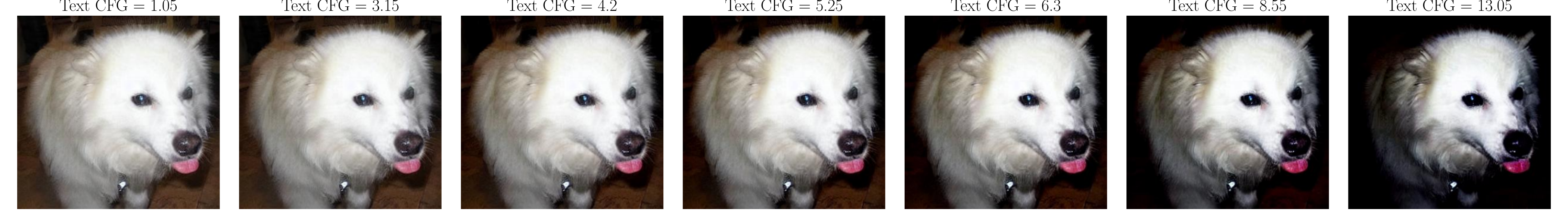}
    \caption{Background intervention for a light-furred dog. }
    \label{fig:bg-intervent}
\end{subfigure}
\begin{subfigure}{0.49\textwidth}
    \includegraphics[width=\linewidth, trim=0.8cm 0.8cm 0.5cm 1cm, clip]{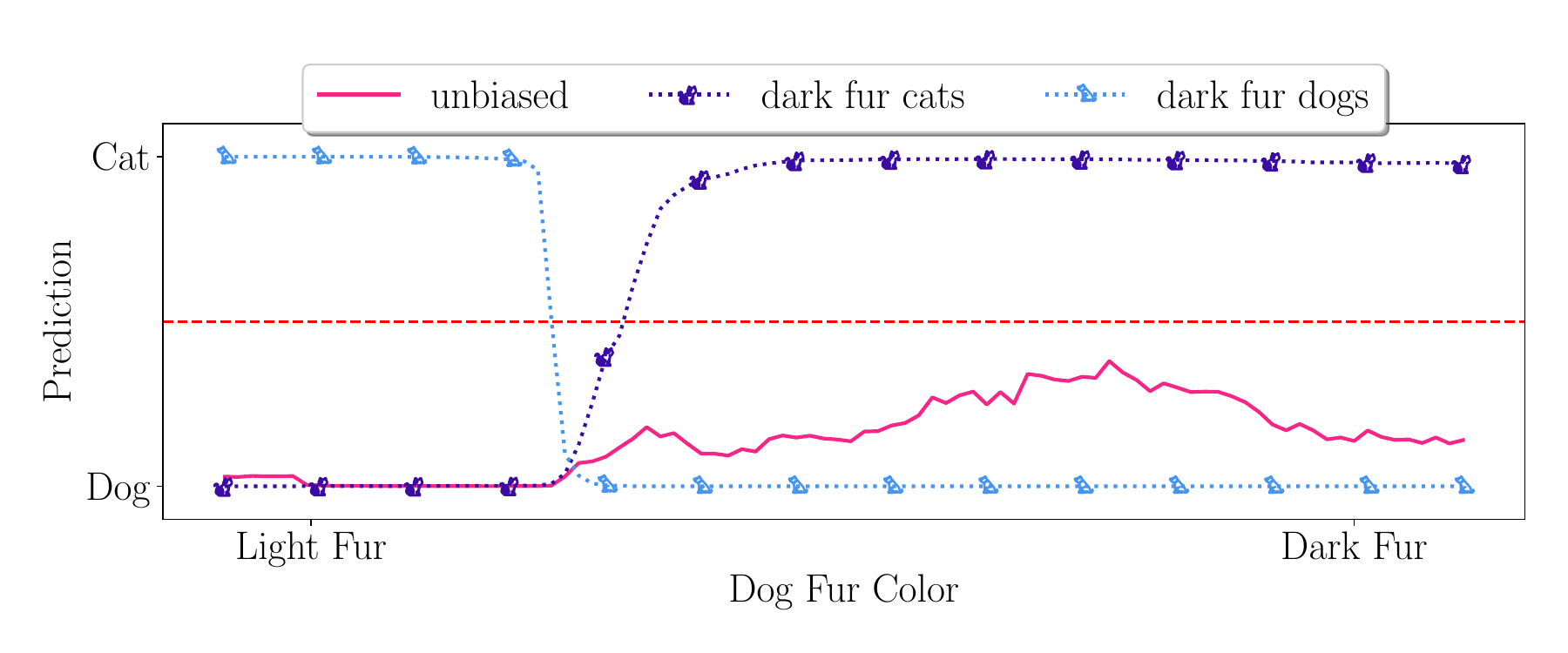}
    \caption{Changes in model behavior for fur color intervention. }
    \label{fig:fur-intervent-results}
\end{subfigure}
\begin{subfigure}{0.49\textwidth}
    \includegraphics[width=\linewidth, trim=0.8cm 0.8cm 0.5cm 1cm, clip]{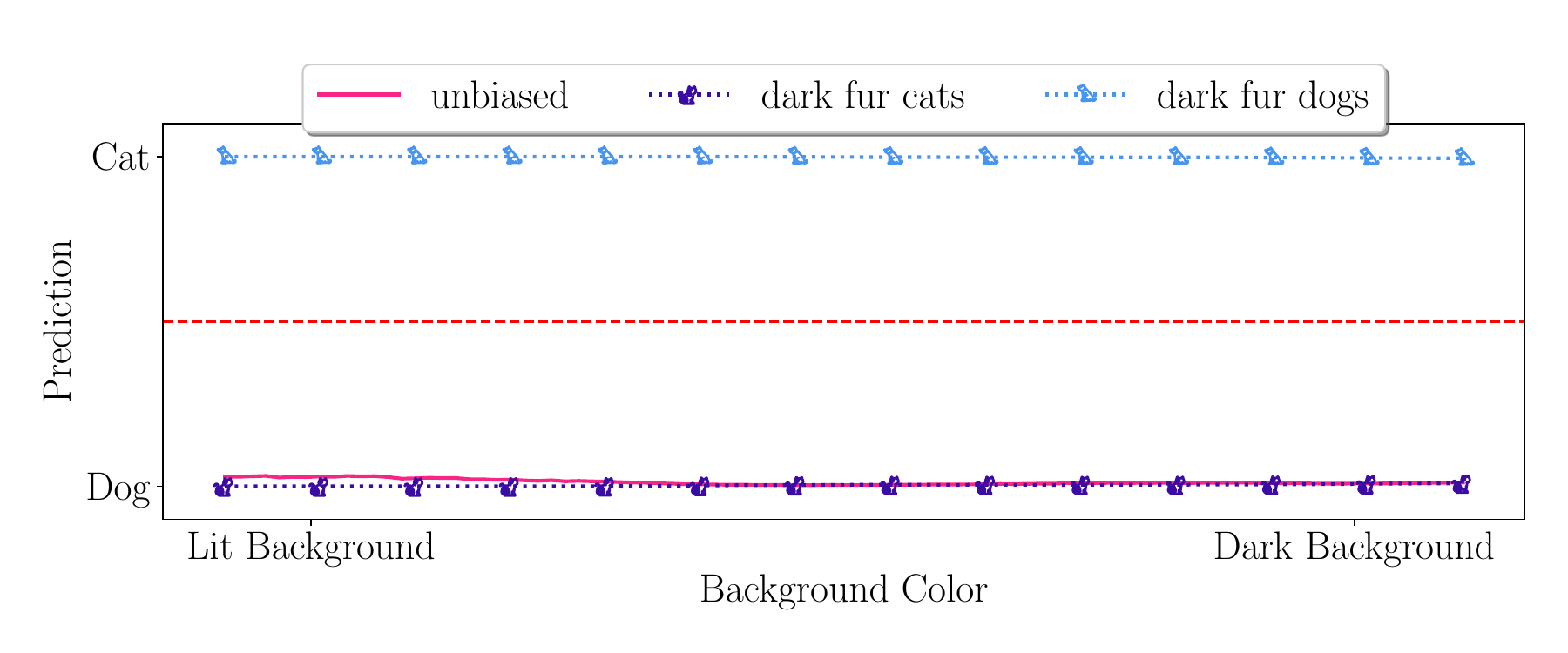}
    \caption{Changes in model behavior for background intervention. }
    \label{fig:bg-intervent-results}
\end{subfigure}

\caption{
\cref{fig:fur-intervent} and \cref{fig:bg-intervent} demonstrate our approach to study model predictions through gradual interventions here for two properties: fur color and background illumination in a cat vs. dog classification task. 
Using \cite{fu2024mgie}, we synthetically adjust these properties, i.e., darkening the fur or the background, while tracking prediction changes.
Specifically, \cref{fig:fur-intervent-results} and \cref{fig:bg-intervent-results} show the resulting shifts in model outputs for three networks: an unbiased one, one trained only on dark-furred cats and light-furred dogs, and one trained on the opposite bias.
The \textcolor{red}{red line} marks the decision threshold, where predictions flip between classes during the intervention.
}
\label{fig:1d-results}
\end{figure*}

Modern deep learning models are complex data-centric systems that achieve high predictive performance but lack intrinsic interpretability.
Hence, many explainability (XAI) methods were proposed to interpret trained model behavior, especially for vision models. 
Post-hoc XAI includes local methods, often generating pixel-wise attributions, e.g., \cite{selvaraju2020grad,lundberg2017unified,ribeiro2016should,sundararajan2017axiomatic,shrikumar2017learning} and global methods focused on human interpretable concepts or properties \cite{kim2018interpretability,yeh2020completeness,reimers2020determining,cunninghamSparseAutoencodersFind2023,scherlisPolysemanticityCapacityNeural2023,conmyAutomatedCircuitDiscovery2023,gaoScalingEvaluatingSparse2024,marksSparseFeatureCircuits2024a,kazimi2025explaining}.
Unfortunately, global insights can be deceiving for individual inputs.
On the local level, properties can be occluded or overshadowed by independent visual elements.

To address this limitation, we introduce a novel approach to locally explain neural network prediction behavior based on input interventions at the property level.
We propose to leverage recent breakthroughs in image-to-image editing models, e.g., \cite{brooks2022instructpix2pix,mokady2023null,fu2024mgie}. 
These models are trained conditionally using Classifier-Free Guidance (CFG) scaling \cite{ho2021classifier}.
Hence, we can gradually control the alignment with the corresponding interventional instruction during inference.
Our key insight is that this paradigm facilitates the smooth transition between two property states (see \cref{fig:1d-results}, top row), also for complex features.
Consequently, by utilizing CFG scaling, we intervene on semantic properties for an individual input and study the shift in prediction behavior of trained neural networks (see \cref{fig:1d-results}, bottom row).
To quantify the impact of a property on a model, we propose approximating the corresponding expected property gradient magnitude.
This score is naturally connected to the intuition of measuring the change along the property axis in \cref{fig:1d-results}.
Additionally, our expected property gradient magnitude can be seen as an extension of the causal concept effect \cite{goyal2019explaining} for gradual interventions.
Finally, to ensure robustness, we suggest a corresponding permutation significance test.

We empirically validate our framework for local interventional explanations on a wide variety of architectures and tasks.
First, we explore a biased scenario (cats vs. dogs \cite{dogs-vs-cats}) where we synthetically correlate a property (fur color) with the label to validate that our approach can locally identify the causal factor (see \cref{fig:1d-results}).
Additionally, our gradual interventions allow for a direct quantification of how the outputs change in response to shifts in the selected property, a key distinction from local methods highlighting image regions, e.g., \cite{selvaraju2020grad,lundberg2017unified,ribeiro2016should,sundararajan2017axiomatic,shrikumar2017learning,fong2017interpretable,goyal2019counterfactual,popescu2021counterfactual,khorram2022cycle}.
Further, our derived score quantitatively outperforms baselines for indicating locally biased behavior during interventions.
We corroborate these findings for real-world skin lesion classification, analyzing a known bias.
Afterward, we study the training dynamics of eight modern classifiers by tracking a property correlated with the label.
Here we find oscillations also in later epochs, depending on the weight initialization.
Finally, we demonstrate that our model-agnostic approach works with diverse sources of interventions by capturing real-life interventional data to analyze a CLIP \cite{radford2021learning} model.

Our key contributions can be summarized as follows: 
(1) We introduce a new framework for local network prediction explanations based on gradual interventions, e.g., by leveraging Classifier-Free Guidance (CFG) scaling \cite{ho2021classifier}.
(2) We derive a novel score to quantify the shift in model behavior with respect to a property by approximating the expected property gradient magnitude.
(3) We provide a corresponding hypothesis test to verify statistical significance.
(4) We conduct experiments on a wide range of architectures and tasks to demonstrate the effectiveness of our approach.

%% file: sec/2_related_work.tex
\section{Related Work}
\label{sec:related}

Many methods to derive local explanations aim to find important regions in the input, e.g., \cite{zeiler2014visualizing,selvaraju2020grad,lundberg2017unified,ribeiro2016should,sundararajan2017axiomatic,shrikumar2017learning}.
A subset of these methods, most closely related to our approach, employs input perturbations to estimate importance.
These perturbations or interventions are often done by replacing patches and, therefore, occluding parts of the input.
Such occlusion patches can be constructed using noise, e.g., \cite{zeiler2014visualizing,petsiuk2018rise}.
Other approaches use similar image regions \cite{zintgraf2017visualizing} or generative infilling \cite{chang2018explaining,korsch2023simplified,khorram2022cycle}.
Similarly related are methods that use causal terminology and generate visual counterfactual explanations \cite{stepin2021survey} by posing questions of the form ``Which parts of the input need to change to result in a specific prediction?''
Examples of this approach include \cite{fong2017interpretable,goyal2019counterfactual,popescu2021counterfactual,khorram2022cycle,augustin2022diffusion}.
Nevertheless, such visual explanations necessitate additional semantic interpretations by experts to identify specific properties responsible for the measured importance.
In contrast, our approach explains the prediction behavior directly on the level of semantic properties.

Related in that regard are, often global, explanation methods, e.g., \cite{lang2021explaining,hendricks2018generating,kim2018interpretability,yeh2020completeness,reimers2020determining,cunninghamSparseAutoencodersFind2023,scherlisPolysemanticityCapacityNeural2023,conmyAutomatedCircuitDiscovery2023,gaoScalingEvaluatingSparse2024,marksSparseFeatureCircuits2024a,schmalwasser2024exploiting,kazimi2025explaining}, which discover abstract concepts learned by a trained model.
However, these methods require direct access to the model parameters or probing datasets.
Further, while they are often explorative, they have difficulty determining whether a certain property is unused and can suffer from confounding, e.g., \cite{kim2018interpretability}, see \cite{goyal2019explaining}.
The approach described in \cite{reimers2020determining} can test for the usage of human-defined properties by trained models.
To do this, they assume usage and either confirm or reject the null hypothesis using conditional independence tests.
However, the results are binary and, unfortunately, do not allow actionable interpretations of the changes in prediction behavior on a local level.
While other works, e.g., \cite{penzel2023analyzing,buechner2024power}, use probing datasets to tackle these limitations, the explanations are strictly associational.
While we similarly test for significance, we propose an interpretable score for the local impact of specific properties.
Additionally, our approach is inherently interventional.

Related to the interventional nature of our work is \cite{buechner2024facing}.
The authors generate synthetic data with selected interventions to investigate emotion classifiers.
We extend their work to more general input properties and provide a structured way to generate local interventional explanations.
Other synthetic analysis datasets, e.g., \cite{hesse2023funnybirds,ramesh2024synthetic}, can be used to study model behavior.
In contrast, we directly intervene in inputs to remove the domain shift necessary for the synthetic analysis of pre-trained models.
The related approach \cite{kazimi2025explaining} explains models globally by intervening on image semantics, while we focus in our work on locally attributing the importance of properties.
Regarding the estimation of the impact with respect to a property, our work is most closely related to \cite{goyal2019explaining}.
In fact, our measure can be seen as an extension of the causal concept effect for gradual interventions.
Lastly, we discuss the interaction with Pearl's causal hierarchy and the causal hierarchy theorem \cite{bareinboim2022onpearl}, specifically for visual interventions \cite{pan2024counterfactual}.
Other works study the link between causality and explanations by intervening on training hyperparameters \cite{karimi2022relationship} or by casting explanations as falsifiable hypotheses to be verified \cite{schuhmacher2023framework}.

%% file: sec/3_method.tex
\begin{figure}[t]
    \centering
    \resizebox{1.0\columnwidth}{!}{
    \input{figs/scm_small}
    }
    \caption{
    Our structural causal model \cite{pearl2009causality} for the property dependence of trained networks.
    Dashed connections potentially exist depending on the specific task/property combination, and the sampled (S$_{train}$) data.
    In this work, we study the \textcolor{red}{red dashed link} between $\foi$ and $\hat{Y}$. 
    By intervening on $\foi$, i.e., \textcolor{blue}{$do(\foi:= \rfoi)$}, we induce changes in $\hat{Y}$, which are fully determined by the network $\class_\cweights$.
    }
    \label{fig:scm}
\end{figure}

\section{Method}
\label{sec:method}

\subsection{Causal Preliminaries}
\label{sec:causal-prelim}

We employ structural causal models (SCMs) \cite{pearl2009causality} to describe the data-generating process underlying our analysis. 
SCMs provide a flexible framework to model complex relationships between variables.
In this work, we aim to understand the decision-making and prediction behavior of deep neural networks for individual inputs.
Consequently, we model the outputs of trained networks as results in a data-generating system and visualize our proposed SCM in \cref{fig:scm}.
In there, the network outputs $\hat{Y}$ are deterministically produced by a parameterized model $\class_\cweights$.
The weights are optimized on a collection of training data $\data$.
This training data is not arbitrary. 
Instead, the sampled inputs strongly depend on task-related reference annotations $Y$.

In the literature, different alternatives to describe the inputs in such a system are discussed, e.g., \cite{goyal2019explaining,reimers2020determining,karimi2022relationship,buechner2024facing}.
In our work, we follow \cite{reimers2020determining,buechner2024facing} and model inputs as a possibly infinite collection of semantic properties.
These properties are not necessarily independent and can be causally related or spuriously correlated.
Note that the reference annotation $Y$ is such a semantic property, and $\class_\cweights$ primarily aims to extract it.
Nevertheless, to interpret the prediction behavior of $\class_\cweights$, we study the influence of other properties of interest $\foi$ on $\hat{Y}$.
To do this, we propose to intervene in individual input properties and measure the changes induced in the outputs.

\subsection{Why Do We Need Interventions?}
\label{sec:y-interv}

Causal insights can be hierarchically ordered in the so-called causal ladder \cite{pearl2009causality}.
This ladder, formally Pearl's Causal Hierarchy (PCH) \cite{bareinboim2022onpearl}, contains three distinct levels: associational, interventional, and counterfactual (see \cite{bareinboim2022onpearl} for a formal definition).
The first level, associational, is characterized by correlations observed in a given system. 
It focuses on statistical patterns and relationships within the data.
The second level, interventional, involves actively changing variables within the system to study the resulting effects. 
This is formally represented using the $do$-operator \cite{pearl2009causality}, which allows researchers to examine the causal impact of interventions.
The third level, counterfactual, deals with hypothetical scenarios, where researchers consider the potential outcome if an intervention had been made, given specific observations. 
Crucially, the causal hierarchy theorem \cite[Thm.~1]{bareinboim2022onpearl} states that the three levels are distinct, and the PCH almost never collapses in the general case.
Hence, to answer questions of a certain PCH level, data from the corresponding level is needed \cite[Cor.~1]{bareinboim2022onpearl}.

Consequently, our work falls into the second level and generates insights beyond associations.
Related works on the interventional level either focus on pixel attributions, e.g., \cite{zeiler2014visualizing,petsiuk2018rise,zintgraf2017visualizing,chang2018explaining}, or global explanations of semantic properties \cite{goyal2019explaining,kazimi2025explaining}.
Such globally identified causal factors do not necessarily hold locally.
To be specific, for a particular input, selected properties could be occluded or overshadowed. 
Our approach closes this gap and provides local interventional insights for specific semantic properties. %

\subsection{Generating Interventional Data}
\label{sec:inv-data}

To locally explain prediction behavior in the vision domain, we propose intervening directly on a property of interest $\foi$ (\textcolor{blue}{$do(\foi:= \rfoi)$} in \cref{fig:scm}).
To perturb $\foi$ in an image, we identify three options: 
Capturing new interventional data, designing synthetic interventions, and interventions via generative models. 
The first two approaches involve collecting new interventional data \cite{bareinboim2022onpearl} and are suited for specialized tasks in complex domains.
And while we empirically assess them in \cref{sec:experiments}, we agree with \cite{goyal2019explaining} and argue that generative models offer broad applicability with reduced human labor.

Consequently, for local interventions, we propose to leverage recent breakthroughs in image-to-image editing models, e.g., \cite{brooks2022instructpix2pix,mokady2023null,fu2024mgie}. 
These models are based on latent diffusion \cite{rombach2022high} and align inputs with text prompts through classifier-free guidance (CFG) scaling \cite{ho2021classifier}. %
Following \cite{brooks2022instructpix2pix} for timesteps $t$, CFG scaling utilizes
\begin{equation}
\label{eq:cfg}
    \Bar{e}(z_t,c_T) = e(z_t, \varnothing) + s_T (e(z_t, c_T)- e(z_t, \varnothing)),
\end{equation}
during inference, to steer a parameterized score network $e$ away from the unconditional distribution, $e(z_t, \varnothing)$, when predicting the noise in latents $z_t$.
Hence, increasing the CFG scale $s_T$ increases alignment with the conditioning text instruction $c_T$.
In practice, \cite{brooks2022instructpix2pix,fu2024mgie} include a second conditioning term, which we discuss in Appx.~\ref{app:cfg-img}.

Crucially, in \cref{eq:cfg}, $e$ is optimized jointly as a conditional model \cite{ho2021classifier}.
Therefore, the generative model learns to interpolate between property states during training.
This is particularly important for achieving non-linear and gradual transitions, allowing us to study more complex properties.
Hence, these prompting capabilities facilitate targeted and controllable interventions via text instructions.
Note that our focus on semantic properties separates our approach from existing works providing visual counterfactual explanations, e.g., \cite{fong2017interpretable,goyal2019counterfactual,popescu2021counterfactual,khorram2022cycle} or see \cite{stepin2021survey}.

In addition, intervening in the input has distinct advantages. 
First, the explanations are model-agnostic and do not depend on a specific architecture. 
Second, we do not need access to a model's weights, only its outputs. 
Finally, users can visually inspect the interventions and potentially include prior knowledge. 
Consequently, it allows for manual validation of interventional data, which is related to the idea of care sets proposed as a relaxation of the causal hierarchy theorem in \cite{pan2024counterfactual}.
We discuss alternatives to input space interventions in Appx.~\ref{app:lat-interv}.
Having established how to intervene, we now turn our attention to measuring the impact of these interventions on a model's predictions. 
Specifically, given interventional data with respect to a property $\foi$, we aim to quantify the corresponding changes in $\hat{Y}$.

\subsection{Measuring Systematic Change}
\label{sec:propgrad}

We utilize CFG scaling (\cref{eq:cfg}) to gradually intervene in a property of an original input image to generate interventional data.
Next, to measure the changes induced in the network outputs, we approximate the magnitude of the gradient with respect to the property $\foi$, i.e., $|\nabla_\foi \class_\cweights(I_\rfoi)|$. 
Here, $I_\rfoi$ is an input with a specific realization $\foi=\rfoi$.

Gradients as a measure of change or impact with respect to $\foi$ are related to the causal concept effect \cite{goyal2019explaining} and can be seen as an extension for gradual interventions.
We provide a detailed discussion of this connection in Appx.~\ref{app:cace}.
Nevertheless, given our gradual approach, this extension is important as periodical or parabolic effects of properties are potentially possible.
To illustrate this possibility, consider the following example.
Imagine a young person with brown hair, and suppose we gradually intervene on their hair color, transitioning from brown to gray to white. 
At first, an age classifier may become increasingly uncertain as the hair color changes, predicting an older age as the hair becomes grayer. 
However, once the hair is completely white, the classifier may again correctly predict a younger age due to hair color trends, e.g., platinum blonde or white hair amongst young people. 
By approximating gradient magnitudes for gradual variations in the hair color property, we can capture this non-linear shift in behavior.

Specifically, we subsample interventions and create a discrete ordered list of interventional inputs $I_\rfoi$, e.g., by using \cite{fu2024mgie} and linearly increasing the CFG scale.
We then compute the output of the trained model for each interventional input sample.
Finally, assuming a set $\fset$ of equidistant property realizations $\rfoi$ $(*)$ \cite{buechner2024facing}, we approximate the expected gradient magnitude with respect to $\foi$ with
\begin{align}
\begin{split}\label{eq:exp_grad}
\mathbb{E}_{\foi}[|\nabla_\foi \class_\cweights(I_\foi)|] &= \int |\nabla_{\rfoi} \class_\cweights(I_\rfoi)|\cdot p(\rfoi) d\rfoi\\
\overset{(*)}&{=} \frac{1}{|\fset|} \sum_{\rfoi\in \fset} |\nabla_{\rfoi} \class_\cweights(I_\rfoi)|,
\end{split}    
\end{align}
where $p(\rfoi)$ is the probability density of realizations $\rfoi$.

We refer to scores measured using \cref{eq:exp_grad} as expected property gradient magnitudes, or \propgrad as a short-form.
To accurately approximate \cref{eq:exp_grad} given our discrete list of intervened input samples, we employ Fornberg's finite differences \cite{fornberg1988generation}, as provided in \cite{harris2020array}. 
The expected property gradient magnitude is an interpretable score of the systematic change with respect to variations in $\foi$.
To illustrate this, a score of \propgrad$= 0.01$ indicates an average deviation of one percent for each discrete step of $\foi$.

Nevertheless, \propgrad has limitations, and a high score does not necessarily indicate significance.
In fact, noise can lead to high gradient magnitudes even if no systematic behavior exists.
Hence, we need to differentiate between systematic and random changes in the prediction behavior.

\subsection{Testing for Statistical Significance}

A high effect size, measured with statistics such as \propgrad or Pearson's correlation \cite{pearson1895notes}, does not imply significance.
Hence, to determine significance, we follow \cite{buechner2024facing} and employ shuffle hypothesis testing, e.g., \cite{good2013permutation}.
This approach compares a test statistic from the original observations to $K$ randomly shuffled versions.
Here, the interventional values of $\foi$ and the corresponding model outputs constitute the original correspondence.
We use \propgrad as our test statistic, which connects our measure of behavior changes to the hypothesis test.
Permuting the observations destroys the systematic relationship between $\foi$ and the model outputs and facilitates approximating the null hypothesis (pseudo-code in Appx.~\ref{app:th-exp}).
In our experiments, we use a significance level $0.01$ and perform 10K permutations.

%% file: figs/scm_small.tex
\begin{tikzpicture}[
        > = stealth,
        shorten > = 1pt,
        auto,
        node distance = 3cm,
        semithick
    ]
    \tikzstyle{every state}=[
        draw = black,
        thick,
        fill = white,
        minimum size = 10mm
    ]
    
    \tikzset{X/.style={
        circle, draw=blue, fill=blue!15, thick,
        }
    }
    \tikzdeclarepattern{
      name=mylines,
      parameters={
          \pgfkeysvalueof{/pgf/pattern keys/size},
          \pgfkeysvalueof{/pgf/pattern keys/angle},
          \pgfkeysvalueof{/pgf/pattern keys/line width},
      },
      bounding box={
        (0,-0.5*\pgfkeysvalueof{/pgf/pattern keys/line width}) and
        (\pgfkeysvalueof{/pgf/pattern keys/size},
    0.5*\pgfkeysvalueof{/pgf/pattern keys/line width})},
      tile size={(0.5\pgfkeysvalueof{/pgf/pattern keys/size},
    \pgfkeysvalueof{/pgf/pattern keys/size})},
      tile transformation={rotate=\pgfkeysvalueof{/pgf/pattern keys/angle}},
      defaults={
        size/.initial=5pt,
        angle/.initial=45,
        line width/.initial=.4pt,
      },
      code={
          \draw [line width=\pgfkeysvalueof{/pgf/pattern keys/line width}]
            (0,0) -- (\pgfkeysvalueof{/pgf/pattern keys/size},0);
      },
    }
    
    \node
        [state]
        (gt)
        {$Y$};
    \node
        [X,align=center, line width=0.5mm, dotted]
        (x)
        [below = 0.2cm of gt]
        {Property of\\ Interest $\foi$};
    
    \node
        [rectangle,pattern={mylines[size=2pt,line width=.6pt,angle=45]}, pattern color=black!25, draw=black, align=center, anchor=south west]
        (lat)
        [below left = -0.9cm and 1cm of gt]
        {Latent Data\\ Distribution\\\vspace{1.2cm}};{
            \node
                [X,align=center,  line width=0.5mm, dotted, overlay]
                (int)
                [below left = 0.3cm and 1.5cm of gt]
                {local\\ Int};
        };
    \node
        [state]
        (ts)
        [right = 1.5cm of gt]
        {$\data$};
        
    \node
        [state]
        (p)
        [right = 5cm of x]
        {$\hat{Y}$};
    
    \node
        [state]
        (w)
        [right = 1.5cm of ts]
        {$\theta$};
    
    \path[black, ->, draw, above, sloped] (lat) -- node {} (gt);
    \path[black, ->, draw, above, sloped] (lat) -- node {} (x);
    \path[black, ->, draw, above, sloped] (gt) -- ($(ts.west)+(-0.8,0)$) -- node {\small$\text{S}_{\text{train}}$} (ts);
    \path[black, ->, draw, below, dashed, sloped] (x) -- node {\small$\text{S}_{\text{train}}$} (ts);
    \path[black, ->, draw, above, sloped] (gt) -- ($(gt.east)+(0.4,-0.25)$) -- node {$\class_\cweights$} (p);
    \path[black, dashed, <->, draw, above, rounded corners=0.2cm] (gt) -- ($(gt.south west)+(-0.4,-0.1)$) -- node {} ($(x.north west)+(-0.2,0.2)$) -- (x);
    \path[black, ->, draw, above, sloped] (ts) edge node {\small Optimizer} (w);
    \path[black, ->, draw, above, sloped] (w) -- node {$\class_\cweights$} (p);
    \path[red, dashed, line width=0.5mm, ->, draw, above, sloped] (x) -- node {$\class_\cweights$} (p);
    \path[blue, dotted, line width=0.5mm, ->, draw, below, rounded corners=0.45cm] (int) -- ($(int.south east)+(-0.8,-0.5)$) -- node {$do(\foi := \rfoi)$} ($(x.south west)+(-1,0)$) -- (x);

    \draw[thick,dotted,rounded corners] ($(gt.north west)+(-0.83,0.3)$)  rectangle ($(x.south east)+(0.45,-0.4)$) node[pos=0.007] {\hspace{0.53cm}Inputs};
    \end{tikzpicture}

%% file: sec/4_experiments.tex
\section{Experiments}
\label{sec:experiments}
We demonstrate the effectiveness of our local interventional approach in various experiments.
First, we validate it in a synthetic biased scenario, where we correlate a property with the label. 
Second, we investigate a realistic skin lesion classification task regarding a known bias.
Afterward, we study the training dynamics of eight modern image classification models and, finally, a large pre-trained CLIP \cite{radford2021learning} model using real-life interventional data.

\begin{figure}[tb]
\centering
\includegraphics[width=\linewidth, trim=0.8cm 0.8cm 0.5cm 1.5cm, clip]{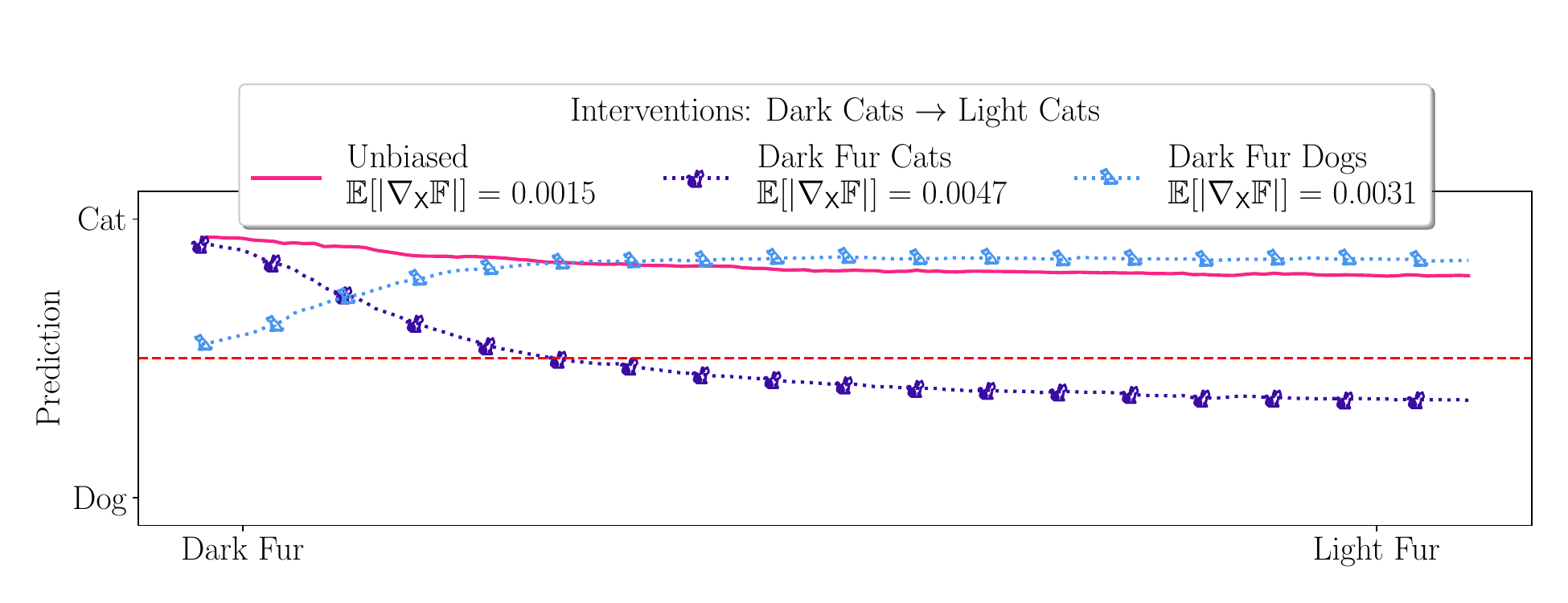}
\caption{
Average changes in model outputs (softmaxed cat-logit) for an intervention on the fur color in all \textbf{dark-furred cat} test images.
Here, we use three models: an unbiased one, one trained on only dark-furred cats and light-furred dogs, and one trained on the opposite bias.
The legend lists the mean \propgrad per model.
}
\label{fig:fur-mean-results}
\end{figure}

\begin{table}[t]
    \centering
    \caption{\propgrad of the fur color and background property for our three CvD models.
    Additionally, we report significance ($p<0.01$) abbreviated as ``S'' and prediction flips denoted as ``F''.
    }
    \label{tab:cvd-impact}
    \begin{tabular}{lcccccc}
    \toprule
        & \multicolumn{3}{c}{Fur Color} & \multicolumn{3}{c}{Background}\\
        \cmidrule(lr){2-4} \cmidrule(lr){5-7}
        Model & \propgrad & S & F & \propgrad & S & F \\
    \midrule
        Unbiased        & \bfseries .0099 & \cmark & \xmark & .00060 & \cmark & \xmark\\
        Dark Cats & \bfseries .0109 & \cmark & \cmark & .00006 & \cmark & \xmark\\
        Dark Dogs & \bfseries .0110 & \cmark & \cmark & .00013 & \cmark & \xmark\\
    \bottomrule
    \end{tabular}
\end{table}

\subsection{Synthetic Biased Scenario}
\label{sec:exp1}

We begin our empirical evaluation by constructing a biased scenario from the Cats vs. Dogs (CvD) dataset \cite{dogs-vs-cats}. 
Using \cite{liu2024improved}, we create two additional variations of the original distribution, where the fur color strongly correlates with the label. 
After manually verifying this approach, we obtain three training and test data splits: the original unbiased split, a split with only dark-furred dogs and light-furred cats, and the reverse. 
We hypothesize that models trained on the biased data splits will strongly rely on the fur color.

To test this hypothesis, we train a ConvMixer \cite{trockman2022patches} model on each split until it achieves high performance (see Appx.~\ref{app:cvd} for concrete numbers). 
Biased models exhibit strong performance degradation on out-of-distribution test splits, while the unbiased model achieves consistent accuracy across all three scenarios.
However, it is unclear whether the fur color is locally the dominant property driving predictions for individual inputs. 
Other properties, such as background illumination, may also influence model behavior. 
To investigate local prediction behavior, we use \cite{fu2024mgie} to intervene on both fur color and background, increasing the CFG scale from 1 to 15. 
Finally, we quantify the impact using \propgrad and test for significance.
We provide the full hyperparameters and additional ablations in Appx.~\ref{app:cvd}.

\begin{figure}
    \centering
    \includegraphics[width=\linewidth, trim=0.4cm 2cm 3.3cm 1.4cm, clip]{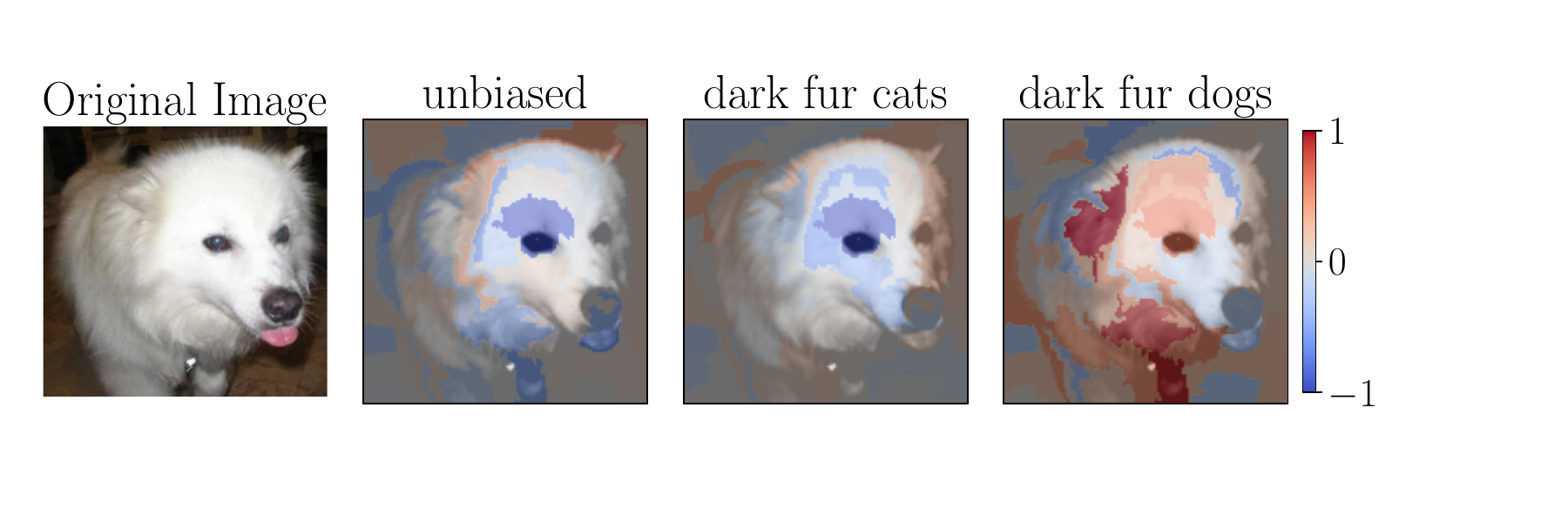}
    \includegraphics[width=\linewidth, trim=0.4cm 2cm 3.3cm 1.4cm, clip]{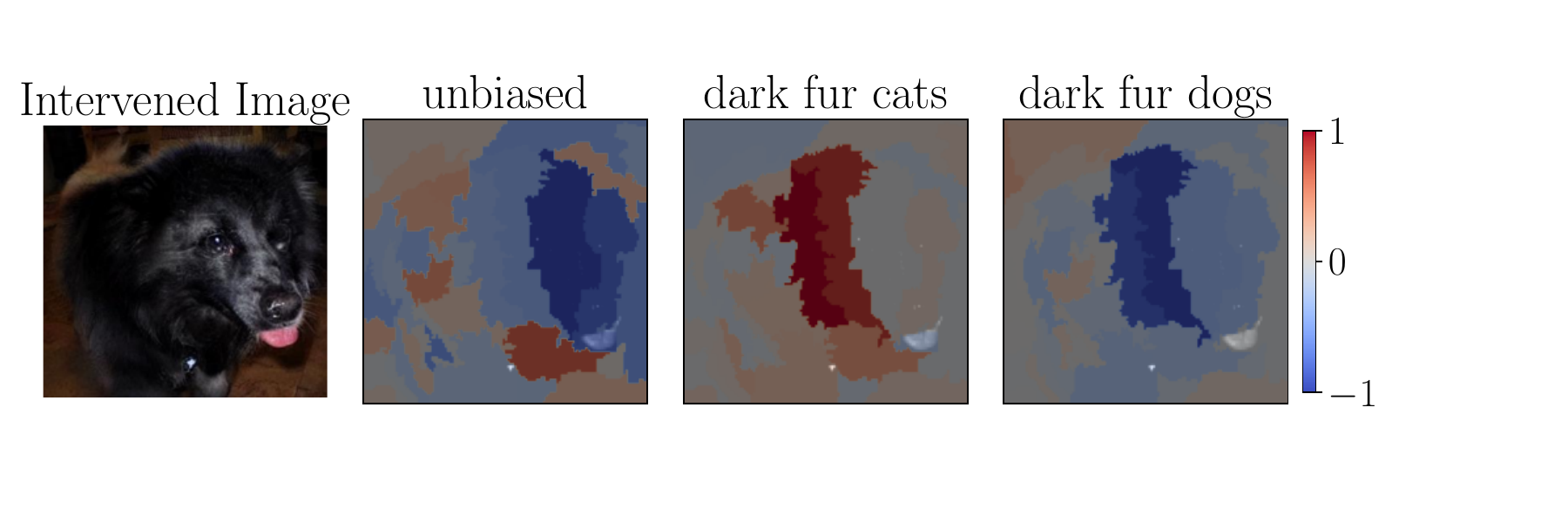}
    \caption{
    Local explanations using LIME \cite{ribeiro2016should} for the three CvD models (compare to \cref{fig:1d-results}).
    We explain with respect to the cat-logit, i.e., \textcolor{red}{red} areas indicate cat, while \textcolor{blue}{blue} signals dog.
    }
    \label{fig:cvd-lime}
\end{figure}

\begin{table*}[t]
    \centering
    \caption{%
    Mean accuracy ($\uparrow$) and standard deviation in percent (\%) of local XAI methods when predicting locally biased model behavior for an intervention.
    The first column denotes the dataset, i.e., Cats vs. Dogs \cite{dogs-vs-cats} (CvD) and ISIC archive \cite{isic-archive} (ISIC).
    For CvD, we evaluate the three ConvMixer \cite{trockman2022patches} from the separate training datasets.
    We investigate the interventions discussed in \cref{sec:exp1} and \cref{sec:exp4}, respectively.
    }
    \label{tab:quant}
    \resizebox{1.0\textwidth}{!}{
    \begin{tabular}{c|lccccccc}
    \toprule
    &
    Model & \textbf{Ours} & G-CAM \cite{selvaraju2020grad} & Int. Grad. \cite{sundararajan2017axiomatic} & Occlusion \cite{zeiler2014visualizing} & LIME \cite{ribeiro2016should} & K-SHAP \cite{lundberg2017unified} & DeepLift \cite{shrikumar2017learning} \\
    \midrule
    \multirow{3}{*}{\rotatebox{90}{CvD}} 
    
    & Unbiased & \textbf{86.13 {\footnotesize $\pm$ 2.0}} & 84.70 {\footnotesize $\pm$ 2.1} & 84.77 {\footnotesize $\pm$ 2.1} & 84.73 {\footnotesize $\pm$ 2.0} & 84.70 {\footnotesize $\pm$ 2.2} & 84.67 {\footnotesize $\pm$ 2.1} & 84.83 {\footnotesize $\pm$ 2.1} \\
    & Dark Cats Bias & \textbf{84.27 {\footnotesize $\pm$ 1.7}} & 61.43 {\footnotesize $\pm$ 3.3} & 64.77 {\footnotesize $\pm$ 1.8} & 67.70 {\footnotesize $\pm$ 1.9} & 58.63 {\footnotesize $\pm$ 1.9} & 60.87 {\footnotesize $\pm$ 1.0} & 67.13 {\footnotesize $\pm$ 2.2} \\
    & Dark Dogs Bias & \textbf{82.20 {\footnotesize $\pm$ 0.8}} & 64.20 {\footnotesize $\pm$ 2.7} & 64.10 {\footnotesize $\pm$ 2.2} & 67.10 {\footnotesize $\pm$ 1.2} & 56.50 {\footnotesize $\pm$ 2.1} & 58.50 {\footnotesize $\pm$ 2.2} & 64.77 {\footnotesize $\pm$ 2.1} \\
    \midrule
    \multirow{4}{*}{\rotatebox{90}{ISIC}} 
    
    &ResNet18 \cite{he2016deep} & \textbf{95.50 {\footnotesize $\pm$ 1.2}} & 80.00 {\footnotesize $\pm$ 4.0} & 78.63 {\footnotesize $\pm$ 2.7} & 78.75 {\footnotesize $\pm$ 4.0} & 76.12 {\footnotesize $\pm$ 5.1} & 75.88 {\footnotesize $\pm$ 4.6} & 76.88 {\footnotesize $\pm$ 6.5} \\
    &EfficientNet-B0 \cite{tan2019efficientnet} & \textbf{94.25 {\footnotesize $\pm$ 2.4}} & 79.00 {\footnotesize $\pm$ 4.9} & 73.75 {\footnotesize $\pm$ 5.6} & 75.37 {\footnotesize $\pm$ 6.3} & 73.88 {\footnotesize $\pm$ 5.1} & 72.75 {\footnotesize $\pm$ 5.5} & 76.38 {\footnotesize $\pm$ 6.5} \\
    &ConvNeXt-S \cite{liu2022convnet} & \textbf{92.88 {\footnotesize $\pm$ 1.6}} & 78.00 {\footnotesize $\pm$ 4.7} & 75.50 {\footnotesize $\pm$ 3.2} & 74.88 {\footnotesize $\pm$ 4.7} & 75.00 {\footnotesize $\pm$ 3.2} & 74.75 {\footnotesize $\pm$ 2.7} & 74.88 {\footnotesize $\pm$ 5.8} \\
    &ViT-B/16 \cite{dosovitskiy2020image} & \textbf{95.12 {\footnotesize $\pm$ 1.6}} & 80.25 {\footnotesize $\pm$ 3.4} & 75.87 {\footnotesize $\pm$ 2.5} & 77.25 {\footnotesize $\pm$ 3.5} & 76.00 {\footnotesize $\pm$ 3.4} & 75.88 {\footnotesize $\pm$ 2.9} & 77.50 {\footnotesize $\pm$ 3.9} \\

    \bottomrule
    \end{tabular}
    }
\end{table*}

\paragraph{Results: }
\cref{fig:1d-results} visualizes the model output behavior of all three classifiers for both the fur color intervention (\cref{fig:fur-intervent}) and the background intervention (\cref{fig:bg-intervent}) for an example.
For the fur color intervention, we can confirm our hypothesis: both biased models flip their predictions to follow the change in fur color, locally disregarding the actual animal.
While the unbiased model always predicts the correct class (dog), its logit for cat still increases. 
In contrast, the background intervention has a minimal effect, and no model crosses the prediction threshold.

To quantify these findings, we approximate the expected gradient magnitudes with respect to the intervened properties (see \cref{tab:cvd-impact}).
For all models, we measure at least one order of magnitude higher \propgrad for the fur color compared to the background property, corroborating the observed behavior changes.
Further, we measure the highest \propgrad of the fur color property for the two biased models, which coincides with the only observed prediction flips.

Next, to ensure robustness, we repeat the experiment with all test images again, intervening in the fur colors.
\cref{fig:fur-mean-results} visualizes the average model outputs for the dark furred cat split, and we provide the full details, additional ablations for the remaining splits, background interventions, and a discussion of the intervention failure cases in Appx.~\ref{app:cvd}.
Again, we observe the largest changes for the biased models, which align with flips in the prediction.
Thus, our interventional approach successfully identifies the fur color as a local cause for the observed model outputs. 

To stress its effectiveness, we compare our approach against local \cite{sundararajan2017axiomatic,shrikumar2017learning,selvaraju2020grad,zeiler2014visualizing,ribeiro2016should,lundberg2017unified} and global \cite{reimers2020determining,yeh2020completeness,schmalwasser2024exploiting} XAI baselines (Appx.~\ref{app:cvd-local-baselines}, \ref{app:cvd-global-baselines}).
While global methods find influential properties, they do not quantify their local impact. 
Local attribution methods highlight important areas, but they require semantic interpretation.
For example, although LIME explanations in \cref{fig:cvd-lime} align with fur color for biased models, the distinction from the unbiased model is unclear.
Interventions, i.e., the disparity between the top and bottom row in \cref{fig:cvd-lime}, can help interpret the results.

To formalize this comparison, we propose a quantitative task inspired by insertion/deletion tests \cite{petsiuk2018rise}: predicting whether an intervention on a property will change the model's output. 
This task directly assesses if a method can indicate locally biased behavior.
Further, this setup allows for a quantitative comparison to local baselines, given that our approach does not produce saliency maps but rather estimates the impact of a property directly using \propgrad. 
To adapt saliency methods, we measure the mean squared difference of the explanation pre- and post-intervention (e.g., between the two rows in \cref{fig:cvd-lime}).
For a fair comparison, we select the optimal threshold maximizing the accuracy for both the local baselines and our score \propgrad.
We repeat this task for the fur color property of 150 test images per class, resampling ten times to estimate standard deviations.
\cref{tab:quant} (top) summarizes the results and highlights the advantage of our approach.
While all methods perform consistently well in predicting changes for the unbiased model, \propgrad clearly improves over baselines for models exhibiting local bias.
Following this validation, we next analyze a real-world bias in skin lesion classification.

\subsection{Skin Lesion Classification}
\label{sec:exp4}

\begin{figure}[t]
    \centering
    \includegraphics[width=\linewidth]{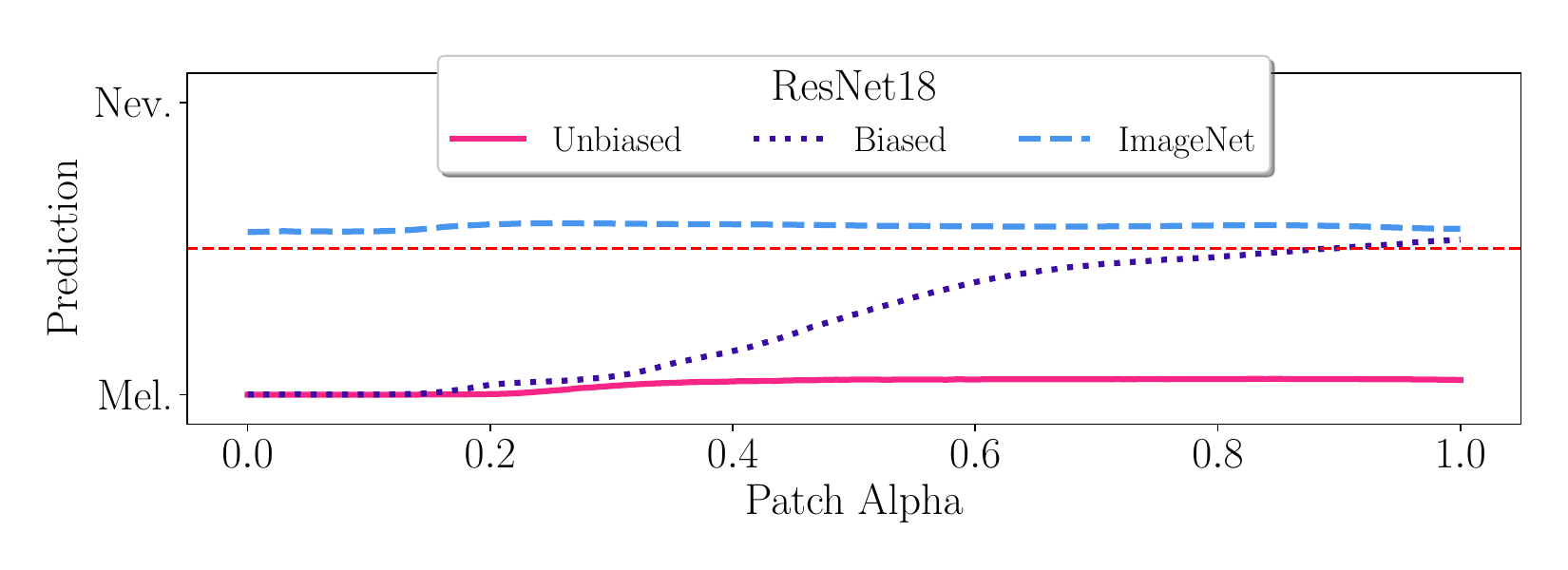}
    \phantom{aaaa}\includegraphics[width=0.9\linewidth, trim=0 0.5cm 0 0.8cm, clip]{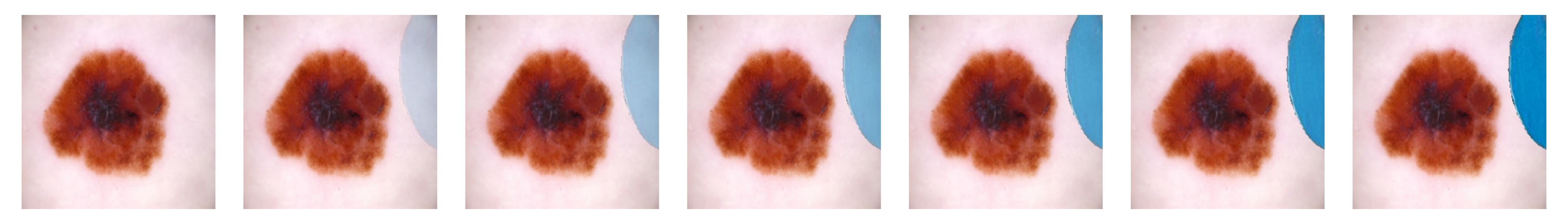}
    \caption{
    We visualize (top) how interventions targeting spurious colorful patches (bottom) \cite{scope2016study} affect three ResNet18 models: trained on biased/unbiased skin lesion data and ImageNet weights.
    }
    \label{fig:isic}
\end{figure}

\begin{table}[t]
\centering
\caption{
\propgrad for colorful patch interventions \cite{scope2016study} in skin lesion classifiers. 
We evaluate different models and training data.
}
\label{tab:isic-impact}
\begin{tabular}{lccc}
\toprule
                & \multicolumn{3}{c}{Training Data} \\
                \cmidrule(lr){2-4}
          Model &      Unbiased &   Biased & ImageNet \\
\midrule
       ResNet18 \cite{he2016deep} &      .00061 & \textbf{.00531} & .00062 \\
EfficientNet-B0 \cite{tan2019efficientnet} &      .00018 & \textbf{.00495} & .00066 \\
     ConvNeXt-S \cite{liu2022convnet} &      .00001 & \textbf{.00519} & .00081 \\
     ViT-B/16 \cite{dosovitskiy2020image} &      .00016 & \textbf{.00208} & .00129 \\
\bottomrule
\end{tabular}
\end{table}

In the domain of skin lesion classification (here, nevus/healthy vs. melanoma), a known bias is the correlation between colorful patches \cite{scope2016study} and the nevus class.
We assess how strongly this property is learned by four architectures: ResNet18 \cite{he2016deep}, EfficientNet-B0 \cite{tan2019efficientnet}, ConvNext-S \cite{liu2022convnet}, and ViT-B/16 \cite{dosovitskiy2020image}. 
For each model, we start with ImageNet \cite{russakovsky2015imagenet} weights.
Then, we either fine-tune on biased skin lesion data (50\% nevi with colorful patches \cite{scope2016study}) or unbiased data (no patches) from the ISIC archive \cite{isic-archive}.

To demonstrate that our approach accommodates diverse sources of interventional data, we build on domain knowledge and intervene synthetically.
Specifically, we blend segmented colorful patches \cite{rieger2020interpretations} into melanoma images (see \cref{fig:isic}, bottom). 
We randomly sample ten correctly classified melanoma images and repeat interventions with five patches each.
Appx. \ref{app:isic} contains detailed hyperparameters, predictive performances, and additional visualizations.

\paragraph{Results:}
The mean \propgrad in \cref{tab:isic-impact} show that models trained on biased data are most impacted by colorful patch interventions, indicating they learn the statistical correlation between the patches and the nevus class. 
Furthermore, the variants with ImageNet \cite{russakovsky2015imagenet} weights show higher patch sensitivity than the unbiased skin lesion models. 
We hypothesize this is because learning color is beneficial for general-purpose pre-training, whereas the unbiased models learn to disregard patches and focus on the actual lesions.

These results are further corroborated in \cref{fig:isic}, where we visualize the average changes in model outputs for the ResNet18 \cite{he2016deep} under the synthetic colorful patch intervention.
Specifically, we observe that the biased model flips its prediction and incorrectly classifies the melanoma images as healthy.
This highlights the ability of our interventional approach to investigate complex scenarios and use domain knowledge to provide actionable local explanations.

To further substantiate the benefit of our approach, we again compare against various local baselines in predicting whether an intervention (adding a colorful patch) will change the model's output. 
We follow the setup described in \cref{sec:exp1}, here using 100 melanoma test images, resampling ten times to estimate standard deviations.
We report accuracies for the four architectures averaged over the biased and unbiased training in \cref{tab:quant} (bottom).
Our score \propgrad significantly outperforms baseline methods for this task ($p<0.002$), highlighting its use case as a robust score to interpret local prediction behavior on the property level.
However, our aim is not to replace saliency methods, but rather to offer a complementary, interventional viewpoint for analyzing local behavior.
We demonstrate these capabilities in the following experiments.

\begin{figure}[t]
    \centering
    \includegraphics[width=\linewidth]{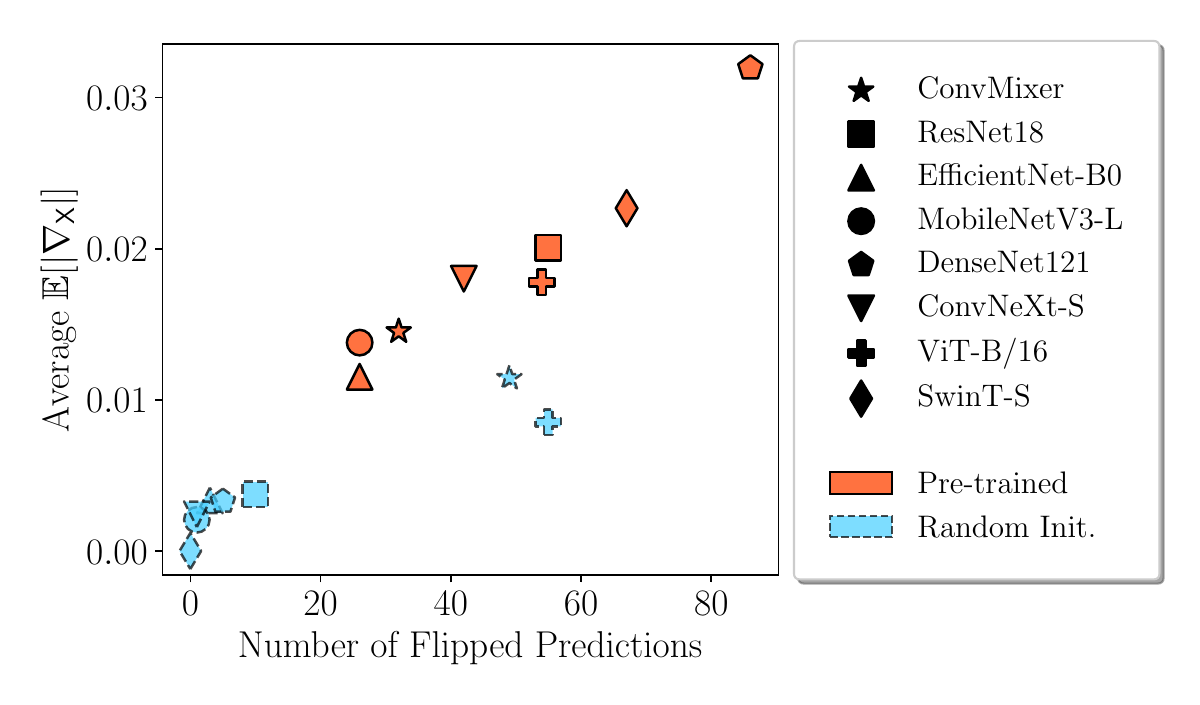}
    \caption{
    Average \propgrad visualized against the number of prediction flips for the training of various architectures with respect to the gray hair intervention. 
    We display both ImageNet \cite{russakovsky2015imagenet} pre-trained and randomly initialized models.
    }
    \label{fig:celebA-impact}
\end{figure}

\subsection{Training Analysis}
\label{sec:exp2}

We investigate how the \propgrad of a property locally develops during the training of various architectures.
This is an important question because it helps us understand how diverse models learn to represent and utilize properties in the data.
Additionally, it is crucial to consider the impact of the initial parameters on the learned properties \cite{penzel2022investigating}.
To address these questions, we select a range of convolutional and transformer-based architectures widely used in computer vision tasks \cite{trockman2022patches,he2016deep,tan2019efficientnet,howard2019searching,huang2017densely,liu2022convnet,dosovitskiy2020image,liu2021swin} (see \cref{fig:celebA-impact}). 
For all of these models, we train a randomly initialized and an ImageNet \cite{russakovsky2015imagenet} pre-trained version for 100 epochs.

Regarding the corresponding task, we construct a binary classification problem from CelebA \cite{liu2015faceattributes}, following an idea proposed in \cite{pan2024counterfactual}.
Specifically, we utilize the attribute young as a label and split the data in a balanced manner.
For this label, the gray hair property is negatively correlated \cite{pan2024counterfactual}, and a well-performing classifier should learn this association during the training process.
To study the training dynamics, we intervene on the hair color of a young test sample using \cite{fu2024mgie} and calculate \propgrad after each epoch.
We include detailed hyperparameters in Appx.~\ref{app:celebA} together with additional ablations and visualizations.

\paragraph{Results: }

\begin{figure}[t]
    \centering
    \includegraphics[width=\linewidth, trim=0 9.3cm 0 0, clip]{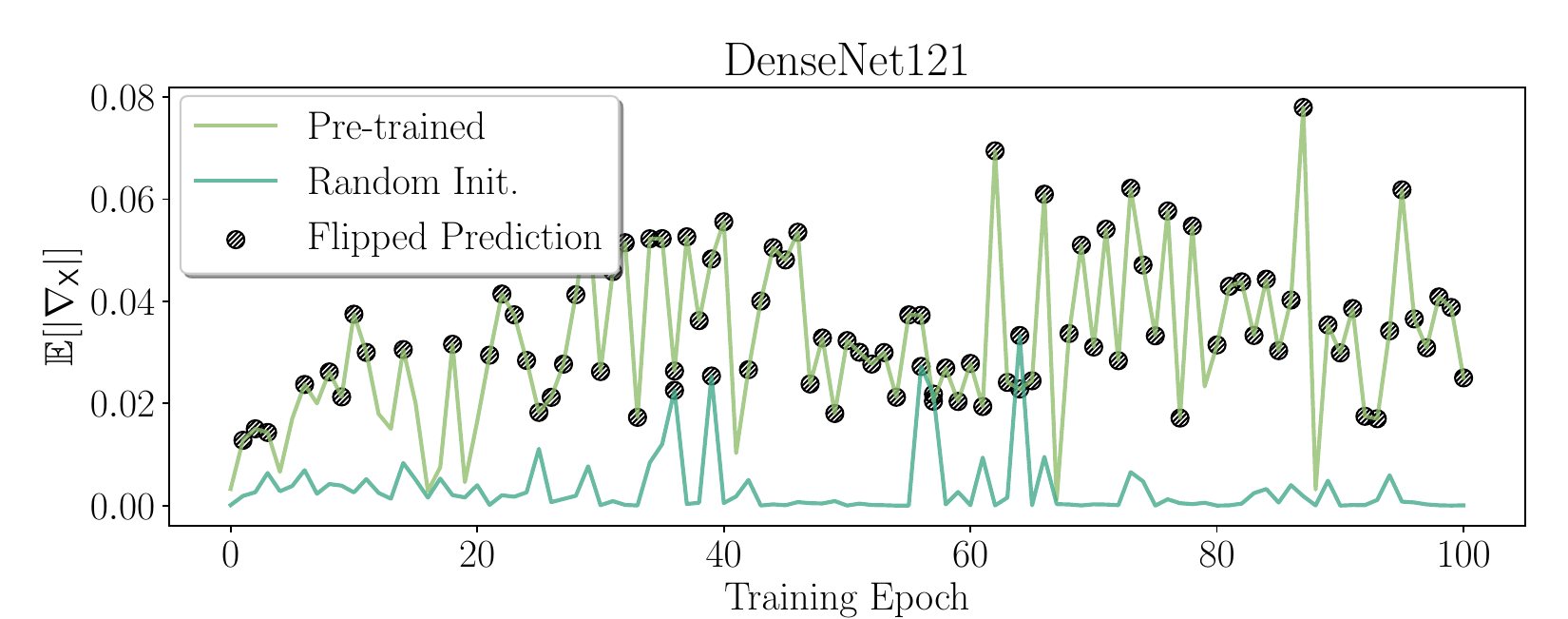}
    \caption{ \propgrad development during training of DenseNet121 \cite{huang2017densely} models with respect to the gray hair property.
    We separate ImageNet pre-trained \cite{russakovsky2015imagenet} and randomly initialized weights, and highlight epochs where predictions flip under the intervention.
    }
    \label{fig:impact-densenet}
\end{figure}

In \cref{fig:celebA-impact}, we visualize the average \propgrad of the hair color for a local example over the training process for both pre-trained and randomly initialized models.
Specifically, we display the average \propgrad against the observed flips in the prediction during the hair color intervention.
In Appx.~\ref{app:celeba-results}, we include the concrete numbers (\cref{tab:celebA-impact-ext}).
Our analysis reveals two key insights:

First, for all architectures, the pre-trained variants locally exhibit higher \propgrad compared to the randomly initialized versions. 
This observation is consistent with the number of times the networks' predictions flip during training.
Notably, regarding prediction changes, the ConvMixer \cite{trockman2022patches} and ViT \cite{dosovitskiy2020image} are outliers.
However, these two models also demonstrate the highest measured \propgrad among the randomly initialized variants. 
In general, we find that increased \propgrad correlates with more flipped predictions in \cref{fig:celebA-impact}.

Our second key finding is illustrated in \cref{fig:impact-densenet}, which reveals that the \propgrad with respect to the hair color exhibits strong local fluctuations during the training for the DenseNets \cite{huang2017densely} (highest \propgrad difference in \cref{fig:celebA-impact}).
Notably, both models classify the original sample correctly after every epoch during training.
This indicates that the differences in \cref{fig:impact-densenet} are not explained by incorrect classifications of the original image for either of the two models.
Nevertheless, the networks do not continuously learn to rely on the hair color property but instead locally ``forget'' it, even in later epochs.
This effect is particularly pronounced for the pre-trained model, whereas the randomly initialized version tends to show low \propgrad values.
We observe similar behavior for other architectures (Appx.~\ref{app:celeba-results}).

\subsection{CLIP Zero-Shot Classification}
\label{sec:exp3}

In our final set of experiments, we investigate the widely used multimodal backbone CLIP ViT-B/32 \cite{radford2021learning} for zero-shot classification.
Our approach is model-agnostic, requiring only access to model outputs, here cosine similarities in the learned latent space.
Specifically, we measure the impact of interventional data by comparing ten different text descriptions against the visual embeddings.
This procedure follows recent trends to increase zero-shot performance for CLIP models, e.g., \cite{pratt2023does,roth2023waffling}.

As the property of interest, we select object orientation, which is a known bias, for example, in ImageNet models \cite{henderson2021biased}.
Additionally, we demonstrate a third type of interventional data and capture real-life interventional images. 
Specifically, we record a rotation of three toy figures (elephant, giraffe, and stegosaurus) using a turntable.
Note that during rotation, the ground truth does not change, i.e., it is the identical object.
We provide the hyperparameters and additional visualizations in Appx.~\ref{app:clip}.

\paragraph{Results: }

\begin{figure}[tb]
\centering
\includegraphics[width=\linewidth, trim=0.7cm 0.5cm 0.5cm 0cm, clip]{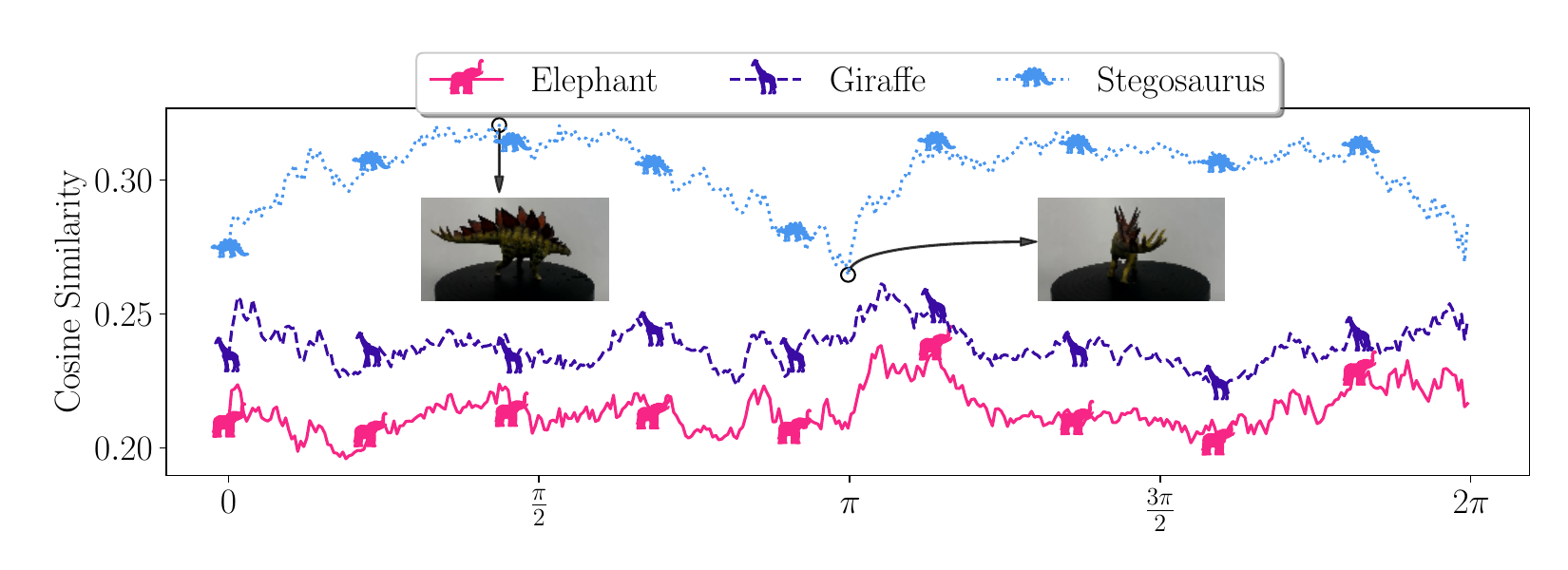}
\caption{Average CLIP \cite{radford2021learning} cosine similarities for \textbf{real-life} interventional data. We mark high and low points for the ground truth.}
\label{fig:clip-real}
\end{figure}

\cref{fig:clip-real} displays the average changes in output behavior (\propgrad scores in Appx.~\ref{app:clip}) for our real-life interventions.
Specifically, we visualize zero-shot classifications of a CLIP model.
Note that the behavior is remarkably consistent between text descriptors with similar standard deviations during the full interventions.
Further, all measured \propgrad are statistically significant ($p < 0.01$), i.e., the CLIP model is influenced by object orientation.
While expected, e.g., \cite{henderson2021biased}, our local interventional approach facilitates direct interpretations of the change in behavior.

\cref{fig:clip-real} reveals that the highest average similarity for the correct class occurs when the toy animal is rotated sideways.
Periodical minima align with the front or back-facing orientations. 
In contrast, the highest similarities for the other classes appear close to the minimum of the ground truth, indicating lower confidence.
In Appx.~\ref{app:clip}, we include the response of CLIP to synthetic rotations of a 3D model around other axes as additional ablations.
We confirm that uncommon, e.g., upside-down positions, lead to lower scores.
Hence, our approach provides actionable guidance to locally select an appropriate input orientation.

%% file: sec/5_limitations.tex
\section{Limitations}
\label{sec:limitations}

The main limitation of our work is related to a point discussed in \cite{goyal2019explaining}.
Specifically, we rely on interventional data, which must be captured or virtually acquired.
While we follow the idea of using generative models, i.e., image editing models \cite{brooks2022instructpix2pix,fu2024mgie} together with CFG scaling \cite{ho2021classifier}, we understand that for these models, the causal hierarchy theorem applies \cite{pan2024counterfactual}.
However, note that \cite{fu2024mgie} is trained on synthetic interventional data using \cite{hertz2022prompt}.
Further, intervening in input space enables visual verification of whether the intervention targets the correct property.
An idea related to the care set of properties in \cite{pan2024counterfactual}.
Nevertheless, we include failure cases for interventions with \cite{fu2024mgie} in Appx.~\ref{app:cvd-results}.

Similar to \cite{reimers2020determining,goyal2019counterfactual,kim2018interpretability}, our approach is non-explorative.
Hence, we need a concrete property to investigate and measure the corresponding \propgrad.
While this has the advantage of being able to investigate unlearned properties, we believe a combination with explorative methods, e.g., \cite{yeh2020completeness,cunninghamSparseAutoencodersFind2023,marksSparseFeatureCircuits2024a,kazimi2025explaining}, is an interesting future direction.

%% file: sec/6_conclusions.tex
\section{Conclusions}
\label{sec:conclusions}

By adopting a causal perspective, we study deep learning models and move beyond local associational explanations of their prediction behavior. 
We leverage recent breakthroughs in image-to-image editing models and Classifier-Free Guidance (CFG) scaling to gradually intervene in semantic properties.
To quantify the impact of the selected properties on the predictions of a trained model, we approximate the expected property gradient magnitude \propgrad and verify statistical significance with a corresponding hypothesis test.
Our approach offers several advantages, including the ability to locally identify causal factors and facilitate direct interpretation and quantification of corresponding output changes.
To demonstrate its effectiveness, we perform an extensive empirical evaluation and study various models and tasks.
First, we validate our approach on synthetically biased data and identify the causal factor before applying it to real-world skin lesion data.
In both scenarios, we find that our approach outperforms local baselines in predicting locally biased behavior.
Then, while investigating the training dynamics of eight classification models, we show that our \propgrad score locally correlates well with the number of flipped predictions.
Finally, we use real-life interventions to study a pre-trained CLIP model, demonstrating that our approach can utilize diverse sources of interventional data.
As black-box models continue to play an increasingly prominent role in a wide range of applications, we believe that our work can aid the development of more trustworthy systems.

\vspace{0.2cm}
\noindent\textbf{Acknowledgements:} 
We thank all our colleagues from the CVG Jena. 
In particular, Tim Büchner, Laines Schmalwasser, Gideon Stein, and Jan Blunk.

%% file: sec/X_suppl.tex
\clearpage
\setcounter{page}{1}
\maketitlesupplementary

\setcounter{section}{0}
\renewcommand{\thesection}{\Alph{section}}
\renewcommand\thesubsection{\thesection.\arabic{subsection}}

\section*{Table of Contents}
\startcontents
\printcontents{ }{1}{}

\section{Method - Additional Details}
\label{app:method}

In this section, we include additional details and discussions for our methodology.
First, we describe the structural causal model (SCM), which is the foundation of our approach.
Then, we comment on the second classifier-free guidance (CFG) scale \cite{ho2021classifier}, which recent image-to-image editing models implement, e.g., \cite{brooks2022instructpix2pix,fu2024mgie}.
While in our main paper, we provide arguments for input space interventions, in \Cref{app:lat-interv}, we discuss alternatives.
To measure the impact of a property under gradual interventions, we propose to estimate the expected magnitude of the corresponding property gradients (\Cref{sec:propgrad}).
This score can be seen as an extension of the causal concept effect \cite{goyal2019explaining}, which we show in \Cref{app:cace}.
Finally, we detail the corresponding shuffle hypothesis test and discuss some finer details in our approach.
We also include example functional dependencies and a comparison of our score to the linear Pearson correlation coefficient \cite{pearson1895notes}.

\subsection{SCM for Property Dependence}
\label{app:scm}

In our main paper, we provide a quick overview (\Cref{sec:causal-prelim} of the SCM that underpins our analysis (\cref{fig:scm}).
Here, we provide a more detailed introduction from a different point of view.

First, to derive a structural causal model (SCM) or causal diagram after \cite{pearl2009causality}, we follow a related approach \cite{reimers2020determining} and frame the structure of supervised learning.
This then enables us to identify the variables determining the prediction behavior.

In supervised learning, the goal is to separate a latent data distribution based on task-specific reference annotations $Y$.
This latent data distribution is an exogenous variable in a supervised learning system that we do not directly observe.
Instead, we observe sampled data points or inputs, such as texts or images.
We adopt the view that network inputs are a collection of properties, which collectively define the input sample \cite{reimers2020determining}.

In this work, we focus on two properties of the inputs: the task-dependent reference annotation $Y$ and a property of interest $\foi$.
These properties are not necessarily independent and can be causally related. 
Alternatively, they may correlate due to the sampling process or a confounding factor.
This is particularly important because deep models are trained on finite, possibly biased samples of the true latent data distribution, where various spurious correlations can occur.
Furthermore, the sampled training data $\data$ is strongly dependent on the task-related reference annotations $Y$ and is not a random sample.

Given such a sample $\data$ of inputs with corresponding annotations and exogenous factors, such as optimizer and hyperparameter choices, we learn parameters $\cweights$ for a model $\class$.
For any input, $\class_\cweights$ deterministically produces a corresponding output $\hat{Y}$.
This output strongly depends on the learned weights $\cweights$, which in turn depends on $\data$.
However, for an input sampled after training, i.e., during inference, we must consider a direct connection between $Y$ and $\hat{Y}$ given that we directly optimize for this relationship.
This connection should hold for any model outperforming random guessing on a held-out test dataset.

In contrast, we cannot be certain about the direct connection between the property of interest $\foi$ and $\hat{Y}$, even if $\foi$ and $Y$ are correlated in $\data$, due to the automatic optimization of $\class_\cweights$.
We summarize the described interactions in the causal diagram in \cref{fig:scm}.

To investigate the influence of a property $\foi$, \cite{reimers2020determining} proposes to test conditional dependence in a collection of corresponding observations. 
While this approach can lead to global insights, they do not necessarily apply to individual local examples.
To be specific, while the prediction behavior of a neural network may be influenced by properties such as hair color overall, other properties can locally dominate.
Therefore, we propose gradually introducing changes in the property of interest $\foi$ for an otherwise fixed input to investigate the influence on $\hat{Y}$ encoded in $\class_\cweights$.
We describe our approach in \Cref{sec:inv-data}.

\subsection{CFG scale for Image Alignment}
\label{app:cfg-img}

In our main paper, we focus on the alignment with the image-edit instruction (\Cref{eq:cfg}) for modern generative models \cite{brooks2022instructpix2pix,fu2024mgie}.
However, in practice, these models implement a second condition: the alignment with the original image.
In the terminology of \cite{brooks2022instructpix2pix}, \Cref{eq:cfg} becomes
\begin{align}
    \begin{split}\label{eq:cfg-ext}
        \Bar{e}(z_t,c_T,c_I) = &e(z_t, \varnothing, \varnothing) \\
        &+ s_I (e(z_t, c_T, \varnothing)- e(z_t, \varnothing, \varnothing))\\
        &+ s_T (e(z_t, c_T, c_I)- e(z_t, c_T, \varnothing)).
    \end{split}
\end{align}
Again, we omit the parameterization of $e$ for brevity.

\Cref{eq:cfg-ext} includes two guiding scales which together determine the intervention.
In \cref{fig:fur-results2d} in Appx.~\ref{app:cvd-results}, we perform a small ablation and find that the CFG text scale predominantly controls the intervention.
Hence, in our work, we focus on $s_T$ and fix $s_I$ depending on the task.

\subsection{Input Intervention Alternatives}
\label{app:lat-interv}

Multiple options exist for intervening on a property $\foi$ of interest.
In principle, we could directly change the property value after extracting it.
However, deep models are not designed to operate on property-level inputs.
Instead, they expect inputs that conform to their learned input domain, such as images represented as pixel matrices for vision models.
Therefore, we can explore two alternative possibilities: intervening in the input space or modifying the latent representations extracted by a model $\class_\cweights$. 
In our main paper, we focus on input interventions.
In this section, we discuss the alternative of intervening in a trained model's latent space.

Specifically, we identify multiple limitations of latent space interventions compared to our approach.
Such latent representations are inherently non-interpretable, and while methods exist to ascribe meaning to changes in neurons, e.g., \cite{kim2018interpretability,yeh2020completeness,schmalwasser2024exploiting}, the representations are polysemantic \cite{elhage2022toy}.
In other words, neurons often encode multiple different semantic properties or concepts simultaneously.
This behavior is problematic because, in contrast to the input space, we cannot simply visualize the interventions.
Hence, we could introduce unknown confounding.

Recent work focuses on disentangling latent representations into human interpretable representations often based on sparse autoencoders, e.g.,  \cite{cunninghamSparseAutoencodersFind2023,scherlisPolysemanticityCapacityNeural2023,conmyAutomatedCircuitDiscovery2023,gaoScalingEvaluatingSparse2024,marksSparseFeatureCircuits2024a}.
However, these approaches are model-specific, require significant implementation overhead, and provide no guarantee that the property of interest $\foi$ is learned or extracted by the model.
To be specific, the latent representation contains no information about an unlearned property from an information-theoretic point of view.
Hence, a latent vector does not uniquely map to one input.
In fact, two inputs that only differ in an unlearned property can result in the same latent vector.
In other words, it might be impossible to investigate many properties deemed important by the user via interventions in latent space.
Among the advantages we list in the main paper, we believe that this last point is a crucial advantage of intervening in input space.

\subsection{Connection to Causal Concept Effect for Binary Properties}
\label{app:cace}

In \cite{goyal2019explaining}, the authors introduce the causal concept effect (CaCE), defined as
\begin{align}
\begin{split}\label{eq:cace}
    \text{CaCE}(\class, \foi) &= \mathbb{E}_g [\class(I) | do(\foi = 1)] \\
    & \hspace{1cm}- \mathbb{E}_g [\class(I) | do(\foi = 0)], 
\end{split}
\end{align}
where $g$ denotes the generative process to perform the intervention.
Similarly to us, the authors discuss various options, including generative models \cite{goyal2019explaining}.
Note that we exchange some of the original symbols by our notation and use $\class$ for the classifier and $\foi$ for the concept or property of interest.
\Cref{eq:cace} describes it for binary properties, but \cite{goyal2019explaining} further extends it to $N$-wise categorical by pairwise comparisons against the observed state for an input.
Nevertheless, we focus here on the version in \Cref{eq:cace}.

To show the connection to our expected gradient magnitude, we assume an intervention with binary property states.
Without loss of generality let $\rfoi \in \{0,1\}$.
Then, \Cref{eq:exp_grad} becomes
\begin{equation}
    \mathbb{E}_{\foi}[|\nabla_\foi \class_\cweights(I_\foi)|] = \frac{1}{2} \sum_{\rfoi\in \{0,1\}} |\nabla_{\rfoi}\class_\cweights(I_\rfoi)|.
\end{equation}
In this limited binary case, we approximate the gradient by calculating the difference between the model outputs for both property states.
Specifically, we get
\begin{equation}
    \frac{1}{2} (|\class_\cweights(I_0)-\class_\cweights(I_1)| + |\class_\cweights(I_1)-\class_\cweights(I_0)|),
\end{equation}
for the two possible property states in the binary example.
Here, the two gradient magnitudes are equal, finally resulting in $|\class_\cweights(I_0)-\class_\cweights(I_1)|$ as the measured effect.

Note that in our work, we focus on local inputs and interventions to not violate the causal hierarchy theorem for image editing \cite{pan2024counterfactual}.
Nevertheless, by taking the expectation over a generative process for $I_\rfoi$, we get $\mathbb{E}_g [|\class_\cweights(I_0)-\class_\cweights(I_1)|]$.
This expectation includes an absolute value.
However, it is clearly a related quantity to the CaCE in \Cref{eq:cace}.
To illustrate this further, assume $\class_\cweights(I_0) > \class_\cweights(I_1)$ and rewrite using the $do$ operator notation, then applying the linearity of expectation, we arrive at the right-hand side of \Cref{eq:cace}.

In our work, we take a further step by analyzing gradual interventions, such as those enabled by \cite{fu2024mgie}. 
Specifically, we utilize CFG scaling \cite{ho2021classifier} to generate ordered variations, allowing us to approximate the respective gradients for various settings of the property of interest using \cite{fornberg1988generation}. 
Here, gradients, as shown above, extend CaCE \cite{goyal2019explaining} and enable a more nuanced understanding of the relationships between inputs and outputs. 
While our main focus is on local explanations, we note that a consequential approach to deriving global explanations would be to take a second expectation over a probing or test dataset of inputs.
However, it is crucial to carefully scrutinize the specific interventions to ensure they do not violate \cite{pan2024counterfactual} and to confirm variations in the property of interest. 
Notably, this approach is similar to our findings in our first experiment (see \Cref{sec:exp1}), where we investigated multiple images (see also \Cref{app:cvd-results}).

\paragraph{Why do we need property gradients?}

To illustrate the advantage of our approach over CaCE, we consider a classic example from the causal literature, as seen in \cite{bareinboim2022onpearl}. 
In this scenario, we aim to estimate the causal effect of administering a drug, which would typically involve collecting interventional data through a randomized control trial.
However, the dosage variable is often not binary, and varying dosages can have different effects.
For instance, administering the drug in extremely high doses may cancel out its beneficial effects, resulting in a negligible or no treatment effect.
In other words, the desired impact is only achieved for specific values of the dosage property.

\begin{figure*}
    \centering
    \begin{subfigure}{0.32\textwidth}
        \includegraphics[width=\textwidth]{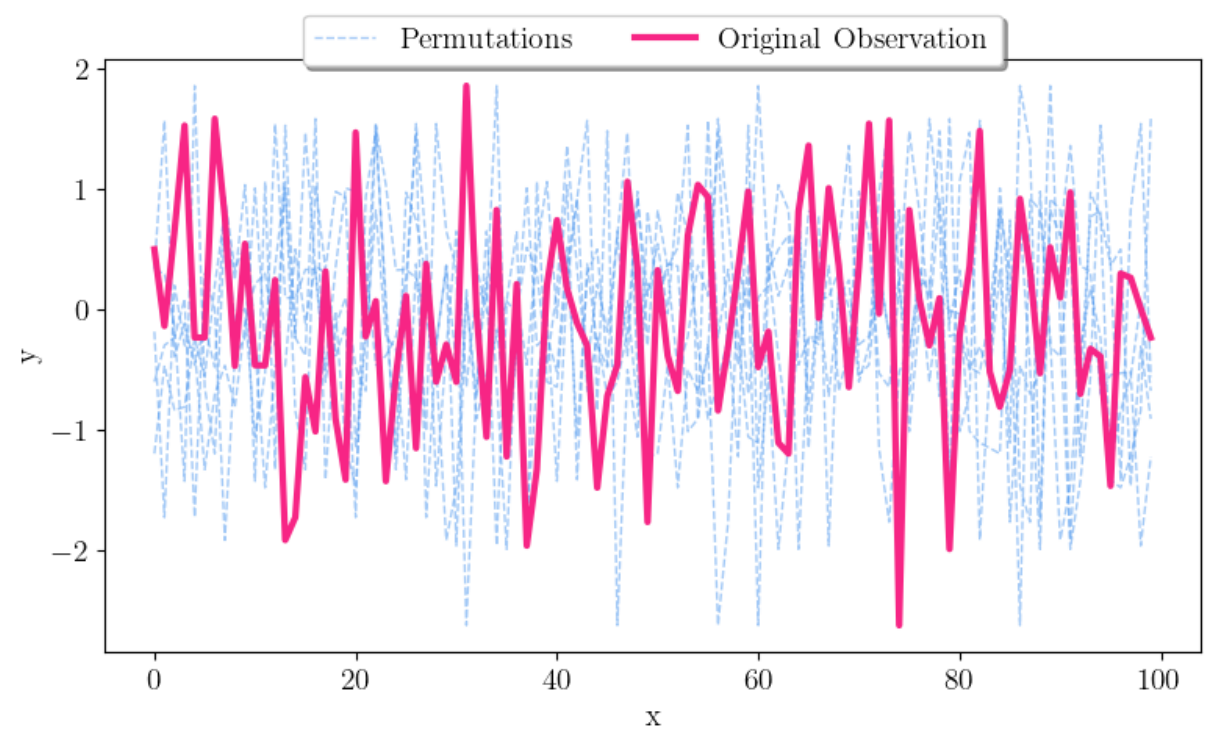}
        \caption{
            Normal distributed noise.
        }
        \label{fig:noise-shuffle}
    \end{subfigure}
    \begin{subfigure}{0.32\textwidth}
        \includegraphics[width=\textwidth]{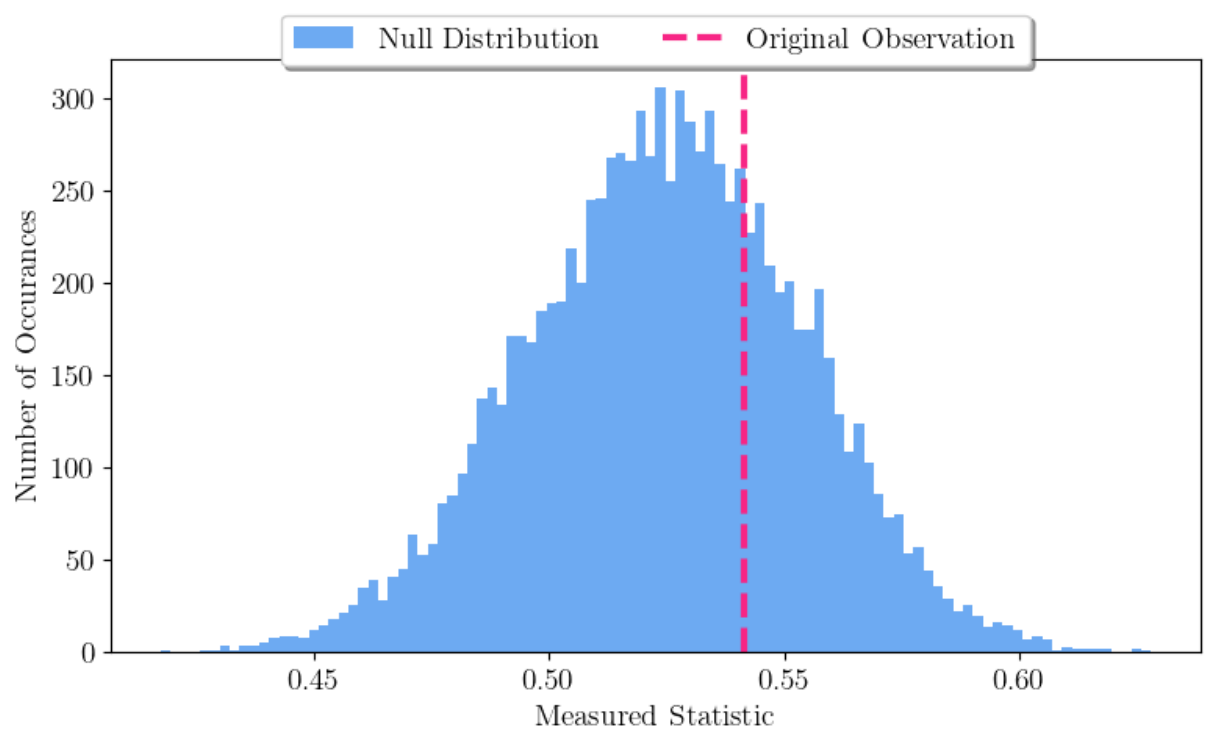}
        \caption{
            Expected $\nabla_\foi$ magnitude.
        }
        \label{fig:noise-abs-grad}
    \end{subfigure}
    \begin{subfigure}{0.32\textwidth}
        \includegraphics[width=\textwidth]{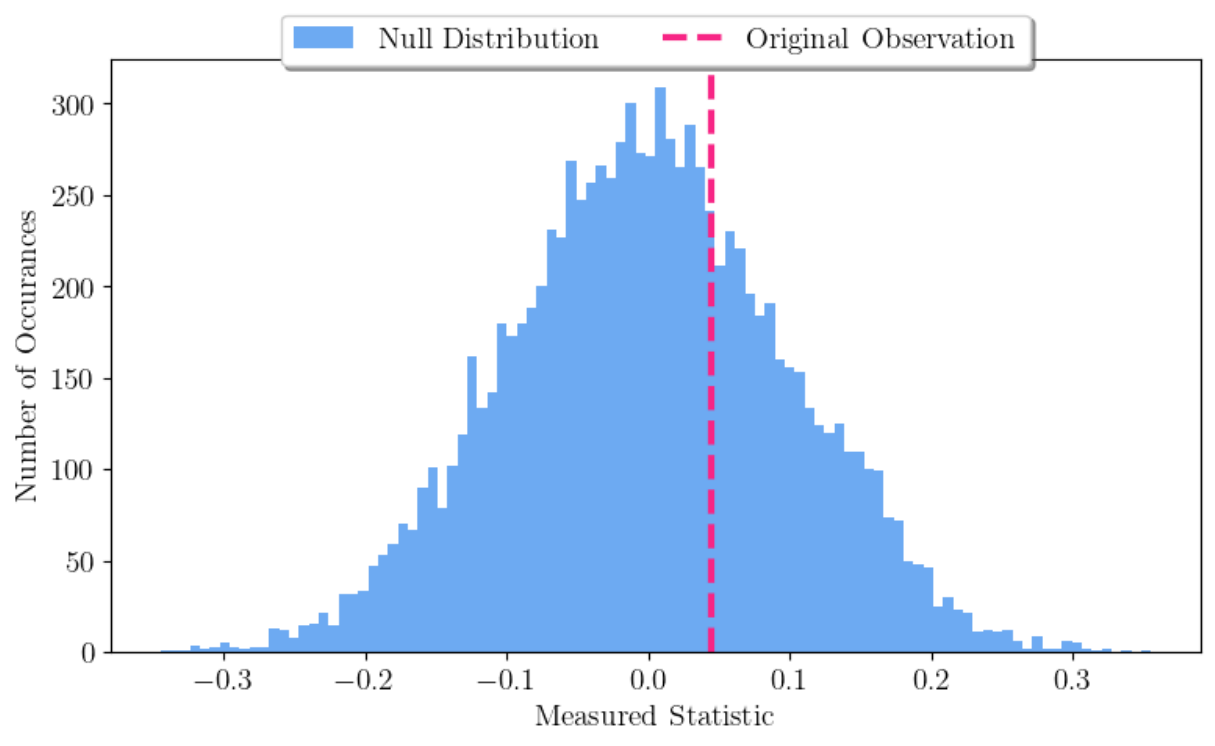}
        \caption{
            Pearson correlation coefficient.
        }
        \label{fig:noise-pcorr}
    \end{subfigure}
    \begin{subfigure}{0.32\textwidth}
        \includegraphics[width=\textwidth]{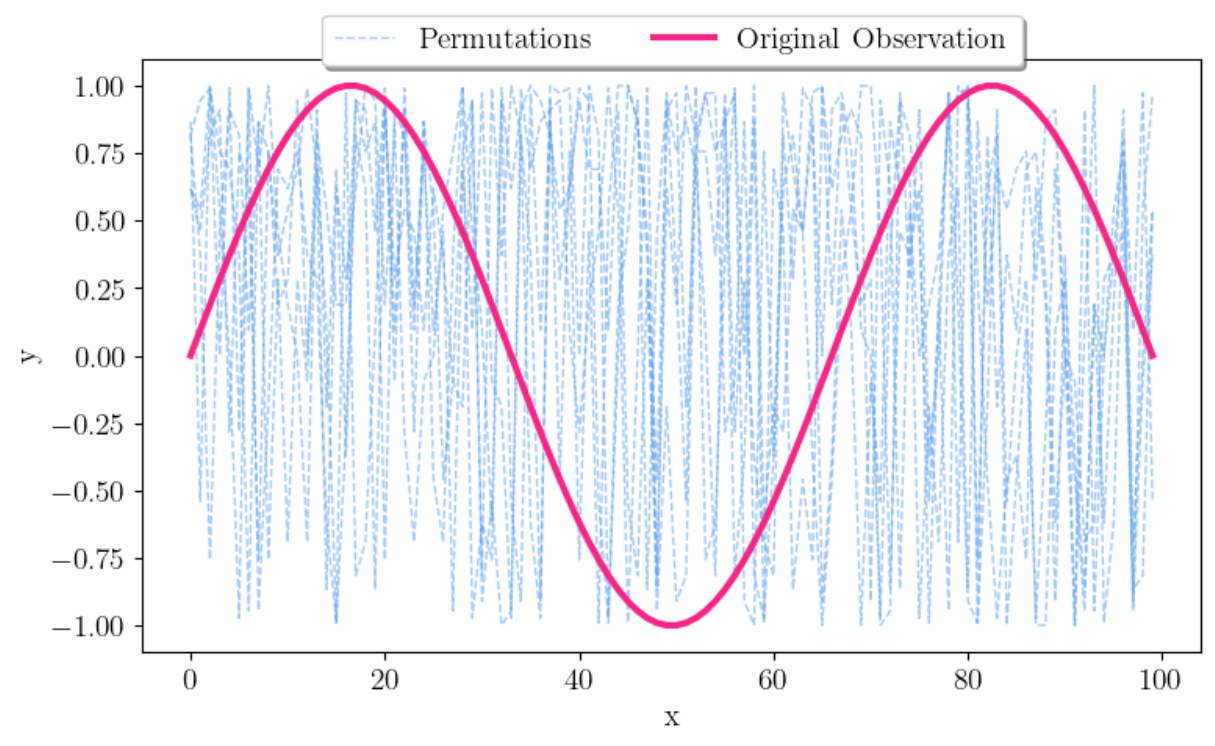}
        \caption{
            Sinus function.
        }
        \label{fig:sinus-shuffle}
    \end{subfigure}
    \begin{subfigure}{0.32\textwidth}
        \includegraphics[width=\textwidth]{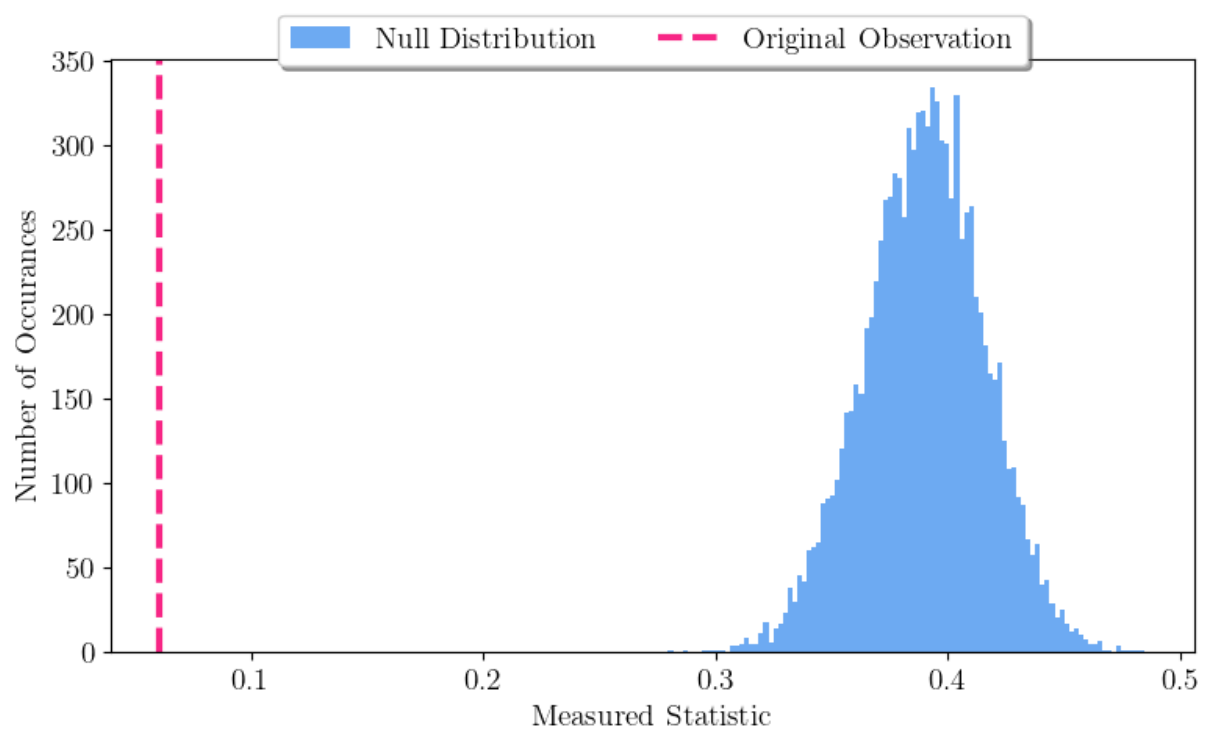}
        \caption{
            Expected $\nabla_\foi$ magnitude.
        }
        \label{fig:sinus-abs-grad}
    \end{subfigure}
    \begin{subfigure}{0.32\textwidth}
        \includegraphics[width=\textwidth]{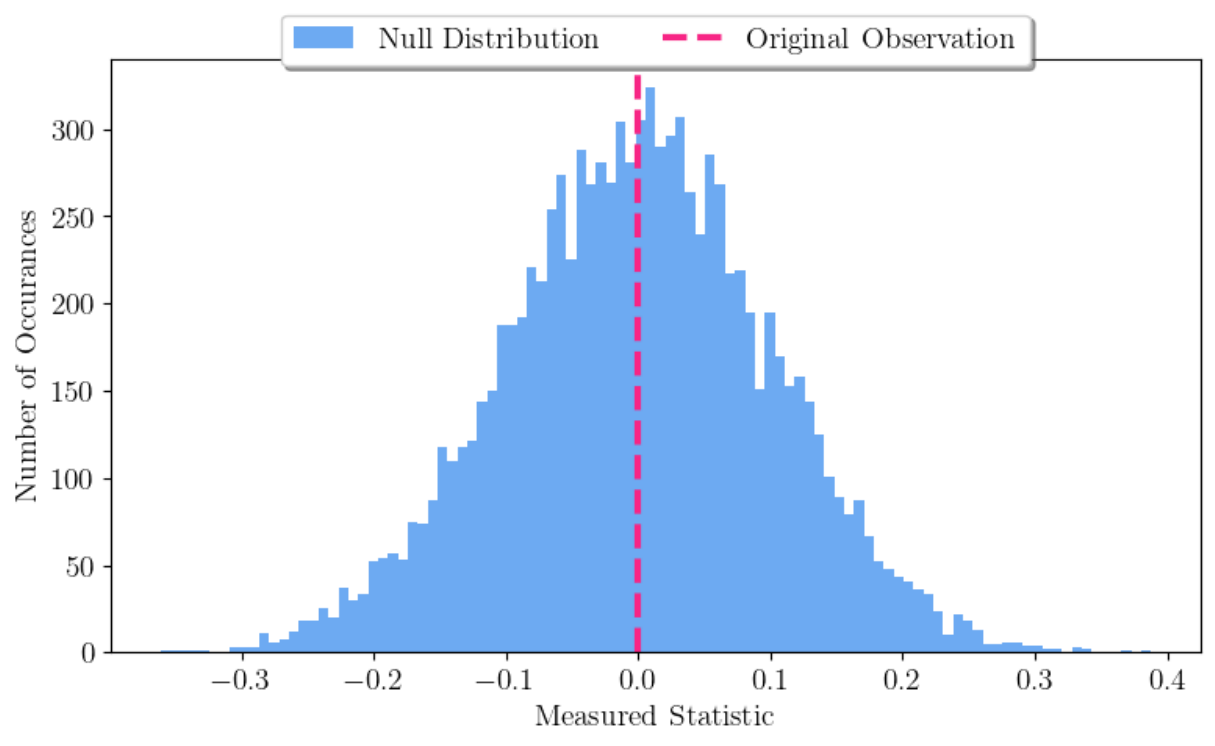}
        \caption{
            Pearson correlation coefficient.
        }
        \label{fig:sinus-pcorr}
    \end{subfigure}
    \begin{subfigure}{0.32\textwidth}
        \includegraphics[width=\textwidth]{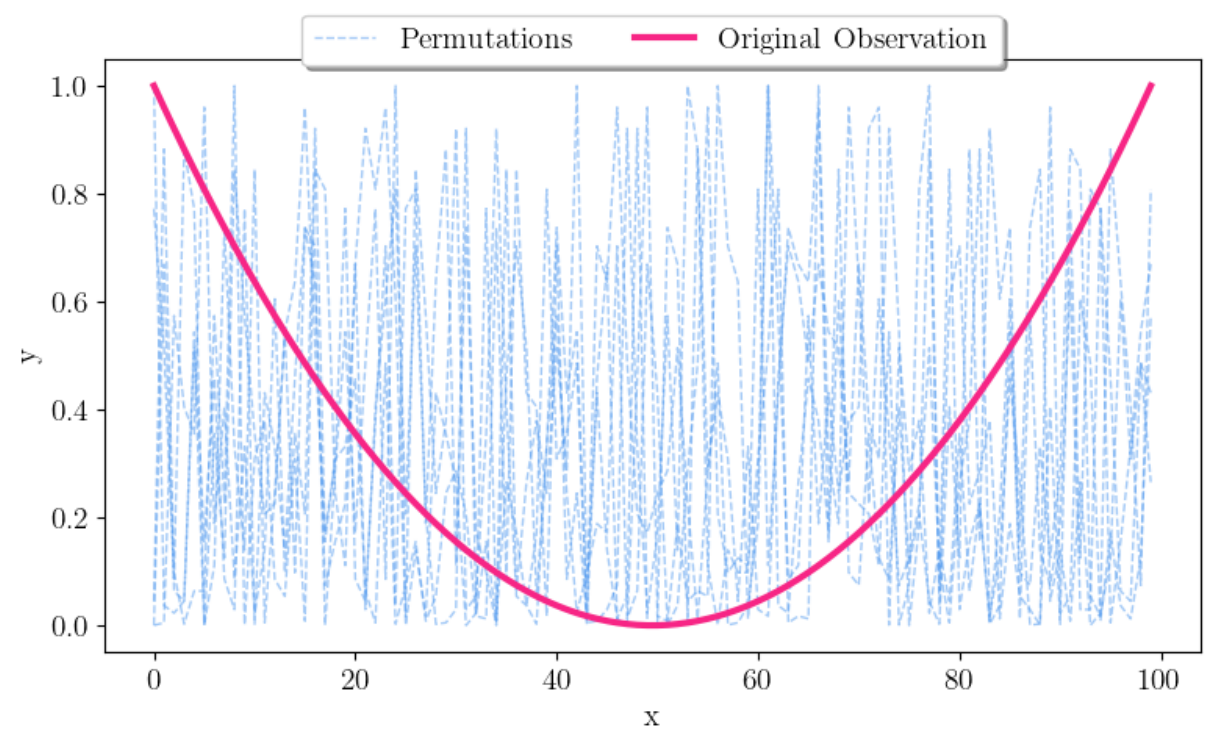}
        \caption{
            Parabola: $x^2$.
        }
        \label{fig:x2-shuffle}
    \end{subfigure}
    \begin{subfigure}{0.32\textwidth}
        \includegraphics[width=\textwidth]{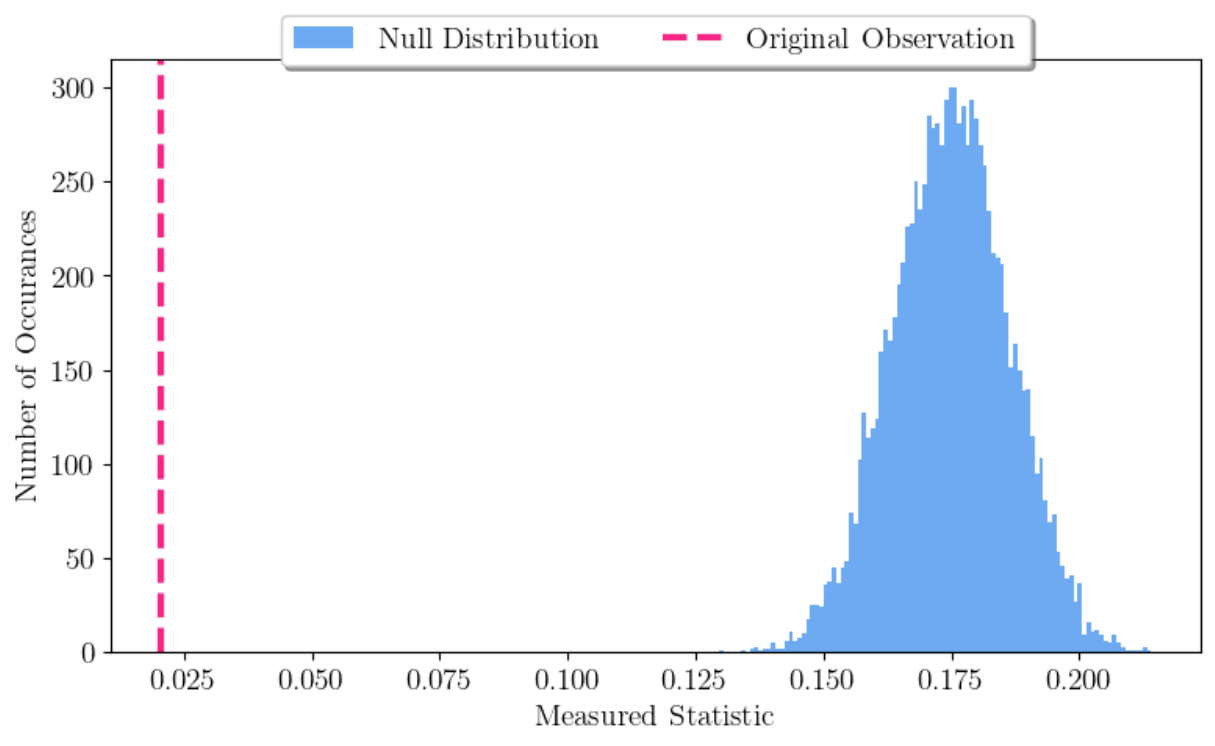}
        \caption{
            Expected $\nabla_\foi$ magnitude.
        }
        \label{fig:x2-abs-grad}
    \end{subfigure}
    \begin{subfigure}{0.32\textwidth}
        \includegraphics[width=\textwidth]{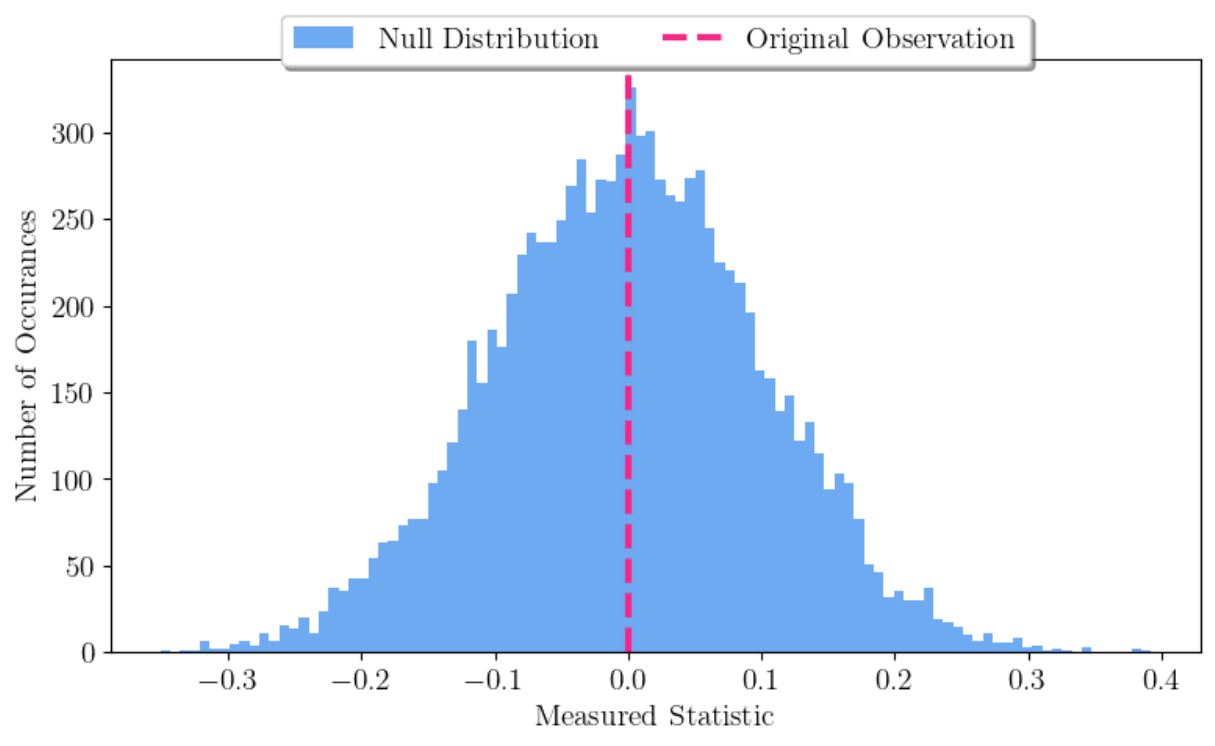}
        \caption{
            Pearson correlation coefficient.
        }
        \label{fig:x2-pcorr}
    \end{subfigure}
    \begin{subfigure}{0.32\textwidth}
        \includegraphics[width=\textwidth]{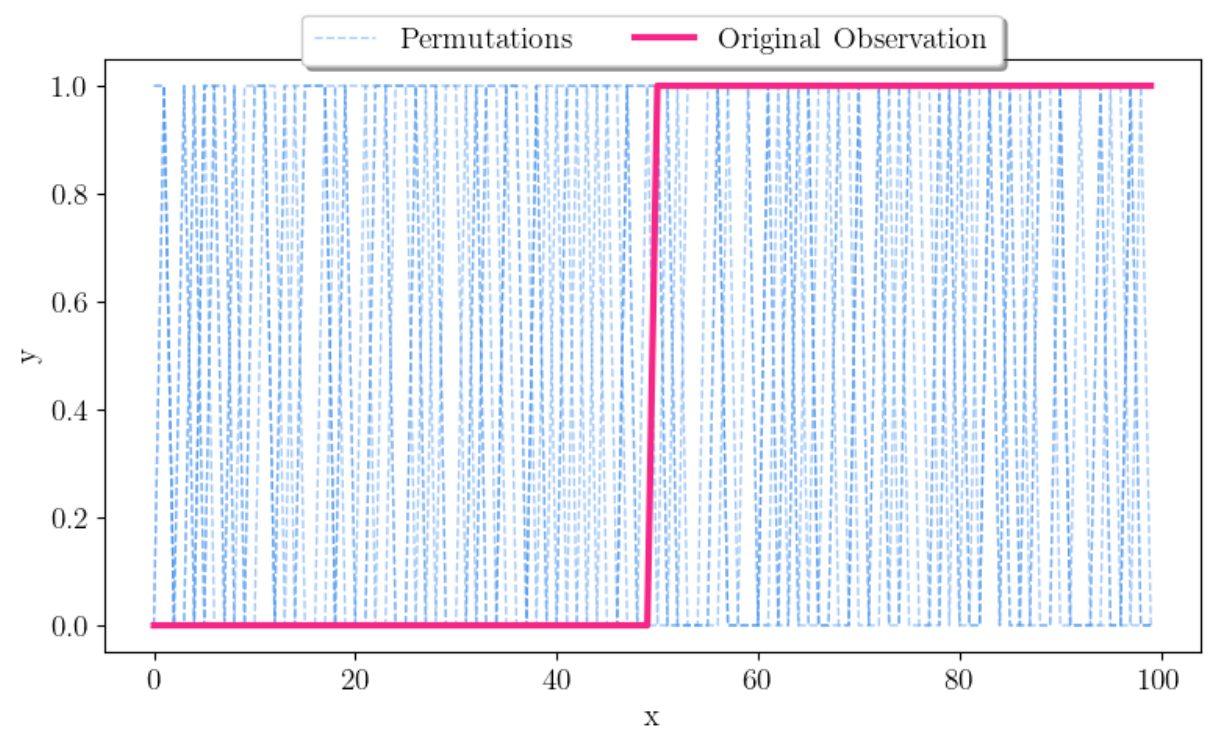}
        \caption{
            Step function.
        }
        \label{fig:step-shuffle}
    \end{subfigure}
    \begin{subfigure}{0.32\textwidth}
        \includegraphics[width=\textwidth]{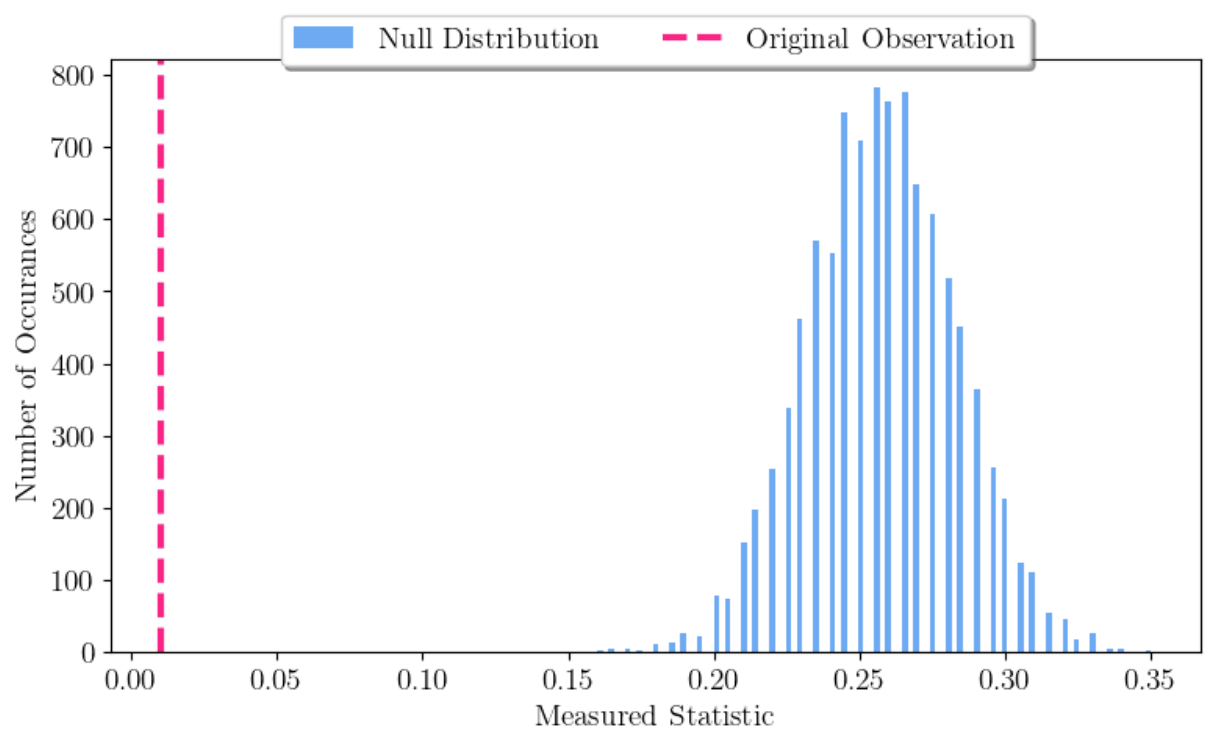}
        \caption{
            Expected $\nabla_\foi$ magnitude.
        }
        \label{fig:step-abs-grad}
    \end{subfigure}
    \begin{subfigure}{0.32\textwidth}
        \includegraphics[width=\textwidth]{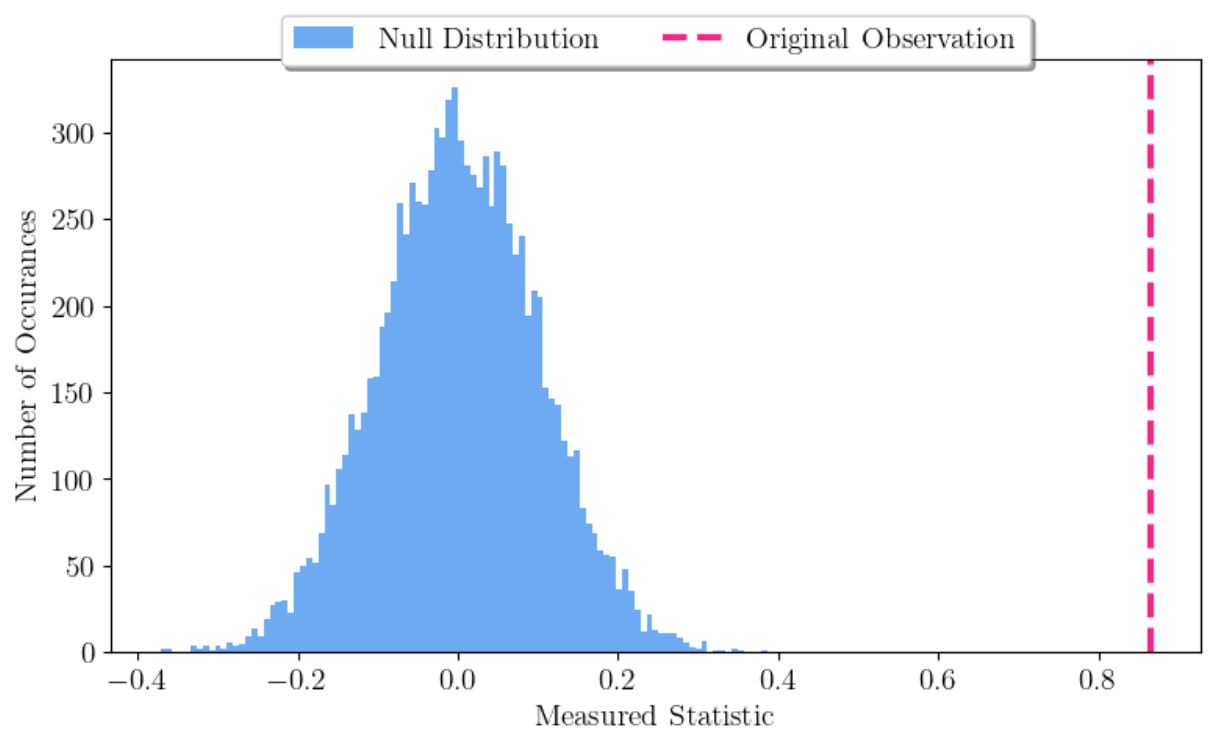}
        \caption{
            Pearson correlation coefficient.
        }
        \label{fig:step-pcorr}
    \end{subfigure}
    \begin{subfigure}{0.32\textwidth}
        \includegraphics[width=\textwidth]{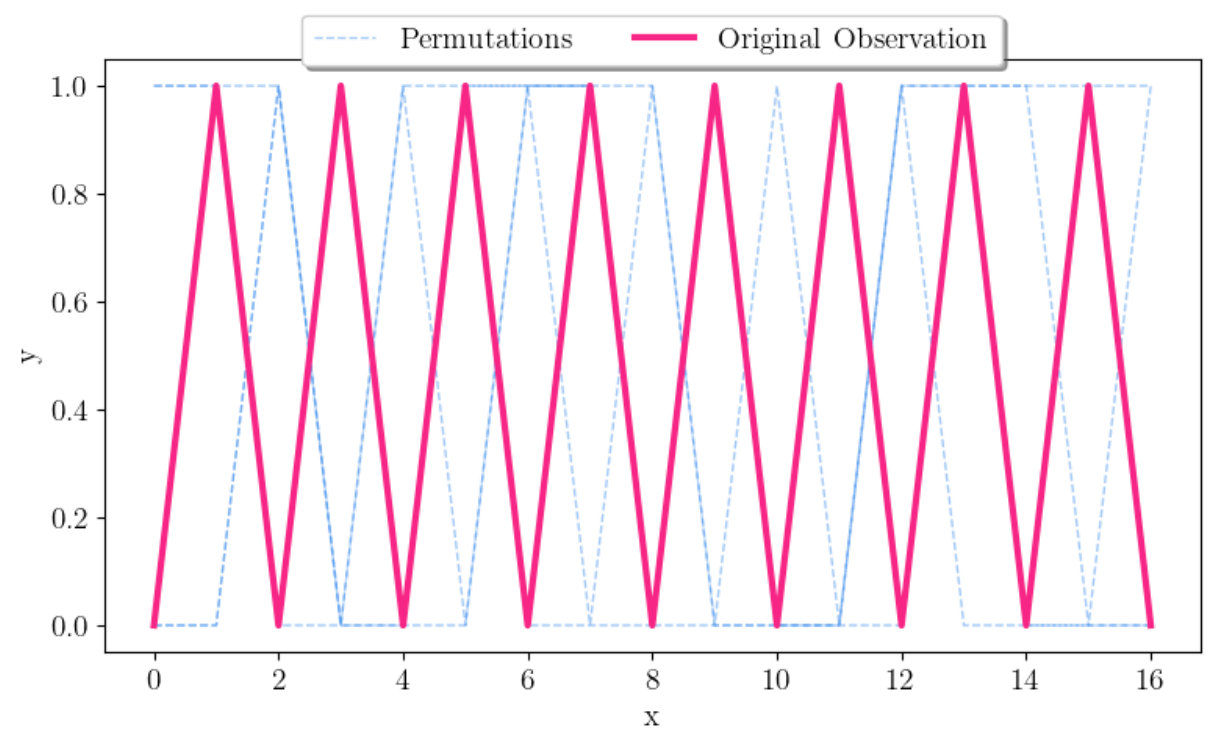}
        \caption{
            Zigzag function.
        }
        \label{fig:zigzag-shuffle}
    \end{subfigure}
    \begin{subfigure}{0.32\textwidth}
        \includegraphics[width=\textwidth]{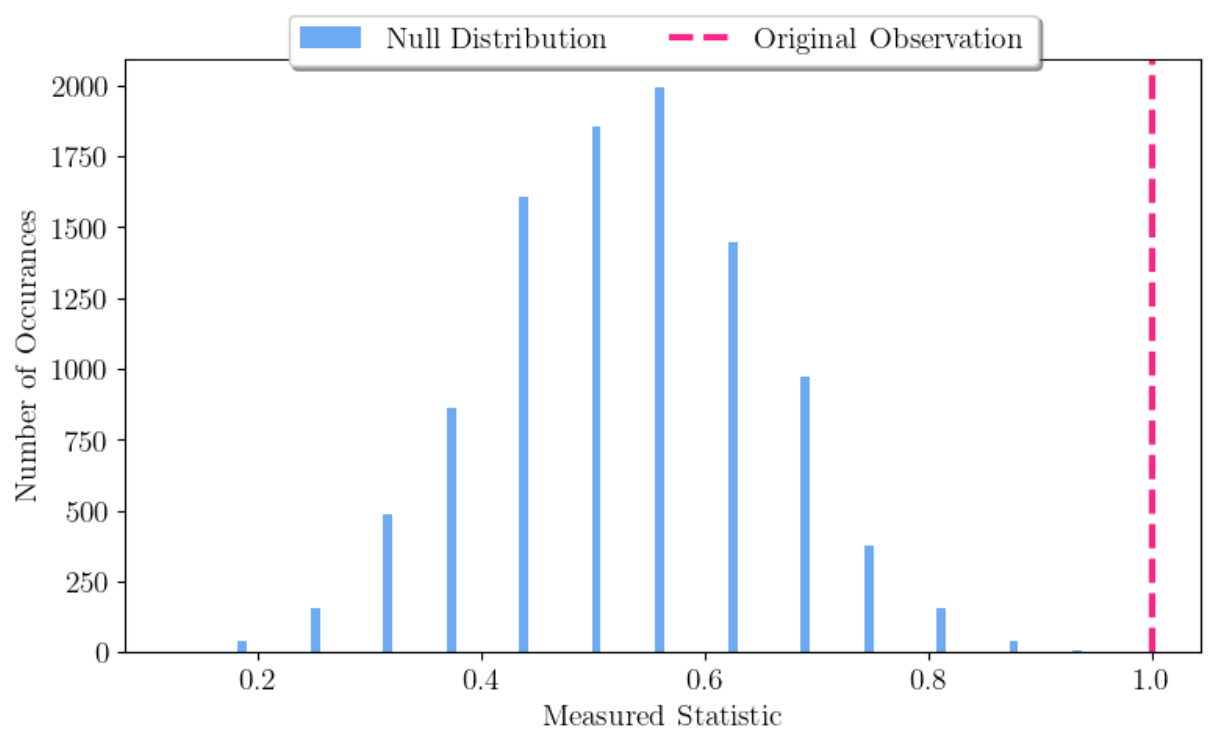}
        \caption{
            Expected $\nabla_\foi$ magnitude.
        }
        \label{fig:zigzag-abs-grad}
    \end{subfigure}
    \begin{subfigure}{0.32\textwidth}
        \includegraphics[width=\textwidth]{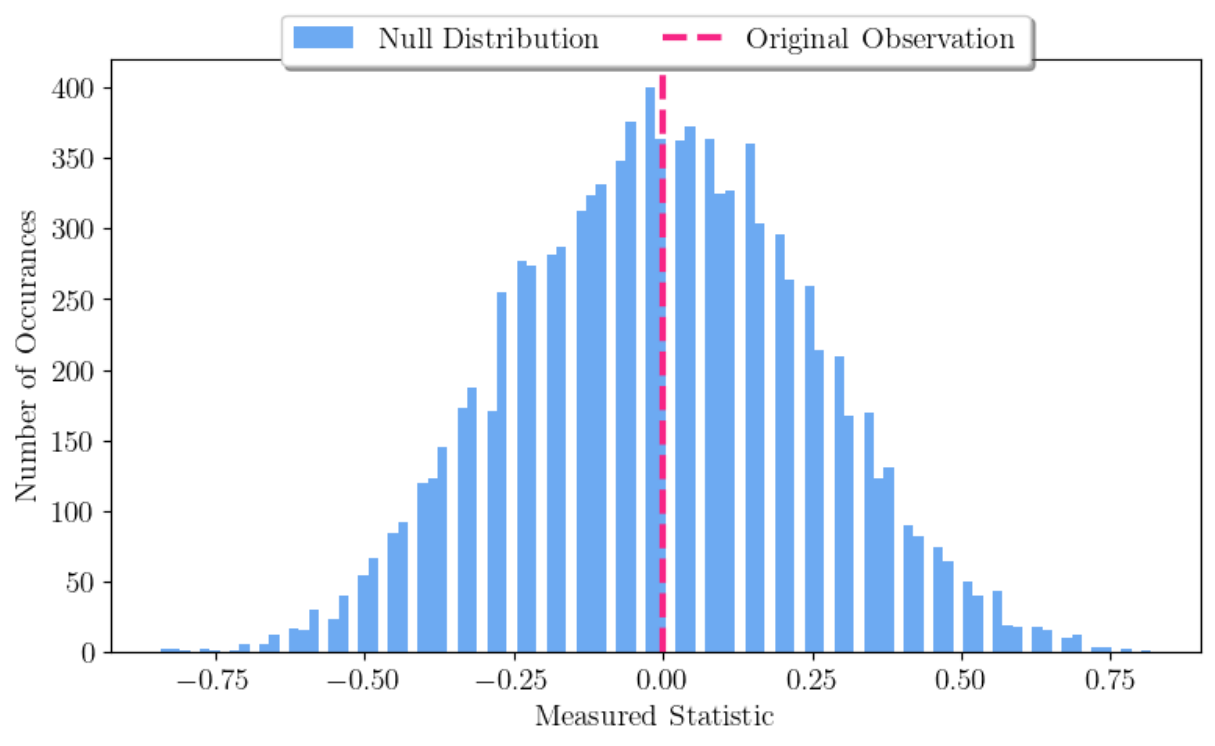}
        \caption{
            Pearson correlation coefficient.
        }
        \label{fig:zigzag-pcorr}
    \end{subfigure}
    \caption{
    Visualization of the null distributions generated by \Cref{alg:sig} for some example functions.
    The left-most column contains the observed functions versus five permuted instances.
    The middle column uses \Cref{eq:exp_grad} as a test statistic, while the right-most column employs the Pearson correlation coefficient \cite{pearson1895notes}.
    Note that in the last row, for the zigzag function, we use the forward difference quotient instead of the central difference quotient to estimate $\nabla_\foi$ \cite{fornberg1988generation}.
    }
    \label{fig:sig-expamples}
\end{figure*}

Our property gradients capture these nuanced effects (see \cref{fig:x2-abs-grad}). 
Furthermore, using visualizations similar to those in our experiments enables the identification of the desired property band, in this case, the optimal dosage range. 
While we use medicine as an example, similar behavior can occur for properties learned by neural networks. 
In fact, hair color is likely to exhibit a similar pattern in real life. 
Specifically, we note that gray hair is correlated with high age, whereas completely white or platinum blond hair is a common hair dye choice for younger individuals.

\subsection{Expected Property Gradient Magnitude Estimates \& Hypothesis Test}
\label{app:th-exp}

\begin{algorithm}
\caption{Hypothesis test for changes in prediction behavior for variations in a property $\foi$.}\label{alg:sig}
\begin{algorithmic}
\Require ordered list of predictions $\class_\cweights(I_\foi)$ \Comment{$N$ elements}
\Require test statistic $\statistic$ \Comment{for the outputs}
\Require integer $K > 0$ \Comment{Number of Permutations}
\Require $\siglevel \in (0,1)$ \Comment{Significance Level}
\State $p \gets 0.0$
\State $\sigma_{orig.} \gets \statistic(\class_\cweights(I_\foi))$ \Comment{Estimate the original statistic}
\For{\texttt{$i \in \{1, ..., K\}$}}
    \State $\class_\cweights(I^{(perm.)}_\foi) \gets \texttt{permute}(\class_\cweights(I_\foi))$
    \State $\sigma_{perm.} \gets \statistic(\class_\cweights( I^{(perm.)}_\foi))$
    \If{$\texttt{compare}(\sigma_{orig.},\sigma_{perm.})$}\\ 
    \Comment{Comparison depends on $\statistic$ ($<$, $>$, two-sided, etc.)}
        \State $p \gets p + \nicefrac{1}{K}$   \Comment{Increment the $p$-value}
    \EndIf
\EndFor

\If{$p < \delta$}
    \State \texttt{return} significant.
\Else
    \State \texttt{return} not significant.
\EndIf
\end{algorithmic}
\end{algorithm}

In our main paper, we propose estimating the expected gradient magnitude, as shown in \Cref{eq:exp_grad}. 
To achieve this, we employ \cite{fornberg1988generation} with a finite discrete list of sampled inputs using an interventional strategy, such as \cite{fu2024mgie}. 
To assess significance, we utilize the procedure outlined in \Cref{alg:sig}. 
In \cref{fig:sig-expamples}, we illustrate various example relationships, including independent random noise. 
Additionally, we visualize the estimated null distributions of \Cref{alg:sig} using both \Cref{eq:exp_grad} and Pearson's correlation coefficient \cite{pearson1895notes} as statistics. 
Notably, \Cref{eq:exp_grad} can detect periodic changes and changes with no linear trend while rejecting random noise, even with high effect strength.
We highlight two key observations: 
First, as shown in \cref{fig:zigzag-abs-grad}, our test statistic is two-sided. 
Second, we can investigate relationships where theoretically no gradient exists, i.e., $\nabla_\foi$ is infinite. 
Specifically, we consider step functions a relevant case, such as \cref{fig:step-shuffle}, where models exhibit categorical changes in behavior upon reaching a certain threshold. 
However, by approximating gradients for discrete observations using \cite{fornberg1988generation}, we can circumvent the issue of nonexisting gradients, enabling us to approximate \Cref{eq:exp_grad} even in such cases.

\section{Cats vs. Dogs - Additional Details}
\label{app:cvd}

In this section, we include additional details regarding our first experiment (\Cref{sec:exp1}).
First, we detail the creation of the biased training and test splits of the Cats versus Dogs (CvD) \cite{dogs-vs-cats} dataset.
Next, we present the training details and hyperparameters of the classifiers under analysis. 
Additionally, we describe the interventional data generation process and extended results that reinforce the claims made in our main paper.
Finally, we conduct a comparative analysis of multiple local and global XAI baselines in this scenario, highlighting the advantages of our local interventional approach in interpreting prediction behavior.

\subsection{Creating a Biased Scenario}
\label{app:cvd-data}

\begin{table}[t]
    \centering
    \caption{Number of samples in the different splits for the CvD \cite{dogs-vs-cats} dataset.
    With dark cat bias, we indicate dark-furred cats and light-furred dogs, while dark dog bias refers to the opposite.}
    \label{tab:cvd-data}
    \begin{tabular}{llcc}
    \toprule
    && Training & Test\\
        Split & Class & Samples & Samples \\
        \midrule
        \multirow{2}{*}{Unbiased}  
         & \texttt{cat} & 4000 & 1011 \\
         & \texttt{dog} & 4005 & 1012 \\
        \midrule
        \multirow{2}{*}{Dark cats bias}
         & \texttt{cat} & 1740 & 457 \\
         & \texttt{dog} & 2141 & 546 \\
         \midrule
        \multirow{2}{*}{Dark dogs bias}
         & \texttt{cat} & 2260 & 554 \\
         & \texttt{dog} & 1864 & 466 \\
    \bottomrule
    \end{tabular}
\end{table}

Our first experiment is based on a binary classification task between cats and dogs \cite{dogs-vs-cats}, where we intentionally introduce a correlation between the reference annotation and the fur color of the pictured animals. 
To create biased training and test splits, we leverage recent advances in multimodal models, specifically LLaVA 1.6 \cite{liu2024improved}. 
We prompt LLaVA with a yes/no question regarding the fur color of the reference annotation.
Specifically, we use ``Answer the question with yes or no: Does the \{reference annotation\} have dark fur?'' and sort the images into corresponding biased splits based on the response.
We assume that a "no" answer implies a light fur color, which is supported by our manual verification. 
The resulting sizes of the three training and test splits are summarized in \Cref{tab:cvd-data}. 
In the following section, we will detail the hyperparameter choices for the classification models.

\begin{figure}[t]
    \centering
    \begin{subfigure}{0.23\textwidth}
        \includegraphics[width=\linewidth]{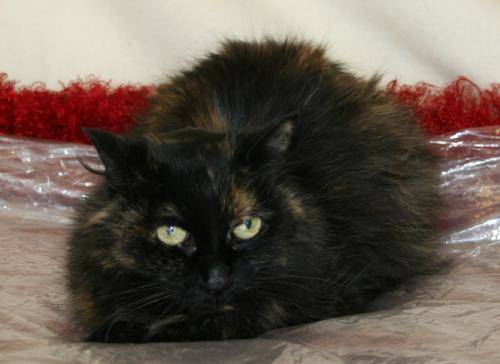}
    \end{subfigure}
    \begin{subfigure}{0.23\textwidth}
        \includegraphics[width=\linewidth]{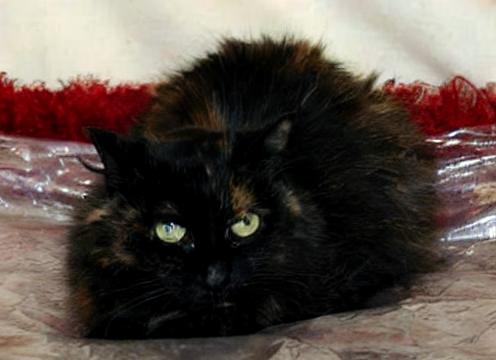}
    \end{subfigure}
    \begin{subfigure}{0.23\textwidth}
        \includegraphics[width=\linewidth]{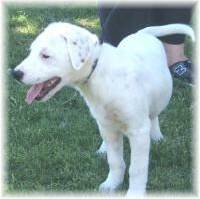}
    \end{subfigure}
    \begin{subfigure}{0.23\textwidth}
        \includegraphics[width=\linewidth]{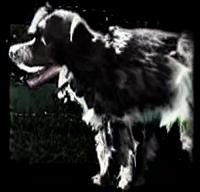}
    \end{subfigure}
    
    \caption{Two failure cases we observed for the background interventions.
    In both, we tried various settings. 
    Here, we report them using a 2.9 image guiding scale and 13.50 and 8.85 CFG text scale for the cat and dog, respectively.
    The left-hand side in both rows shows the original input image.}
    \label{fig:bg-fails}
\end{figure}

\subsection{Training Details}
\label{app:cvd-setup}

For the first experiment, we select the ConvMixer architecture \cite{trockman2022patches}, a simple yet effective convolution-only model class.
Specifically, our ConvMixer configuration consists of an initial patch size of 5, a depth of 8, kernels with a width of 7, and a latent representation size of 256. 
For a detailed explanation of these hyperparameters, we refer the reader to the original paper \cite{trockman2022patches}.

During both training and inference, we preprocess the images by resizing them to an input size of $128\times 128$ and normalizing the pixel values to the interval $[-1,1]$. 
Furthermore, we apply the TrivialAugment data augmentation technique \cite{muller2021trivialaugment} with the wide augmentation space during training to enhance model robustness.
We optimize the models using AdamW \cite{loshchilov2019decoupled}, setting the learning rate to 0.001, weight decay to 0.0005, and momentum to 0.9.
After training for 100 epochs with batch size 64, we save the final model weights, which achieve the performances disclosed in \Cref{tab:cvd-accs}.

\begin{table}[t]
\centering
\caption{
Final test set accuracies in percent (\%) achieved by our models trained to differentiate cats and dogs.
Here, the columns signify the test data, and the rows denote the training data split.}
\label{tab:cvd-accs}
\begin{tabular}{lccc}
\toprule
 &  & dark cats & dark dogs\\
 & unbiased & split & split \\
\midrule
Unbiased & 90.71 & \bfseries 92.52 & 89.22 \\
Dark Cats split& 69.40 & \bfseries  93.92 & 45.78 \\
Dark Dogs split& 70.93 & 48.45 & \bfseries 92.75 \\
\bottomrule
\end{tabular}
\end{table}

\begin{figure*}[tb]
    \centering
    \includegraphics[width=\linewidth]{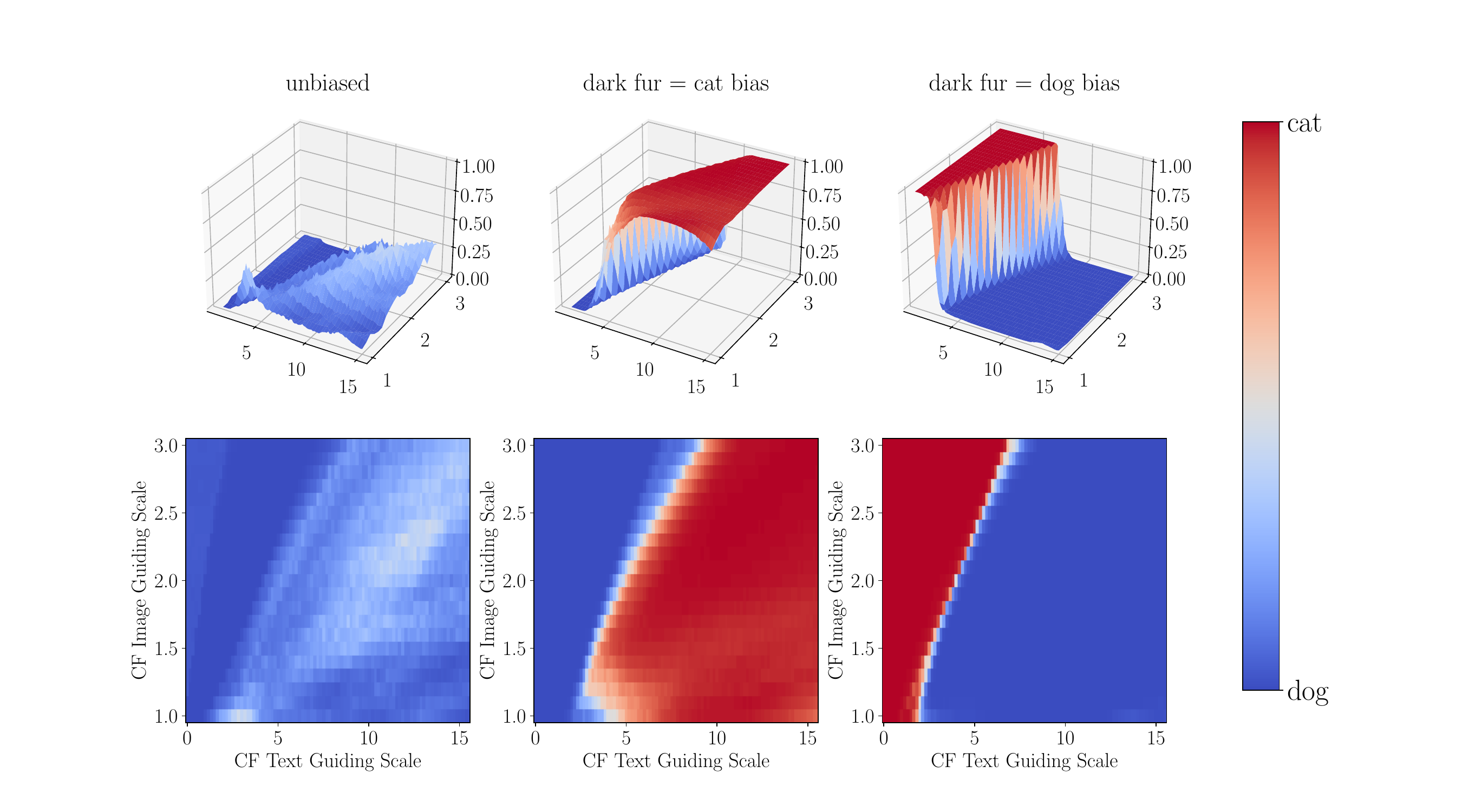}
    \caption{
    Changes in model predictions for an intervention on the fur color.
    Here, we display one ConvMixer model per column, specifically an unbiased one, one trained on only dark-furred dogs and light-furred cats (``dark fur = dog''), and one trained only on the opposite split (``dark fur = cat'').
    The x and y axes indicate the CF guiding scales for text and images, respectively.
    The white color (logit = 0.5) indicates the decision threshold between the two classes.
    }
    \label{fig:fur-results2d}
\end{figure*}

\subsection{Additional Results}
\label{app:cvd-results}

\paragraph{Fur Color Intervention - 2D Results}

To perform the interventions on the fur color, we utilize a pre-trained version of \cite{fu2024mgie}, an image-to-image edit model that can be controlled using text prompts.
This model utilizes a multimodal large language model (LLaVA \cite{liu2024improved}) together with an adapter network to provide expressive and focused edit instructions for instruct pix2pix \cite{brooks2022instructpix2pix}.
The authors find that it performs especially well for local edits, meaning color changes, for example, do not change global illumination.
Further, it includes two hyperparameters to control the alignment with the provided text instruction.
Both are implemented and trained as classifier-free guidance (CFG) scales \cite{ho2021classifier}.
The first one, which we call the CFG text scale, generally controls the alignment with the instruction, while the CFG image scale controls the similarity with the input image.
We provide more details in \Cref{app:cfg-img}.

In our main paper (\cref{fig:1d-results}), we utilize a CFG image scale of 2.0 and interpolate the text scale between $[1.05, 14.7]$ using a stepsize of $0.15$.
We use ``change the fur color to black'' as our instruction.
As an ablation, we visualize the results for the fur color intervention again in \cref{fig:fur-results2d} for both guiding scales, varying the image scale between 1.0 and 3.0, following \cite{fu2024mgie}.
Notably, the CFG image scale primarily controls the point at which the prediction flip occurs. Specifically, we observe that the order of the predictions remains unchanged. 
The unbiased model does not exhibit a prediction flip but instead increases the activation for the cat class logit as the text guiding scale increases. 
In contrast, the model trained on dark dogs showcases the most abrupt change, while the dark cat model transitions slightly more gradually.

In summary, our results indicate that the actual intervention is controlled by the alignment with the image edit instruction. 
Therefore, in our subsequent experiments, we focus on the CFG text scale while selecting a suitable CFG image scale.

\begin{figure*}[t]
\begin{subfigure}{\textwidth}
    \includegraphics[width=\linewidth, trim=0.8cm 0.6cm 0.5cm 1.5cm, clip]{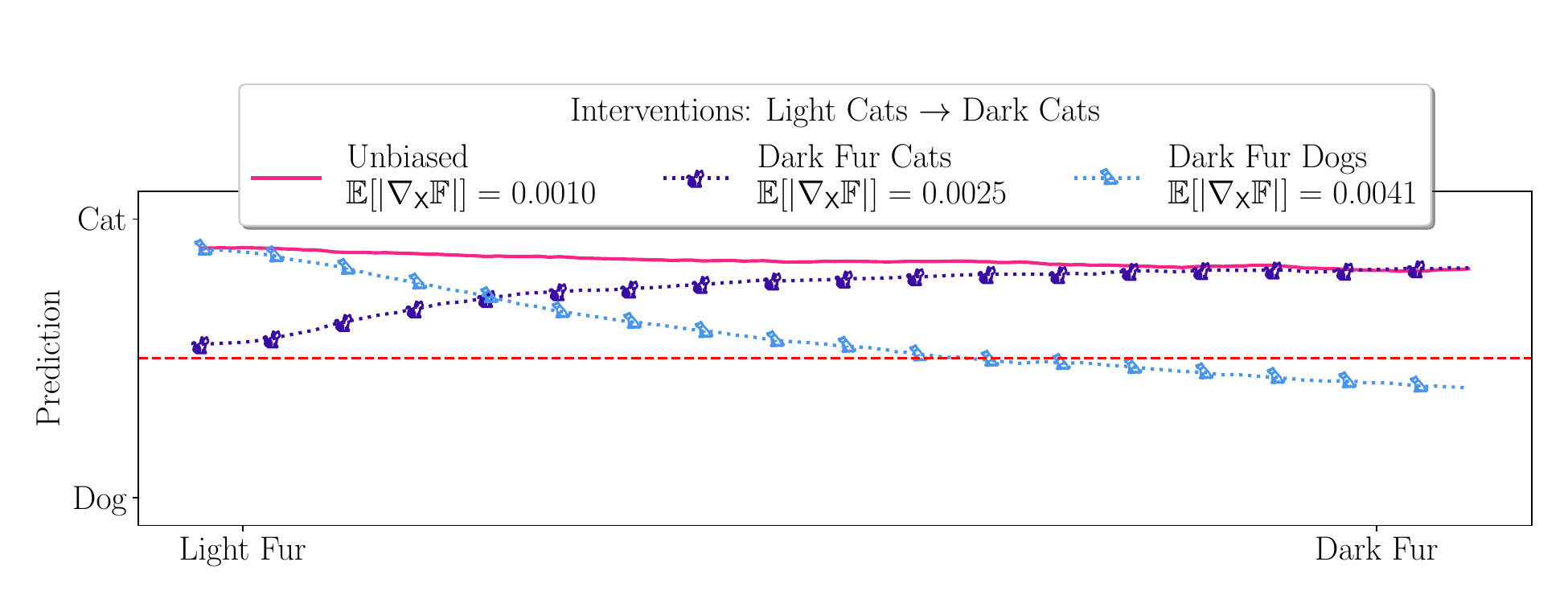}
    \caption{Light furred cats $\rightarrow$ Dark furred cats.}
    \label{fig:fur-mean-lcats-dcats}
\end{subfigure}
\begin{subfigure}{\textwidth}
    \includegraphics[width=\linewidth, trim=0.8cm 0.6cm 0.5cm 1.5cm, clip]{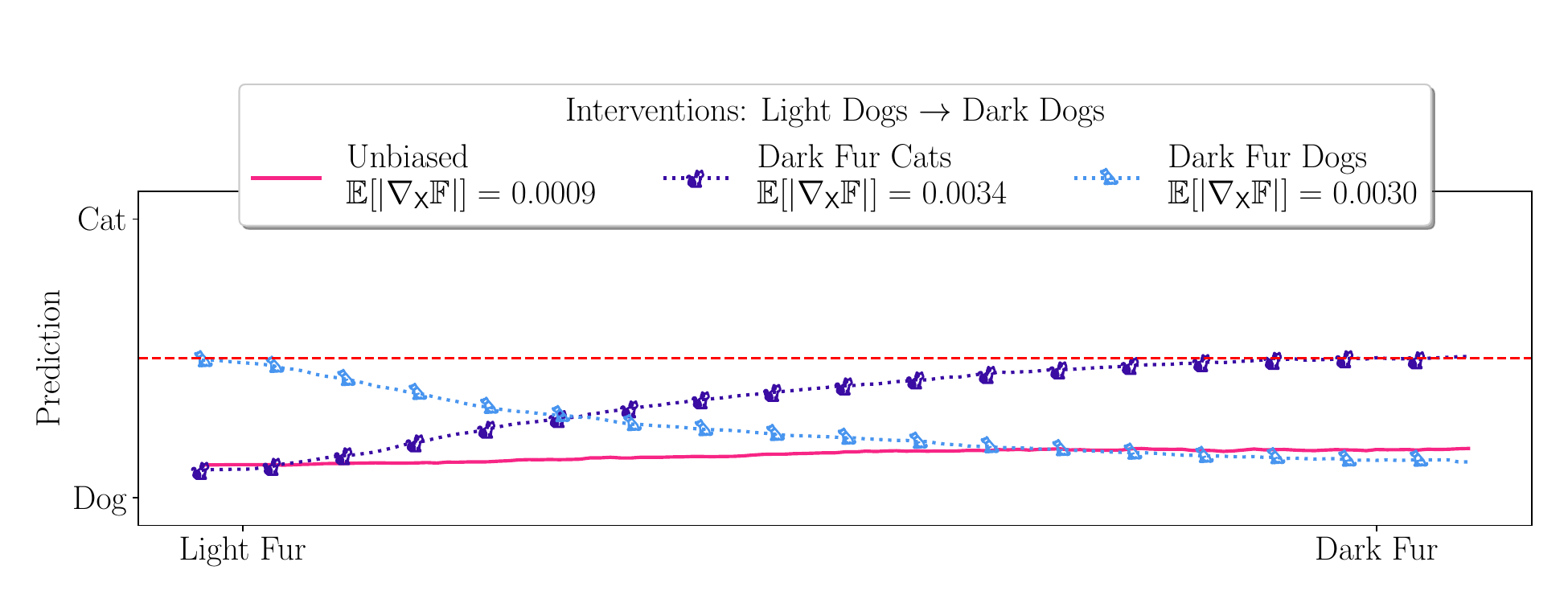}
    \caption{Light furred dogs $\rightarrow$ Dark furred dogs.}
    \label{fig:fur-mean-lcats-dcats}
\end{subfigure}
\begin{subfigure}{\textwidth}
    \includegraphics[width=\linewidth, trim=0.8cm 0.6cm 0.5cm 1.5cm, clip]{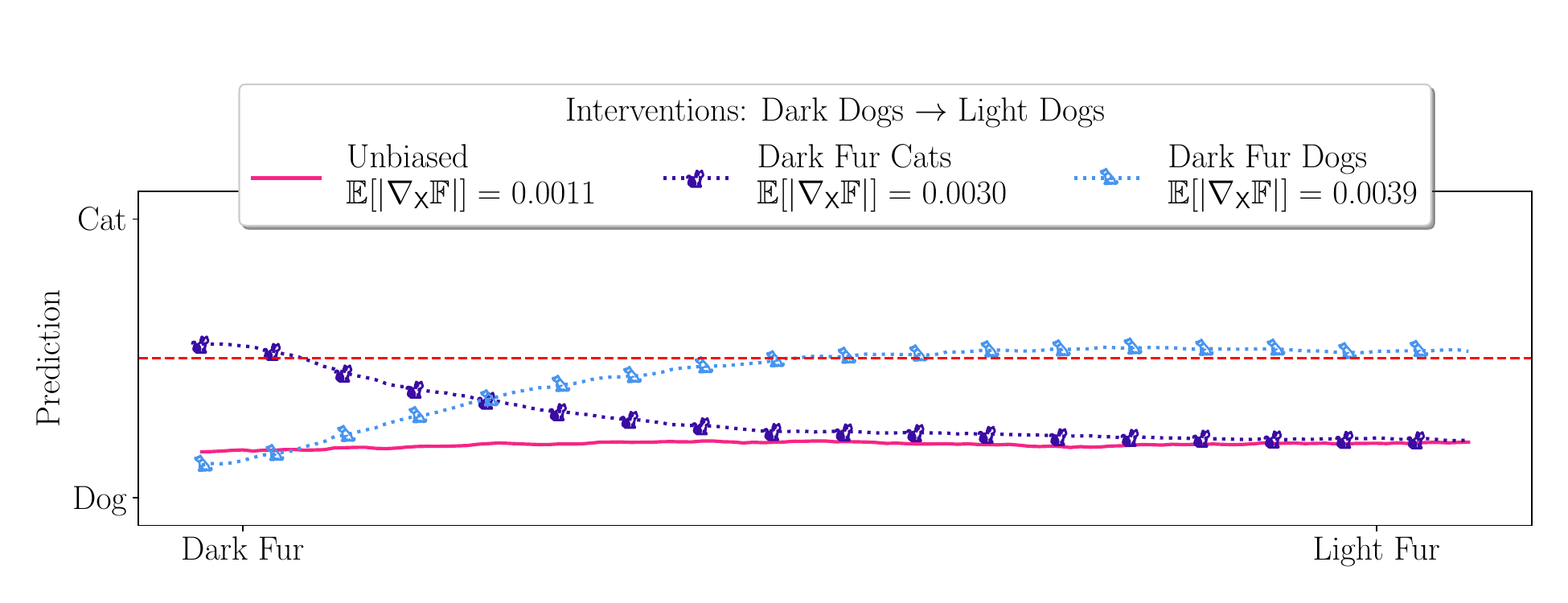}
    \caption{Dark furred dogs $\rightarrow$ light furred dogs.}
    \label{fig:fur-mean-lcats-dcats}
\end{subfigure}

\caption{Average model output behavior for the property label combinations not contained in the main paper (compare to \cref{fig:fur-mean-results}).
Here, we use three ConvMixer models: an unbiased one, one trained on only dark-furred dogs and light-furred cats (``Dark Fur Dogs''), and one trained only on the opposite split (``Dark Fur Cats'').
The \textcolor{red}{red dotted line} indicates the threshold where the model prediction flips.
We include the average \propgrad per model in the legend.
In all cases, we observe the lowest fur color impact for the unbiased model, while the models trained on the biased splits showcase the expected changes in prediction behavior.
We provide \textbf{ten} local explanations for light-furred dogs and dark-furred cats in \cref{fig:fur-schar-results}
}
\label{fig:fur-dogs-mean}
\end{figure*}

\begin{figure*}[t]
    \centering
    \begin{subfigure}{0.48\textwidth}
        \includegraphics[width=\linewidth]{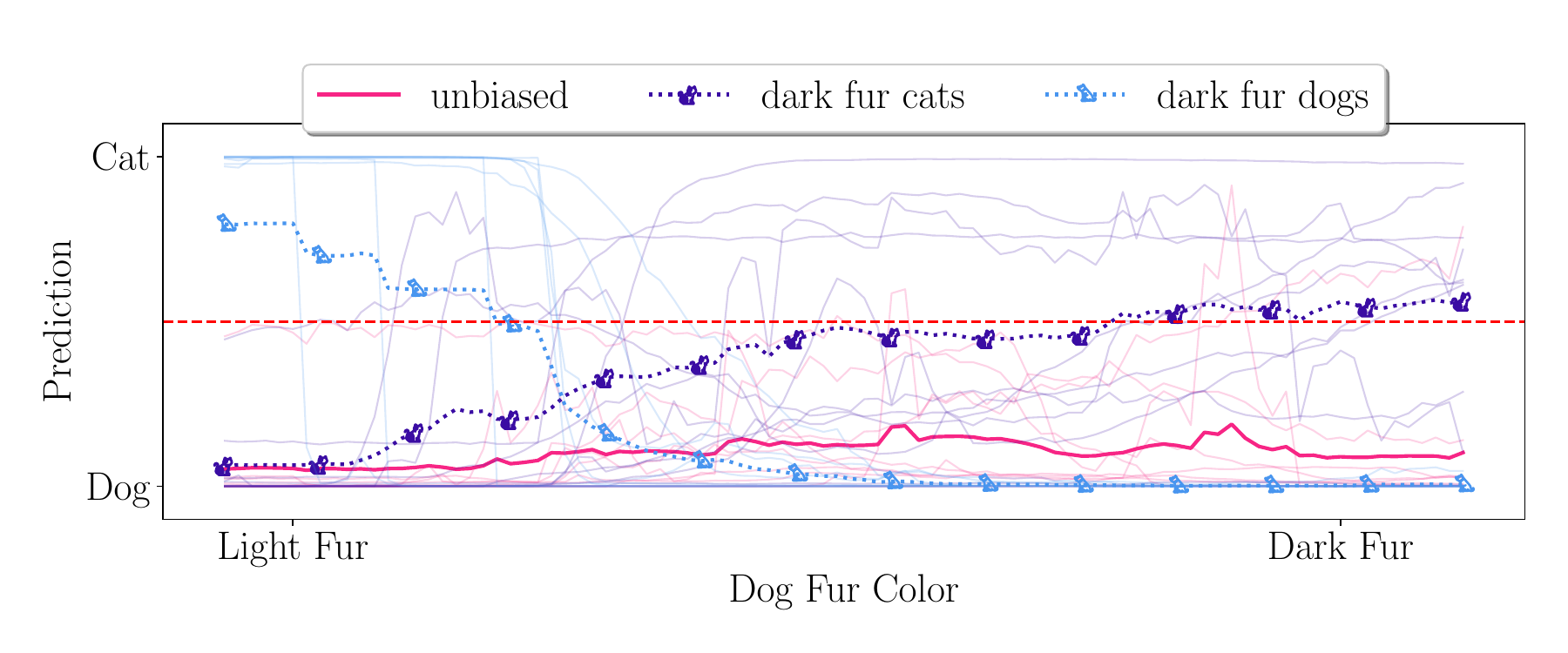}
        \caption{Ten \textbf{dog} images with fur color interventions.}
        \label{fig:fur-dogs-schar}
    \end{subfigure}
    \begin{subfigure}{0.48\textwidth}
        \includegraphics[width=\linewidth]{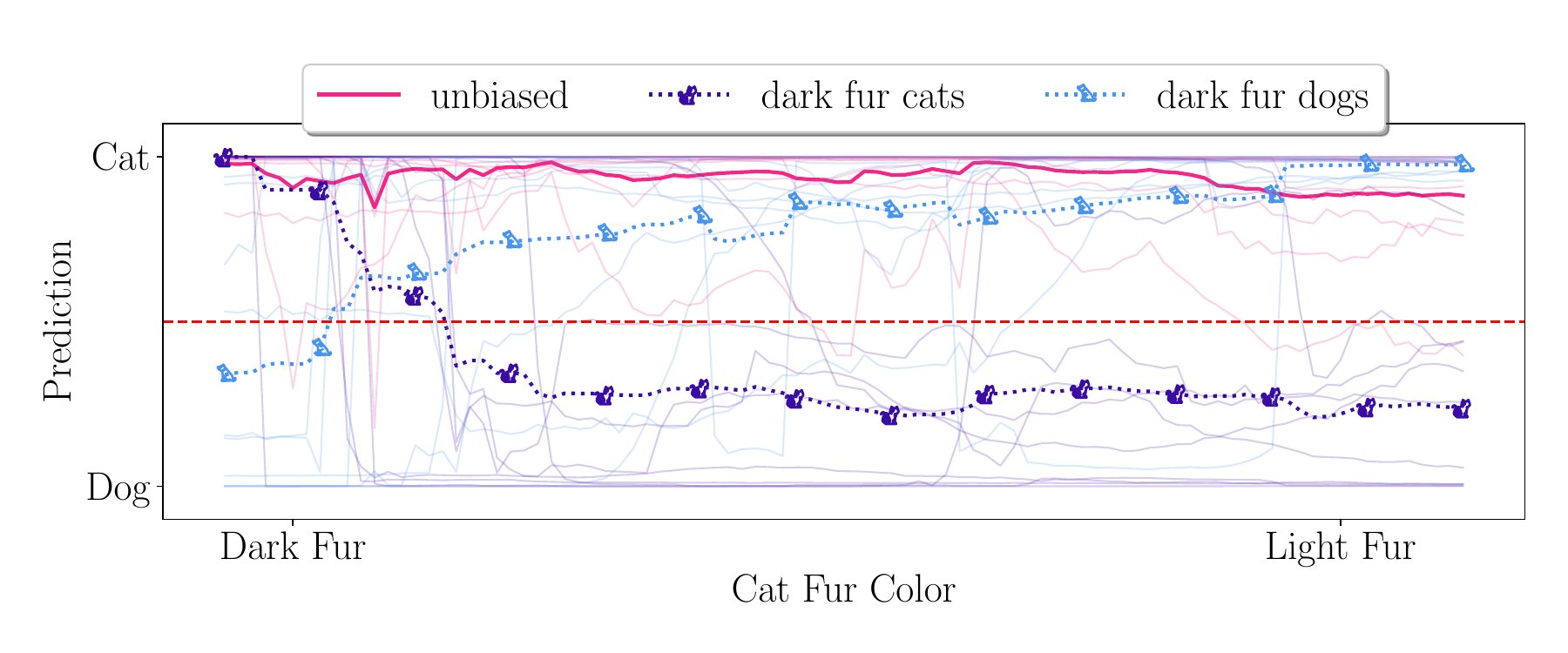}
        \caption{Ten \textbf{cat} images with fur color interventions.}
        \label{fig:fur-cats-schar}
    \end{subfigure}
    
    \caption{
    Changes in model predictions for an intervention on the fur color for ten respective images.
    Here, we use three ConvMixer models: an unbiased one, one trained on only dark-furred dogs and light-furred cats (``dark fur = dog''), and one trained only on the opposite split (``dark fur = cat'').
    The \textcolor{red}{red dotted line} indicates the threshold where the model prediction flips.
    }
    \label{fig:fur-schar-results}
\end{figure*}

\paragraph{Additional Fur Color Interventions}

\begin{figure*}
    \centering
    \begin{subfigure}{\textwidth}
    \includegraphics[width=\textwidth]{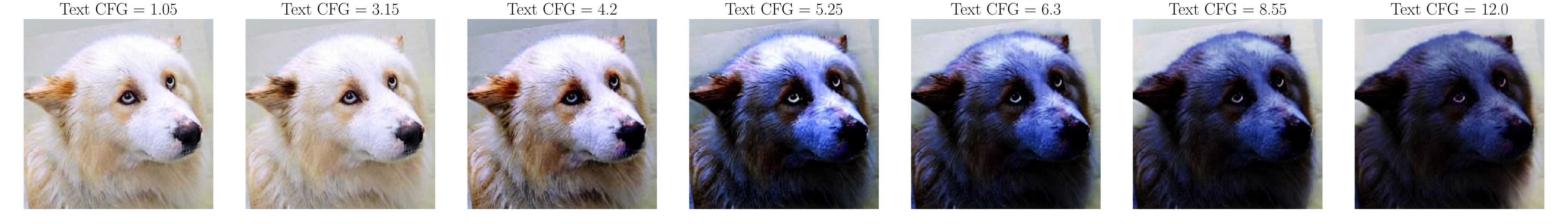}
    \end{subfigure}
    \begin{subfigure}{\textwidth}
    \includegraphics[width=\textwidth]{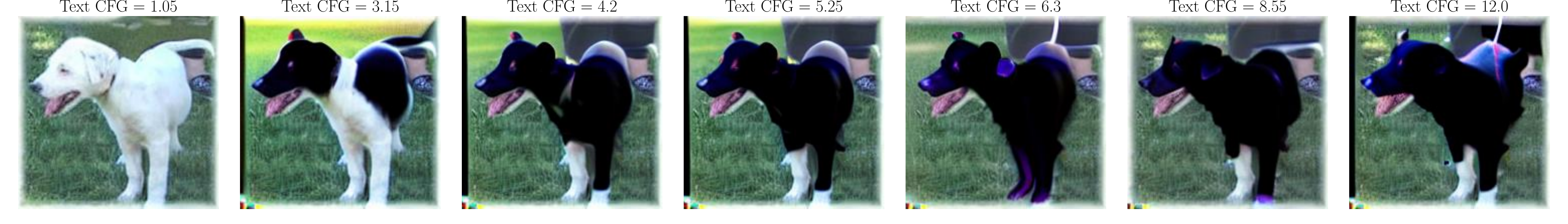}
    \caption{Two additional example interventions for dogs with light fur.}
    \end{subfigure}
    \begin{subfigure}{\textwidth}
    \includegraphics[width=\textwidth]{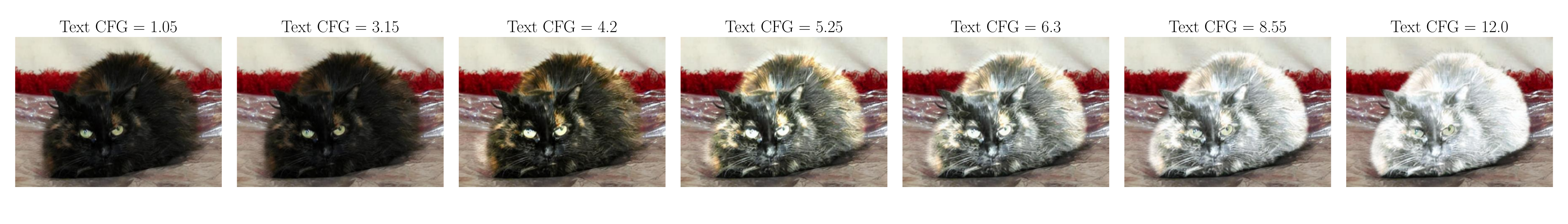}
    \end{subfigure}
    \begin{subfigure}{\textwidth}
    \includegraphics[width=\textwidth]{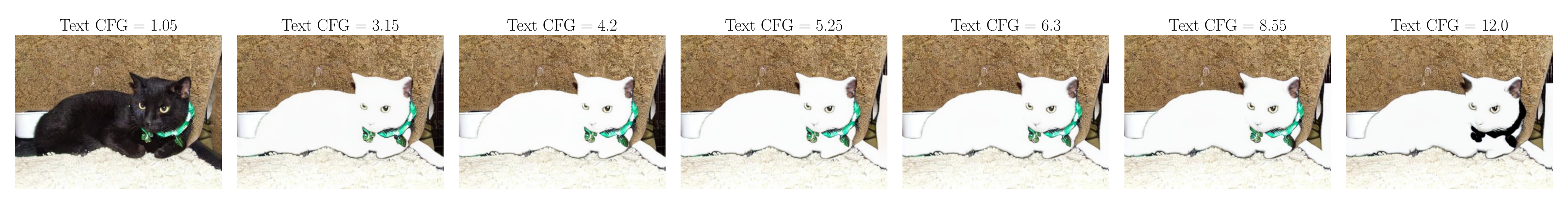}
    \caption{Two example interventions for cats with dark fur.}
    \end{subfigure}
    \caption{
    Interventions on the fur color for both cats and dogs. 
    We display two additional examples per class.
    Note the difference in the onset for both examples per animal, given that we utilize the same hyperparameters and prompt for \cite{fu2024mgie}.
    }
    \label{fig:add-interv}
\end{figure*}

In our main paper, we present the average model predictions for all dark-furred cat images in the test dataset (see \cref{fig:fur-mean-results}).
Here we provide the remaining splits.
To generate these images, we select the dog and cat images according to their initial fur color as described in \cref{sec:exp1}.
We use the following editing prompts: "change the fur color to dark black" for images of light-furred animals and "change the fur color to bright white" for images of dark-furred animals. 
Specifically, we employ an image guiding scale of $2.8$ and vary the text guiding scale between $1.05$ and $14.7$.
We provide examples of these edited images in \cref{fig:add-interv}.
In all cases, we observe the lowest fur color impact for the unbiased model, while the models trained on the biased splits showcase the expected changes in prediction behavior.

Further, in \cref{fig:fur-schar-results}, we zoom in and visualize not only the mean but also the local behavior for ten selected images, where we verified correct interventions.
While the overall trend remains consistent, indicating that the biased models flip their predictions on average, we occasionally observe a weaker relationship. 
We report the concrete measured \propgrad values for these ten images in \Cref{tab:cvd-impact-mean}. 
Notably, here, we find generally higher \propgrad scores compared to the average over all images, indicating that manually verified interventions benefit our measurements.
Nevertheless, the ordering between the unbiased and the biased models persists, and we consistently measure a higher \propgrad for the two biased models.
We find a stronger decrease in the intervention effect for light-furred dogs compared to dark-furred cats. 
Furthermore, the unbiased model (see \cref{fig:fur-mean-results}) is the only model in our extended analysis where the averaged prediction never flips.

\begin{table}[t]
    \centering
    \caption{\propgrad of the fur color property for our three CvD models.
    Here, we utilize the mean behavior ( \cref{fig:fur-mean-results}) over ten images.
    Additionally, we report significance ($p<0.01$) and prediction flips.
    For examples of the interventional data, see \cref{fig:add-interv}.
    }
    \label{tab:cvd-impact-mean}
    \begin{tabular}{llccc}
    \toprule
        & & \multicolumn{3}{c}{Model Behavior}\\
        \cmidrule(lr){3-5}
      Data  & Model & \propgrad & $p<0.01$ & Pred. Flips \\
    \midrule
    \multirow{3}{*}{\shortstack[l]{Light \\Dogs}} 
        & unbiased  & 0.00598 & \cmark & \xmark \\
        & dark cats & 0.00863 & \cmark & \cmark \\
        & dark dogs & 0.00888 & \cmark & \cmark \\
    \midrule
    \multirow{3}{*}{\shortstack[l]{Dark \\Cats}} 
        & unbiased  & 0.00638 & \cmark & \xmark \\
        & dark cats & 0.01207 & \cmark & \cmark \\
        & dark dogs & 0.01084 & \cmark & \cmark \\
    \bottomrule
    \end{tabular}
\end{table}

Finally, the discrepancies between the \propgrad values in the averaged case and the individual local case can be attributed to the varying onsets of the behavior changes. 
This phenomenon is also reflected in the variations observed in the interventions (\cref{fig:add-interv}). 
Specifically, as shown in \cref{fig:fur-dogs-schar}, the local flips occur at different CFG scales, highlighting the limitations of the behavior visualization.
This observation underscores the non-linear mapping learned by \cite{fu2024mgie} to intervene on the original image.
Further, this mapping locally varies, which supports our claims made in the main paper.

\paragraph{Additional Background Interventions}

\begin{figure*}
    \centering
    \begin{subfigure}{\textwidth}
    \includegraphics[width=\textwidth]{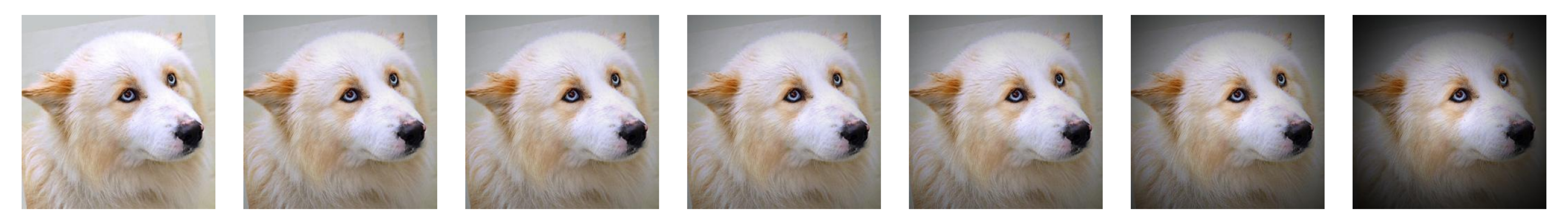}
    \end{subfigure}
    \begin{subfigure}{\textwidth}
    \includegraphics[width=\textwidth]{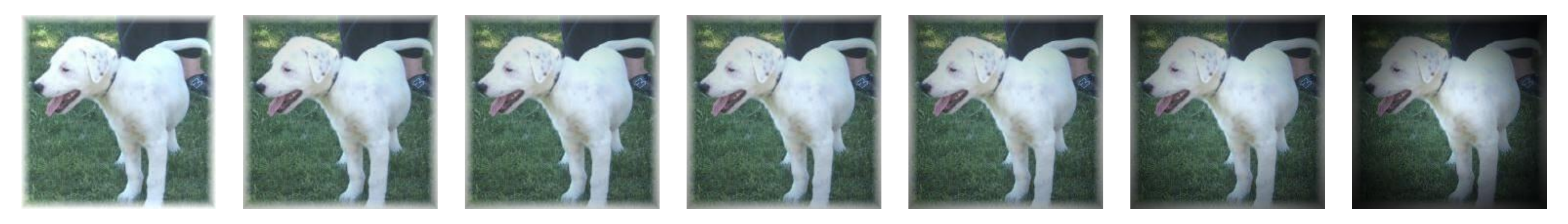}
    \caption{Two additional example interventions for dogs with light fur.}
    \end{subfigure}
    \begin{subfigure}{\textwidth}
    \includegraphics[width=\textwidth]{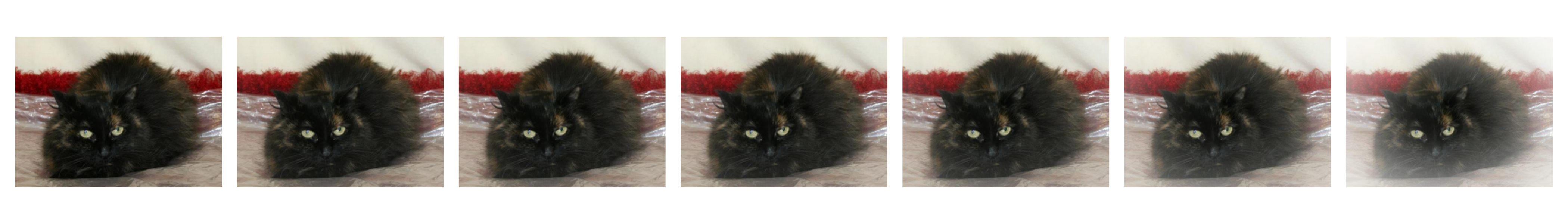}
    \end{subfigure}
    \begin{subfigure}{\textwidth}
    \includegraphics[width=\textwidth]{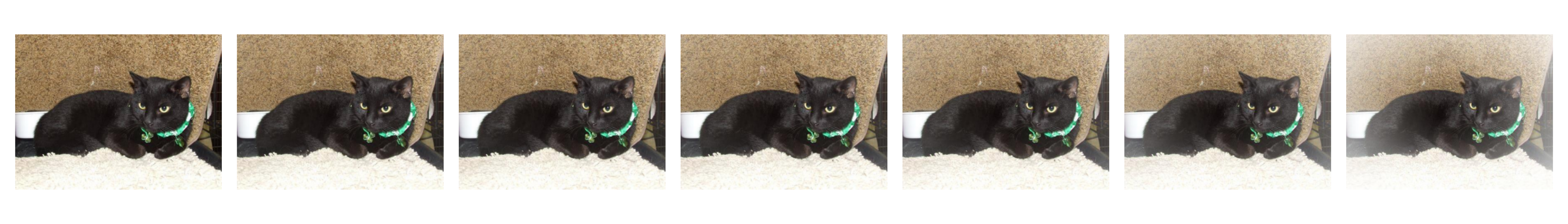}
    \caption{Two example interventions for cats with dark fur.}
    \end{subfigure}
    \caption{
    Interventions on the background for both cats and dogs. 
    For both classes, we include two examples.
    After observing failure cases (see \cref{fig:bg-fails}) using \cite{fu2024mgie}, we design the intervention using Gaussian kernels to scale pixels close to the image edges.
    }
    \label{fig:add-interv-bg}
\end{figure*}

To intervene on the background illumination, we utilize the following prompt with \cite{fu2024mgie}: ``[Darken/Brighten] the background color, keep the fur color unchanged''.
However, in contrast to the foreground interventions, we observe various failure cases.
We illustrate two problems we encountered in \cref{fig:bg-fails}.

In the first case, we observe minimal changes in the background, while instead, we notice more pronounced changes in the foreground, which we explicitly aim to avoid. 
In the second failure case, we witness a complete shift in the subject, in addition to the correct intervention in the background.
We hypothesize that these observations are an expression of the causal hierarchy theorem for image edits \cite{pan2024counterfactual}. 
Specifically, in \cite{pan2024counterfactual}, the authors demonstrate that even if a model correctly learns the training distributions, it does not guarantee that it learns the correct causal structure at a higher level of the PCH \cite{bareinboim2022onpearl}.

The backbone of \cite{fu2024mgie}, Instruct-Pix2Pix \cite{brooks2022instructpix2pix}, was trained using synthetic interventional data produced using \cite{hertz2022prompt}. 
We believe that the background interventions are examples of image edits that are uncommon in the synthetic training data.

Nevertheless, these failure cases reinforce our conviction that human oversight is currently necessary to elevate explanations of local behavior to the interventional level, as discussed in the main paper in \Cref{sec:limitations}. 
Therefore, to further investigate the background interventions, we employ our second identified approach to generate interventional data and utilize image processing.
Specifically, we assume a centered subject and reduce and utilize a centered Gaussian kernel to multiplicatively increase or decrease the pixels that are closer to the image edge. 
We provide examples in \cref{fig:add-interv-bg}.

\begin{table}[t]
    \centering
    \caption{\propgrad of the background property for our three CvD models.
    Here, we utilize the mean behavior (\cref{fig:bg-mean-results}) over ten images.
    Additionally, we report significance ($p<0.01$) and prediction flips.
    For examples of the interventional data, see \cref{fig:add-interv-bg}.
    }
    \label{tab:cvd-impact-mean-bg}
    \begin{tabular}{llccc}
    \toprule
        & & \multicolumn{3}{c}{Model Behavior}\\
        \cmidrule(lr){3-5}
      Data  & Model & \propgrad & $p<0.01$ & Pred. Flips \\
    \midrule
    \multirow{3}{*}{\shortstack[l]{Light \\Dogs}} 
        & unbiased  & 0.00289 & \cmark & \xmark \\
        & dark cats & 0.00015 & \cmark & \xmark \\
        & dark dogs & 0.00193 & \cmark & \xmark \\
    \midrule
    \multirow{3}{*}{\shortstack[l]{Dark \\Cats}} 
        & unbiased  & 0.00080 & \cmark & \xmark \\
        & dark cats & 0.00004 & \cmark & \xmark \\
        & dark dogs & 0.00188 & \cmark & \xmark \\
    \bottomrule
    \end{tabular}
\end{table}

\begin{figure*}[tb]
\centering
\begin{subfigure}{0.48\textwidth}
\includegraphics[width=\linewidth]{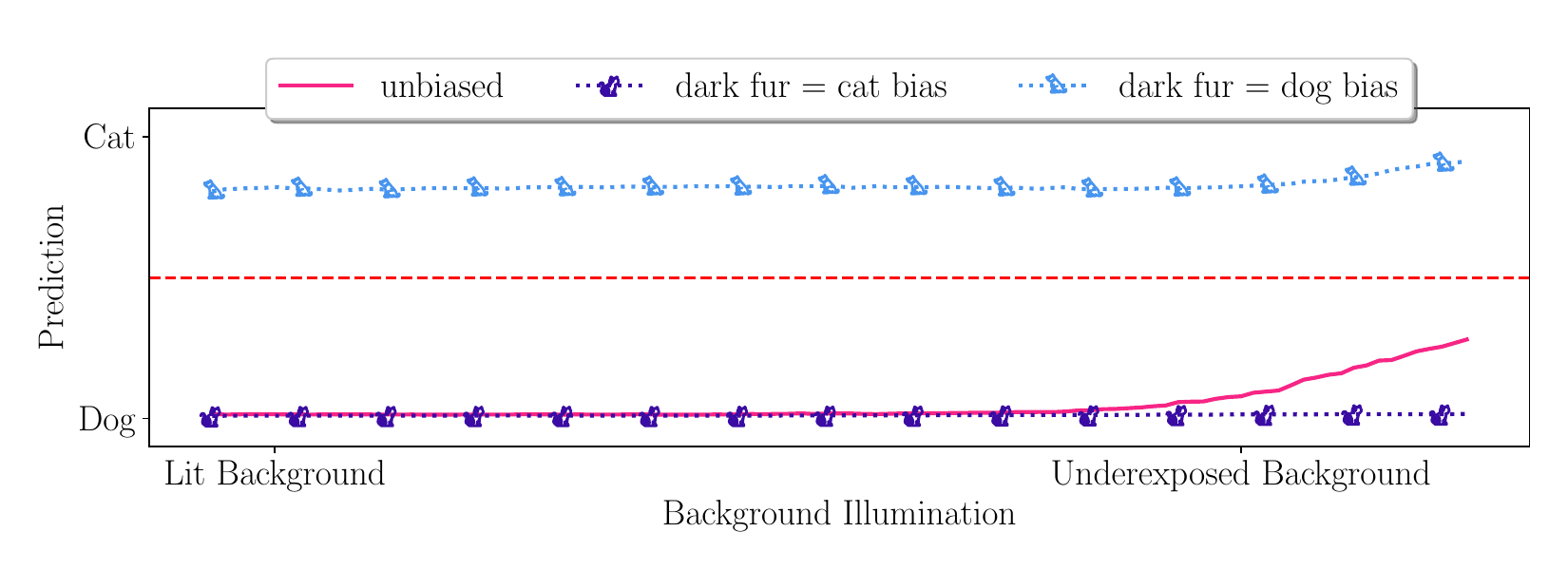}
\caption{Average of ten \textbf{dog} images with background interventions.}
\label{fig:bg-dogs-mean}
\end{subfigure}
\begin{subfigure}{0.48\textwidth}
\includegraphics[width=\linewidth]{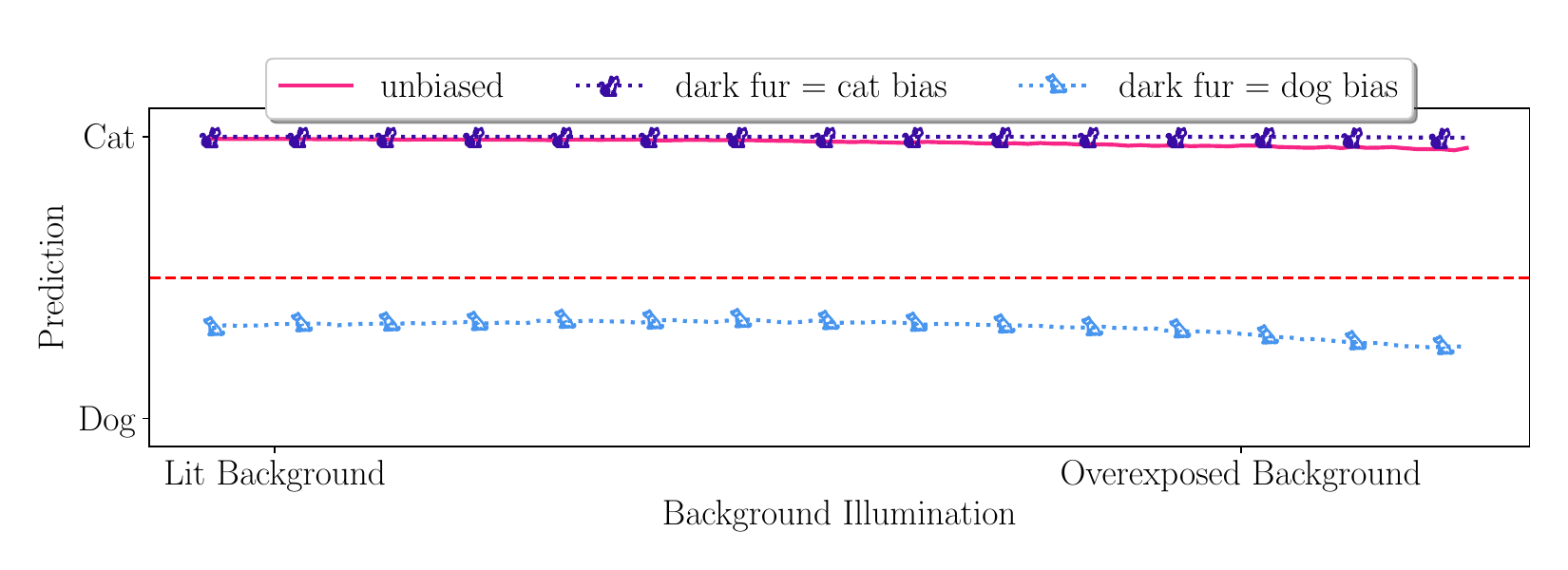}
\caption{Average of ten \textbf{cat} images with background interventions.}
\label{fig:bg-cats-mean}
\end{subfigure}

\caption{
Changes in model predictions for an intervention on the background illumination for ten respective images.
Here, we use three ConvMixer models: an unbiased one, one trained on only dark-furred dogs and light-furred cats (``dark fur = dog''), and one trained only on the opposite split (``dark fur = cat'').
The \textcolor{red}{red dotted line} indicates the threshold where the model prediction flips.
}
\label{fig:bg-mean-results}
\end{figure*}

\begin{figure*}[tb]
    \centering
    \begin{subfigure}{0.48\textwidth}
        \includegraphics[width=\linewidth]{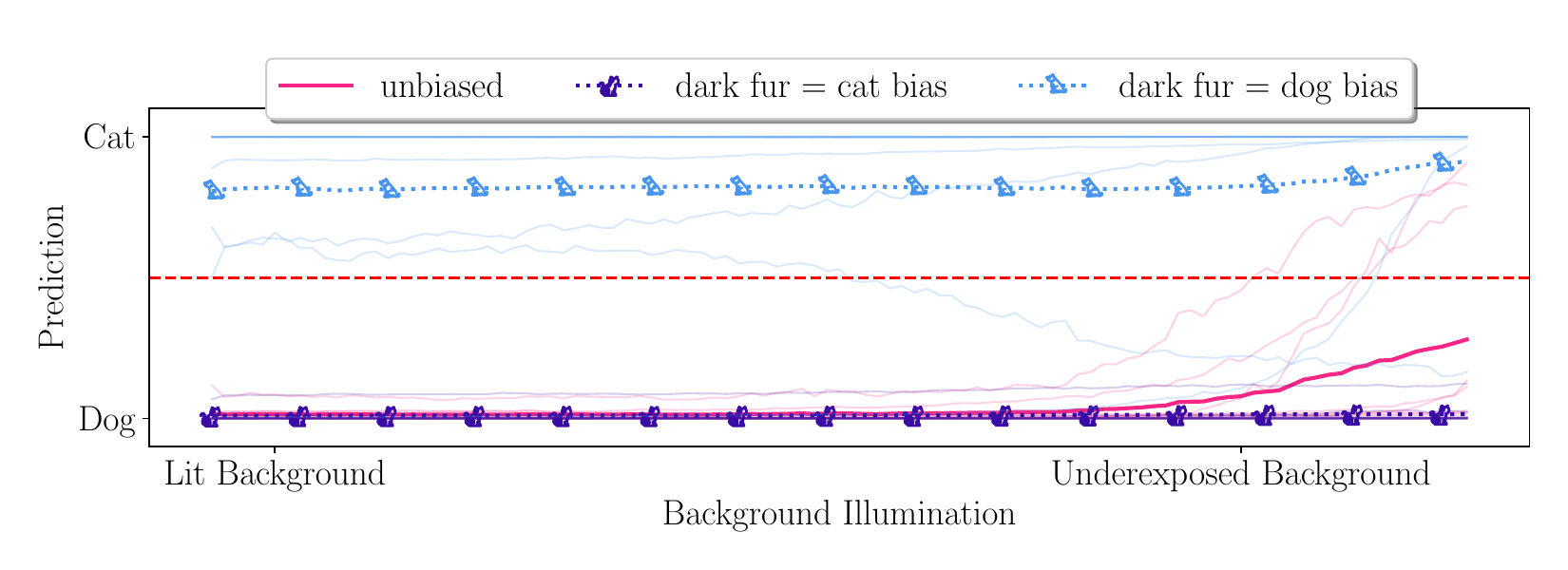}
        \caption{Ten \textbf{dog} images with background interventions.}
        \label{fig:bg-dogs-schar}
    \end{subfigure}
    \begin{subfigure}{0.48\textwidth}
        \includegraphics[width=\linewidth]{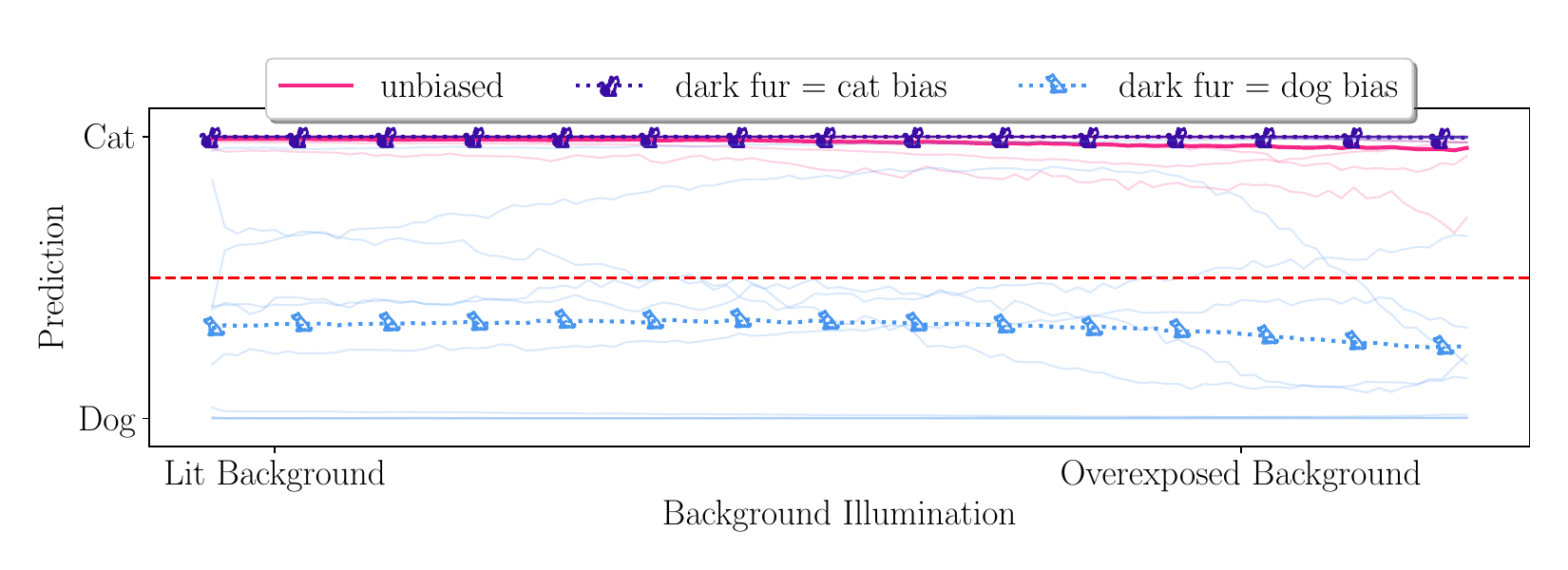}
        \caption{Ten \textbf{cat} images with background interventions.}
        \label{fig:bg-cats-schar}
    \end{subfigure}
    
    \caption{
    Changes in model predictions for an intervention on the background illumination for ten respective images.
    Here, we use three ConvMixer models: an unbiased one, one trained on only dark-furred dogs and light-furred cats (``dark fur = dog''), and one trained only on the opposite split (``dark fur = cat'').
    The \textcolor{red}{red dotted line} indicates the threshold where the model prediction flips.
    }
    \label{fig:bg-schar-results}
\end{figure*}

\cref{fig:bg-mean-results} illustrates the mean behavior changes, and we summarize the corresponding \propgrad values in \Cref{tab:cvd-impact-mean-bg}. 
Furthermore, \cref{fig:bg-schar-results} visualizes the local behavior for the different local images under the background interventions.

Our analysis largely confirms the local observations made in \Cref{sec:exp1}. 
Specifically, we observe significantly lower \propgrad values for all models under the background interventions compared to the fur color interventions. 
Nevertheless, all measured changes in behavior are statistically significant, as determined by \Cref{alg:sig}. 
However, on average, none of the models exhibit flips in their predictions. 
Despite this, we observe in \cref{fig:bg-schar-results} that for some images, the models flip predictions for the stronger stages of the intervention. 
Notably, the unbiased model in \cref{fig:bg-dogs-schar} incorrectly predicts \texttt{cat} for some of the intervened dog images. 
We hypothesize that this is a consequence of our designed interventional strategy, which can also affect the foreground subjects. 
These results provide additional evidence to support the arguments raised in \Cref{sec:limitations} and highlight the importance of visually inspecting the interventional data and corresponding model behavior.

\subsection{Local Baseline XAI Results}
\label{app:cvd-local-baselines}

In this section, we generate multiple other local explanations and highlight the challenges in extracting meaningful semantic insights from the results. 
In contrast, our approach presented in the main paper analyzes the changes in output for variations in a property, yielding actionable explanations. 
We also demonstrate how previous approaches can benefit from incorporating local interventions.

\paragraph{Setup}

We select the following methods to generate visual explanations for individual examples 
\cite{sundararajan2017axiomatic,shrikumar2017learning,selvaraju2020grad,zeiler2014visualizing,ribeiro2016should,lundberg2017unified}.
We choose these methods due to their widespread usage and ease of accessibility.
In any case, for all chosen methods, we utilize the implementations from \cite{kokhlikyan2020captum} to ensure consistency and reproducibility.

These local methods highlight areas in the input that speak for or against the selected class (sometimes both).
Specifically, we select the \texttt{Cat} logit of our models to stay as comparable as possible to our main paper.
We now detail specific hyperparameter choices to ensure reproducibility:
\begin{itemize}
    \item or Guided Grad-CAM \cite{selvaraju2020grad}, we generate explanations with respect to the last convolutional layer of our models.
    \item For the occlusion-based method \cite{zeiler2014visualizing}, we use the average gray scale value with a window size of $9\times 9$. Additionally, we employ a stride length of four in both $x$ and $y$ directions.
    \item For LIME \cite{ribeiro2016should}, and Kernel-SHAP \cite{lundberg2017unified}, we utilize SLIC \cite{achanta2012slic} super-pixels with 100 segments and a compactness of one as a basis.
\end{itemize}
For all other hyperparameters, we use the default settings provided by \cite{kokhlikyan2020captum}.

\begin{figure*}
    \centering
    \begin{subfigure}{0.48\textwidth}
        \includegraphics[width=\textwidth]{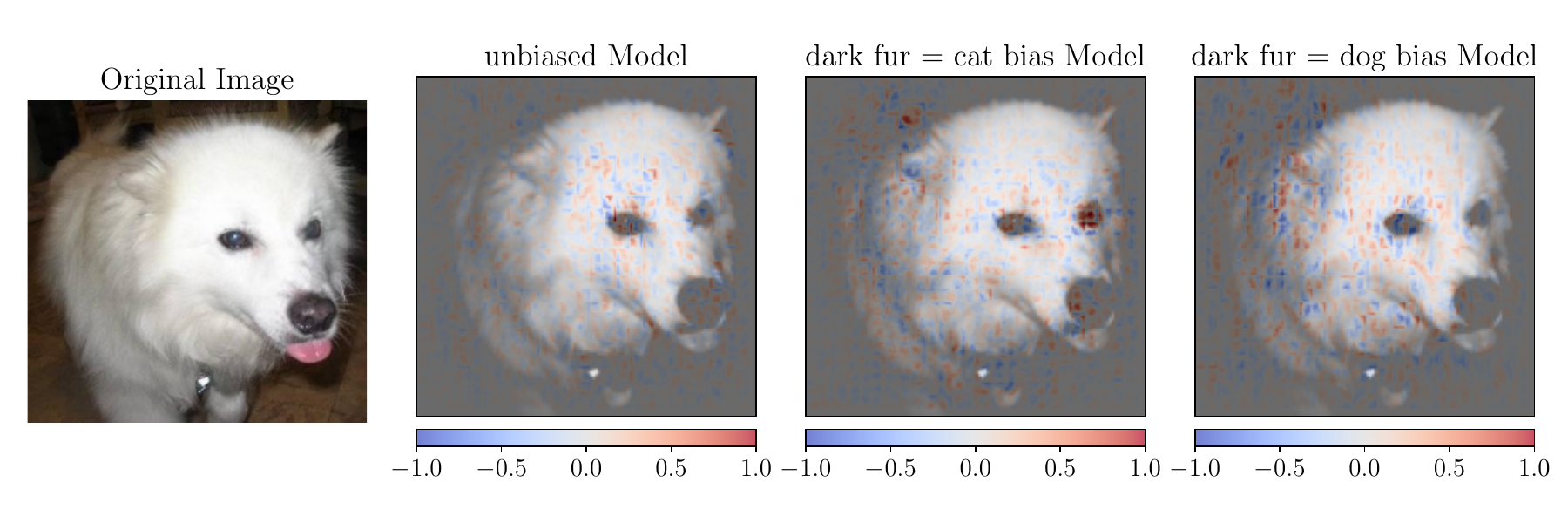}
        \caption{Integrated gradients \cite{sundararajan2017axiomatic} for image of a dog.}
        \label{fig:ig-orig}
    \end{subfigure}
    \begin{subfigure}{0.48\textwidth}
        \includegraphics[width=\textwidth]{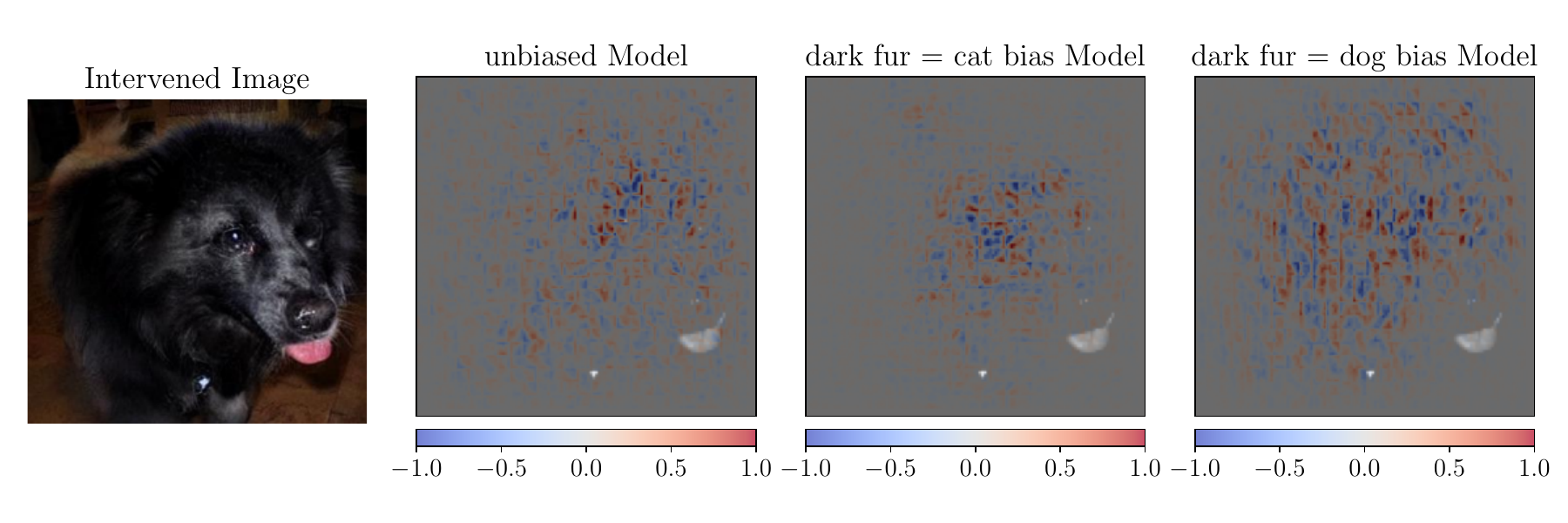}
        \caption{Integrated gradients \cite{sundararajan2017axiomatic} for intervened image.}
        \label{fig:ig-inter}
    \end{subfigure}

    \begin{subfigure}{0.48\textwidth}
        \includegraphics[width=\textwidth]{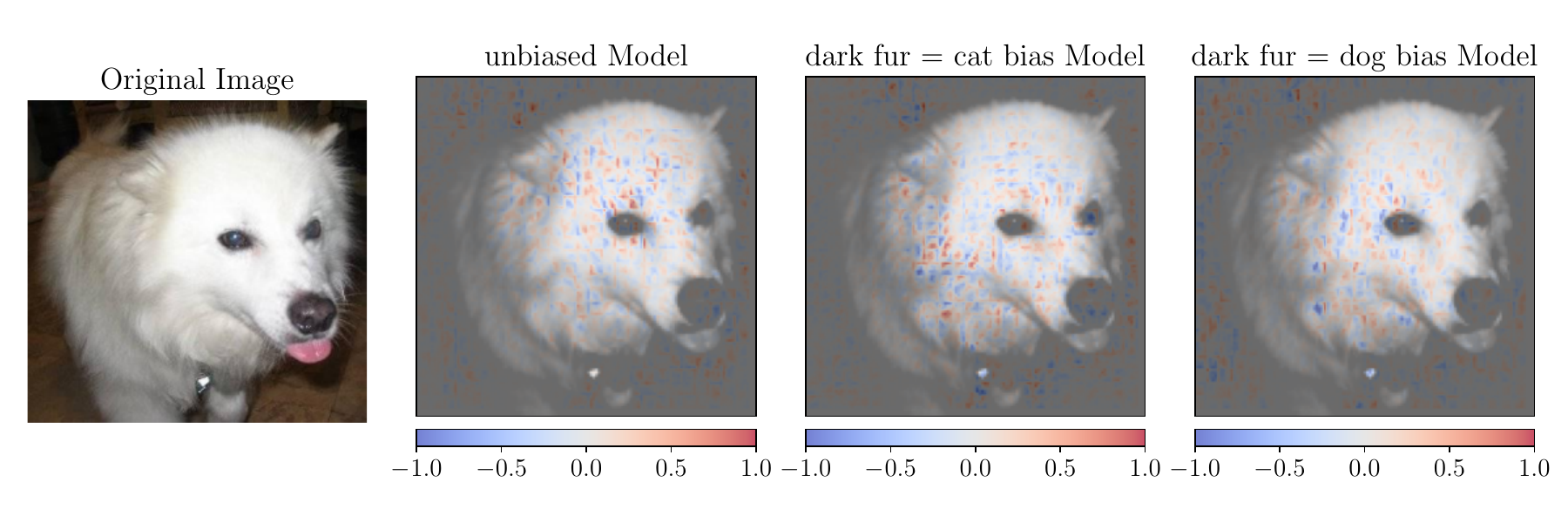}
        \caption{DeepLIFT \cite{shrikumar2017learning} attribution for image of a dog.}
        \label{fig:deeplift-orig}
    \end{subfigure}
    \begin{subfigure}{0.48\textwidth}
        \includegraphics[width=\textwidth]{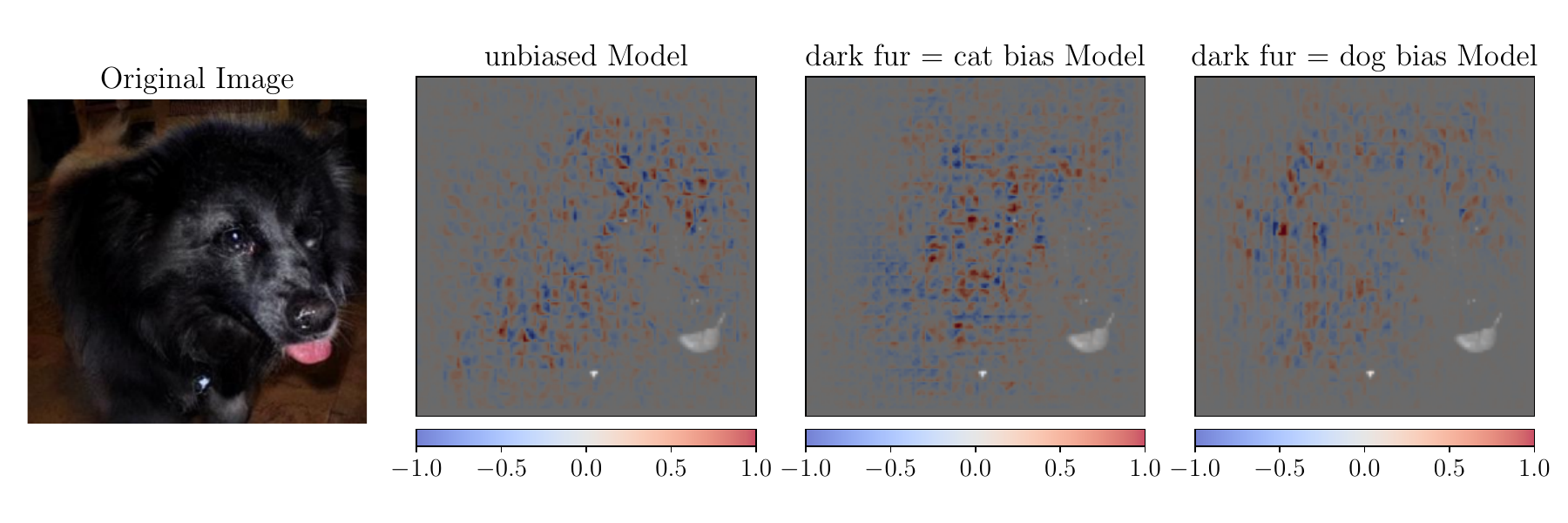}
        \caption{DeepLIFT \cite{shrikumar2017learning} attribution for intervened image.}
        \label{fig:deeplift-inter}
    \end{subfigure}

    \begin{subfigure}{0.48\textwidth}
        \includegraphics[width=\textwidth]{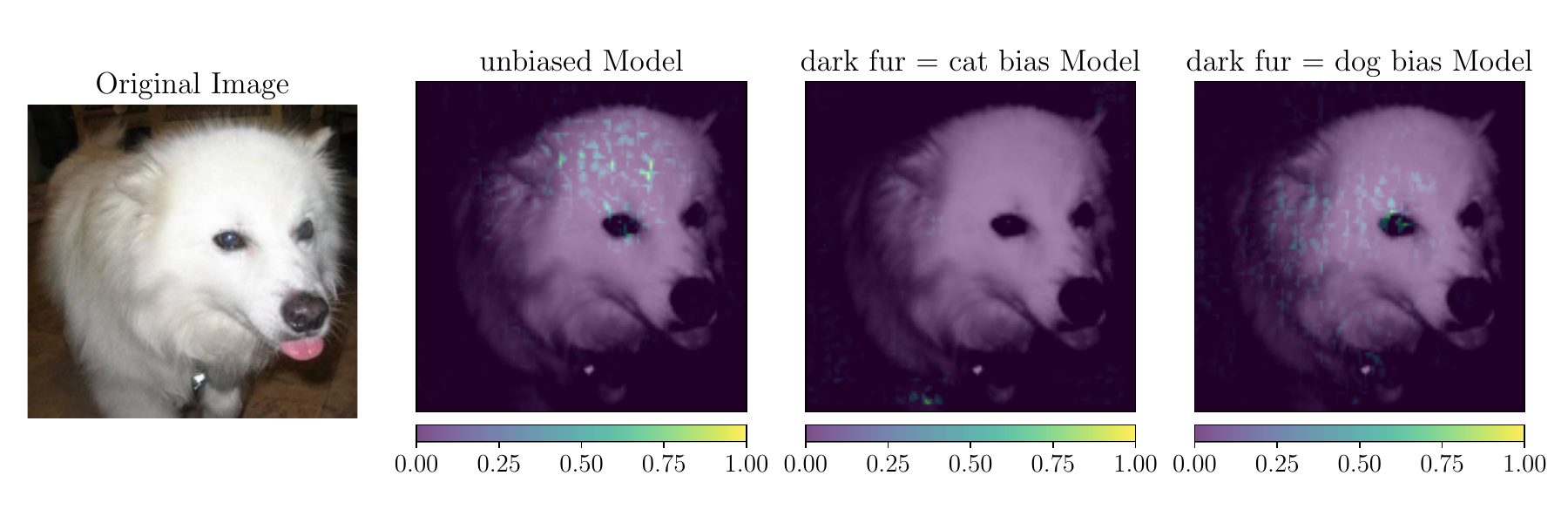}
        \caption{Guided Grad-CAMs \cite{selvaraju2020grad} for image of a dog.}
        \label{fig:gcam-orig}
    \end{subfigure}
    \begin{subfigure}{0.48\textwidth}
        \includegraphics[width=\textwidth]{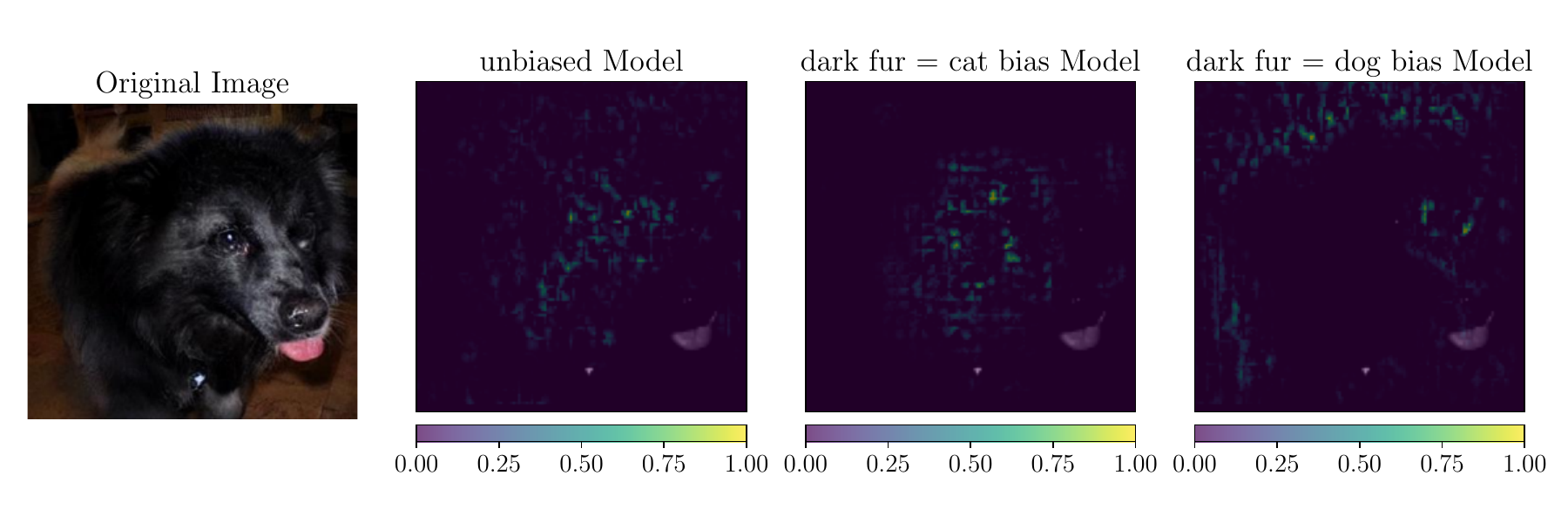}
        \caption{Guided Grad-CAM \cite{selvaraju2020grad} for intervened image.}
        \label{fig:gcam-inter}
    \end{subfigure}

    \begin{subfigure}{0.48\textwidth}
        \includegraphics[width=\textwidth]{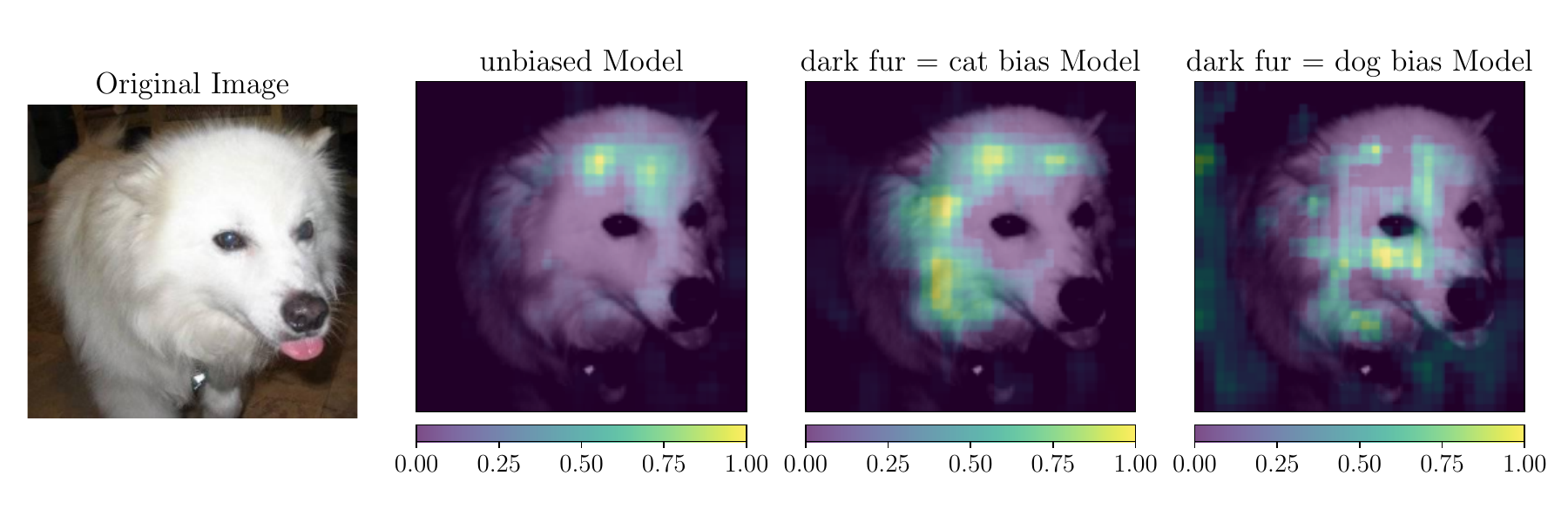}
        \caption{Occlusion based attribution \cite{zeiler2014visualizing} for image of a dog.}
        \label{fig:occ-orig}
    \end{subfigure}
    \begin{subfigure}{0.48\textwidth}
        \includegraphics[width=\textwidth]{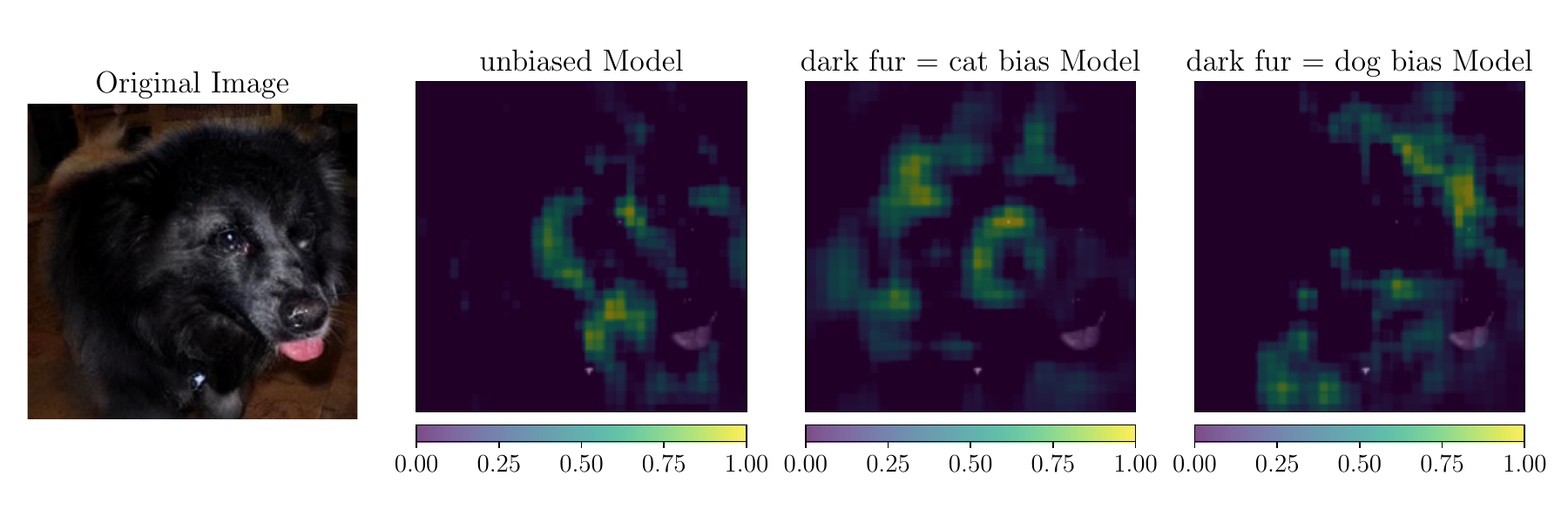}
        \caption{Occlusion based attribution \cite{zeiler2014visualizing} for intervened image.}
        \label{fig:occ-inter}
    \end{subfigure}

    \begin{subfigure}{0.48\textwidth}
        \includegraphics[width=\textwidth]{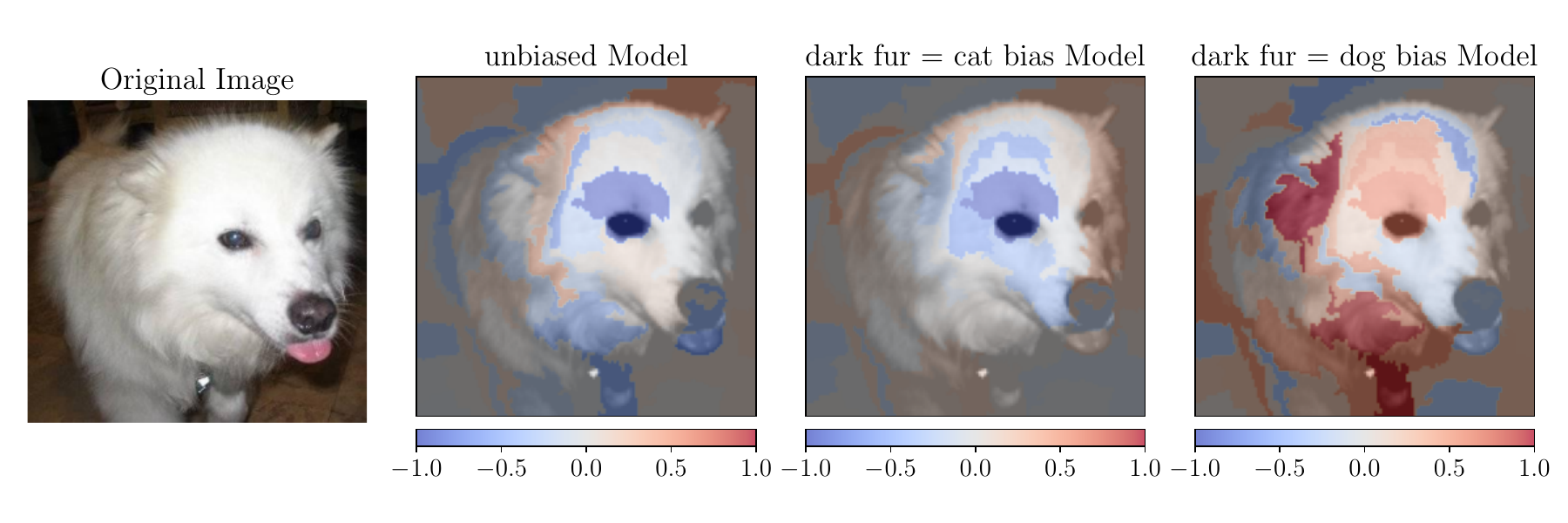}
        \caption{LIME-based \cite{ribeiro2016should} attribution for the image of a dog.}
        \label{fig:lime-orig}
    \end{subfigure}
    \begin{subfigure}{0.48\textwidth}
        \includegraphics[width=\textwidth]{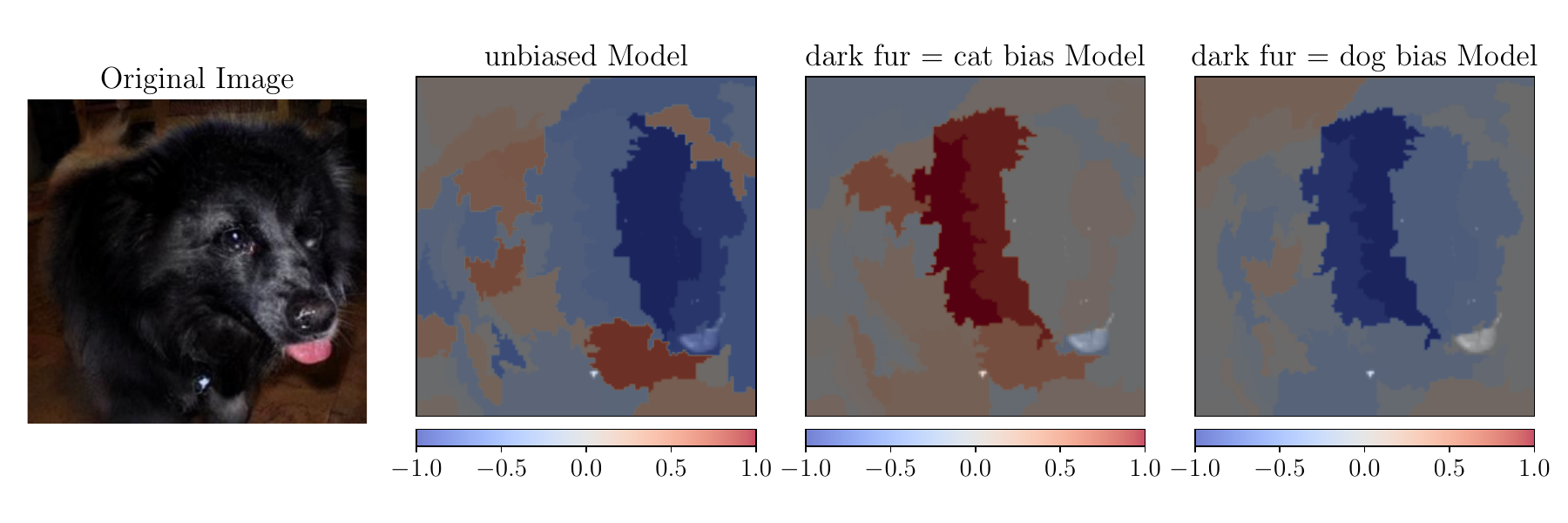}
        \caption{LIME-based \cite{ribeiro2016should} for intervened image.}
        \label{fig:lime-inter}
    \end{subfigure}

    \begin{subfigure}{0.48\textwidth}
        \includegraphics[width=\textwidth]{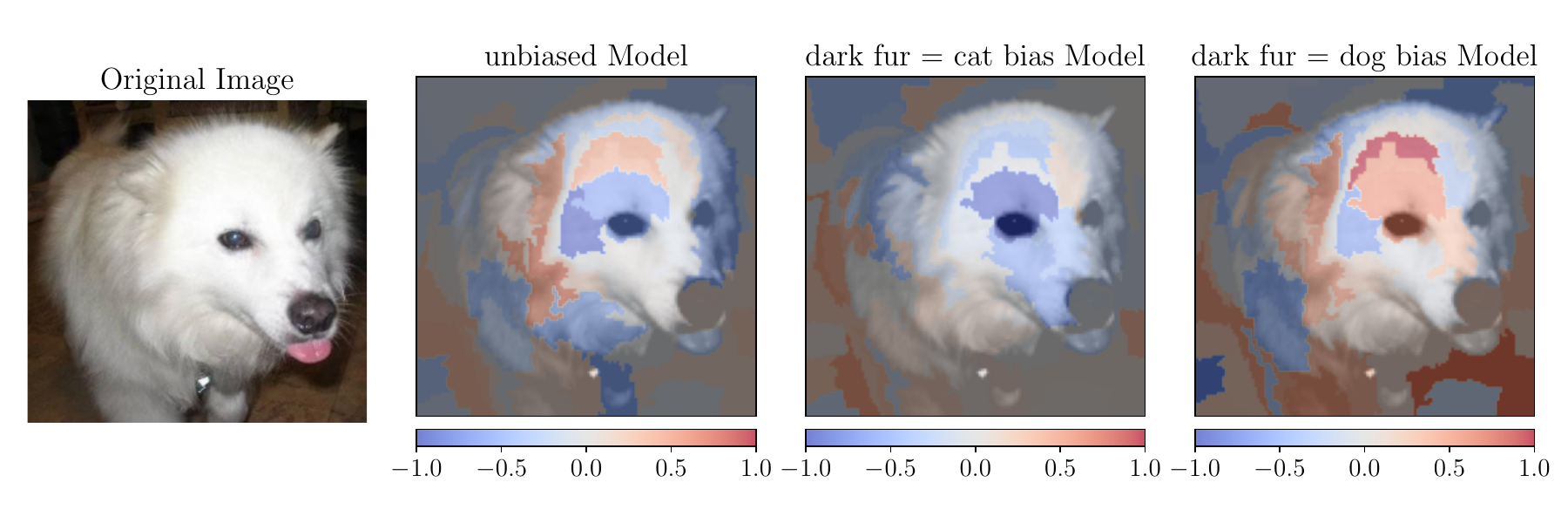}
        \caption{Kernel-SHAP based \cite{lundberg2017unified} attribution for image of a dog.}
        \label{fig:kshap-orig}
    \end{subfigure}
    \begin{subfigure}{0.48\textwidth}
        \includegraphics[width=\textwidth]{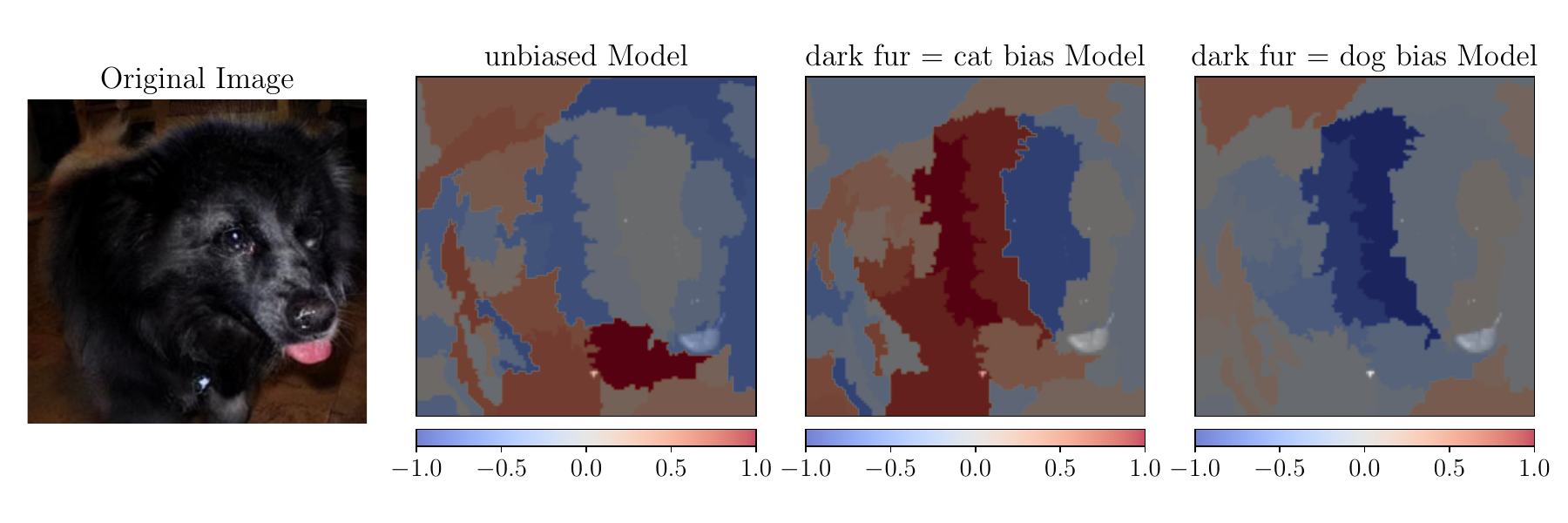}
        \caption{Kernel-SHAP based \cite{lundberg2017unified} attribution for intervened image.}
        \label{fig:kshap-inter}
    \end{subfigure}
    
    \caption{
    Multiple local XAI methods were applied to our dog image example.
    In all cases, we utilized the implementation as provided in \cite{kokhlikyan2020captum}.
    We also exclusively visualize the results for the cat class logit, meaning positive values should correspond to regions important for the decision of the respective models.
    }
    \label{fig:local-baselines}
    \vspace{0.5cm}
\end{figure*}

\paragraph{Results}

\cref{fig:local-baselines} presents visualizations of the selected local XAI baselines, including attribution maps for both the original image and an image with an intervention on the fur color.

Notably, in many cases, the most important regions identified by these methods align with the head of the dog. 
For gradient-based methods, such as \cite{selvaraju2020grad,sundararajan2017axiomatic,shrikumar2017learning}, this suggests that changes in these pixels lead to significant changes in the \texttt{cat} class logit. 
Similarly, the occlusion-based approach \cite{zeiler2014visualizing} indicates that occluding the head has the highest impact. 
LIME \cite{ribeiro2016should} and Kernel-SHAP \cite{lundberg2017unified} also highlight regions that support or contradict the prediction locally, correctly marking the head in the respective color of the corresponding bias.
Furthermore, investigating the intervened images reveals that the colors flip.

This observation suggests a change in behavior at the local level, which can be explained by the highlighted image regions. 
However, visual explanations require semantic interpretation to identify the human-understandable property driving this change. 
By examining only the left-hand side of \cref{fig:local-baselines}, it is challenging to determine whether the head shape, ears, fur color, or another property is the causal factor.
Interventional images can facilitate this interpretation. 
By providing both the original image and an image with a black fur intervention, we simplify the interpretation of the visualizations. 
In contrast, our gradual interventional approach yields a direct explanation with respect to a specific property. 
Further, estimating \propgrad provides a structured way to measure the corresponding local impact.
However, our approach is not mutually exclusive with other local explanation methods but rather can complement them in future works, as seen in \cref{fig:local-baselines}.

\subsection{Global Baseline XAI Results}
\label{app:cvd-global-baselines}

In this section, we evaluate our approach by comparing it to global explanations derived for the cats versus dogs model. 
To facilitate this comparison, we first outline the experimental setup and introduce the chosen methods for generating global explanations.

\paragraph{Setup}

For the comparison with global methods, we select three approaches: \cite{reimers2020determining,yeh2020completeness,schmalwasser2024exploiting}. 
These methods provide a diverse range of techniques for generating global explanations, allowing us to evaluate our approach.

First, we utilize \cite{yeh2020completeness} to identify important concepts in the model. 
This method uses a probing dataset to identify concepts in a specific layer and orders them by importance using Shapley values (SHAP) \cite{shapley1953value}. 
Specifically, it finds concepts that maximize completeness for all classes.
They then remove duplicates, where the concept activation vectors (CAVs) have a similarity of over 95\%.
To generate explanations for a specific class, the top-K image patches closest in the activation space are selected. 
We follow this approach and select the three CAVs with the highest SHAP values and show the top six images per CAV.

Next, we employ \cite{schmalwasser2024exploiting} to identify semantic concepts. 
This method builds on \cite{yeh2020completeness} and aims to identify textual descriptions for specific discovered concepts. 
Towards this goal, it performs a comparison to a set of texts in CLIP \cite{radford2021learning} latent space using cosine similarities. 
For both \cite{yeh2020completeness} and \cite{schmalwasser2024exploiting}, we target the last convolutional layer of our trained cat versus dog networks and use the validation set of ImageNet \cite{russakovsky2015imagenet} as the probing dataset. 
Additionally, for \cite{schmalwasser2024exploiting}, we use the 20K most common Google terms \cite{googl20k} to calculate the textual descriptions.

Finally, we select \cite{reimers2020determining} as a global explanation method because we deem it closely related to our approach for analyzing arbitrary properties.
However, it is an associational approach that aims to uncover changes in behavior on the test dataset with respect to a selected property without interventions. 
Similar to our approach, \cite{reimers2020determining} frames supervised learning as a Structural Causal Model (SCM). 
However, it investigates the dependence between the property and the output statistically. 
Crucially, the authors note that the reference annotation is often a common confounder when the label correlates with the property of interest. 
Hence, conditional independence (CI) is estimated with the reference annotations serving as the conditioning variable. 
The resulting explanation is the binary result of this hypothesis test.

However, there is no universal non-parametric CI test \cite{shah2020hardness}. 
Therefore, selecting a suitable CI test is the essential hyperparameter choice.
We use partial correlation to capture linear relationships and two non-linear tests, namely CMIknn \cite{runge2018conditional} and conditional HSIC \cite{fukumizu2007kernel}. 
For the properties in our test dataset, we select the fur color and background illumination to stay comparable to our experiments in the main paper. 
Specifically, we take the binary fur color as indicated by membership in our biased splits (see \Cref{app:cvd-data}).
Regarding the background color, we calculate the mean of the pixel intensities over the four corner pixels for all images in our test set.
Here, a low value indicates a dark background, while a high value corresponds to bright colors.

\begin{figure*}
    \centering
    \includegraphics[width=\linewidth]{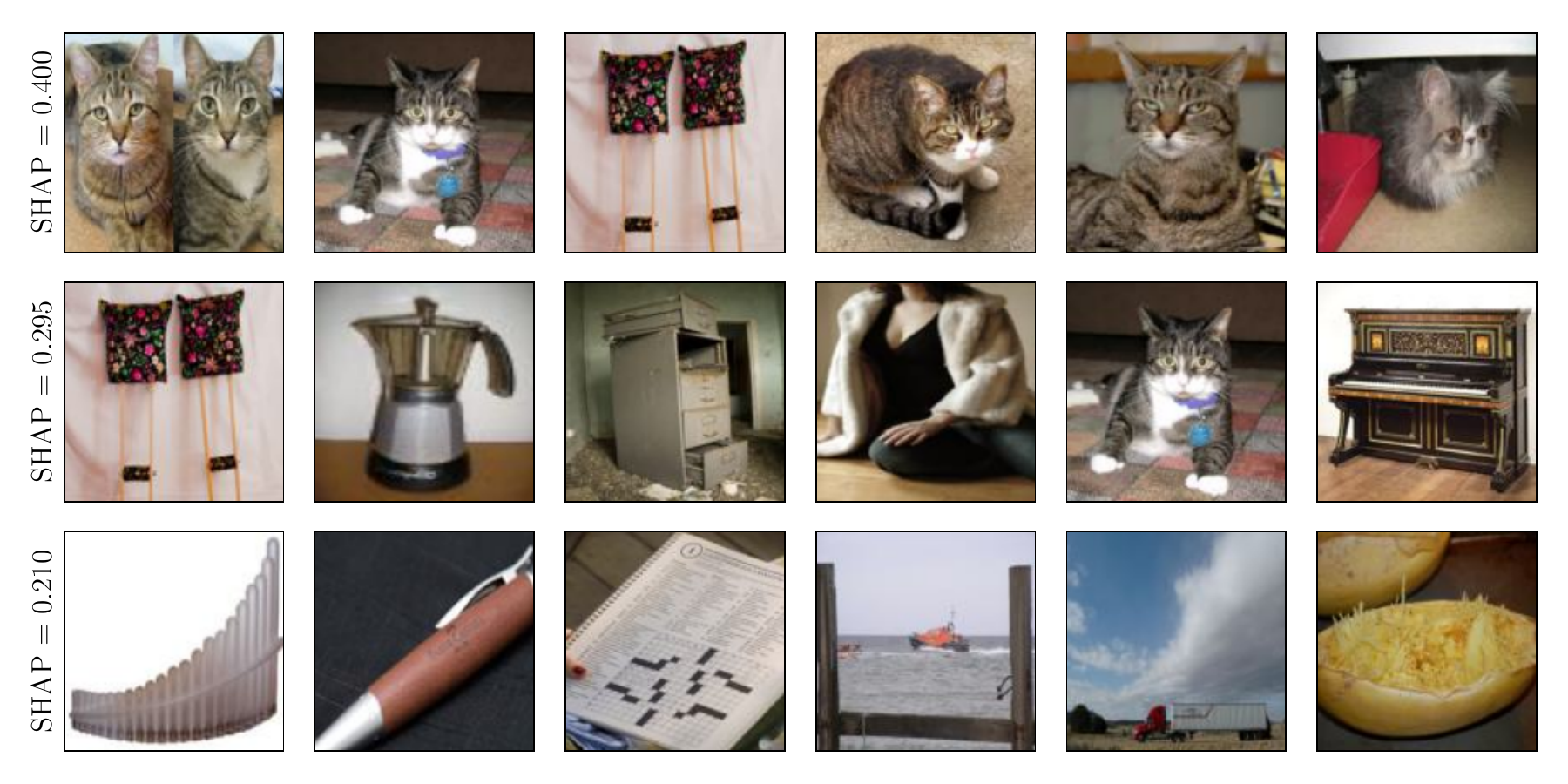}
    \caption{The top three CAVs for the unbiased model using \cite{yeh2020completeness}.
    Here we focus again on the cat class logit, to stay consistent with our other experiments.
    }
    \label{fig:unbiased-cavs}
    \vspace{0.6cm}
\end{figure*}
\begin{figure*}
    \centering
    \includegraphics[width=\linewidth]{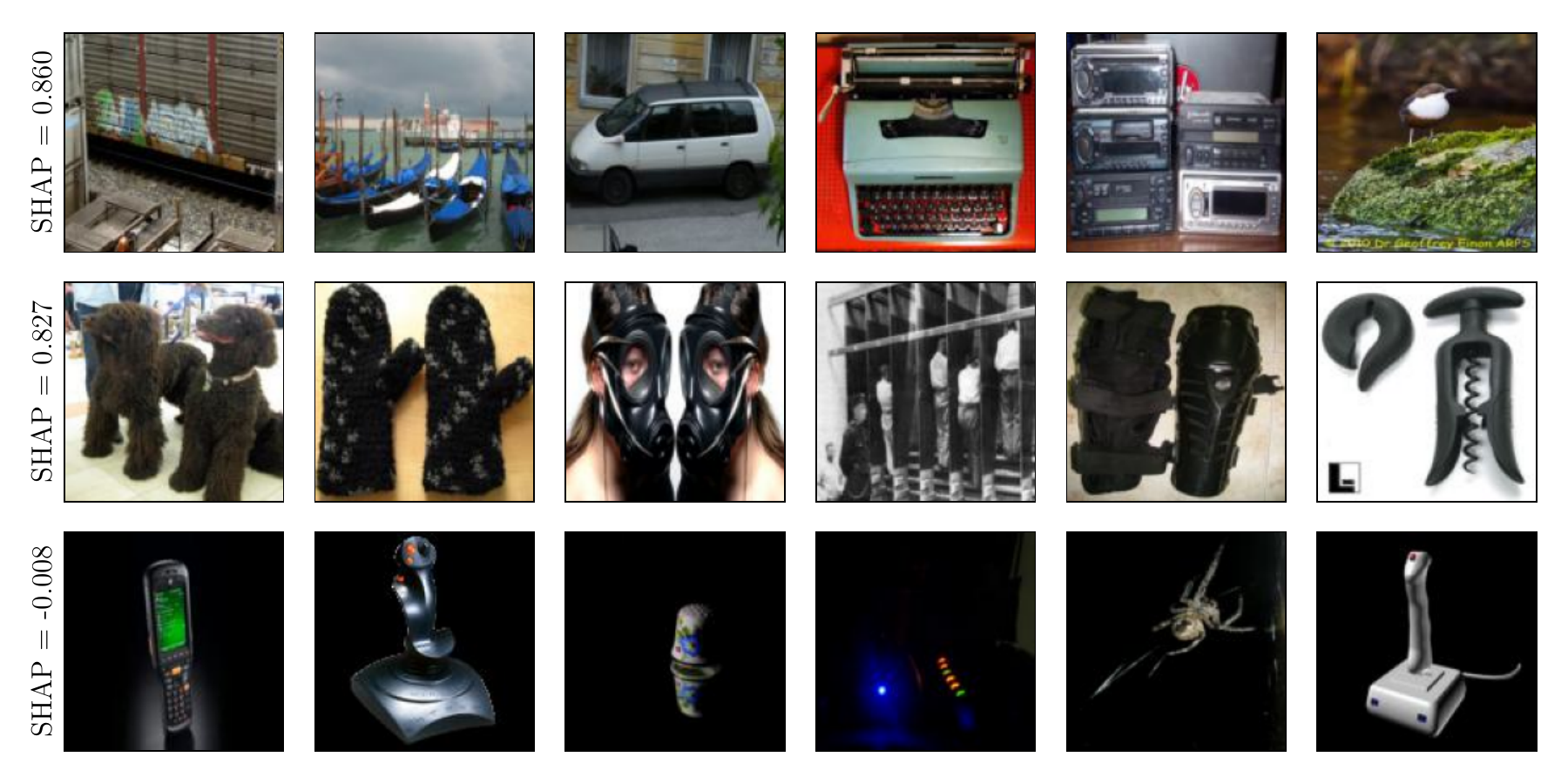}
    \caption{The top three CAVs for the dark cat biased model using \cite{yeh2020completeness}.
    Here we focus again on the cat class logit, to stay consistent with our other experiments.
    }
    \label{fig:dcats-cavs}
    \vspace{0.5cm}
\end{figure*}
\begin{figure*}
    \centering
    \includegraphics[width=\linewidth]{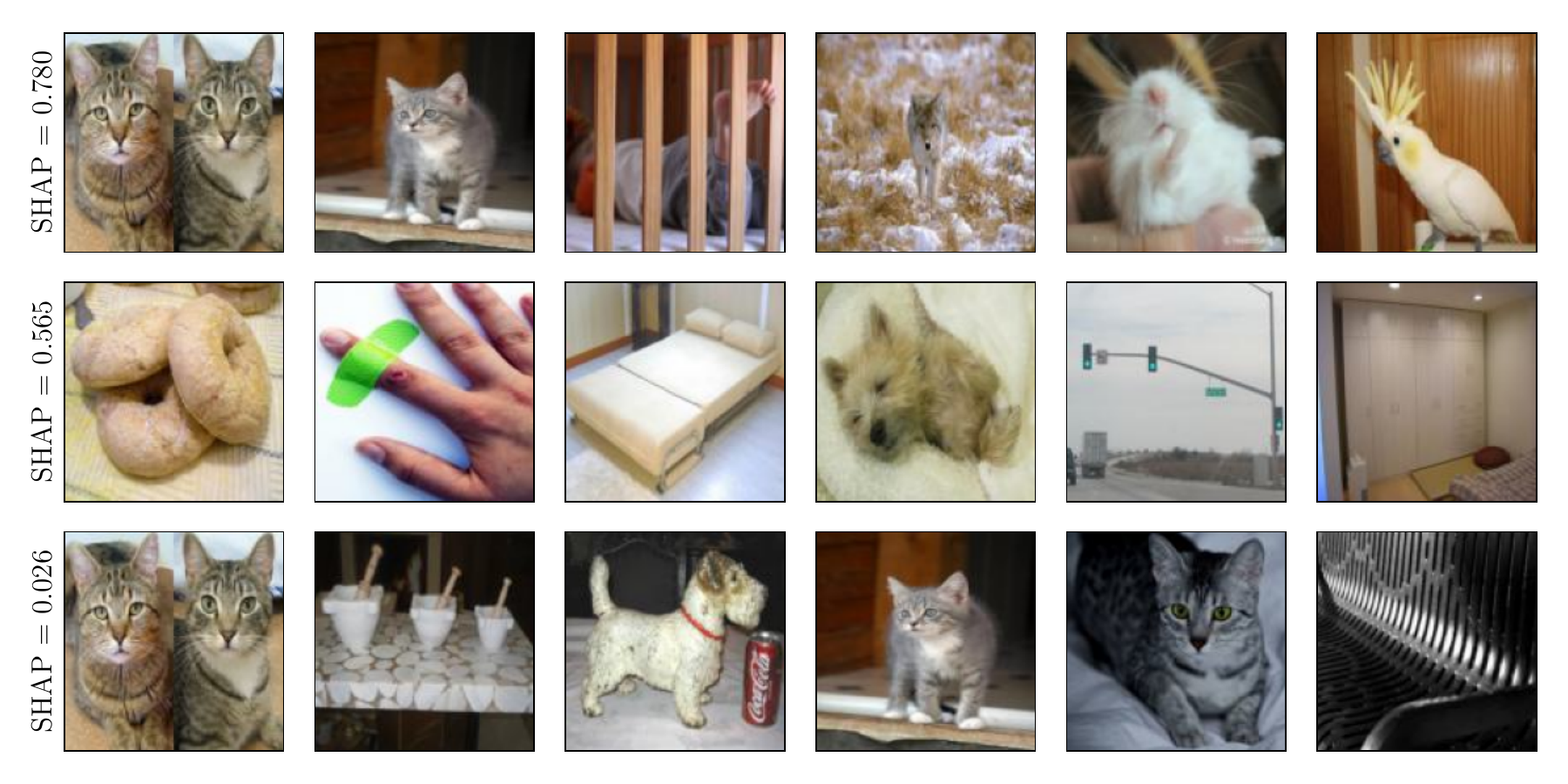}
    \caption{The top three CAVs for the dark dog biased model using \cite{yeh2020completeness}.
    Here we focus again on the cat class logit, to stay consistent with our other experiments.
    }
    \label{fig:ddogs-cavs}
\end{figure*}

\begin{table*}[t]
    \centering
    \caption{Concept descriptions for the CAVs visualized in \cref{fig:unbiased-cavs}, \cref{fig:dcats-cavs}, and \cref{fig:ddogs-cavs} generated using \cite{schmalwasser2024exploiting}.
    Here, Google20K \cite{googl20k} is used as the text probing dataset.}
    \label{tab:cav-words}
    \begin{tabular}{lcll}
    \toprule
       Model & SHAP & Central Word & Closest Words  \\
    \midrule
       \multirow{3}{*}{Unbiased} 
       &  0.400 & cats & kitts, cats, katz, cat, lynx \\
       &  0.295 & cabinets & cabinets, dresser, furniture, cabinet, furnishing \\
       &  0.210 & products & products, shipment, casserole, product, coatings\\
       \midrule
       \multirow{3}{*}{Dark cats bias} 
       &  0.860 & vehicles & vehicles, vehicle, automobiles, wagon, decks \\
       &  0.827 & armor & black, equipment, armor, exhibitor, overstock\\
       &  -0.008 & lights & nightlife, candle, lights, flashlight, darkness\\
       \midrule
       \multirow{3}{*}{Dark dogs bias} 
       &  0.780 & kitten & beige, kitten, persian, cygwin, mimi \\
       &  0.565 & room & room, bed, condosaver, white, resulting \\
       &  0.026 & cat & cat, cats, kitten, kitts, persian\\
    \bottomrule
    \end{tabular}
\end{table*}

\begin{table*}[t]
    \centering
    \caption{
    Results for the global method described in \cite{reimers2020determining} and the fur color and background color properties in the cats versus dogs test set.
    We use a significance level of 0.05 and provide the binary results of the hypothesis tests. Additionally, we mark \underline{significant results}.
    Finally, the columns denote different conditional independence tests, the main hyperparameter of \cite{reimers2020determining}. 
    }
    \label{tab:ci-results}
    \begin{tabular}{lccc ccc}
    \toprule
        & \multicolumn{3}{c}{Fur Color} & \multicolumn{3}{c}{Background Color} \\
    \cmidrule(lr){2-4} \cmidrule(lr){5-7}
       Model & Partial Corr. & CMIknn \cite{runge2018conditional} & cHSIC \cite{fukumizu2007kernel} & Partial Corr. & CMIknn \cite{runge2018conditional} & cHSIC \cite{fukumizu2007kernel}  \\
    \midrule
       unbiased model & \underline{$p<0.05$} & \underline{$p<0.05$} & $p>0.05$ & $p>0.05$ & \underline{$p<0.05$} & $p>0.05$ \\
       dark fur = cats & \underline{$p<0.05$} & \underline{$p<0.05$} & $p>0.05$ & \underline{$p<0.05$} & \underline{$p<0.05$} & $p>0.05$\\
       dark fur = dogs & \underline{$p<0.05$} & \underline{$p<0.05$} & $p>0.05$ & \underline{$p<0.05$} & \underline{$p<0.05$} & $p>0.05$ \\
    \bottomrule
    \end{tabular}
\end{table*}

\paragraph{Results}

\cref{fig:unbiased-cavs}, \cref{fig:dcats-cavs}, and \cref{fig:ddogs-cavs} visualize the three CAVs with the highest SHAP values \cite{shapley1953value} for the \texttt{cat} class in the unbiased model, the dark cats biased model, and the dark dogs biased model, respectively. 
Each CAV is accompanied by the top six images with the highest similarity in the latent space.

Notably, the unbiased model exhibits relatively large positive SHAP values for all three visualized concepts (\cref{fig:unbiased-cavs}). 
In contrast, the biased models (\cref{fig:dcats-cavs} and \cref{fig:ddogs-cavs}) have a third-ranked CAV with a SHAP value close to zero. 
This suggests that two concepts are sufficient to reproduce the outputs of biased models according to \cite{yeh2020completeness}.

However, interpreting the visual explanations semantically is a challenging task, similar to the local methods (\Cref{app:cvd-local-baselines}).
For the unbiased model, we observe multiple cat images, particularly for the concept with the largest SHAP value. 
This may indicate that a broad cat concept is important for the model regarding the \texttt{cat} class. 
In contrast, the dark cats biased model (\cref{fig:dcats-cavs}) does not exhibit any cat images. Instead, we observe many darker images, hinting at the underlying bias of the fur color. 
However, without knowledge of the training setup or additional steps, this is difficult to discover from just the visual explanations of the CAVs.

To gain a deeper understanding, we utilize \cite{schmalwasser2024exploiting} to find textual descriptions for the CAVs found by \cite{yeh2020completeness}.
We list these descriptions in \Cref{tab:cav-words}.
Specifically, we provide the five closest words in CLIP \cite{radford2021learning} latent space and indicate the central word of the corresponding cluster \cite{schmalwasser2024exploiting}.
These descriptions confirm our previous observations. 
The unbiased model and the dark dog-biased model learn a broad cat or kitten concept. 
In contrast, the most important CAV for the dark cats model is described as vehicles. 
Nevertheless, we highlight two findings. 
First, ``darkness'' is part of the five closest words in the third CAV of the dark cats biased model. 
Second, ``white'' is similarly discovered for the second CAV of the dark dogs model.
These results hint again at the underlying bias.

While our local approach is not explorative, it allows us to directly test for specific properties or concepts. 
Hence, it provides an additional tool for investigating model behavior. 
Furthermore, we believe that combining our approach with explorative concept-based methods is promising, as discussed in \Cref{sec:limitations}.

Finally, \cite{reimers2020determining} can detect global behavior changes with respect to a specific property similar to our approach. 
In \Cref{tab:ci-results}, we summarize the binary results for both the fur color and background properties using different conditional independence (CI) tests.
We find strong differences between the CI tests. Specifically, CMIknn \cite{runge2018conditional} always rejects the null hypothesis (property is not used). 
In contrast, conditional HSIC \cite{fukumizu2007kernel} cannot identify the relationship for both properties. 
For partial correlation, we find that the biased models always change behavior for changes in both the fur color and background illumination. 
Taking the majority decision following \cite{reimers2021conditional}, we overall find that all three models learn the fur color, while only the biased models additionally use the background.

While the approach described in \cite{reimers2020determining} enables testing for the usage of arbitrary properties on a global level, we identify two differences to our approach. 
First, our approach allows a direct interpretation of how the outputs change given the local interventions. 
In other words, by utilizing gradual interventions, we can visualize the shift in behavior for specific variations in the property. 
Second, while the global insights in \Cref{tab:ci-results} tell us that background is learned, it is not clear whether it is important for individual inputs. 
With \propgrad, we develop a score to measure the impact of a property under interventions to provide local insights.

Overall, we find that our local approach complements global baselines to investigate how model outputs change for interventions in individual inputs.

\FloatBarrier
\clearpage

\section{ISIC Classification - Additional Details}
\label{app:isic}

\subsection{Setup Details}
\label{app:isic-setup}

To showcase the ability of our approach to measure systematic changes in model prediction behavior in complex settings, we select the real-world task of skin lesion classification.
Specifically, we choose the binary problem to differentiate between healthy skin lesions (nevi) and dangerous melanomata.
As a dataset, we sample an equal amount of both classes from the ISIC archive \cite{isic-archive}.
We train four different architectures: ResNet18 \cite{he2016deep}, EfficientNet-B0 \cite{tan2019efficientnet}, ConvNext-S \cite{liu2022convnet}, and ViT-B/16 \cite{dosovitskiy2020image}.
For each of these datasets, we consider three training datasets: unbiased skin lesion data, biased skin lesion data, and the ImageNet \cite{russakovsky2015imagenet} pre-trained weights.

Regarding the bias, consider that the ISIC archive is a collection of various skin lesion images collected by independent groups and medical researchers.
Hence, there are strong variations depending on the respective data sources.
This includes biases such as the spurious correlation of colorful patches with the class nevus introduced by \cite{scope2016study}.
Specifically, Scope et al. \cite{scope2016study} study nevi in children and apply visually large and distinct colorful patches next to healthy skin lesions.

In our biased training setting, we sample half of the images of class nevus from the set of images containing colorful patches.
In contrast, we exclude these images for our unbiased split.
We proceed similarly for the respective test data, on which we perform our remaining investigation.

\paragraph{Training Hyperparameters: }
Again, we rely on ImageNet \cite{russakovsky2015imagenet} pre-trained weights and normalization statistics.
Hence, we resize the images to an input size of $224\times 224$ during training and inference. 
We apply \cite{muller2021trivialaugment} with the wide augmentation space during training.

We optimize using AdamW \cite{loshchilov2019decoupled}, with a learning rate of 0.0001, weight decay of 0.0005, and momentum of 0.9.
Finally, we train for 50 epochs with a batch size of 32.
The performance of all architecture and training data combinations is contained in \Cref{tab:isic-acc}.
\begin{table}[h]
    \centering
    \caption{Accuracy in percent (\%) of various architectures for melanoma classification trained on data from \cite{isic-archive}.
    We separate the test data into biased data containing colorful patches from \cite{scope2016study} and unbiased data where we sample nevus images without.
    Similarly, the first rotated column indicates the training distribution.    
    }
    \label{tab:isic-acc}
    \begin{tabular}{llcc}
\toprule
                & & \multicolumn{2}{c}{Test Data} \\
\cmidrule(lr){3-4}
           & Model &  Unbiased &    Biased \\
\midrule
\multirow{4}{*}{\rotatebox{90}{Unbiased}} 
    & ResNet18 \cite{he2016deep} & 86.70 & 85.89 \\
    & EfficientNet-B0 \cite{tan2019efficientnet} & 87.15 & 90.12 \\
    & ConvNeXt-S \cite{liu2022convnet} & 88.05 & 88.32 \\
    & ViT-B/16 \cite{dosovitskiy2020image} & 86.25 & 84.37 \\
\cmidrule(lr){2-4}
\multirow{4}{*}{\rotatebox{90}{Biased}} 
    & ResNet18 \cite{he2016deep} & 82.84 & 89.13 \\
    & EfficientNet-B0 \cite{tan2019efficientnet} & 85.09 & 91.37 \\
    & ConvNeXt-S \cite{liu2022convnet} & 83.92 & 92.18 \\
    & ViT-B/16 \cite{dosovitskiy2020image} & 83.47 & 87.42 \\
\bottomrule
\end{tabular}
\end{table}

\paragraph{Colorful Patch Interventions: }
To test how strongly each of the models relies on the colorful patches to derive its prediction, we intervene in images containing melanomata.
Specifically, we randomly sample ten melanoma images correctly classified by both the biased and unbiased networks.
Next, we sample five random images containing colorful patches per melanoma image.
We ensure that these patches are neither part of the training nor the test datasets.
Finally, we use segmentations of the colorful patches provided in \cite{rieger2020interpretations} to alpha blend them with the melanoma images.
In our main paper, we include one example in \cref{fig:isic} (bottom).

We choose synthetic interventions to showcase the ability of our approach to work with diverse sources of interventional data.
This especially holds for expert and domain knowledge, where specifically designed interventions can ensure the correct target, similar to this experiment or \cite{buechner2024facing}.
Hence, our approach facilitates the analysis of complex tasks where it is important to strictly apply the causal hierarchy theorem \cite{pan2024counterfactual}.

\begin{figure*}[t]
\centering
\begin{subfigure}{0.48\textwidth}
\includegraphics[width=\linewidth]{figs/isic_behavior/ResNet18_mean.pdf}
\caption{Resnet18 \cite{he2016deep}: Average of ten colorful patch interventions.}
\label{fig:isic-resnet18-mean}
\end{subfigure}
\begin{subfigure}{0.48\textwidth}
\includegraphics[width=\linewidth]{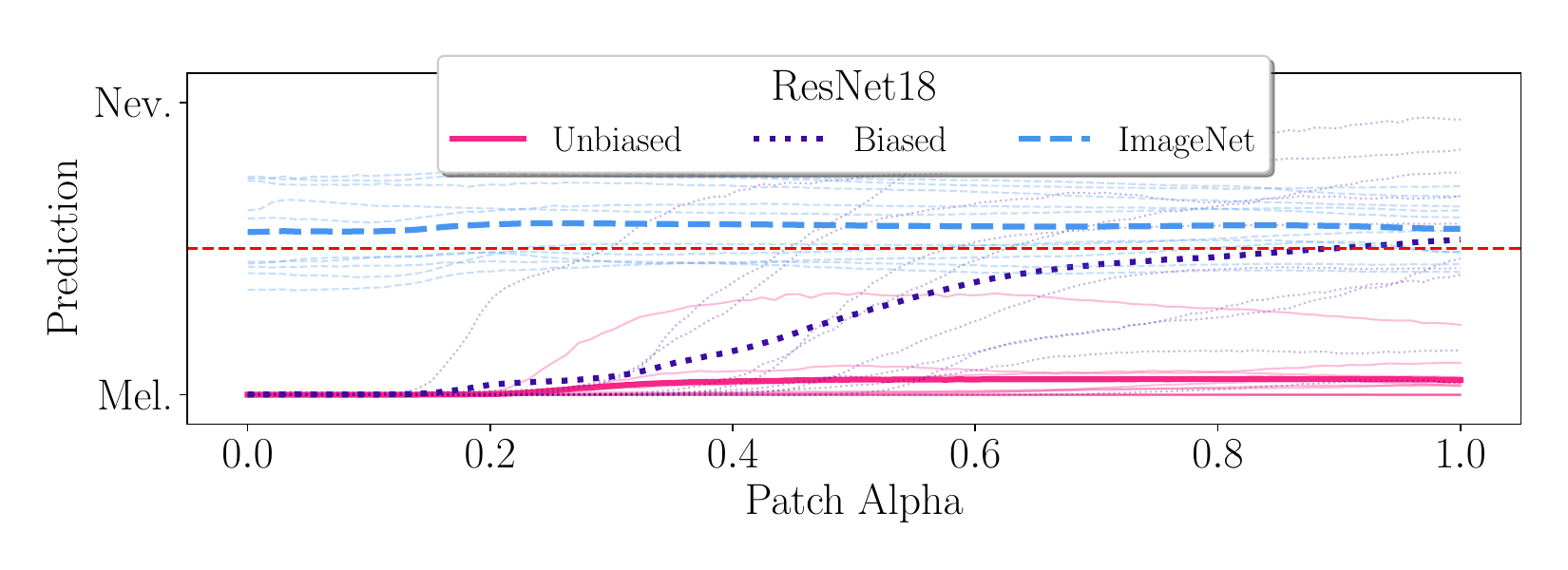}
\caption{Resnet18 \cite{he2016deep}: Ten local colorful patch interventions.}
\label{fig:isic-resnet18-schar}
\end{subfigure}

\begin{subfigure}{0.48\textwidth}
\includegraphics[width=\linewidth]{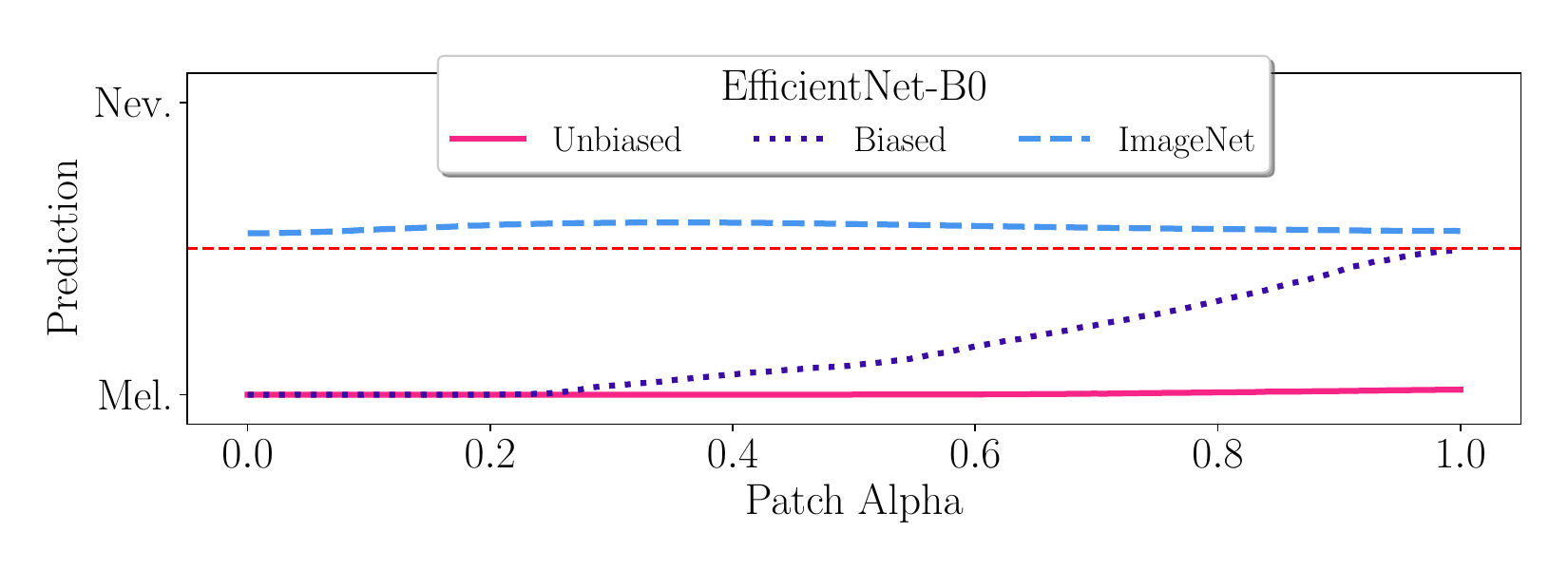}
\caption{EfficientNet-B0 \cite{tan2019efficientnet}: Average of ten colorful patch interventions.}
\label{fig:isic-effnetb0-mean}
\end{subfigure}
\begin{subfigure}{0.48\textwidth}
\includegraphics[width=\linewidth]{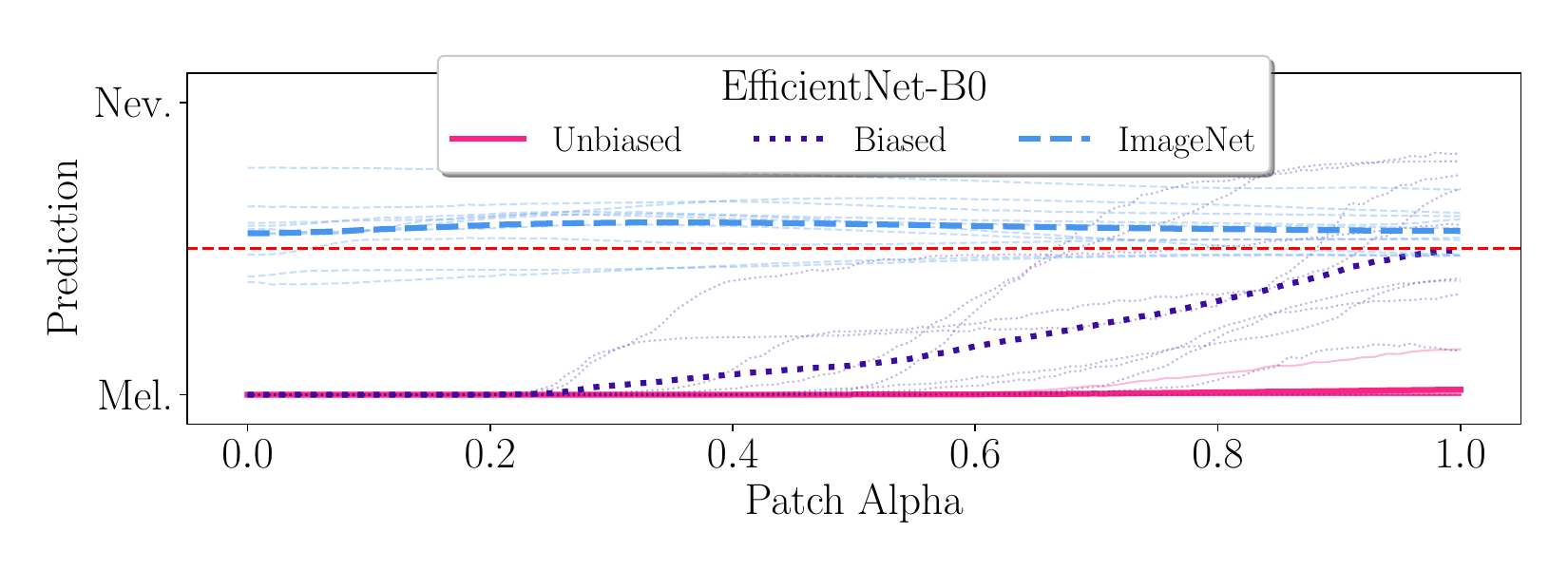}
\caption{EfficientNet-B0 \cite{tan2019efficientnet}: Ten local colorful patch interventions.}
\label{fig:isic-effnetb0-schar}
\end{subfigure}

\begin{subfigure}{0.48\textwidth}
\includegraphics[width=\linewidth]{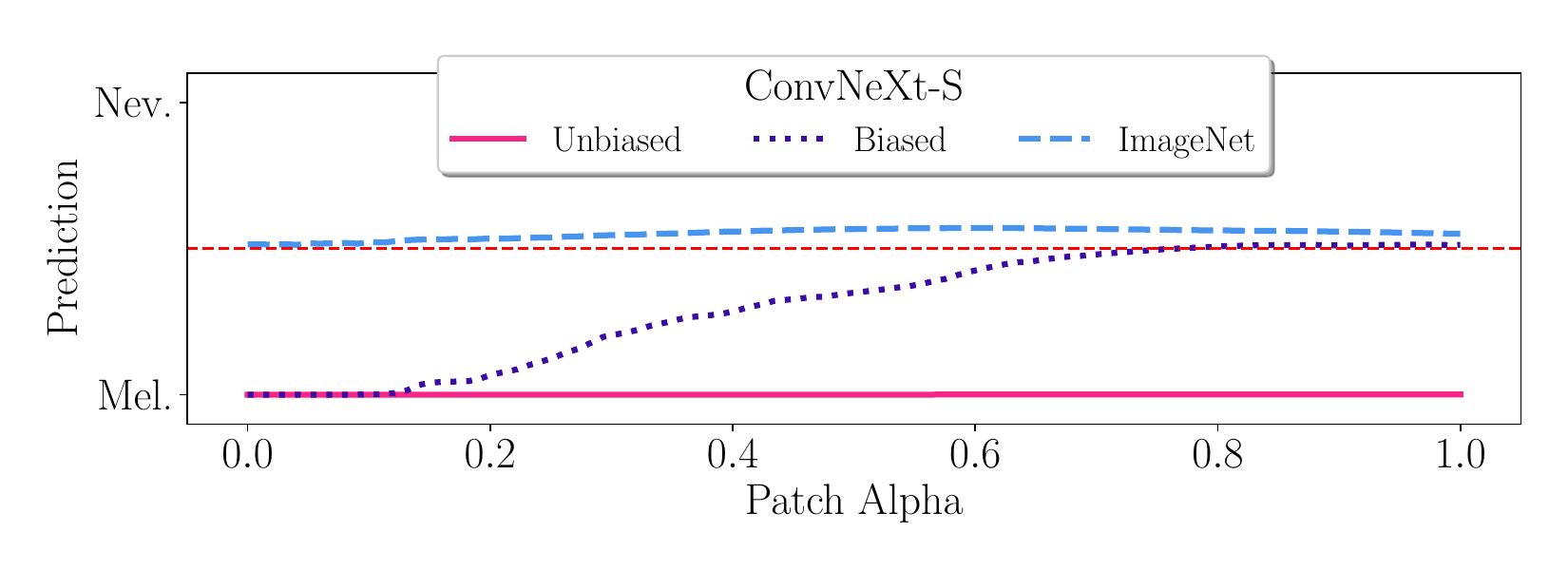}
\caption{ConvNext-S \cite{liu2022convnet}: Average of ten colorful patch interventions.}
\label{fig:isic-convnext-mean}
\end{subfigure}
\begin{subfigure}{0.48\textwidth}
\includegraphics[width=\linewidth]{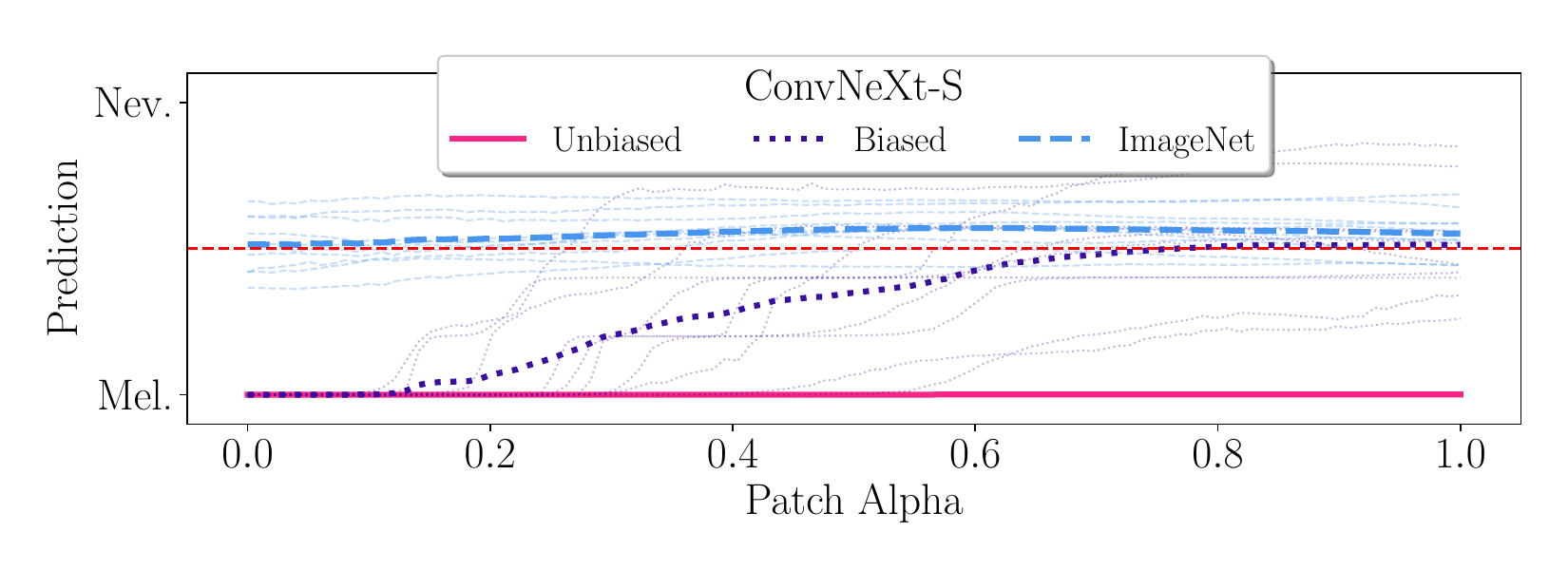}
\caption{ConvNext-S \cite{liu2022convnet}: Ten local colorful patch interventions.}
\label{fig:isic-convnext-schar}
\end{subfigure}

\begin{subfigure}{0.48\textwidth}
\includegraphics[width=\linewidth]{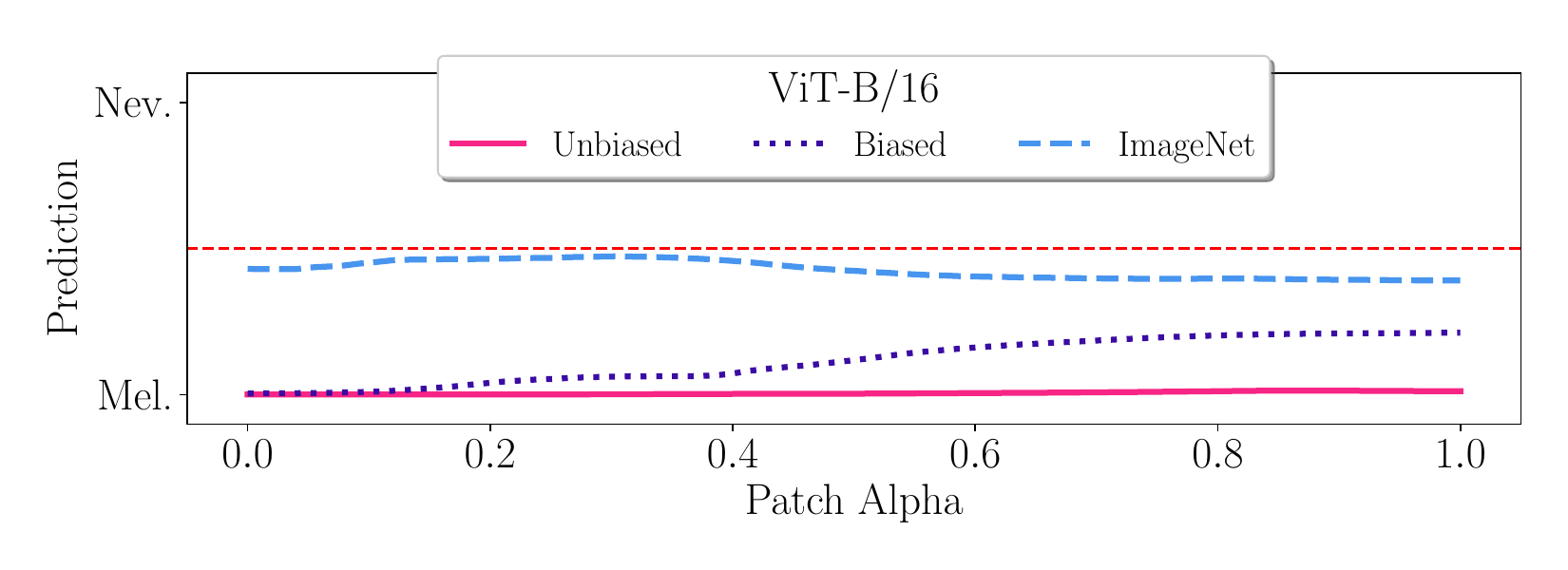}
\caption{ViT-B/16 \cite{dosovitskiy2020image}: Average of ten colorful patch interventions.}
\label{fig:isic-vit16-mean}
\end{subfigure}
\begin{subfigure}{0.48\textwidth}
\includegraphics[width=\linewidth]{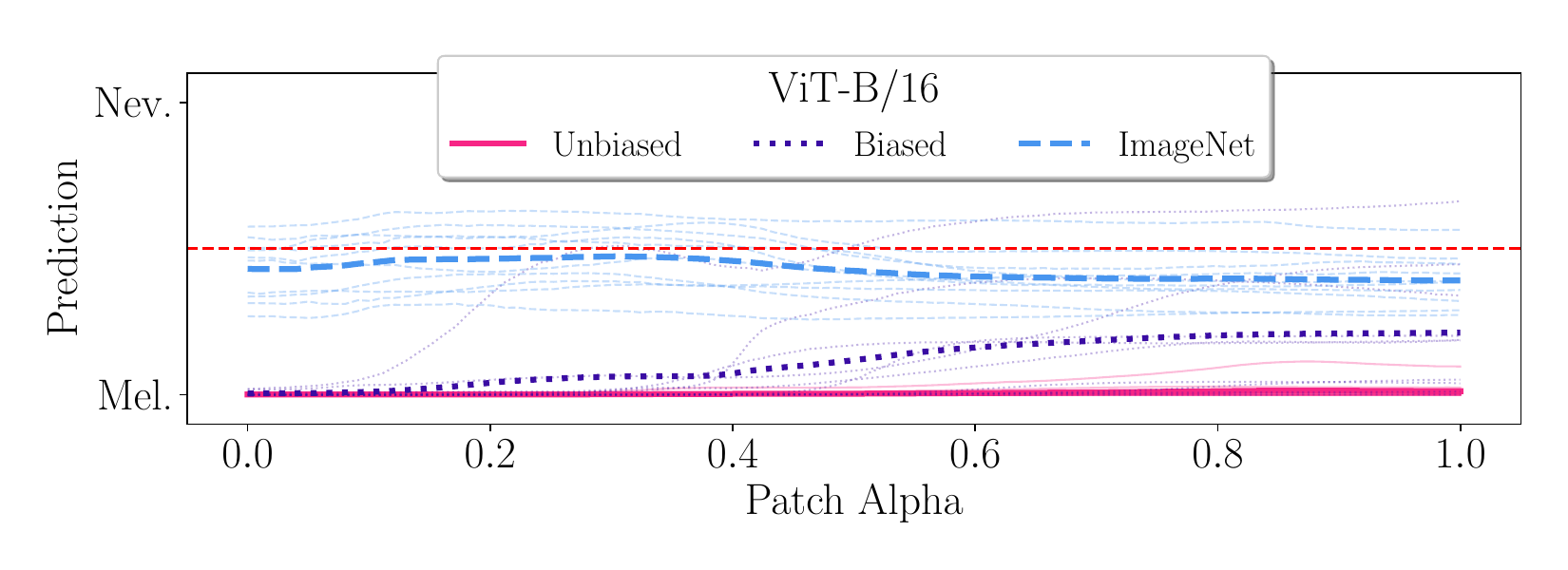}
\caption{ViT-B/16 \cite{dosovitskiy2020image}: Ten local colorful patch interventions.}
\label{fig:isic-vit16-schar}
\end{subfigure}

\caption{
Changes in model predictions of skin lesion classifiers for the intervention of introducing colorful patches \cite{scope2016study}.
We include various architectures and split depending on the training data: unbiased skin lesion images, biased skin lesion images where spurious patches correlate with the class \texttt{nevus}, and ImageNet pre-trained weights.
The \textcolor{red}{red dotted line} is the threshold where the predictions flip.
}
\label{fig:isic-extended}
\end{figure*}

\subsection{Additional Results}
\label{app:isic-results}

\Cref{tab:isic-acc} contains the accuracies for the different training and test splits (biased/unbiased).
There is a clear difference between models trained on the unbiased split and models trained in a biased scenario.
Each of the two paradigms outperforms the other in the test split following their respective training distribution.
However, the models trained on unbiased skin lesions also achieve similarly high performance on the data containing colorful patches.
In contrast, the biased models seem to overfit the training domain and perform drastically worse on the unbiased split.
Here, the ViT \cite{dosovitskiy2020image} achieves the lowest performance on the biased data, which is even outperformed by some of the models trained on the unbiased split.
Nevertheless, the biased ViT \cite{dosovitskiy2020image} loses around 4\% when evaluated on unbiased data, which is the lowest performance drop of all biased models.

This observation is congruent with the estimated \propgrad in \Cref{tab:isic-impact} in our main paper.
Specifically, we measure the lowest expected property gradient magnitude of all biased models for the ViT \cite{dosovitskiy2020image}.
All convolutional models show, on average, a higher impact during the colorful patch interventions.
\cref{fig:isic-extended} visualizes the average for the ten melanoma images with patch interventions and confirms our observation.
The vision transformer is the only architecture where the biased model does not, on average, change its prediction.
This result is an indication that convolutional models learn this spurious visual bias more strongly.
Previous works, e.g., \cite{reimers2021conditional,piater2024medical}, analyze this behavior on an associational level and find significant changes in prediction behavior for colorful patches.
Now, our proposed approach of interventional local explanation enables a more fine-grained analysis.

Next, both in \cref{fig:isic-extended} and \Cref{tab:isic-impact}, we see higher colorful patch impact for the ImageNet \cite{russakovsky2015imagenet} pre-trained models compared to the models trained on unbiased skin lesions.
This is an expected observation, given the large visual changes during the intervention.
Further, the differentiation of large blocks of colors is a useful feature during the pre-training on general-purpose datasets.
In contrast, the models trained on unbiased skin lesions, which are often centered, learn to focus on the actual lesion.
Hence, they are only minimally influenced by the colorful patches.
This result is not obvious from a pure performance analysis (\Cref{tab:isic-acc}).
However, our interventional approach does provide insights into models trained for this complex scenario beyond benchmarking (\Cref{tab:isic-impact}).

\FloatBarrier
\clearpage

\section{CelebA - Additional Details}
\label{app:celebA}

In this section, we provide more details regarding our second experiment, where we study the training dynamics of eight architectures (\Cref{sec:exp2}).
First, \Cref{app:celebA-setup} discusses the corresponding hyperparameters and training details.
Then, we provide additional visualizations in \Cref{app:celeba-results} before finally investigating another local example.

\subsection{Setup Details}
\label{app:celebA-setup}

In our second experiment, we investigate the local training dynamics of eight different architectures: 
ConvMixer \cite{trockman2022patches}, ResNet18 \cite{he2016deep}, EfficientNet-B0 \cite{tan2019efficientnet}, MobileNetV3-L \cite{howard2019searching}, DenseNet121 \cite{huang2017densely}, ConvNeXt-S \cite{liu2022convnet}, ViT-B/16 \cite{dosovitskiy2020image}, and SwinTransformer-S \cite{liu2021swin}. 
We select these architectures to cover a range of model families and design choices and investigate random initialization versus pre-trained weights.

For the ConvMixer model, we use an initial patch size of 14, a depth of 20, kernels with a width of 9, and a latent representation size of 1024. 
These hyperparameters are specifically chosen to utilize the ImageNet pre-trained weights included in \cite{wightman2019timm}. 
For specifics regarding these parameters, we refer the reader to the original paper \cite{trockman2022patches}. 
For all other architectures, we rely on the standard PyTorch \cite{paszke2019pytorch} implementation and parameterizations.

During training and inference, we resize the images to an input size of $224\times 224$. 
For pre-trained models, we use the ImageNet \cite{russakovsky2015imagenet} statistics for normalization. 
In contrast, for random initializations, we normalize the values in the interval of $[-1,1]$. Additionally, we utilize \cite{muller2021trivialaugment} with the wide augmentation space during training, irrespective of the initialization. 
We optimize the models using AdamW \cite{loshchilov2019decoupled}, setting the learning rate to 0.0001, weight decay to 0.0005, and momentum to 0.9.
For all models, we employ a batch size of 32.

\begin{table}[t]
\centering
\caption{
Final accuracies in percent (\%) achieved by various models trained to differentiate young versus old in CelebA \cite{liu2015faceattributes}. 
We split between ImageNet \cite{russakovsky2015imagenet} pre-training (``PT'') and random initialization (``RI'') and calculate the performance delta ($\Delta$).
}
\label{tab:celebA-accs}
\begin{tabular}{lccc}
\toprule
Model & PT & RI & Perf. $\Delta$ \\
\midrule
ConvMixer \cite{trockman2022patches} & 86.08 & 81.49 & -4.59 \\
ResNet18 \cite{he2016deep} & 85.37 & 84.38 & -0.99 \\
EfficientNet-B0 \cite{tan2019efficientnet} & 86.61 & 84.14 & -2.46 \\
MobileNetV3-L \cite{howard2019searching} & 85.94 & 83.05 & -2.88 \\
DenseNet121 \cite{huang2017densely} & 85.75 & 84.20 & -1.55 \\
ConvNeXt-S \cite{liu2022convnet} & 85.98 & 83.63 & -2.35 \\
ViT-B/16 \cite{dosovitskiy2020image} & 85.51 & 72.49 & -13.02 \\
SwinT-S \cite{liu2021swin} & 85.72 & 50.25 & -35.47 \\
\bottomrule
\end{tabular}
\end{table}

After each of the 100 training epochs, we save the model weights, with the final weights achieving the performances disclosed in \Cref{tab:celebA-accs}.
Note that the attention-based models show a stronger decrease in performance for randomly initialized weights.
Especially, the SwinTransformer-S \cite{liu2021swin} diverges to random guessing capabilities.
This observation confirms other works, e.g., \cite{dosovitskiy2020image,piater2024medical}, which find that transformer architectures depend heavily on pre-training.
The divergence is also visible in our property analysis (see \cref{fig:impact-random-init} and \Cref{tab:celebA-impact-ext}), which we will discuss in the next section.

\begin{figure*}
    \centering
    \includegraphics[width=\linewidth]{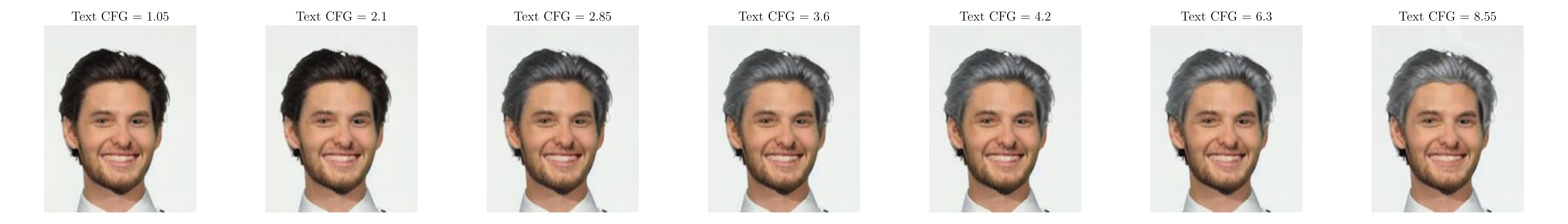}
    \includegraphics[width=\linewidth]{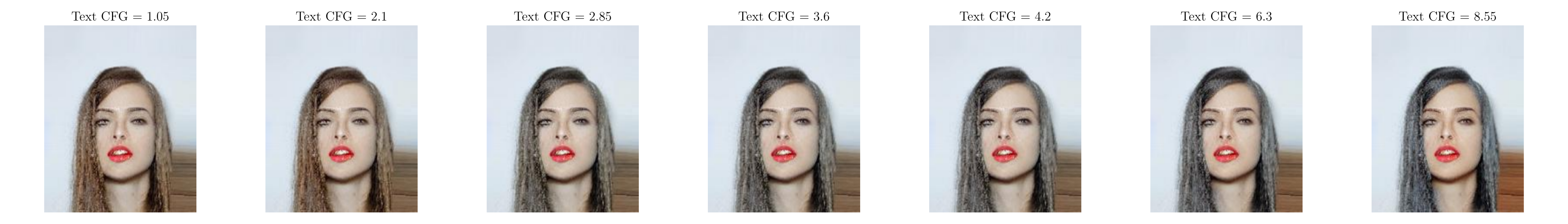}
    \caption{
    Hair color interventions for two people labeled as young.
    Here, we use a pre-trained version of \cite{fu2024mgie}.
    Increasing the CFG text scale indicates a higher alignment with the edit instruction, here ``change the hair to gray-white color''.
    The top row is extensively discussed in the main section of the paper.
    We include additional results about the bottom row in the supplementary material.
    }
    \label{fig:hair-series}
\end{figure*}

\subsection{Additional Results}
\label{app:celeba-results}

\begin{figure*}[t]
    \centering
    \includegraphics[width=\linewidth]{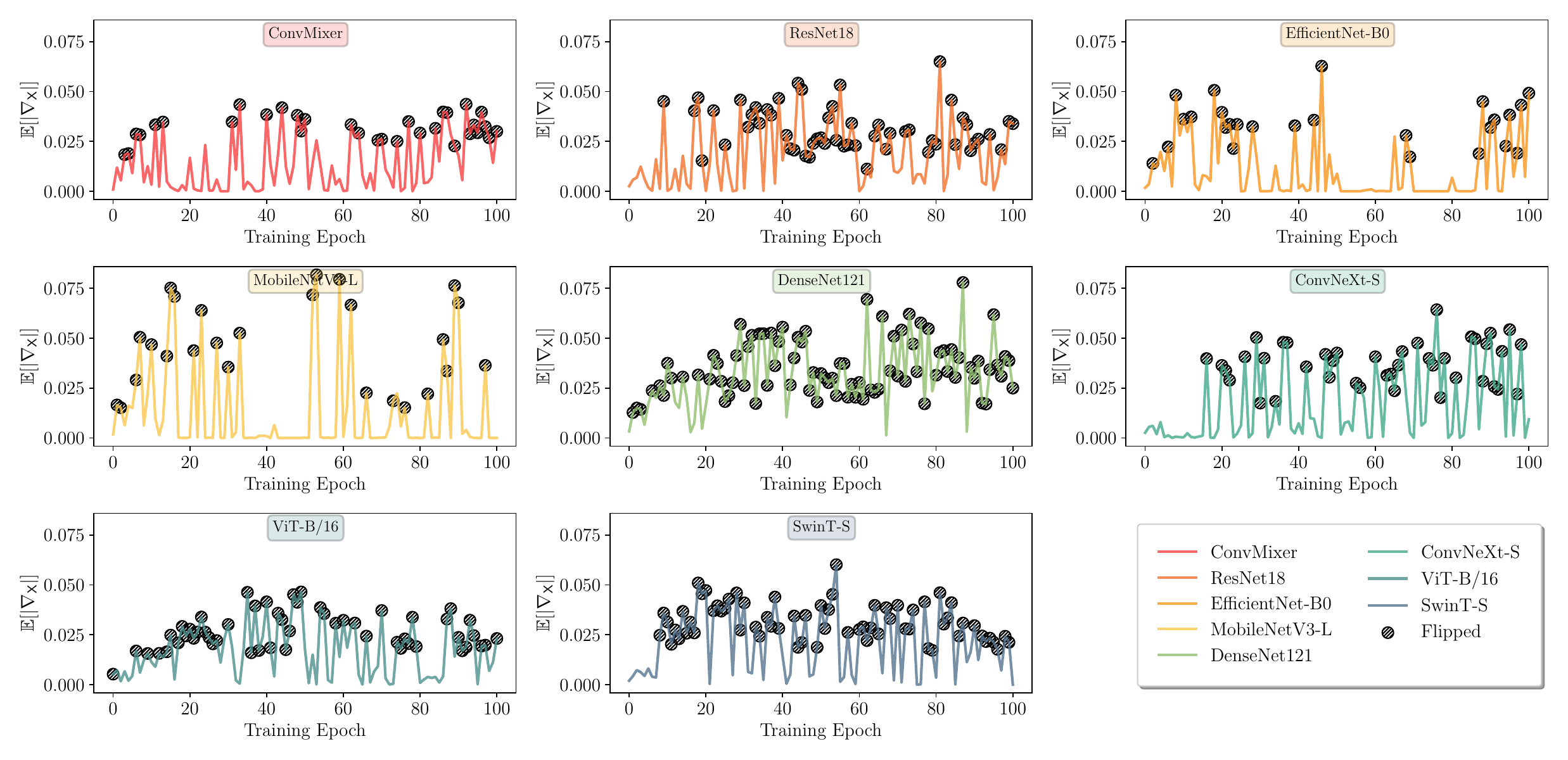}
    \caption{Visualization of how the impact of gray hair changes over the training for all models in \Cref{tab:celebA-accs}, and top row in \cref{fig:hair-series}.
    Here, we focus on ImageNet pre-trained \cite{russakovsky2015imagenet} weights, marking epochs where the network prediction flips during the intervention.}
    \label{fig:impact-pre-trained}
\end{figure*}

\begin{figure*}[t]
    \centering
    \includegraphics[width=\linewidth]{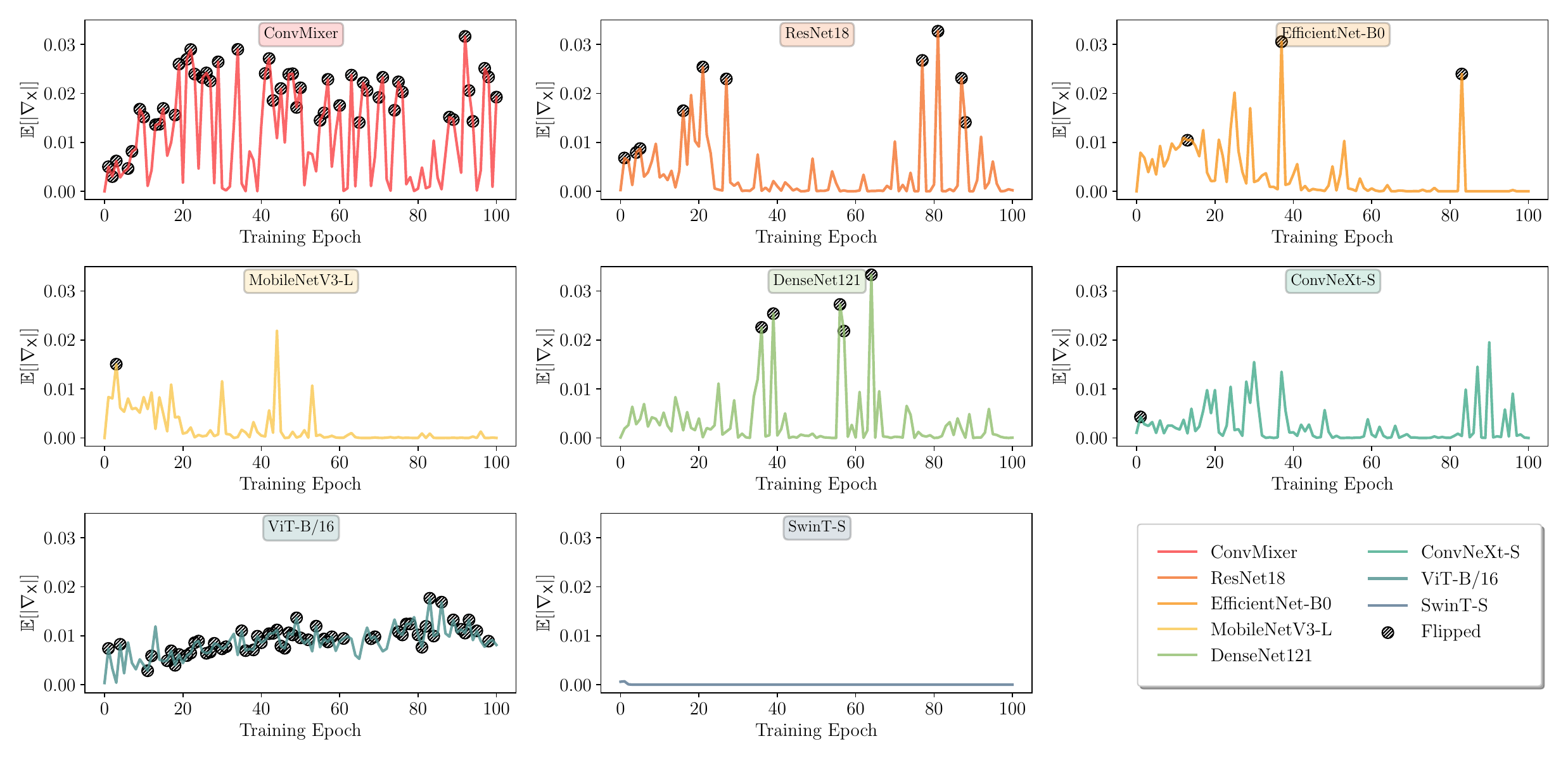}
    \caption{Visualization of how the impact of gray hair changes over the training for all models in \Cref{tab:celebA-accs}, and top row in \cref{fig:hair-series}.
    Here, we focus on randomly initialized weights, mark epochs, where the network prediction flips during the intervention.}
    \label{fig:impact-random-init}
\end{figure*}

\begin{table*}[t]
    \centering
    \caption{
    Average \propgrad for the gray hair color feature over the training process. 
    We list both ImageNet \cite{russakovsky2015imagenet} pre-trained and randomly initialized models.
    In \textbf{bold}, we highlight a greater measured \propgrad for any architecture. 
    Similarly, we use \underline{underline} for more flipped predictions.
    We also denote the number of significant $p$-values ($p<0.01$) over the training following \Cref{alg:sig}.
    Note that the maximum is 101 because we test the initial model and one model after each of the 100 epochs.
    Lastly, we determine the average Pearson correlation coefficient \cite{pearson1895notes} ($\rho$) over the training.
    However, $\rho$ is not defined for constants, leading to missing values for some models.}
    \label{tab:celebA-impact-ext}
    \begin{tabular}{lcccccccc}
    \toprule
    & \multicolumn{4}{c}{Pre-trained} & \multicolumn{4}{c}{Random Init} \\
    \cmidrule(lr){2-5} \cmidrule(lr){6-9}
 Model & \propgrad & \#Flips & \#Sig. & $\rho$ & \propgrad & \#Flips &  \#Sig.  & $\rho$ \\
    \midrule
    ConvMixer \cite{trockman2022patches}
            & \textbf{0.01453} & 32 & 101 & -0.73037
            & 0.01144 & \underline{49} & 100 & ---\\ 
    ResNet18 \cite{he2016deep}
            & \textbf{0.02006} & \underline{55} & 101 & -0.71550
            & 0.00376 & 10 & 101 & -0.55702\\
    EfficientNet-B0 \cite{tan2019efficientnet} 
            & \textbf{0.01153} & \underline{26} & 95 & -0.63042
            & 0.00336 & 3 & 98 & ---\\
    MobileNetV3-L \cite{howard2019searching}  
            & \textbf{0.01379} & \underline{26} & 78 & -0.32585
            & 0.00208 & 1 & 95 & --- \\
    DenseNet121 \cite{huang2017densely}
            & \textbf{0.03194} & \underline{86} & 101 & -0.81263
            & 0.00329 & 5 & 101 & -0.43834\\
    ConvNeXt-S \cite{liu2022convnet}
            & \textbf{0.01801} & \underline{42} & 101 & -0.67621
            & 0.00244 & 1 & 101 & -0.16601 \\
    ViT-B/16 \cite{dosovitskiy2020image}
            & \textbf{0.01780} & 54 & 101 & -0.74494
            & 0.00853 & \underline{55} & 101 & -0.75000 \\
    SwinT-S \cite{liu2021swin}
            & \textbf{0.02268} & \underline{67} & 101 & -0.78029
            & 0.00001 & 0 & 19 & ---\\
    \bottomrule
    \end{tabular}
\end{table*}

\begin{figure*}
    \centering
    \begin{subfigure}{0.49\textwidth}
        \includegraphics[width=\linewidth]{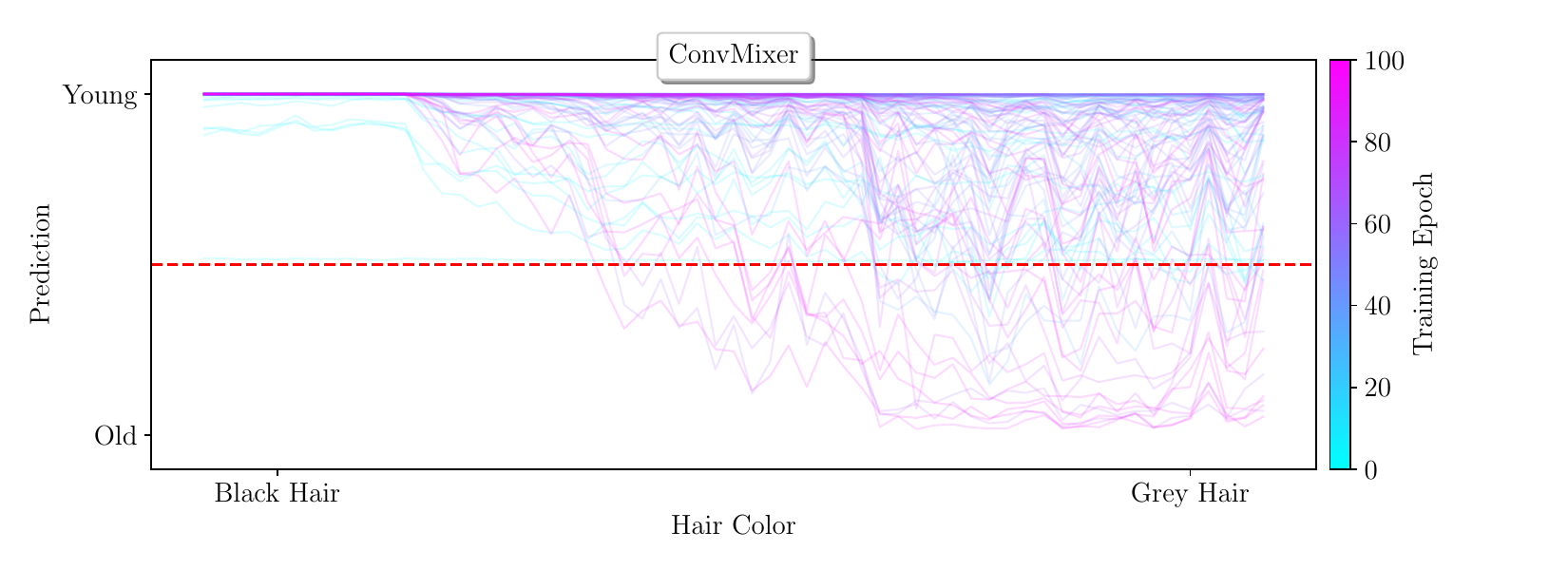}
    \end{subfigure}
    \begin{subfigure}{0.49\textwidth}
        \includegraphics[width=\linewidth]{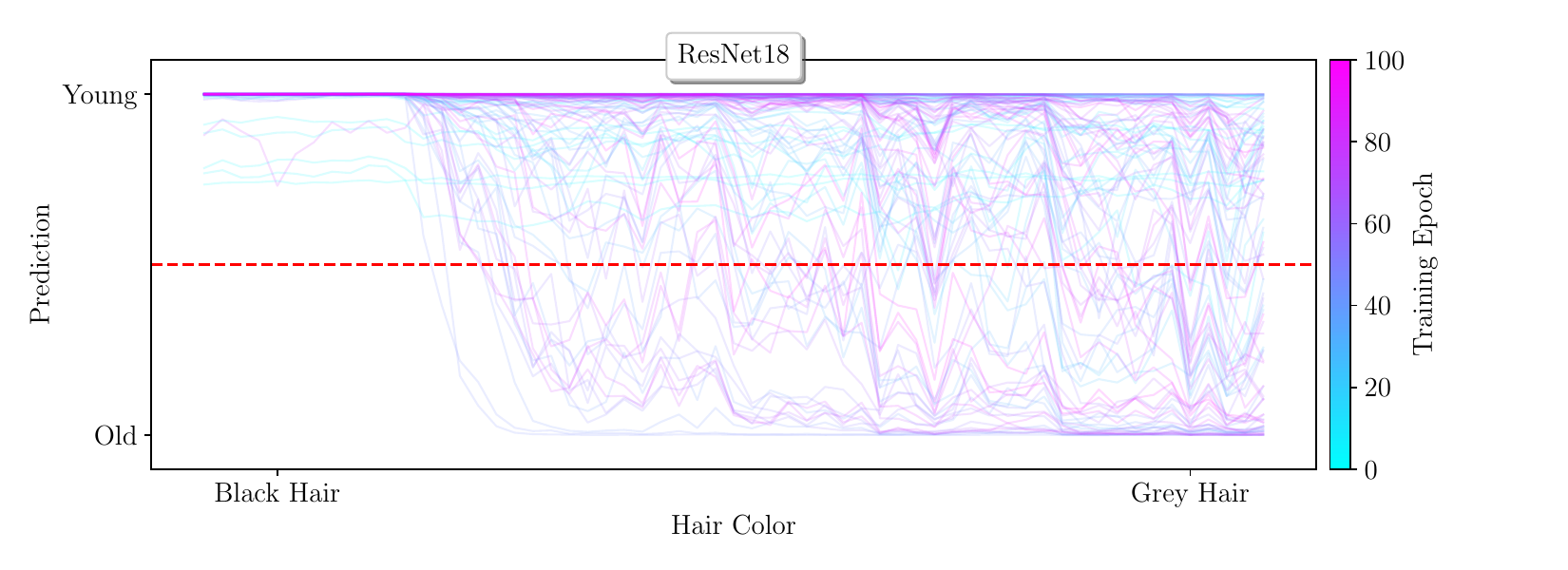}
    \end{subfigure}
    \begin{subfigure}{0.49\textwidth}
        \includegraphics[width=\linewidth]{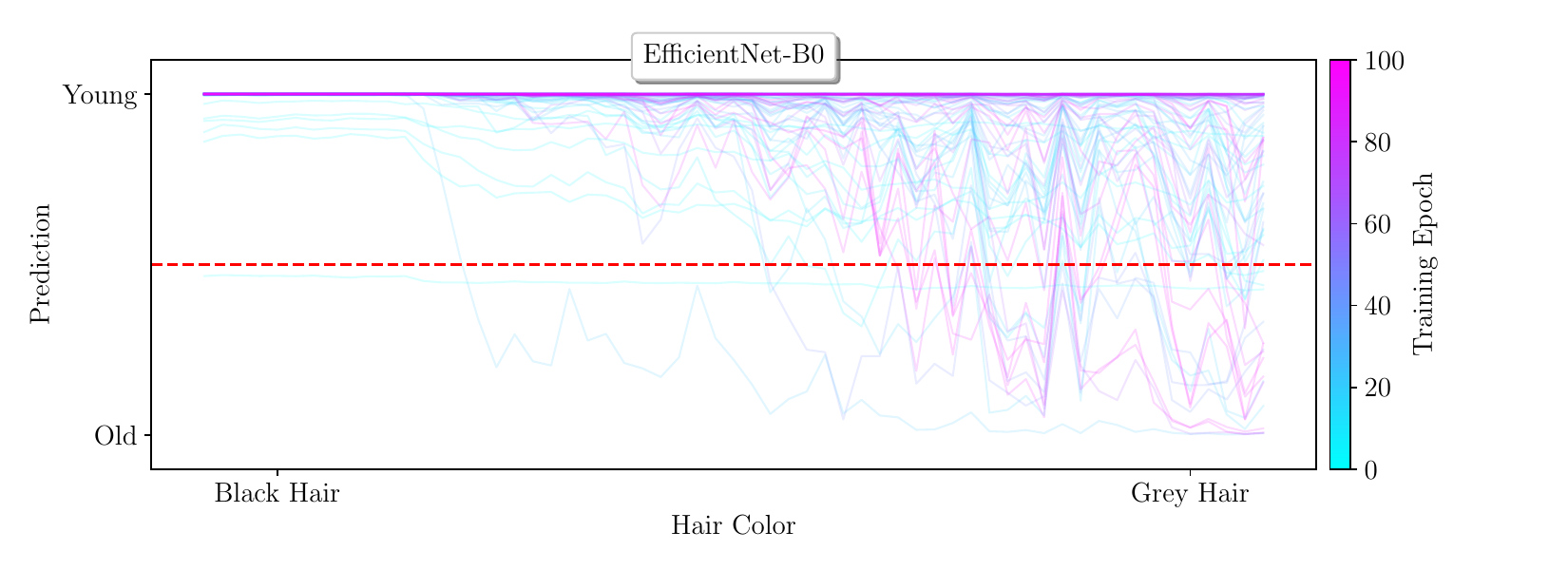}
    \end{subfigure}
    \begin{subfigure}{0.49\textwidth}
        \includegraphics[width=\linewidth]{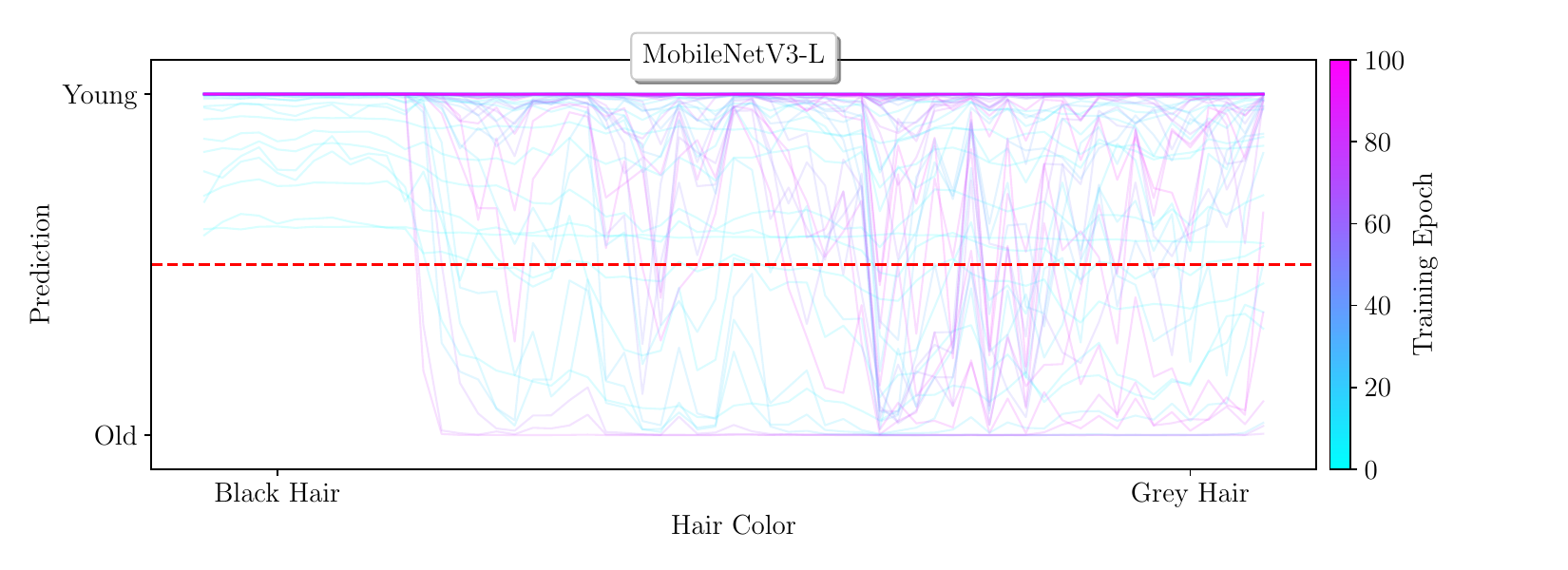}
    \end{subfigure}
    \begin{subfigure}{0.49\textwidth}
        \includegraphics[width=\linewidth]{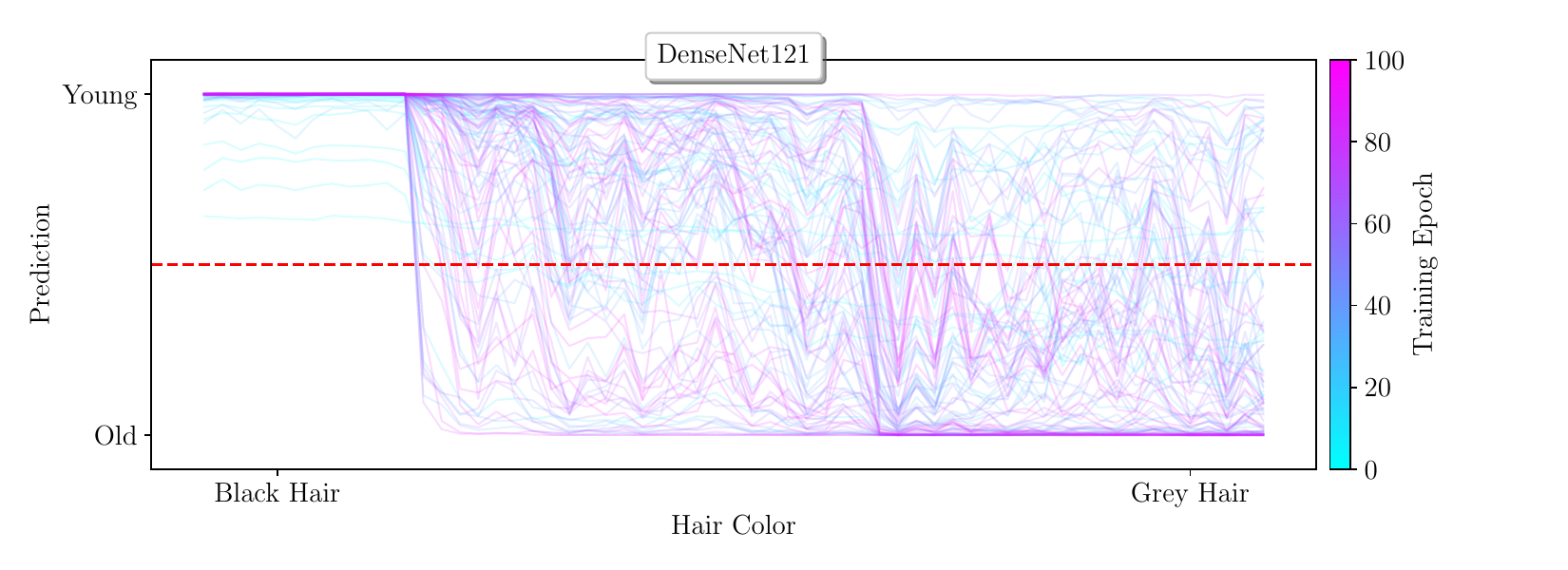}
    \end{subfigure}
    \begin{subfigure}{0.49\textwidth}
        \includegraphics[width=\linewidth]{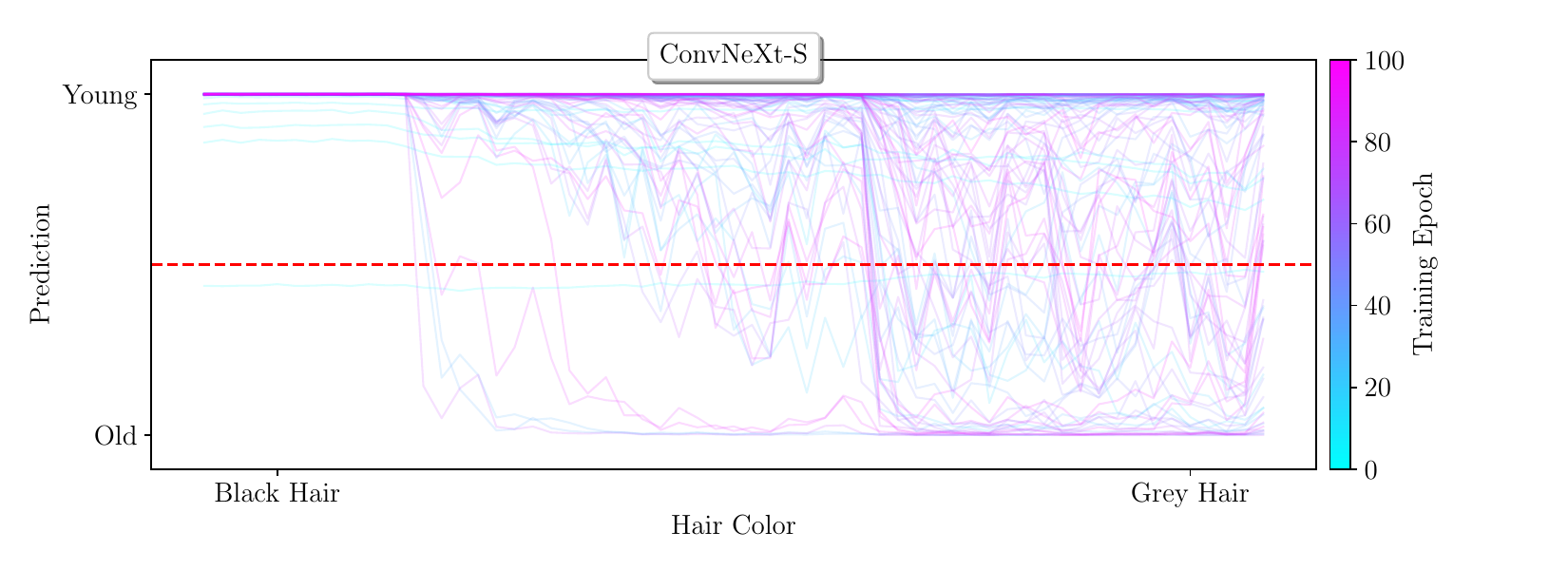}
    \end{subfigure}
    \begin{subfigure}{0.49\textwidth}
        \includegraphics[width=\linewidth]{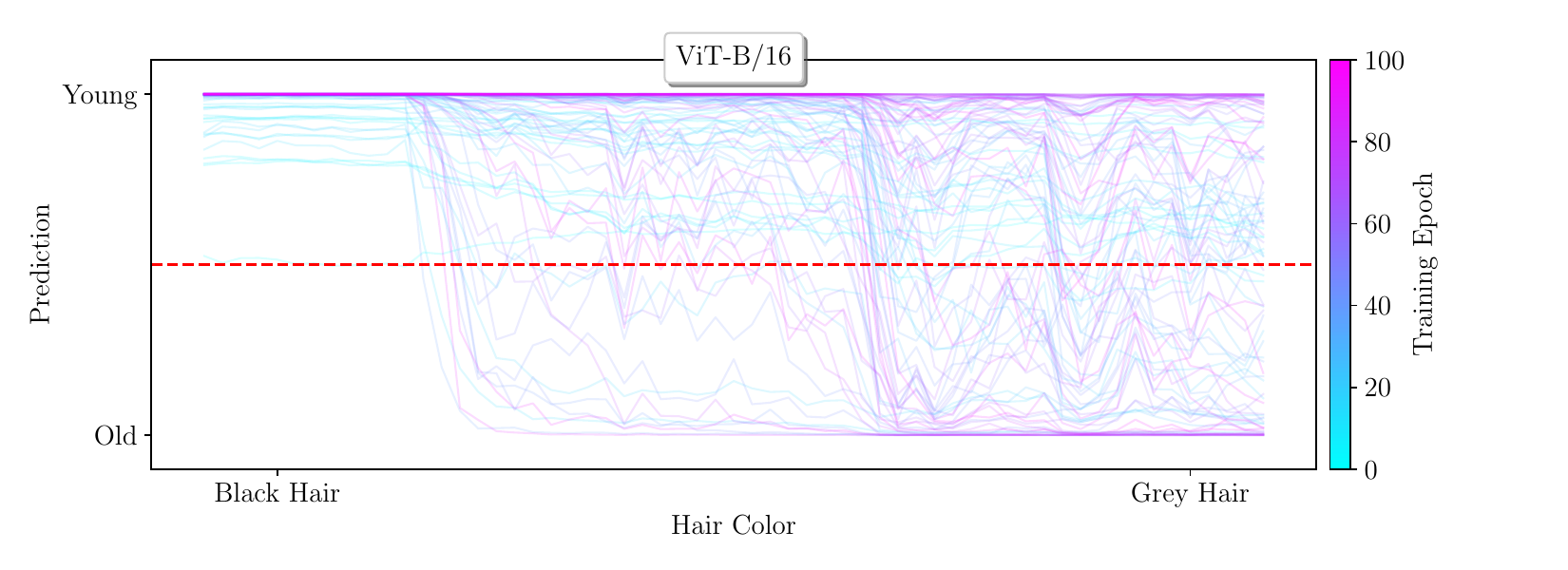}
    \end{subfigure}
    \begin{subfigure}{0.49\textwidth}
        \includegraphics[width=\linewidth]{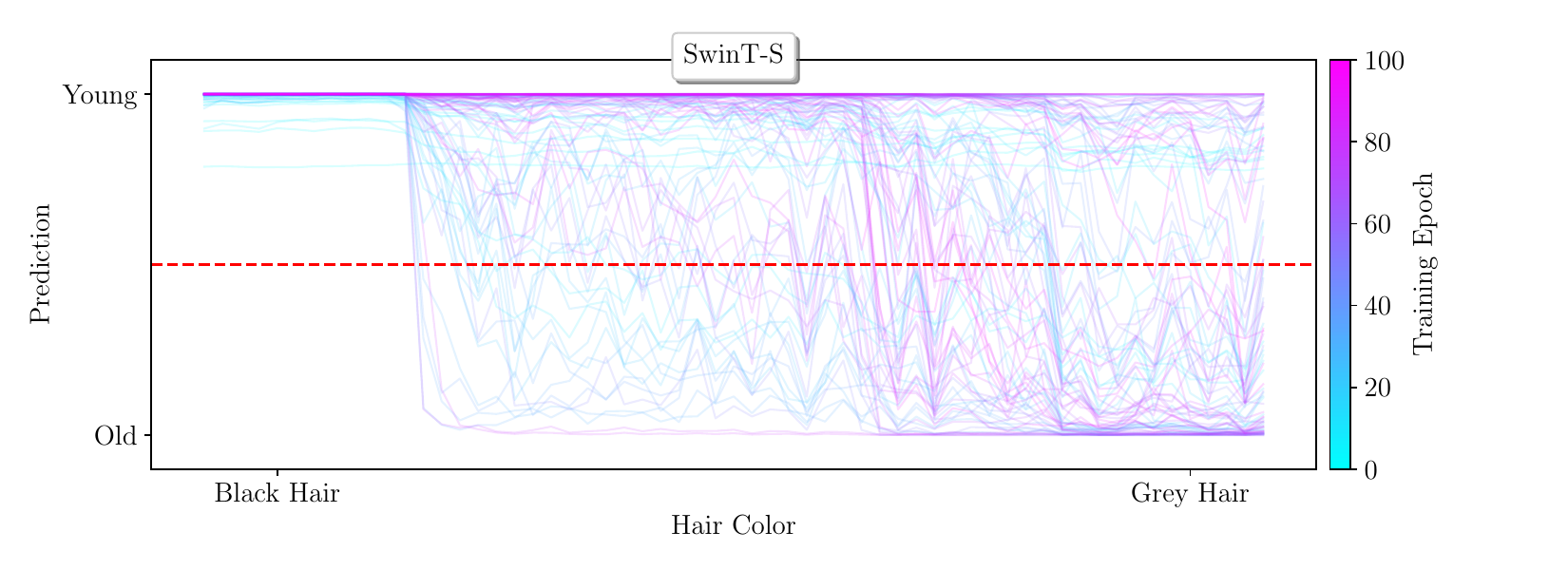}
    \end{subfigure}
    \caption{Model behavior per epoch visualized for the hair color intervention shown in the top row of \cref{fig:hair-series} and discussed in the main part of the paper.
    Here, we only show the behavior for \textbf{pre-trained models} using ImageNet \cite{russakovsky2015imagenet} weight.
    The models include ConvMixer \cite{trockman2022patches}, ResNet18 \cite{he2016deep}, EfficientNet-B0 \cite{tan2019efficientnet}, MobileNetV3-L \cite{howard2019searching}, DenseNet121 \cite{huang2017densely}, ConvNeXt-S \cite{liu2022convnet}, ViT-B/16 \cite{dosovitskiy2020image}, and SwinTransformer-S \cite{liu2021swin}.
    The \textcolor{red}{red dotted line} indicates, in all cases, the threshold where the model prediction flips.}
    \label{fig:training-vis}
\end{figure*}

\begin{figure*}
    \centering
    \begin{subfigure}{0.49\textwidth}
        \includegraphics[width=\linewidth]{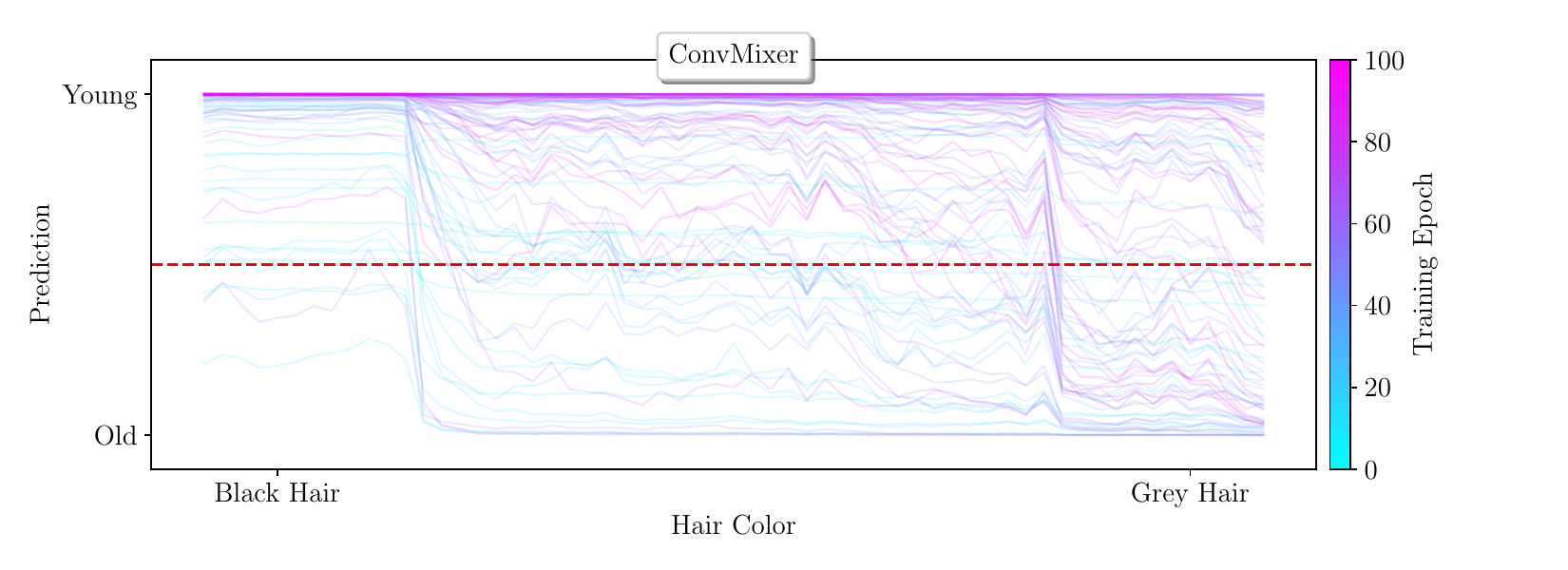}
    \end{subfigure}
    \begin{subfigure}{0.49\textwidth}
        \includegraphics[width=\linewidth]{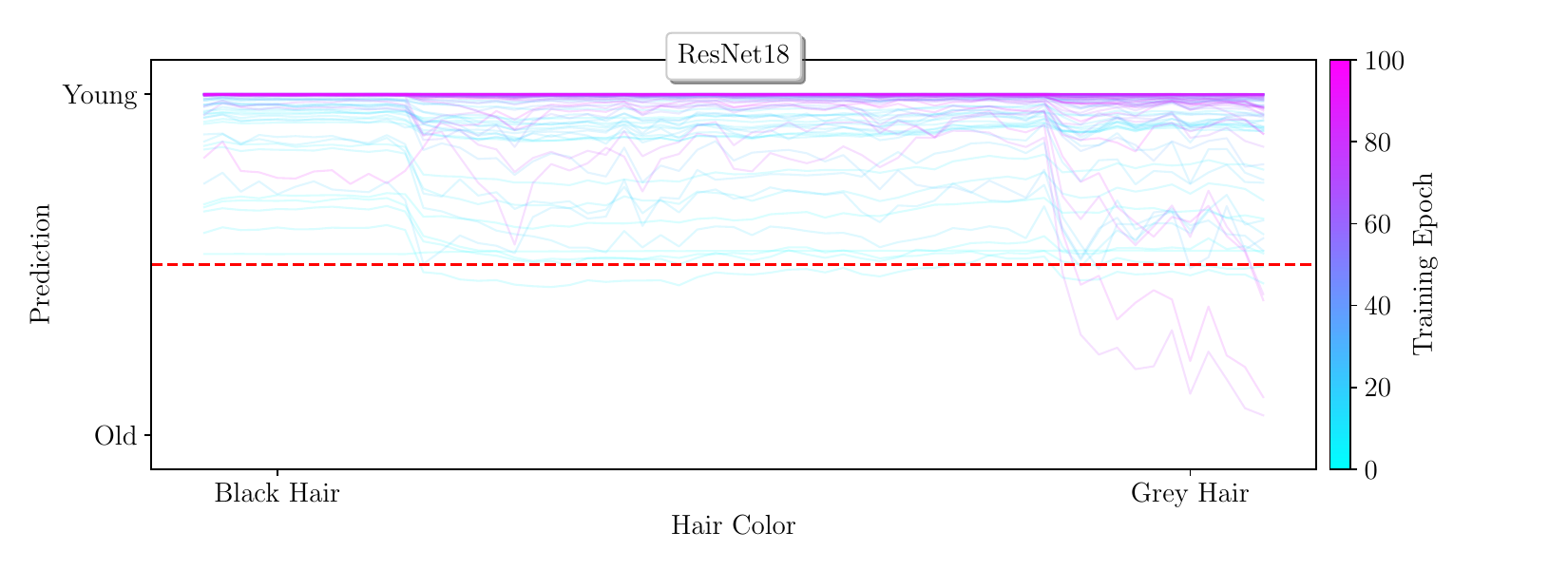}
    \end{subfigure}
    \begin{subfigure}{0.49\textwidth}
        \includegraphics[width=\linewidth]{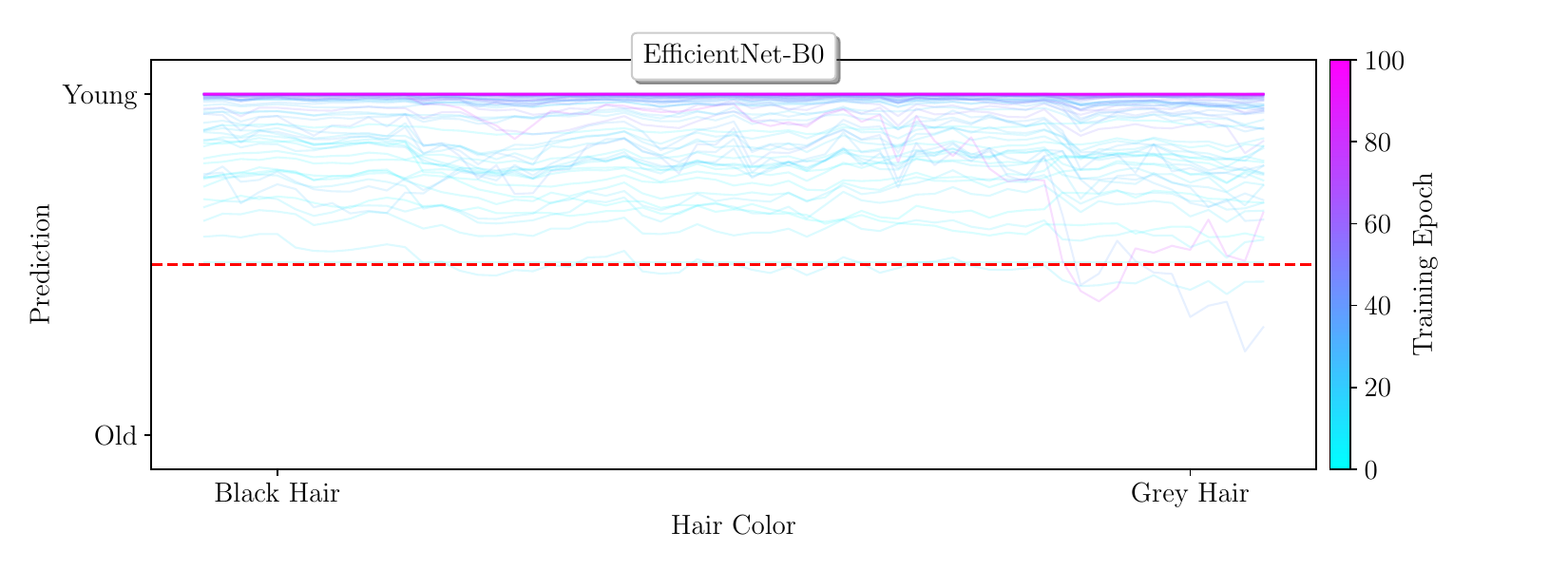}
    \end{subfigure}
    \begin{subfigure}{0.49\textwidth}
        \includegraphics[width=\linewidth]{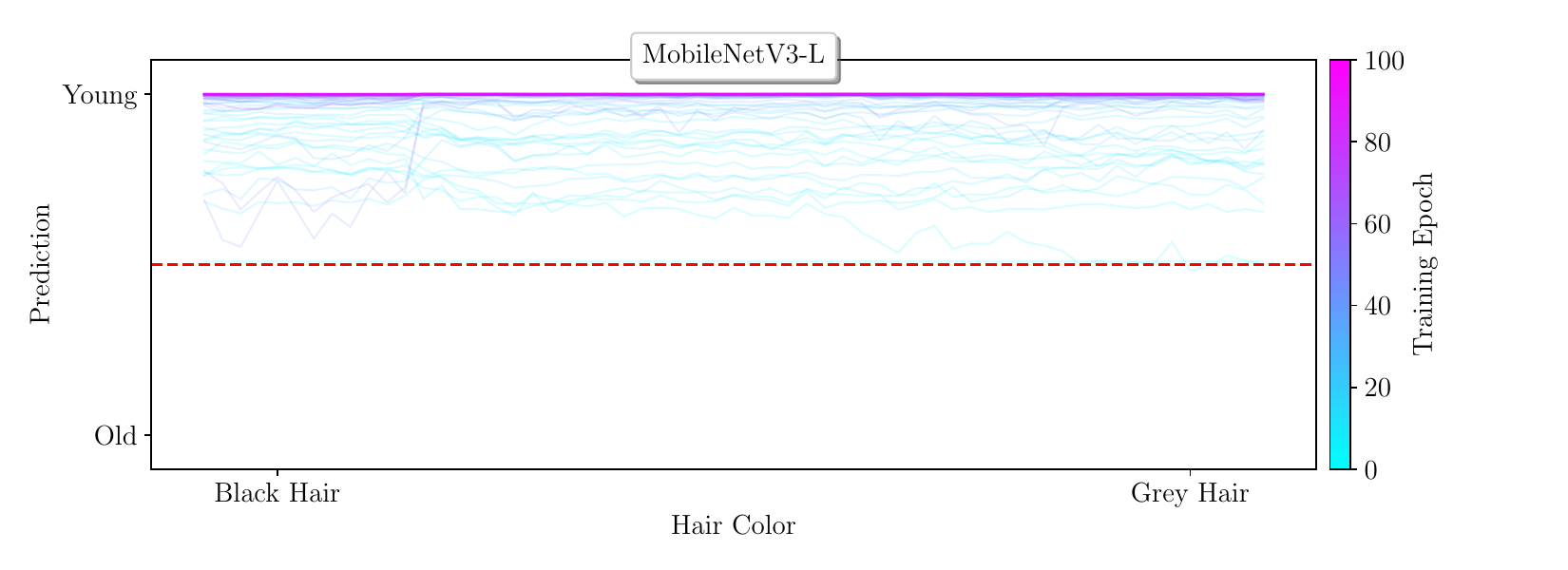}
    \end{subfigure}
    \begin{subfigure}{0.49\textwidth}
        \includegraphics[width=\linewidth]{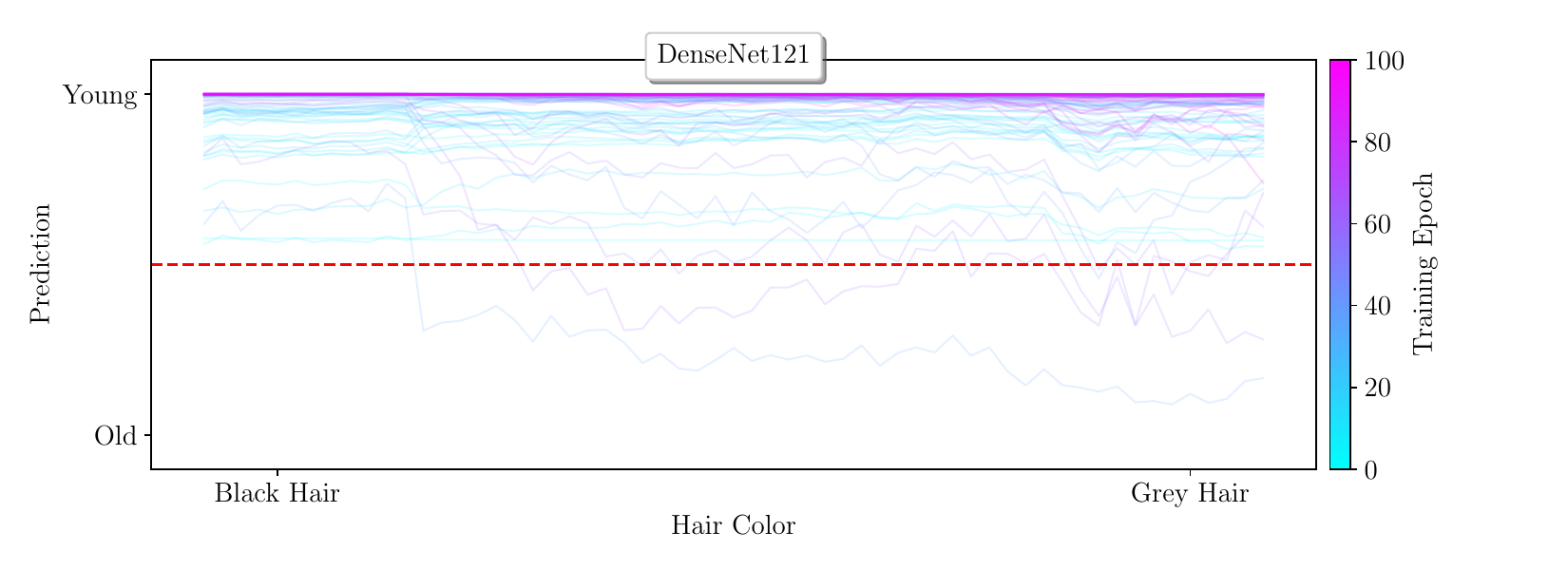}
    \end{subfigure}
    \begin{subfigure}{0.49\textwidth}
        \includegraphics[width=\linewidth]{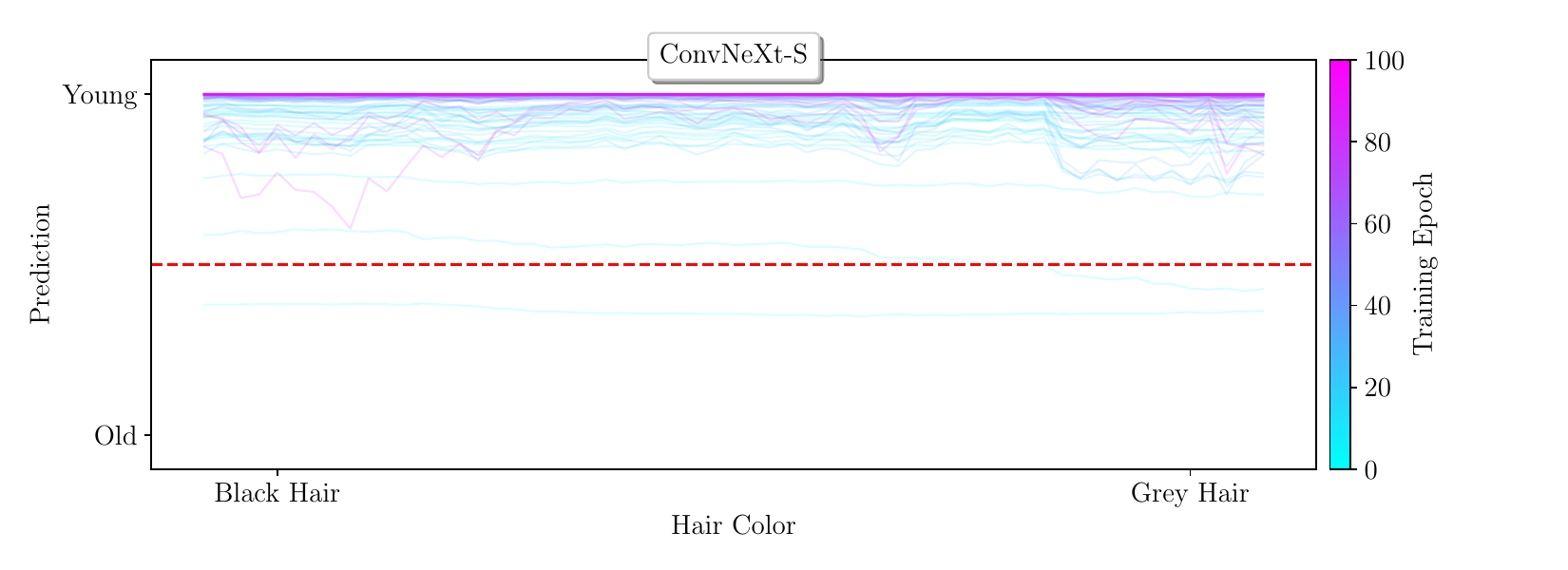}
    \end{subfigure}
    \begin{subfigure}{0.49\textwidth}
        \includegraphics[width=\linewidth]{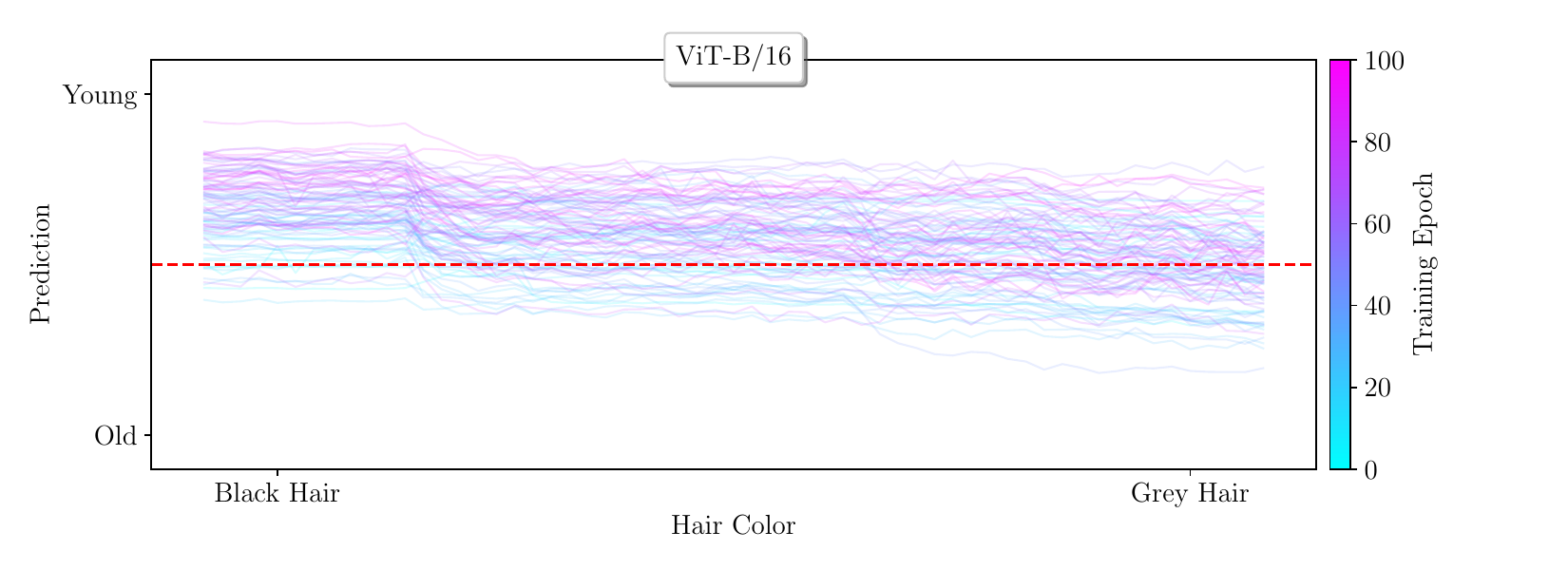}
    \end{subfigure}
    \begin{subfigure}{0.49\textwidth}
        \includegraphics[width=\linewidth]{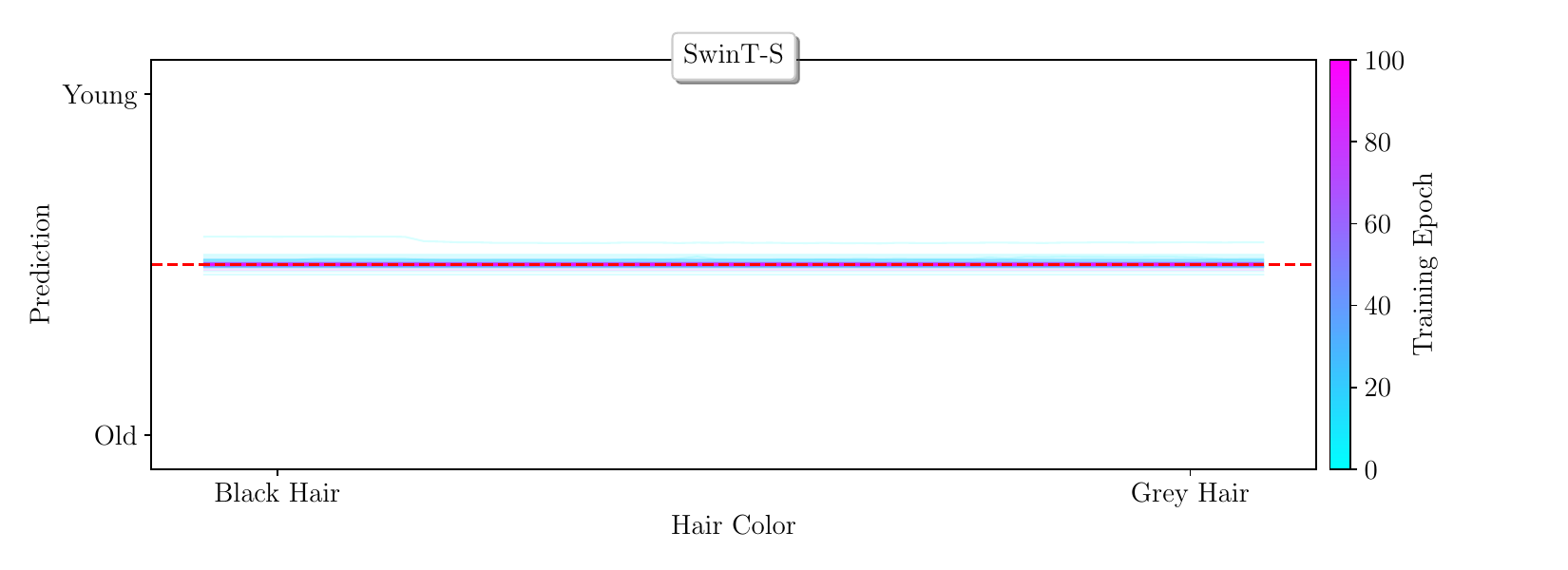}
    \end{subfigure}
    \caption{Model behavior per epoch visualized for the hair color intervention shown in the top row of \cref{fig:hair-series} and discussed in the main part of the paper.
    Here, we only show the behavior for \textbf{randomly initialized} models.
    The models include ConvMixer \cite{trockman2022patches}, ResNet18 \cite{he2016deep}, EfficientNet-B0 \cite{tan2019efficientnet}, MobileNetV3-L \cite{howard2019searching}, DenseNet121 \cite{huang2017densely}, ConvNeXt-S \cite{liu2022convnet}, ViT-B/16 \cite{dosovitskiy2020image}, and SwinTransformer-S \cite{liu2021swin}.
    The \textcolor{red}{red dotted line} indicates, in all cases, the threshold where the model prediction flips.}
    \label{fig:training-vis-ri}
\end{figure*}

We provide the visualizations of the interventional data corresponding to the results in the main paper in \cref{fig:hair-series} (top row).
Specifically, we again use \cite{fu2024mgie} with a CFG image scale of 2.5 and increase the corresponding text scale starting from 1.05 up to 9.75.
Here, we find that higher scales result in unwanted artifacts beyond the targeted intervention.
As an editing phrase, we employ ``change the hair to gray-white color''.
In this section, we provide the corresponding analysis for the other model architectures omitted in the main text and concrete measurements for the average \propgrad.

In \cref{fig:impact-pre-trained} and \cref{fig:impact-random-init}, we present the development of the \propgrad during training for the local hair color intervention across all eight architectures in our analysis. 
We split the visualizations between ImageNet \cite{russakovsky2015imagenet} pre-trained (\cref{fig:impact-pre-trained}) and randomly initialized weights (\cref{fig:impact-random-init}). 
These results strongly support our previous observations.

Notably, there is a stark difference between the pre-trained models and the randomly initialized versions.
The former strongly learn the hair color property and change their behavior based on the intervention, whereas the latter show mostly lower \propgrad scores.
We highlight the different scales for the respective $y$-axes. Additionally, we draw attention to the results for the SwinTransformer-S \cite{liu2021swin}. 
While the pre-trained model often strongly learns the hair color property, the randomly initialized variant diverges completely (\Cref{tab:celebA-accs}). 
This observation is further corroborated by the average \propgrad in \Cref{tab:celebA-impact-ext}.

We also calculate the corresponding Pearson correlation coefficients \cite{pearson1895notes}, which we include in \Cref{tab:celebA-impact-ext}. 
Note that in all cases where we can calculate the correlation, we measure lower effect strength for the randomly initialized models. 
This provides further evidence of the effectiveness of our approach in capturing the strength of the behavior changes under gradual interventions.

To gain further insight into the learned behavior, we provide visualizations similar to our behavior plots (e.g., \cref{fig:1d-results}), where we show the output changes under the interventions after every epoch during training. 
In \cref{fig:training-vis} and \cref{fig:training-vis-ri}, we show the pre-trained and randomly initialized models, respectively.

We highlight two key observations.
First, for both model variants, we observe that the selected example is nearly always correctly classified. 
For example, as noted in our main paper, the DenseNets \cite{huang2017densely} both classify the original image correctly in all cases but differ in their behavior under the intervention. 
An exception is the randomly initialized SwinTransformer-S \cite{liu2021swin}, which achieves random guessing accuracy (\Cref{tab:celebA-accs}) and shows outputs nearly independent of the inputs for the complete training.

Second, given that most models correctly classify the original image, we note that changes in behavior often lead to incorrect predictions during the intervention. 
Specifically, for grayer hair colors (according to \cref{fig:hair-series}, top row), we observe lower activations in the \texttt{Young} logits of our classifiers. 
In fact, we often find a rapid decline after a certain state in the gradual intervention. 
However, the specific threshold varies depending on the epoch. 
This behavior is, for example, visible for the ConvMixer \cite{trockman2022patches} in \cref{fig:training-vis-ri}.

In general, we can confirm the lower average \propgrad for the randomly initialized models in \Cref{tab:celebA-impact-ext} using \cref{fig:training-vis-ri}. 
Many models only show slight deviations under the hair color interventions. 
This again highlights the difference between effect size and significance of our \propgrad scores.
Further, these results show that pre-trained and randomly initialized models differ on the level of properties they employ for decisions on a local level. 
Additionally, this change in local behavior is not directly connected to the classification of the original sample.

Although the relationship between hair color and age is not causal, with gray hair not necessarily indicating older age, we would expect well-performing classifiers to capture the statistical correlation present in the CelebA dataset \cite{liu2015faceattributes}. 
Indeed, the results in \Cref{tab:celebA-accs} show that pre-trained models achieve higher predictive performance, likely due to their ability to exploit such correlations. 
Our local analysis supports this finding, revealing a stronger dependence on the hair color property for pre-trained models. 
However, it is crucial to note that these results may not generalize to other local inputs, as other factors can influence the model's behavior and lead to different observations. 
To illustrate this, we provide an additional example and re-examine the local model outputs, highlighting again how global correlations can be misleading when analyzing local model behavior.

\FloatBarrier

\paragraph{Additional Sample with Confounding Properties}

\begin{figure*}[t]
    \centering
    \includegraphics[width=\linewidth]{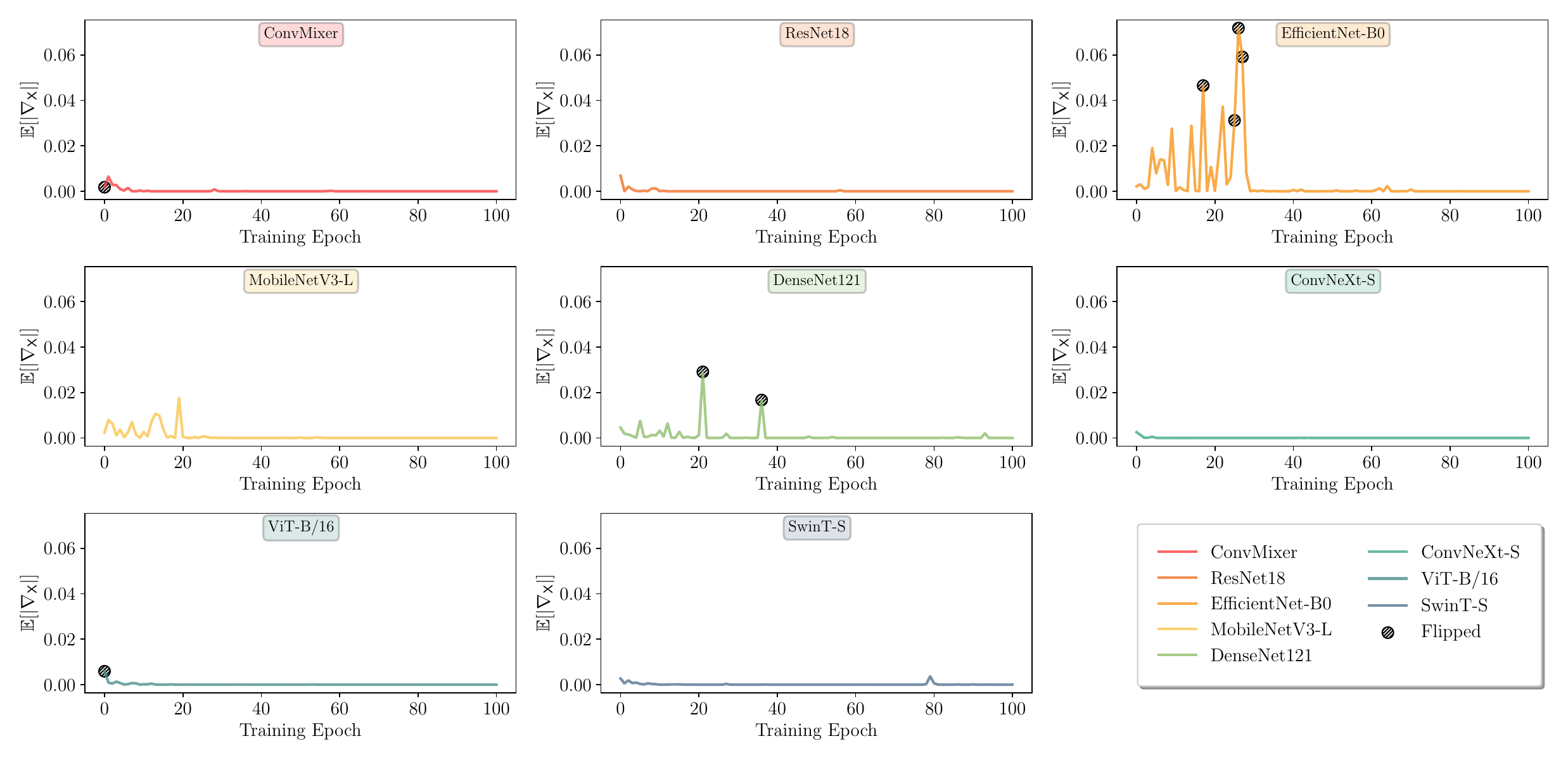}
    \caption{Visualization of how the impact of gray hair changes over the training for various pre-trained architectures.
    Here, we utilize the bottom row of \cref{fig:hair-series} and find little impact of the hair color, hinting at confounding properties.
    }
    \label{fig:impact-pre-trained2}
\end{figure*}

\begin{figure*}[t]
    \centering
    \includegraphics[width=\linewidth]{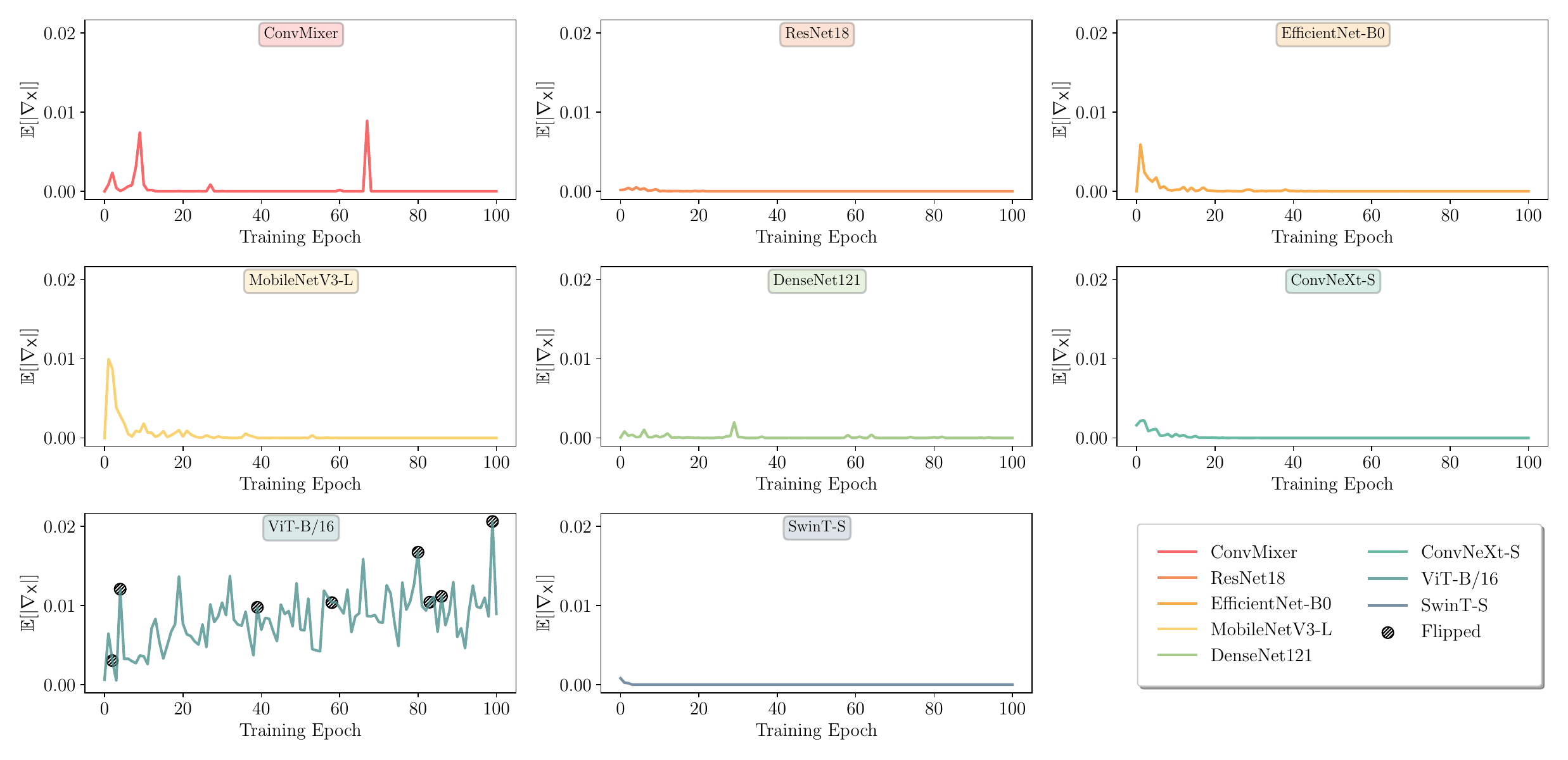}
    \caption{Visualization of how the impact of gray hair changes over the training for various randomly initialized architectures.
    Here, we utilize the bottom row of \cref{fig:hair-series} and find little impact of the hair color, hinting at confounding properties.
    }
    \label{fig:impact-random-init2}
\end{figure*}

\begin{figure*}
    \centering
    \begin{subfigure}{0.49\textwidth}
        \includegraphics[width=\linewidth]{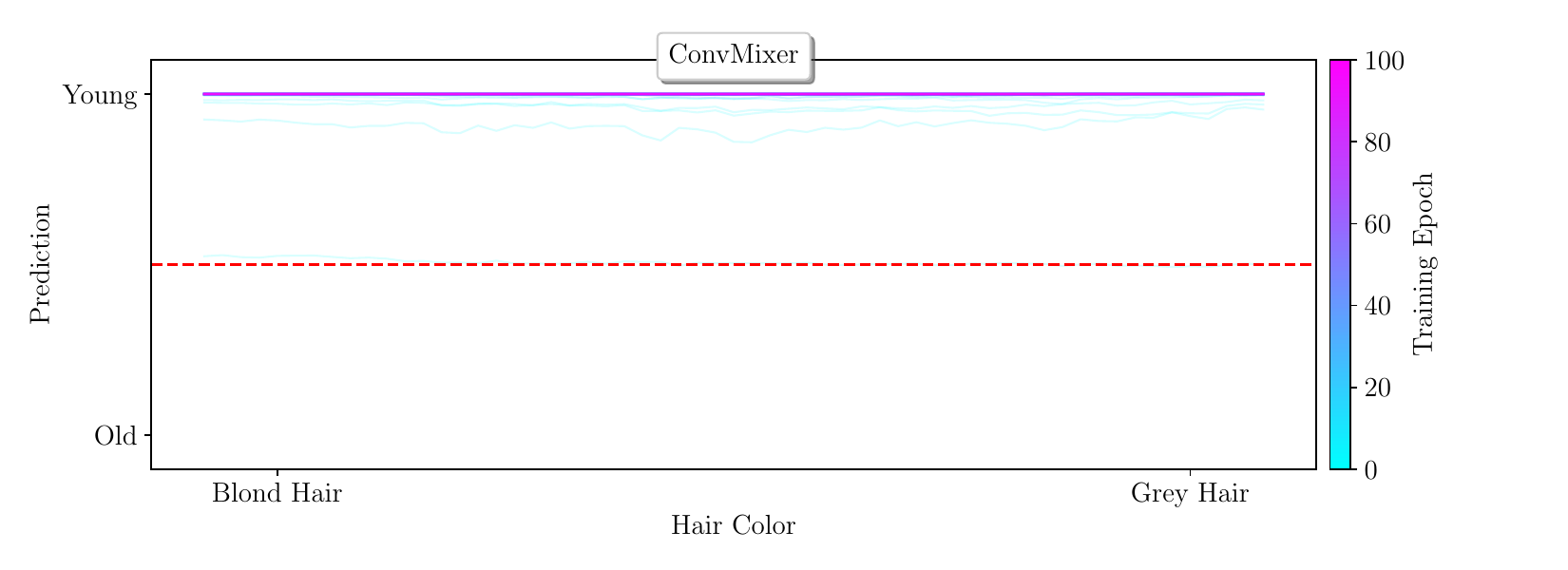}
    \end{subfigure}
    \begin{subfigure}{0.49\textwidth}
        \includegraphics[width=\linewidth]{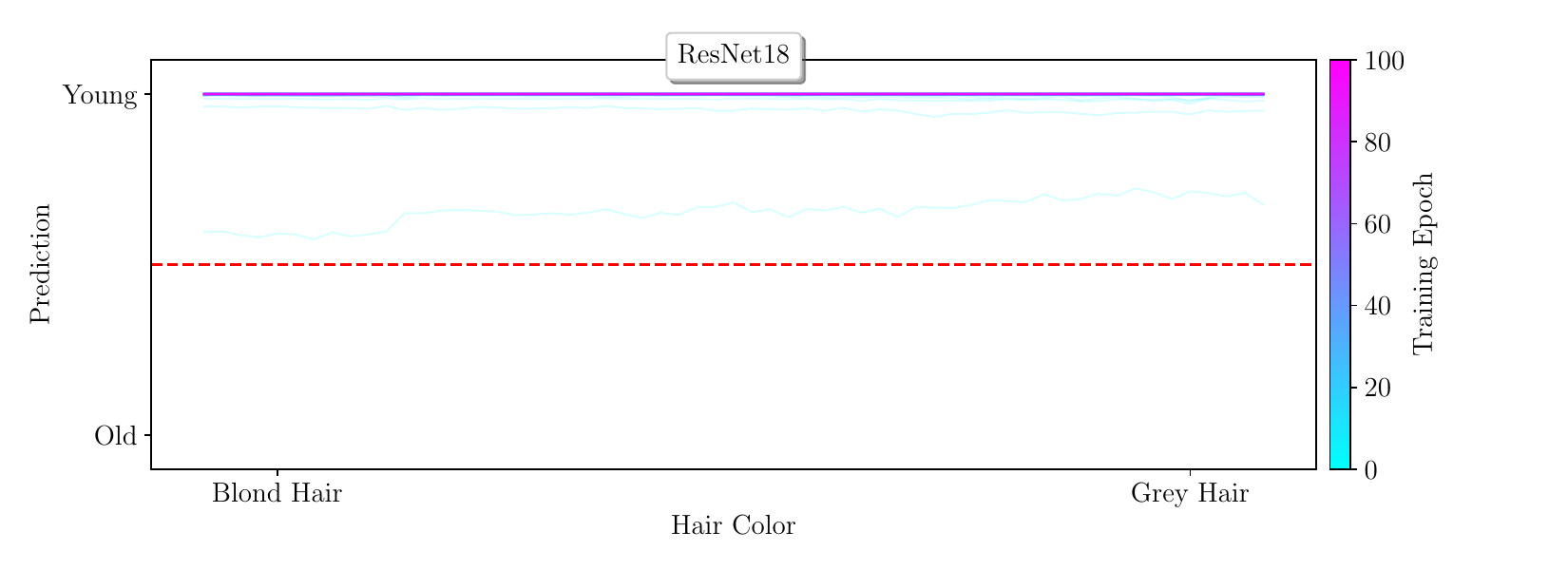}
    \end{subfigure}
    \begin{subfigure}{0.49\textwidth}
        \includegraphics[width=\linewidth]{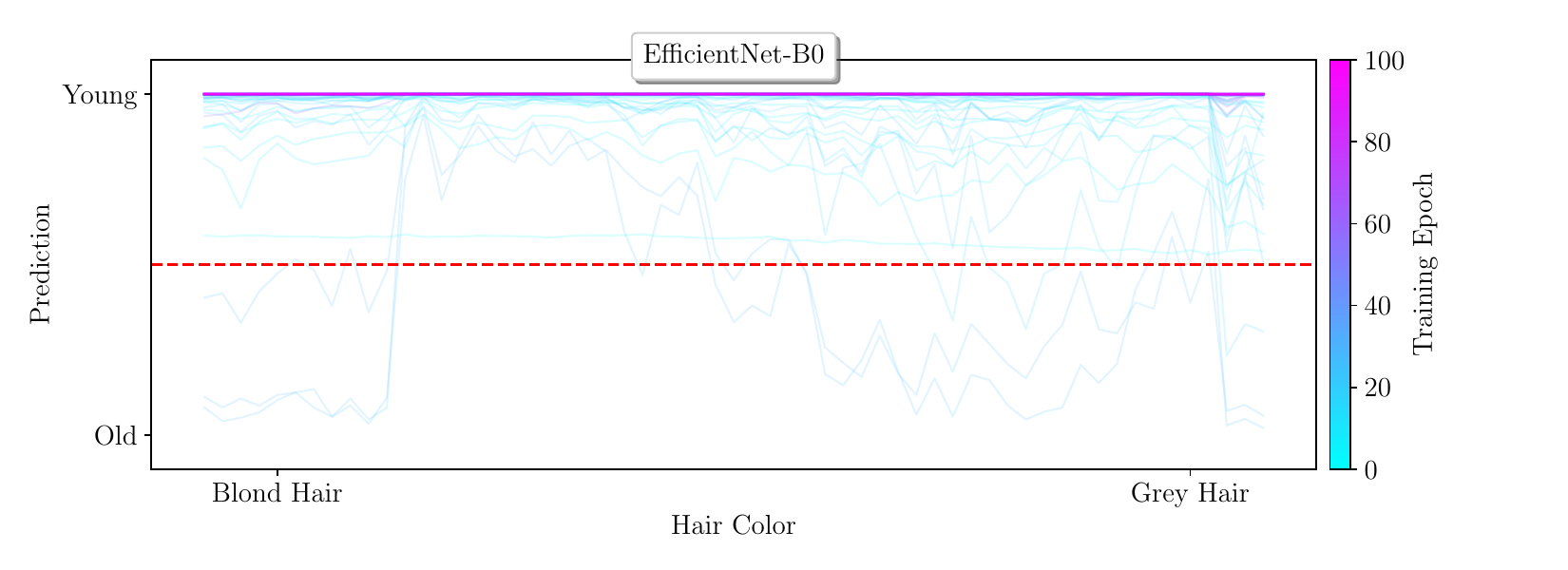}
    \end{subfigure}
    \begin{subfigure}{0.49\textwidth}
        \includegraphics[width=\linewidth]{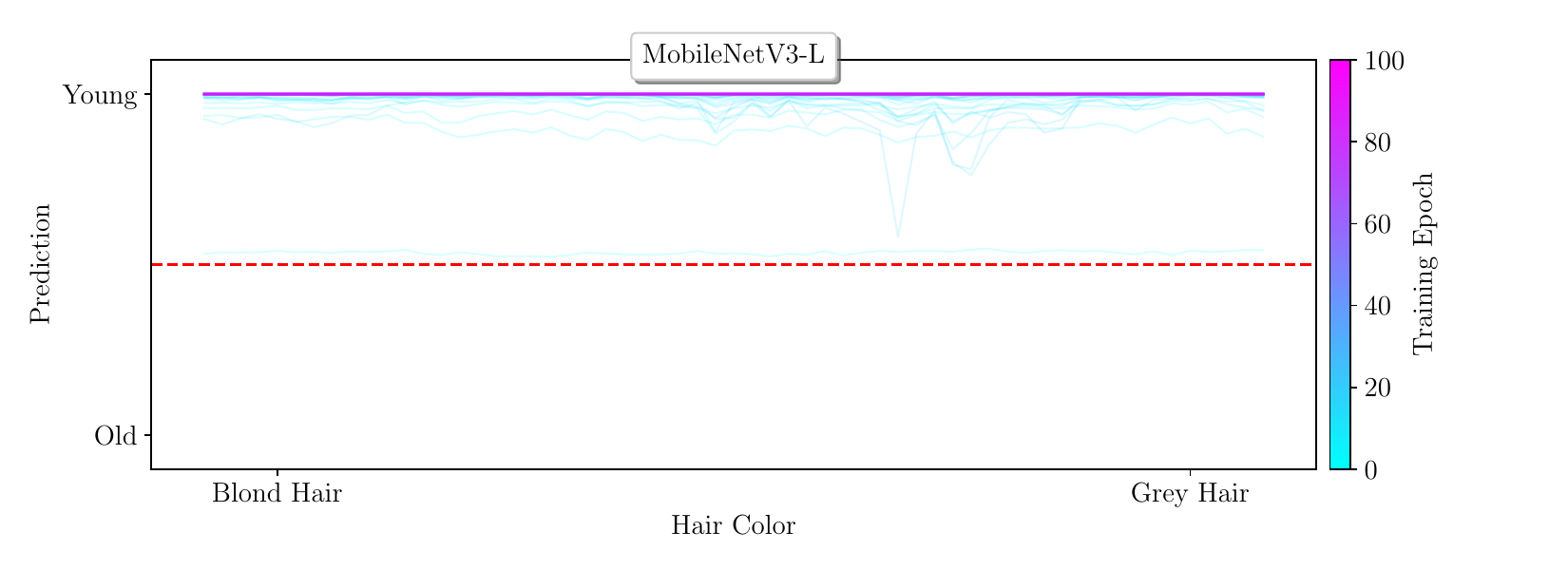}
    \end{subfigure}
    \begin{subfigure}{0.49\textwidth}
        \includegraphics[width=\linewidth]{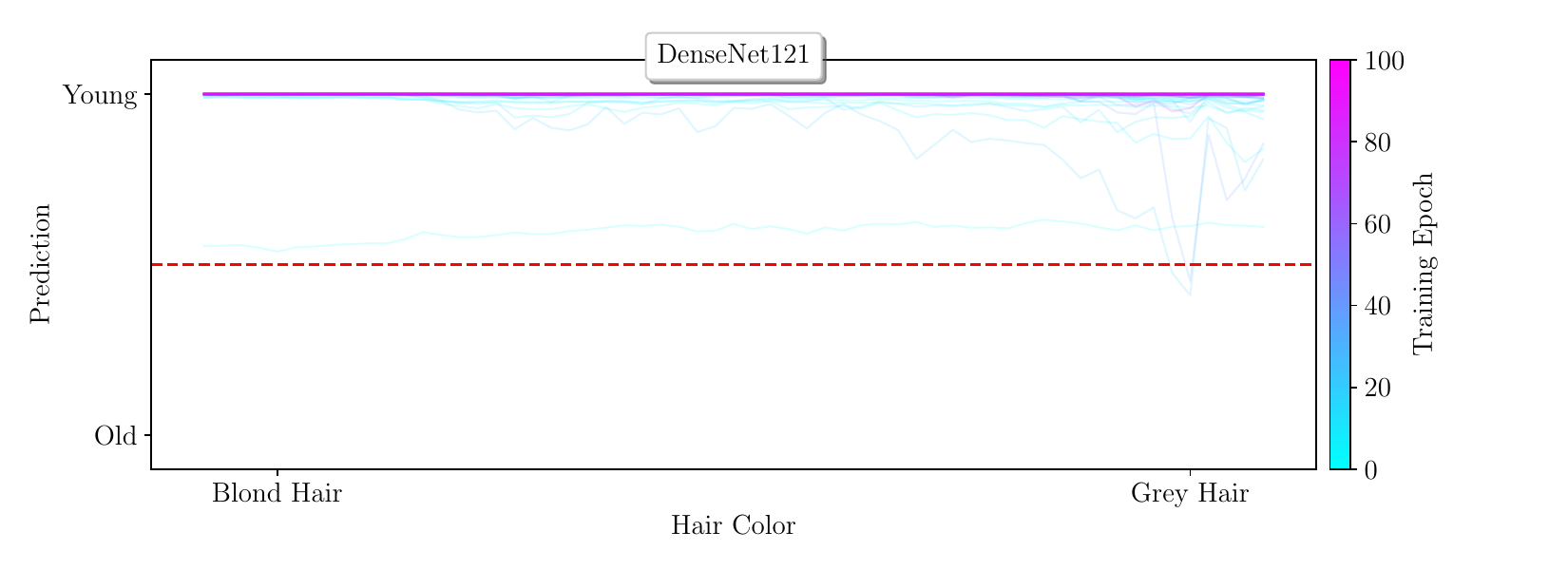}
    \end{subfigure}
    \begin{subfigure}{0.49\textwidth}
        \includegraphics[width=\linewidth]{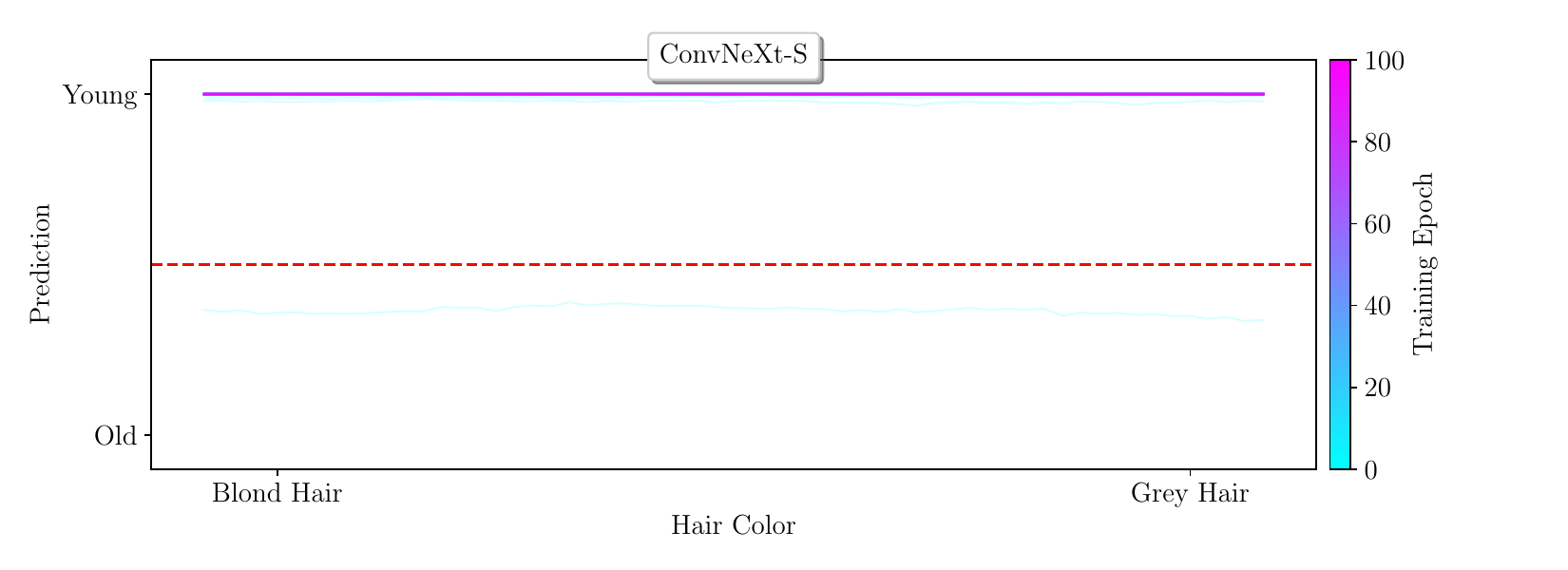}
    \end{subfigure}
    \begin{subfigure}{0.49\textwidth}
        \includegraphics[width=\linewidth]{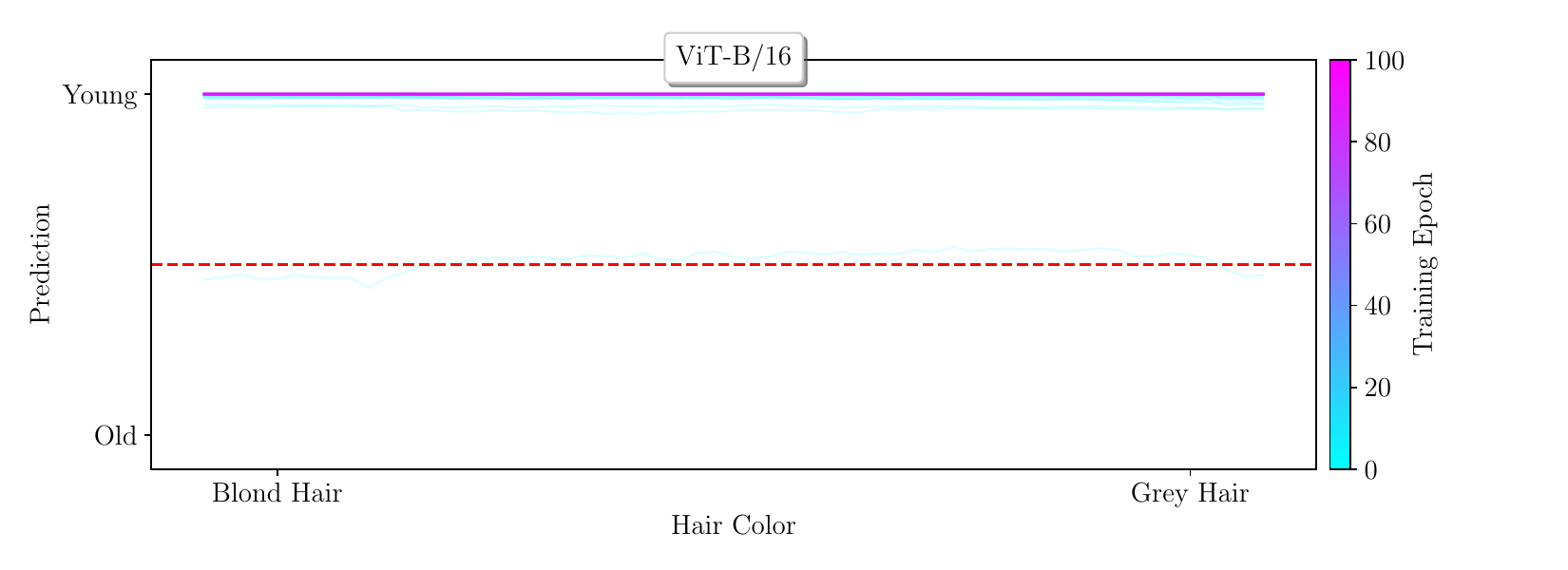}
    \end{subfigure}
    \begin{subfigure}{0.49\textwidth}
        \includegraphics[width=\linewidth]{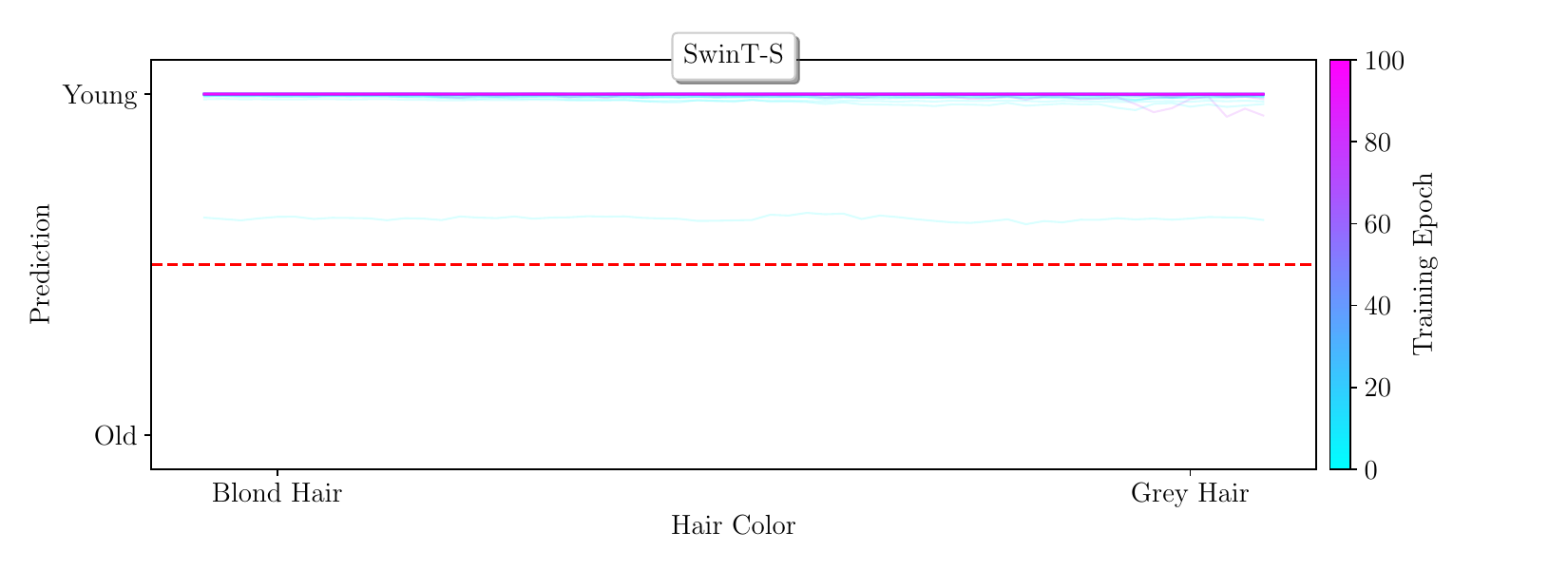}
    \end{subfigure}
    \caption{Model behavior per epoch visualized for the hair color intervention shown in the bottom row of \cref{fig:hair-series}.
    Here, we only show the behavior for \textbf{pre-trained models}.
    The \textcolor{red}{red dotted line} indicates, in all cases, the threshold where the model prediction flips.
    Note the near-constant model outputs, which are also reflected in the low impacts of \cref{fig:impact-pre-trained2}, meaning the networks do not change behavior for hair color interventions for the selected example.}
    \label{fig:training-vis2}
\end{figure*}

\begin{figure*}
    \centering
    \begin{subfigure}{0.49\textwidth}
        \includegraphics[width=\linewidth]{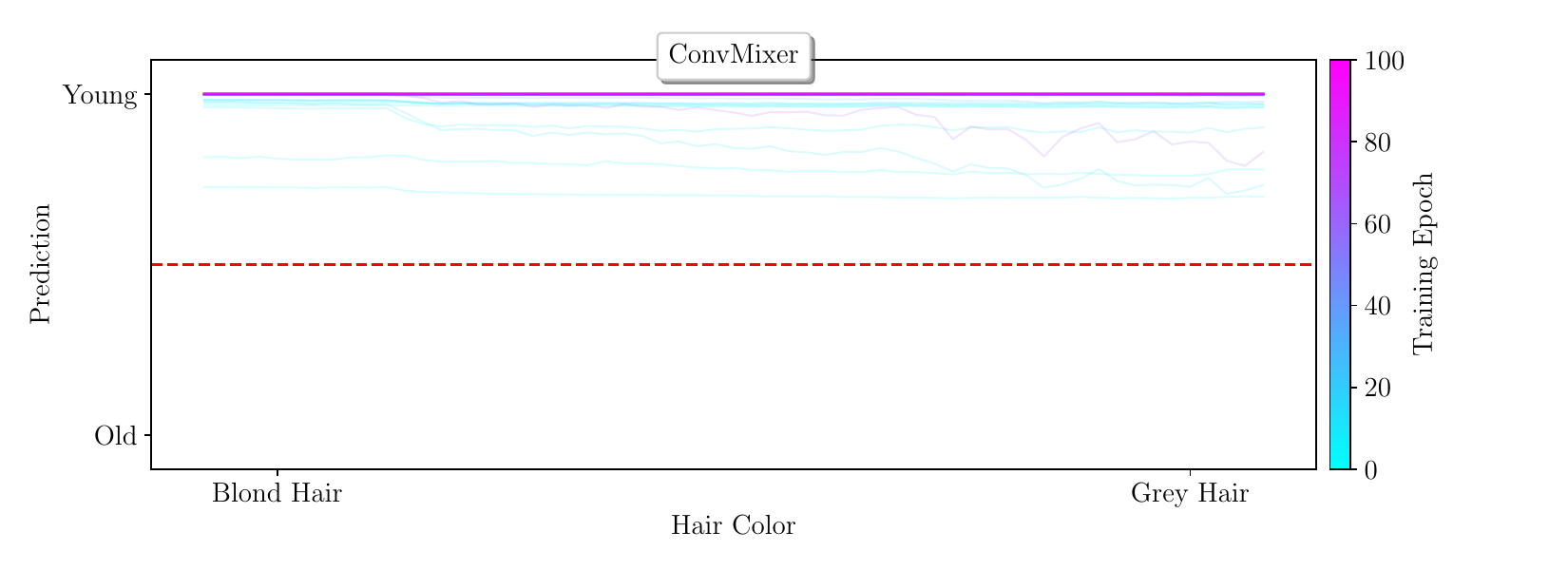}
    \end{subfigure}
    \begin{subfigure}{0.49\textwidth}
        \includegraphics[width=\linewidth]{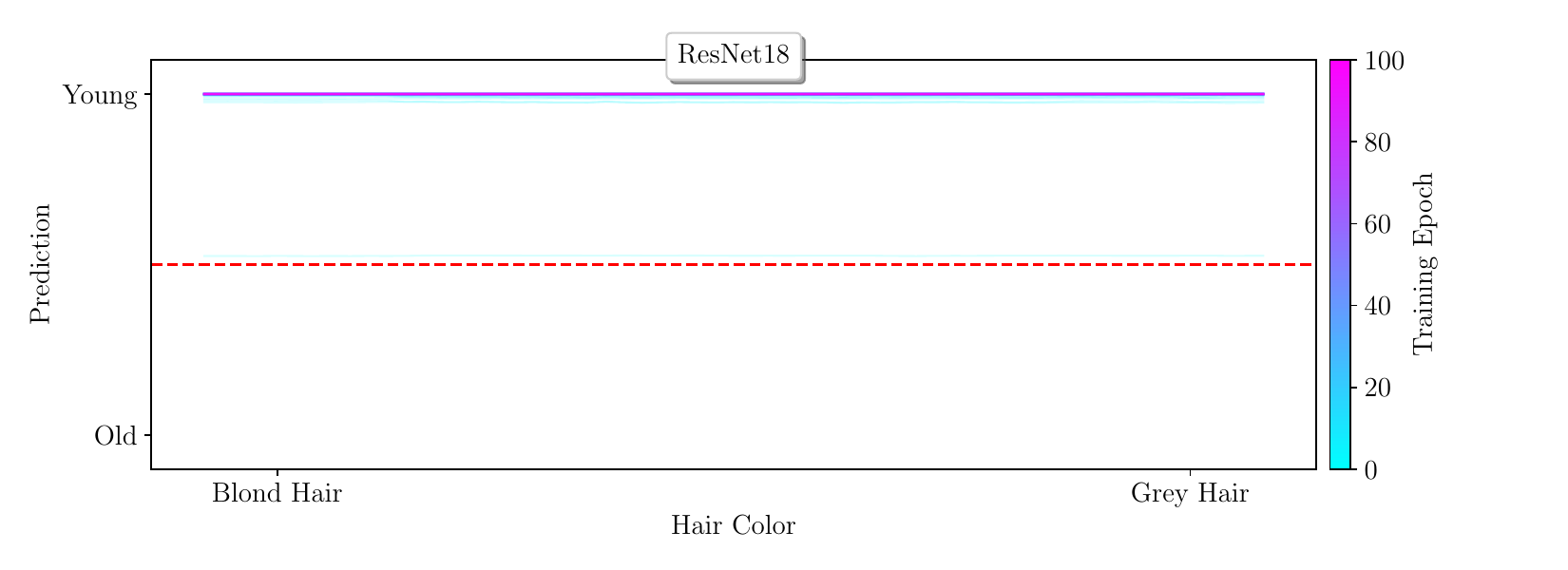}
    \end{subfigure}
    \begin{subfigure}{0.49\textwidth}
        \includegraphics[width=\linewidth]{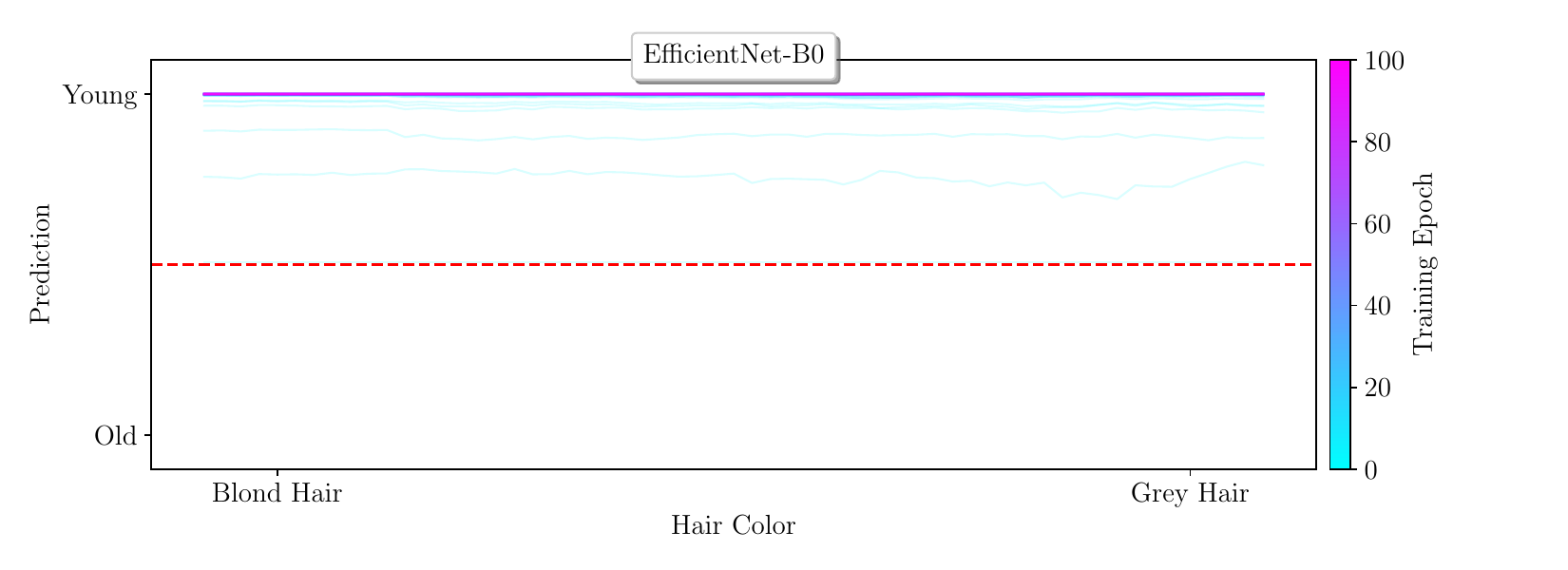}
    \end{subfigure}
    \begin{subfigure}{0.49\textwidth}
        \includegraphics[width=\linewidth]{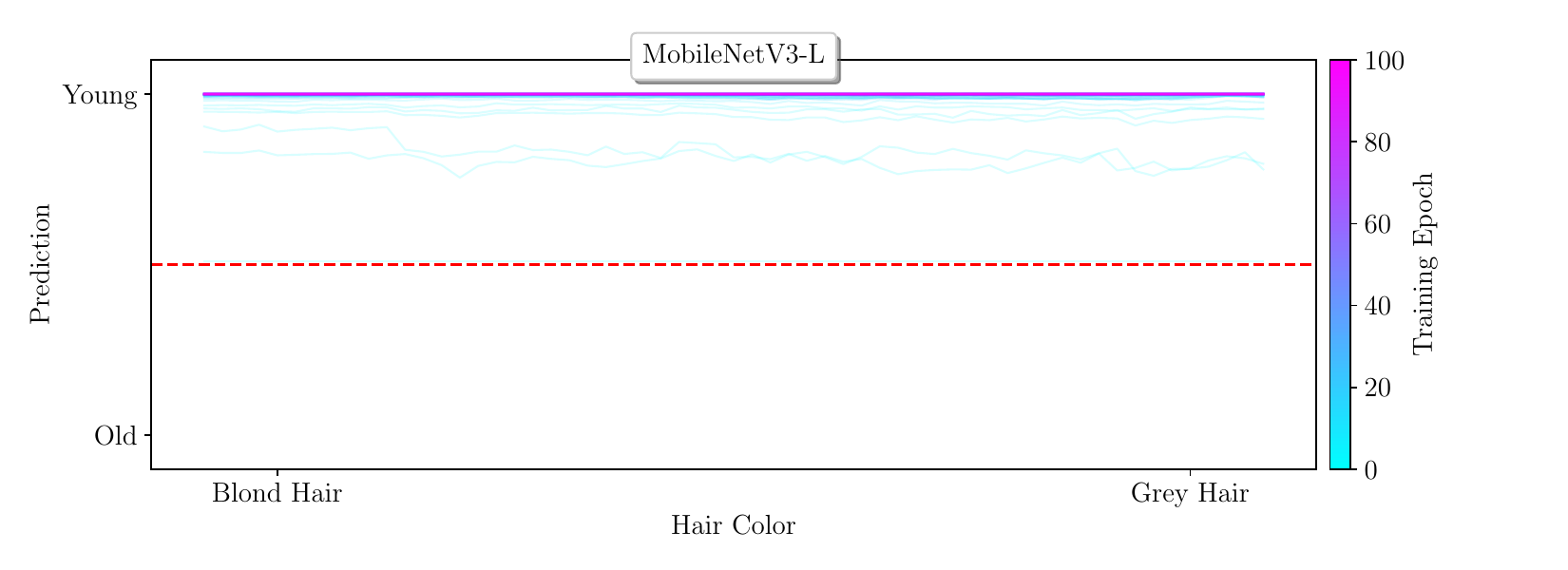}
    \end{subfigure}
    \begin{subfigure}{0.49\textwidth}
        \includegraphics[width=\linewidth]{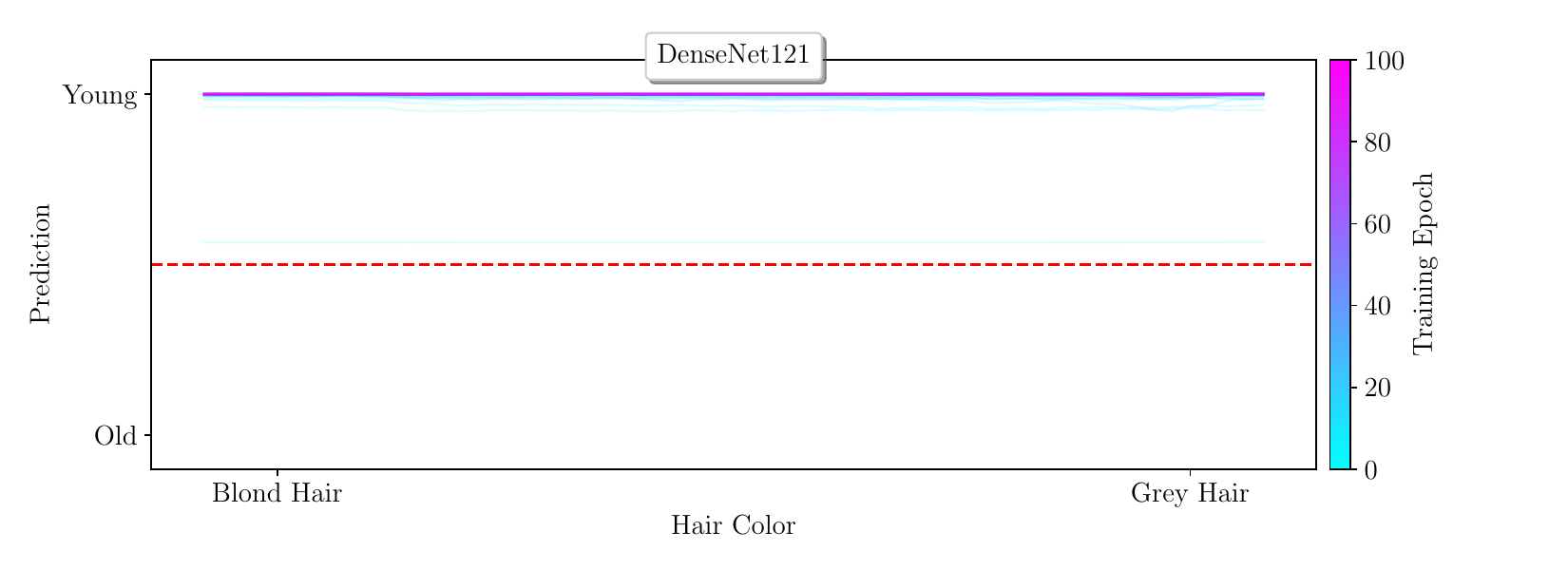}
    \end{subfigure}
    \begin{subfigure}{0.49\textwidth}
        \includegraphics[width=\linewidth]{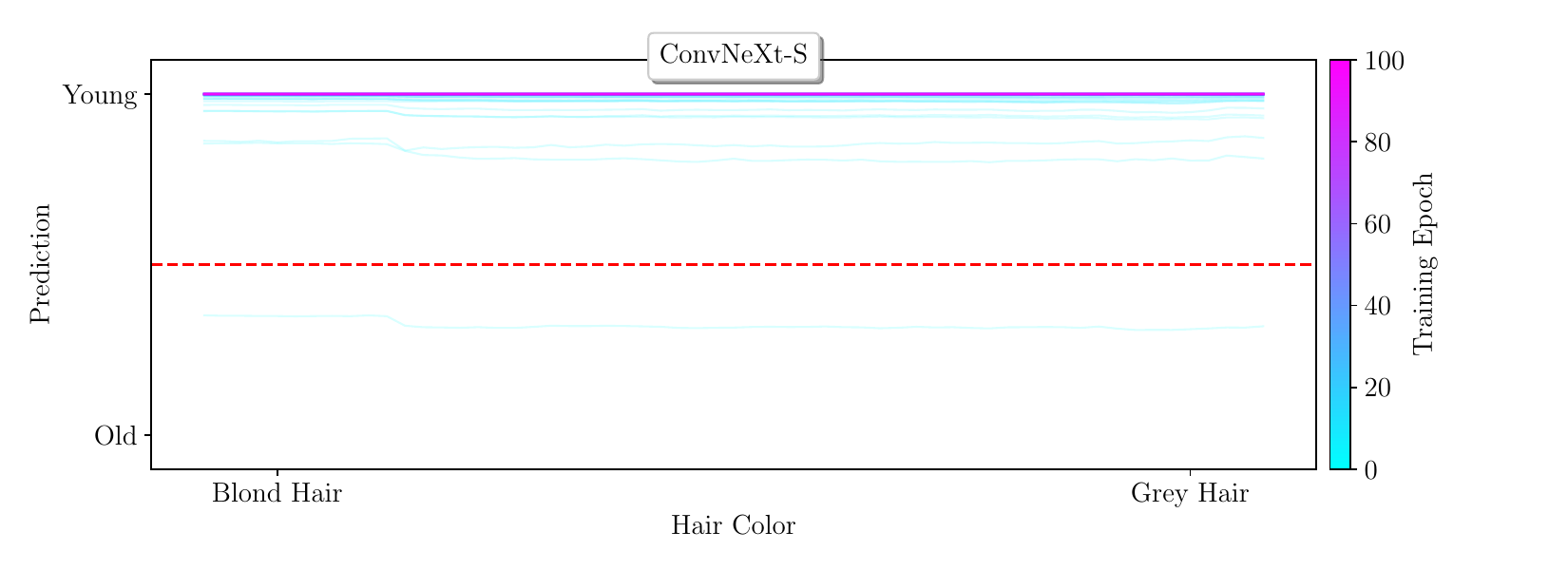}
    \end{subfigure}
    \begin{subfigure}{0.49\textwidth}
        \includegraphics[width=\linewidth]{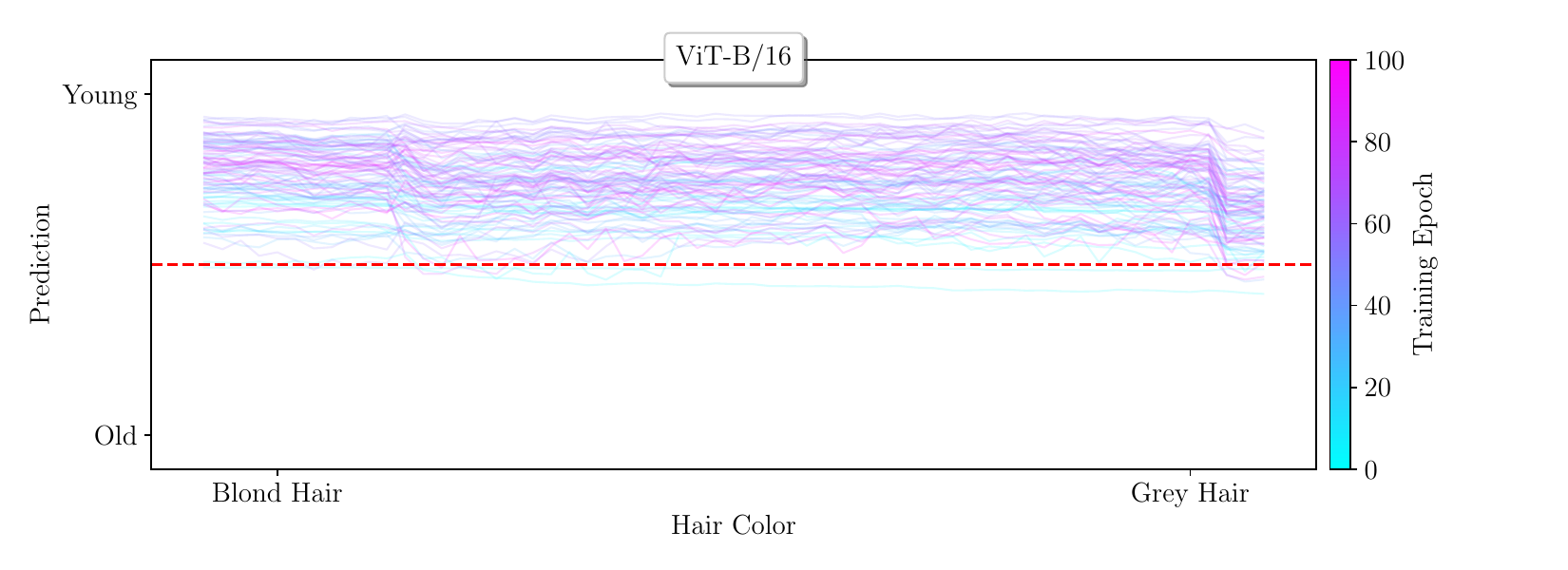}
    \end{subfigure}
    \begin{subfigure}{0.49\textwidth}
        \includegraphics[width=\linewidth]{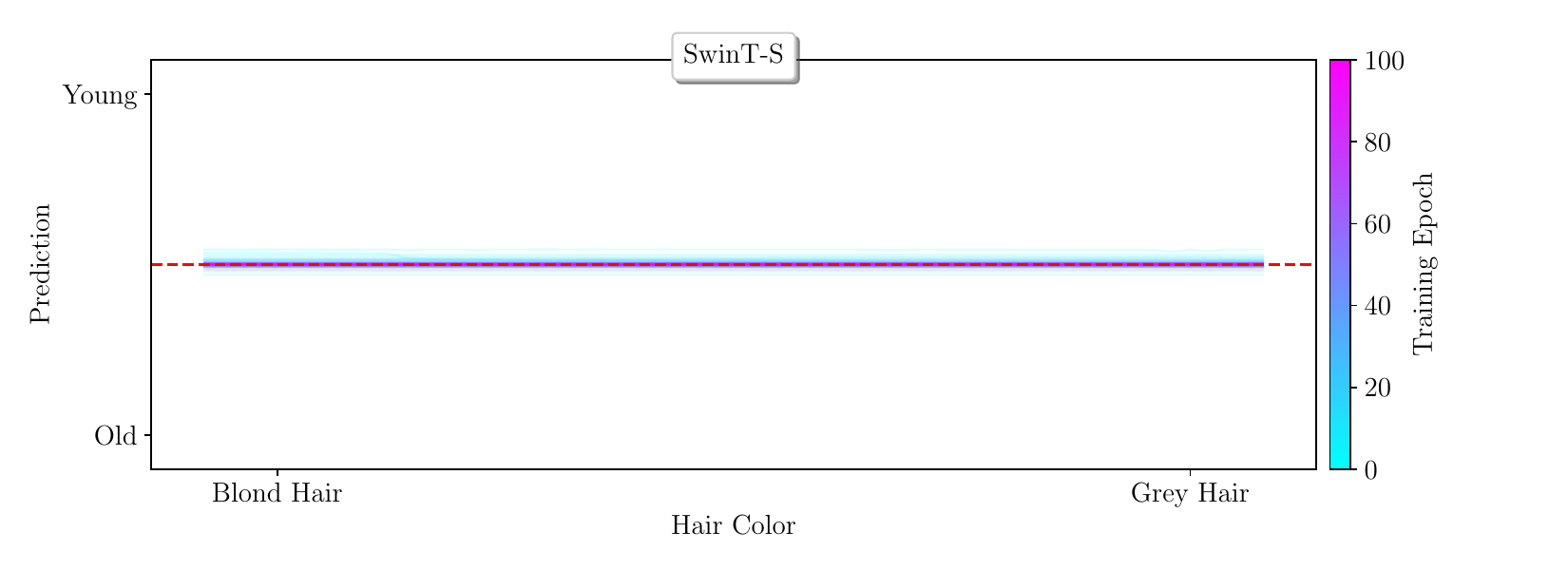}
    \end{subfigure}
    \caption{Model behavior per epoch visualized for the hair color intervention shown in the bottom row of \cref{fig:hair-series}.
    Here, we only show the behavior for \textbf{randomly initialized} models.
    The \textcolor{red}{red dotted line} indicates, in all cases, the threshold where the model prediction flips.
    Note the near-constant model outputs, which are also reflected in the low impacts of \cref{fig:impact-random-init2}, meaning the networks do not change behavior for hair color interventions for the selected example.
    }
    \label{fig:training-vis-ri2}
\end{figure*}

In \cref{fig:hair-series}, we present a second example used for local training analysis (bottom row), where we employ the same prompt and hyperparameters as before. 
Notably, many properties, such as makeup, perceived gender, and hair length, differ between the two individual samples.
We visualize the development of the local \propgrad for the hair color intervention over training in \cref{fig:impact-pre-trained2} and \cref{fig:impact-random-init2} for the pre-trained and randomly initialized models, respectively. 
The corresponding changes in behavior are showcased in \cref{fig:training-vis2} and \cref{fig:training-vis-ri2}.
In all cases, we observe very small \propgrad scores, indicating that the models are not locally influenced by the gray hair color for this individual sample. 
Although minor exceptions exist, such as the pre-trained EfficientNet early in the training or the randomly initialized ViT \cite{dosovitskiy2020image}, we find only minimal differences between the initializations. 
The visualizations in \cref{fig:training-vis2} and \cref{fig:training-vis-ri2} provide an explanation, showing that the models, with rare exceptions, correctly classify the sample throughout the complete intervention and training.

We hypothesize that other properties correlated with age may lead to this phenomenon. 
For instance, makeup is strongly correlated with the \texttt{Young} label in CelebA \cite{liu2015faceattributes}. 
The visible makeup in \cref{fig:hair-series} (bottom row) could potentially be more influential for this specific input. 
Future work should investigate the local interactions of properties to further interpret local prediction behavior.

Our observations underscore the importance of local interventional explanations, which can provide additional insights for individual inputs that go beyond global insights.

\FloatBarrier
\clearpage

\section{CLIP Analysis - Additional Details}
\label{app:clip}

In this section, we structure the content as follows.
First, we provide the full details of our inference setup for zero-shot classification using a pre-trained CLIP \cite{radford2021learning} model.
Then, we highlight additional results and ablations.

\begin{table}[H]
    \centering
    \caption{Recently, researchers utilize multiple text descriptions when performing zero-shot classification with CLIP \cite{radford2021learning} models, see, for example, \cite{pratt2023does,roth2023waffling}.
    Here, we list the text descriptors used for the corresponding objects in our real-life and virtual interventional data to calculate the cosine similarities.
    }
    \label{tab:clip-texts}
    \begin{tabular}{ll}
    \toprule
        Obj. & Text Descriptors \\
    \midrule
         \multirow{6}{*}{\rotatebox{90}{Toy Elephant}} 
            & \texttt{toy elephant, elephant, african}\\
            & \texttt{elephant, picture of an}\\
            & \texttt{elephant, gray elephant,}\\
            & \texttt{standing elephant, elephant}\\
            & \texttt{model, small elephant,}\\
            & \texttt{indian elephant, elephant tusk}\\
            \cmidrule{2-2}
         \multirow{6}{*}{\rotatebox{90}{Toy Giraffe}}  
            & \texttt{toy giraffe, giraffe, african}\\
            & \texttt{giraffe, picture of a giraffe,}\\
            & \texttt{spotted giraffe, standing}\\
            & \texttt{giraffe, giraffe model,}\\
            & \texttt{small giraffe,tall giraffe,}\\
            & \texttt{giraffe bull}\\
            \cmidrule{2-2}
         \multirow{7}{*}{\rotatebox{90}{Toy Stegosaurus}}  
            & \texttt{toy stegosaurus, stegosaurus,}\\
            & \texttt{dinosaur, picture of a }\\
            & \texttt{stegosaurus, toy dinosaur,}\\
            & \texttt{standing stegosaurus,}\\
            & \texttt{stegosaurus model, small}\\
            & \texttt{stegosaurus, green stegosaurus,}\\
            & \texttt{stegosaurus plates}\\
    \midrule
         \multirow{6}{*}{\rotatebox{90}{3D Frog Model}} 
         & \texttt{frog model, toy frog, frog,}\\
         & \texttt{frog rendering, picture of}\\
         & \texttt{a frog, 3d frog model,}\\
         & \texttt{green frog, photo of a}\\
         & \texttt{large frog, sitting frog,}\\
         & \texttt{still frog}\\
    \bottomrule
    \end{tabular}
\end{table}

\begin{figure*}
    \centering
    \begin{subfigure}{\textwidth}
        \includegraphics[width=\textwidth]{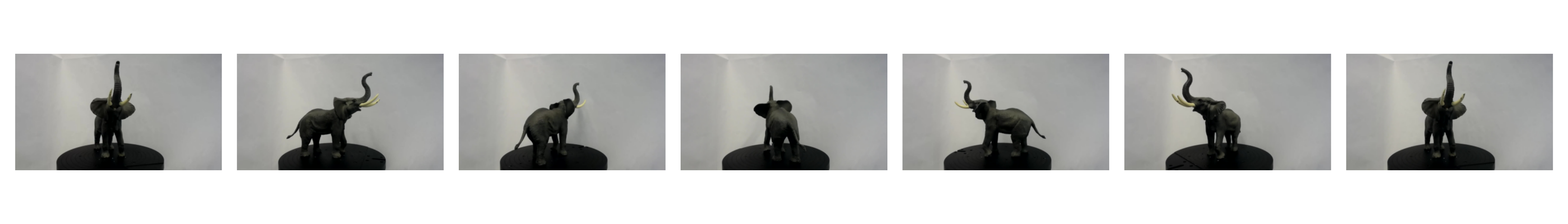}
        \caption{Toy elephant.}
    \end{subfigure}
    \begin{subfigure}{\textwidth}
        \includegraphics[width=\textwidth]{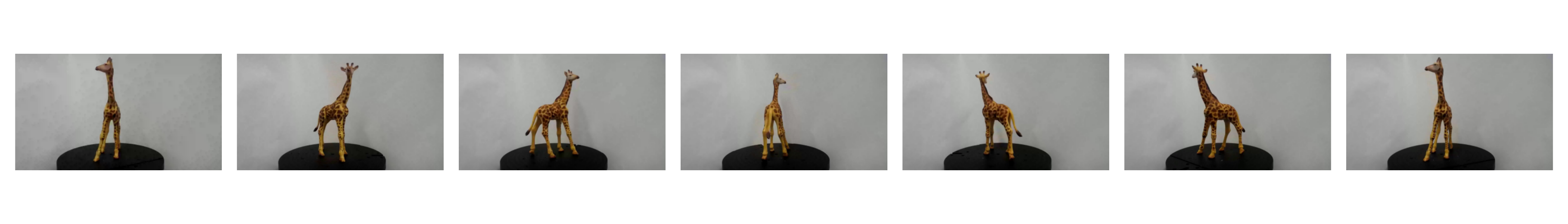}
        \caption{Toy giraffe.}
    \end{subfigure}
    \begin{subfigure}{\textwidth}
        \includegraphics[width=\textwidth]{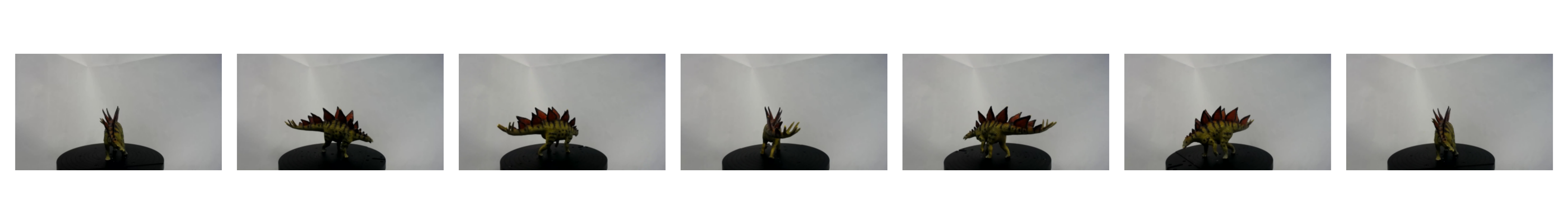}
        \caption{Toy stegosaurus.}
    \end{subfigure}
    \caption{
    Real-life interventions on the position of toy animals.
    The interventions here use a turn table, meaning we intervene in the rotational position compared to the fixed camera.
    }
    \label{fig:real-intervent}
    \vspace{0.5cm}
\end{figure*}

\begin{figure*}
    \centering
    \begin{subfigure}{\textwidth}
        \includegraphics[width=\textwidth]{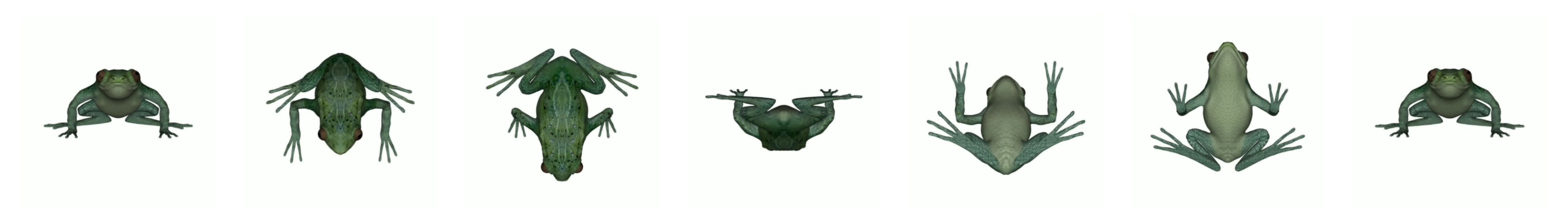}
        \caption{Rotation around the x-axis.}
    \end{subfigure}
    \begin{subfigure}{\textwidth}
        \includegraphics[width=\textwidth]{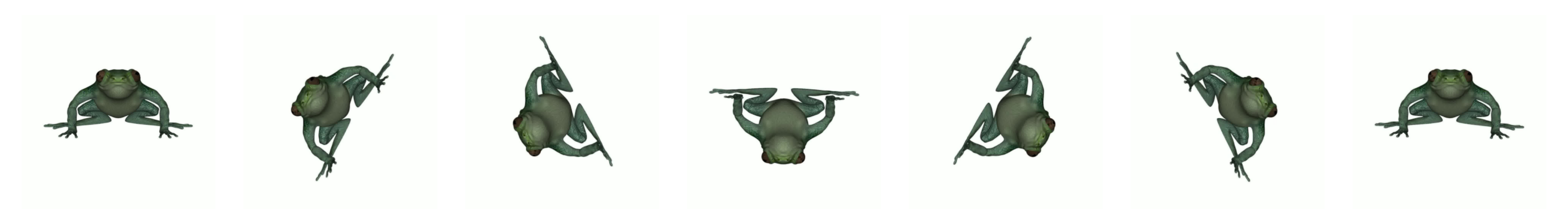}
        \caption{Rotation around the y-axis.}
    \end{subfigure}
    \begin{subfigure}{\textwidth}
        \includegraphics[width=\textwidth]{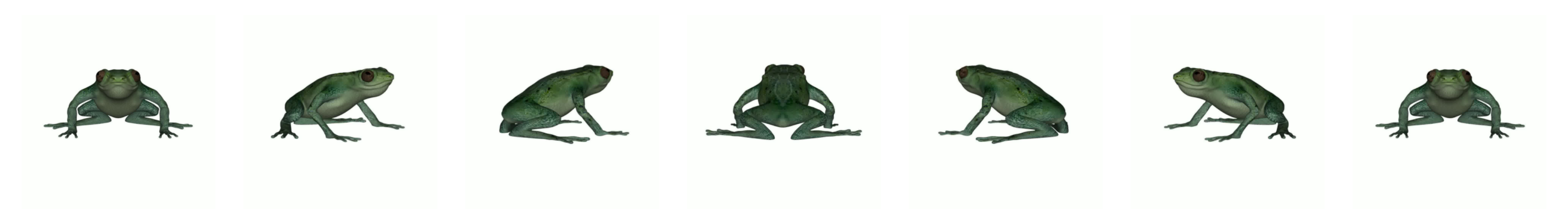}
        \caption{Rotation around the z-axis.}
    \end{subfigure}
    \caption{
    Virtual interventions on the position of a rendered frog model.
    Here, we intervene by rotating around all three axes of orientation while keeping the camera fixed.
    }
    \label{fig:virtual-intervent}
    \vspace{0.5cm}
\end{figure*}

\begin{figure*}
    \centering
    \begin{subfigure}{0.32\textwidth}
        \includegraphics[width=\textwidth]{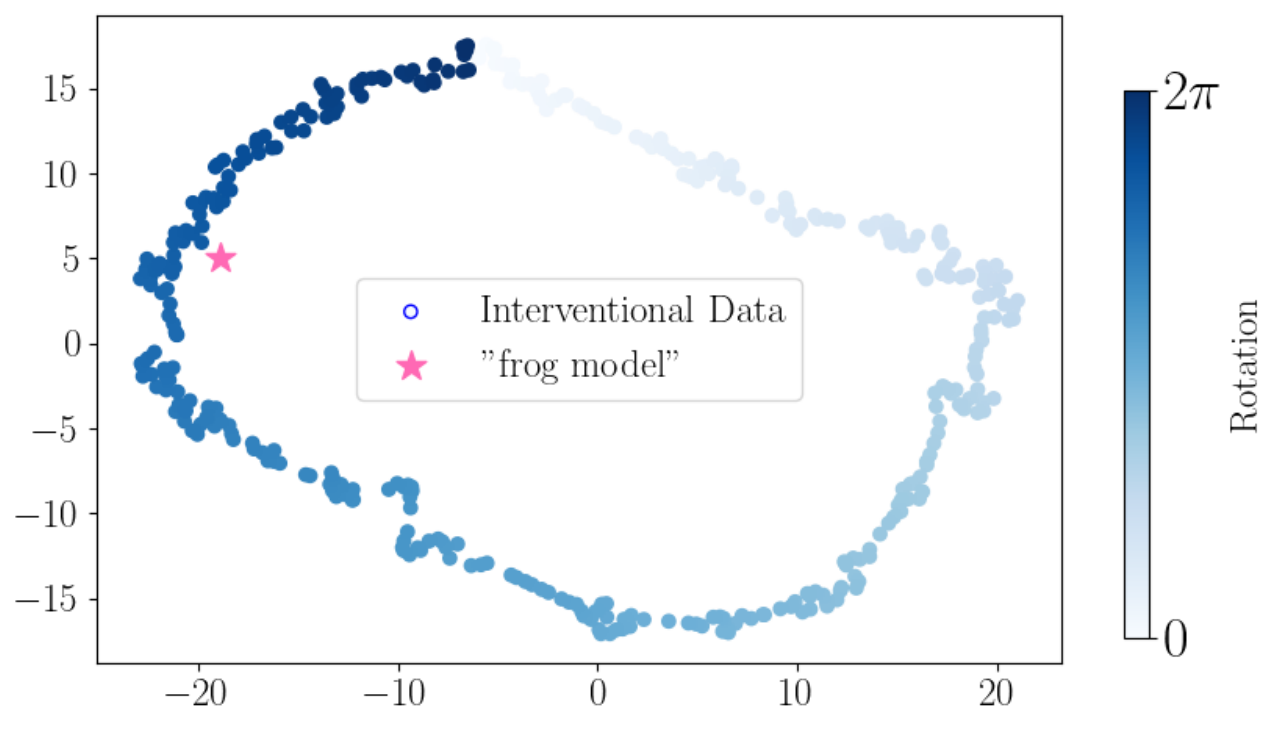}
        \caption{Frog model, rotation around the $x$-axis.}
    \end{subfigure}
    \begin{subfigure}{0.32\textwidth}
        \includegraphics[width=\textwidth]{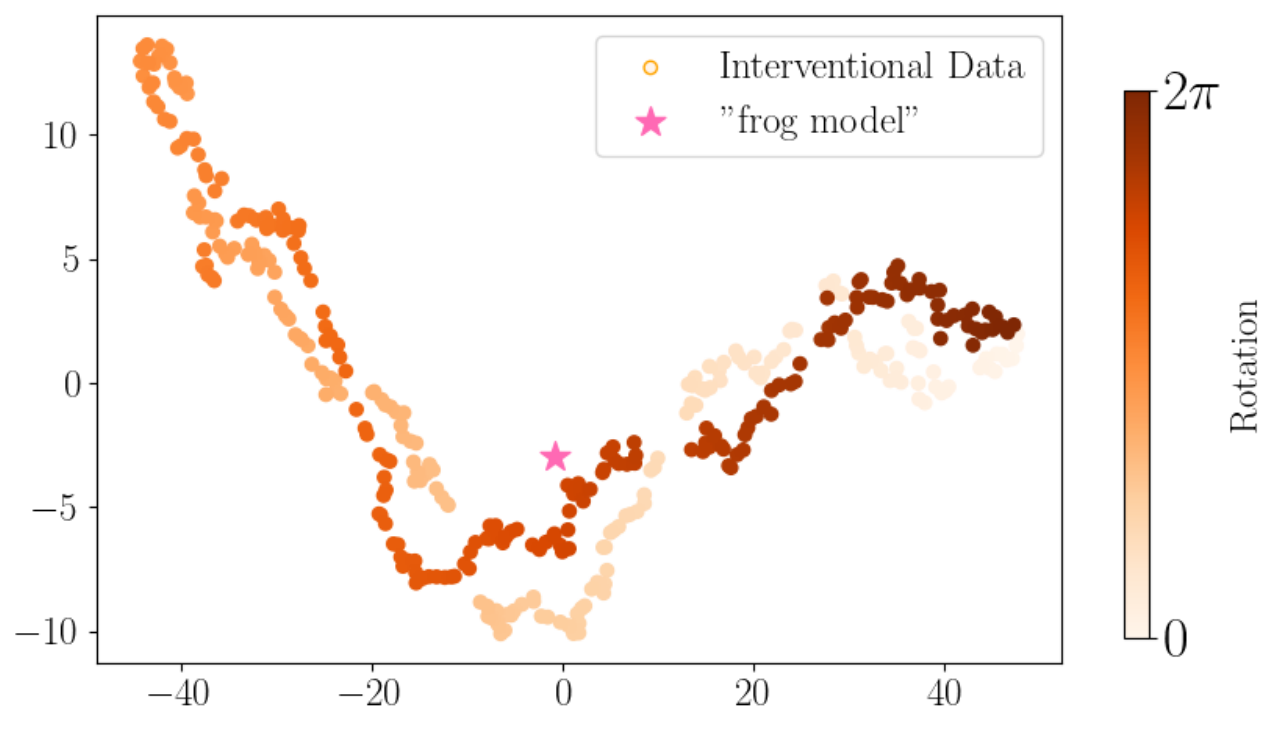}
        \caption{Frog model, rotation around the $y$-axis.}
    \end{subfigure}
    \begin{subfigure}{0.32\textwidth}
        \includegraphics[width=\textwidth]{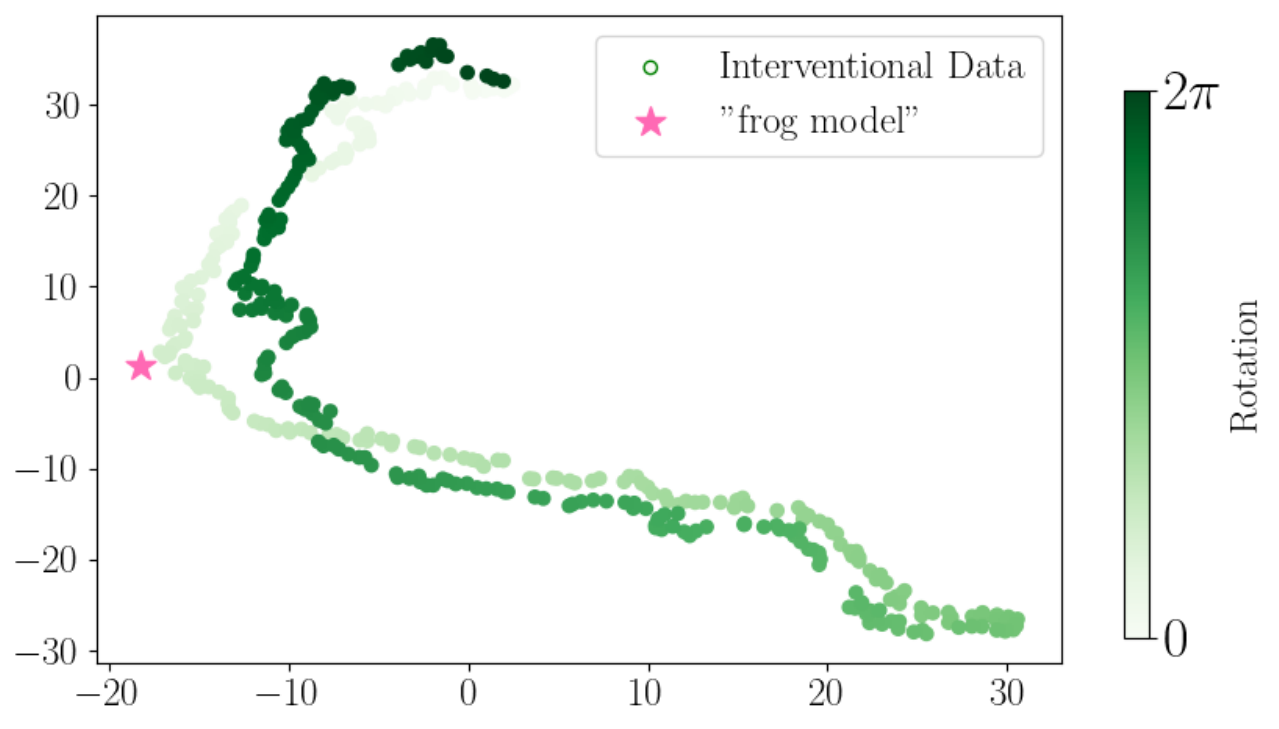}
        \caption{Frog model, rotation around the $z$-axis.}
    \end{subfigure}
    \caption{
    CLIP \cite{radford2021learning} latents visualized in 2D using t-SNE \cite{maaten2008visualizing}.
    In all cases, we add the embedding for the corresponding description.
    We encode the rotation angle using color in all three visualizations.
    }
    \label{fig:real-latents}
\end{figure*}

\begin{figure*}
    \centering
    \begin{subfigure}{0.32\textwidth}
        \includegraphics[width=\textwidth]{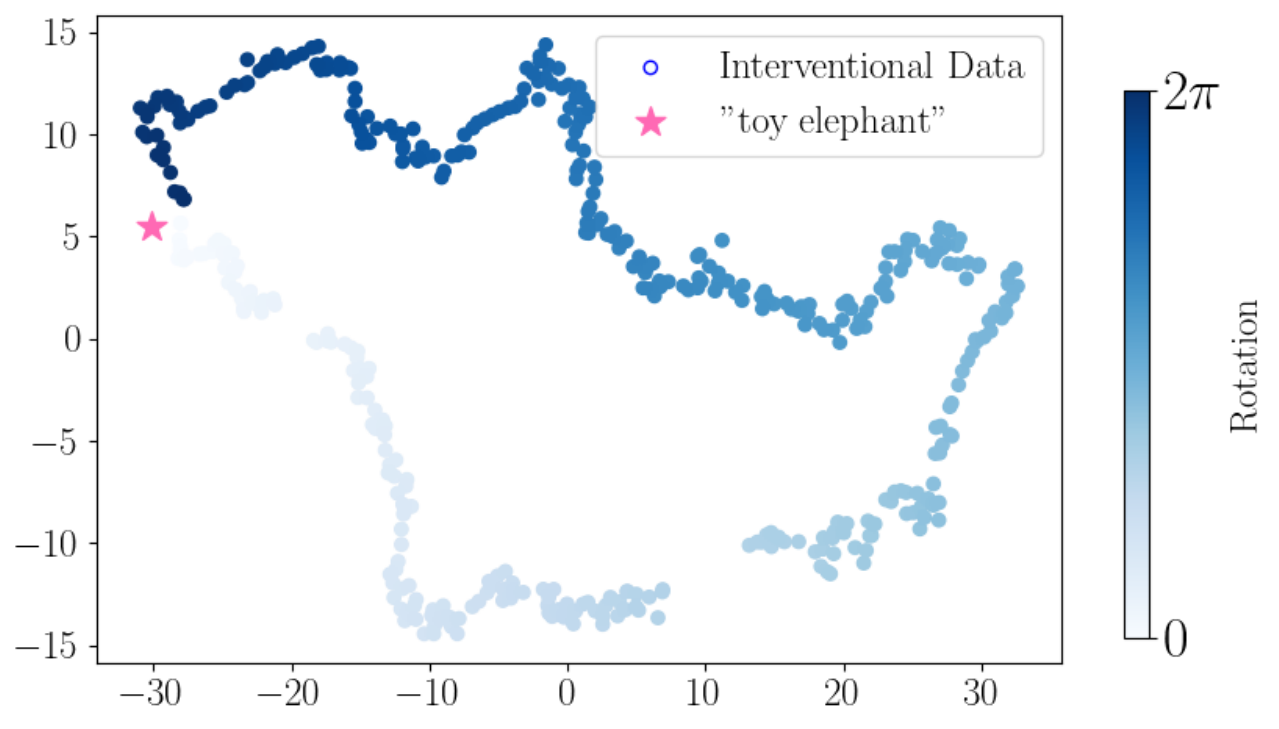}
        \caption{Toy elephant.}
    \end{subfigure}
    \begin{subfigure}{0.32\textwidth}
        \includegraphics[width=\textwidth]{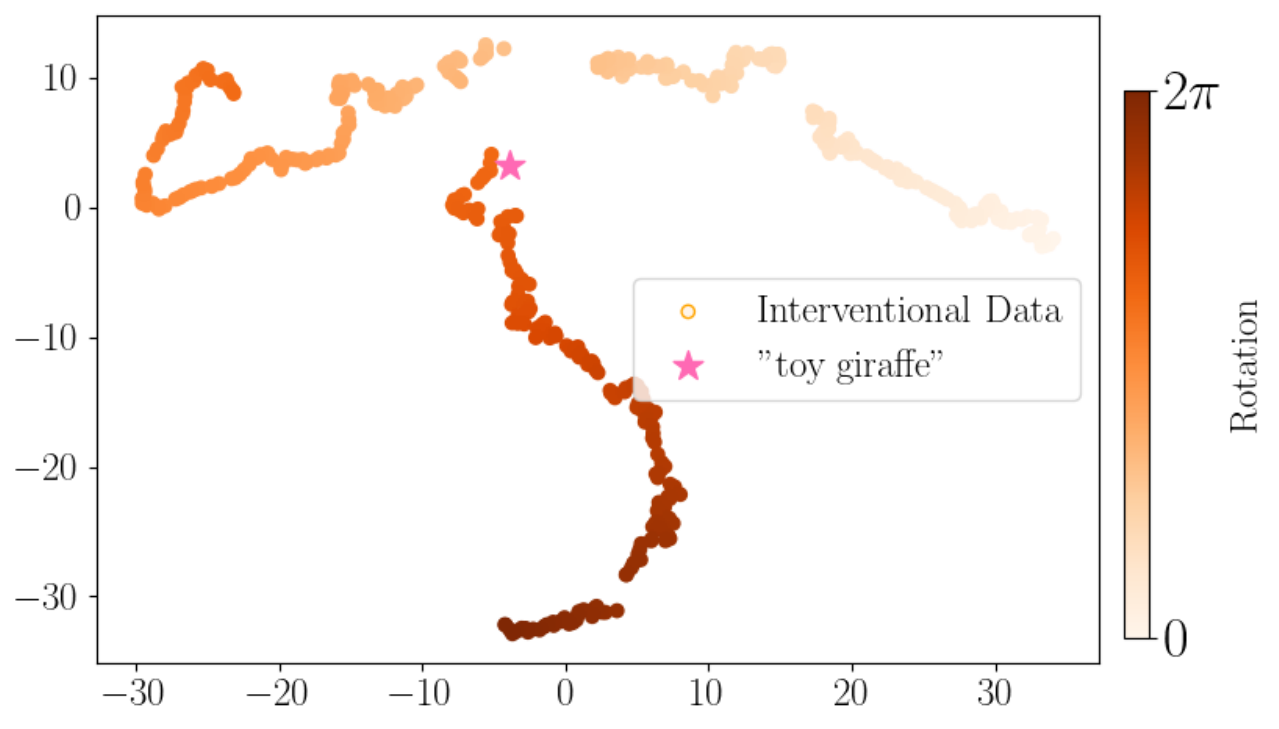}
        \caption{Toy giraffe.}
    \end{subfigure}
    \begin{subfigure}{0.32\textwidth}
        \includegraphics[width=\textwidth]{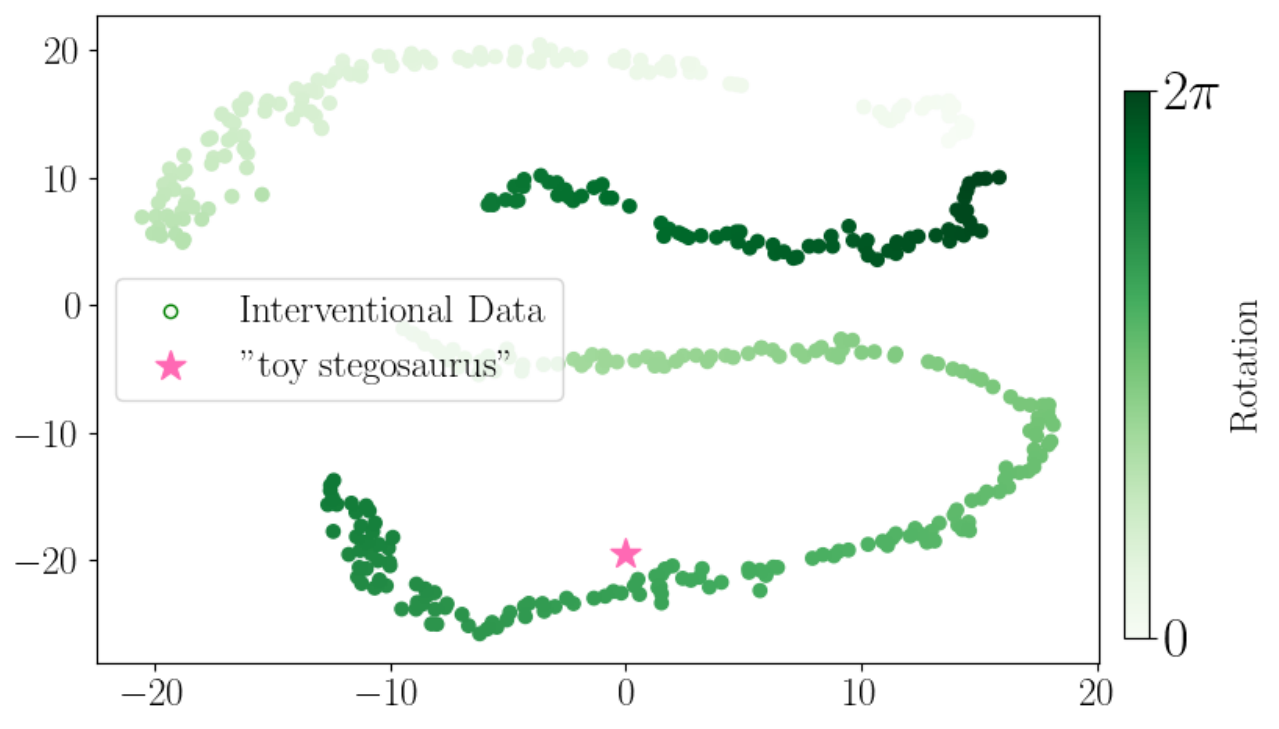}
        \caption{Toy stegosaurus.}
    \end{subfigure}
    \caption{
    CLIP \cite{radford2021learning} latents visualized in 2D using t-SNE \cite{maaten2008visualizing}.
    In all cases, we add the embedding for the corresponding description.
    We encode the rotation angle using color in all three visualizations.
    }
    \label{fig:virtual-latents}
\end{figure*}

\subsection{Setup Details}
\label{app:clip-setup}

As mentioned in our main paper, we focus on real-life interventional data.
Additionally, we separately perform a virtual rotation of a 3D frog model\footnote{\url{https://skfb.ly/6USP7}} around all three axes, rendering one image per degree to also exclude photon noise.
This setup enables us to systematically evaluate the local impact of orientation on the CLIP \cite{radford2021learning} model's outputs.

Specifically, we construct a zero-shot classification scenario by capturing images of three toy figures: an elephant, a giraffe, and a stegosaurus.
For real-life interventions, these objects do not change. 
Therefore, the ground truth remains fixed.
Nevertheless, we intervene in the input orientation using a turn table and capture one complete rotation of the toy figures in front of a neutral background.
We do this specifically because we expect animals to be most often photographed upright and facing the camera.
In other words, we expect pre-trained models to show behavioral changes for uncommon object positions.
We visualize parts of the data in \cref{fig:real-intervent}

The rotation of the turn table is an in-plane rotation around the $z$-axis, where the figurines do not flip upside down.
Hence, we strengthen our analysis by additionally incorporating virtual interventions of a 3D animal model.
Specifically, we choose a model of a frog and perform interventions by rotating it around all three axes, rendering one image per degree.
We visualize the resulting data in \cref{fig:virtual-intervent}.

To demonstrate the cyclic nature of both the real-life and virtual interventional data, we utilize t-SNE \cite{maaten2008visualizing} to display the CLIP \cite{radford2021learning} model latent vectors as generated by the visual encoder.
Specifically, we perform a dimensionality reduction to 2D using t-SNE and show the results in \cref{fig:real-latents} and \cref{fig:virtual-latents}, respectively. 
We also include a corresponding text phrase to provide an intuition of the latent space for our cosine similarity analysis. 
While these visualizations are inherently limited in the insights they provide due to the strong reduction in dimensionality, the periodicity of the data is visible, particularly for the virtual interventions. 
This observation is consistent with the behavior of the cosine similarities in our main paper (see \Cref{sec:exp3}).

For the actual zero-shot classification, we follow recent approaches, e.g., \cite{pratt2023does,roth2023waffling}, and compare them to multiple text phrases or, rather, the corresponding embeddings.
The specific text descriptors are listed in \Cref{tab:clip-texts}.

\begin{table}[tb]
    \centering
    \caption{
    The approximated \propgrad scores for the selected CLIP \cite{radford2021learning} model using the three real-life rotation interventions (see \cref{fig:real-intervent}).
    The columns signify the text embeddings we use to calculate the cosine similarities in the latent space.
    To be specific, row one contains the scores pertaining to \cref{fig:clip-real-elephant}, and
    rows two and three correspond to \cref{fig:clip-real-giraffe} and \cref{fig:clip-real}, respectively.  
    Note that all scores are statistically significant following \Cref{alg:sig}.
    }
    \label{tab:clip-mean-scores-real}
    \begin{tabular}{l|ccc}
        \toprule
        & \multicolumn{3}{c}{Texts Similarity \propgrad}\\
        \cmidrule{2-4}
         Interv. Data & Elephant & Giraffe & Stegosaurus\\
         \midrule
         Elephant & 0.00141 & 0.00141 & 0.00178\\
         Giraffe & 0.00096 & 0.00086 & 0.00118\\
         Stegosaurus & 0.00157 & 0.00142 & 0.00140 \\
         \bottomrule
    \end{tabular}
\end{table}

\begin{table}[tb]
    \centering
    \caption{
    The approximated \propgrad scores for the selected CLIP \cite{radford2021learning} model using the three virtual rotation interventions  (see \cref{fig:virtual-intervent}).
    The rows contain interventional data pertaining to the three axes of rotation.
    Specifically, they correspond to the scores achieved by the visualized means in \cref{fig:clip-virtual-split}.
    Note that all scores are statistically significant following \Cref{alg:sig}.
    }
    \label{tab:clip-mean-scores-virtual}
    \begin{tabular}{l|ccc}
        \toprule
         Interv. Data &  Frog Texts Similarity \propgrad\\
         \midrule
         $x$-axis & 0.00161\\
         $y$-axis & 0.00156\\
         $z$-axis & 0.00095\\
         \bottomrule
    \end{tabular}
\end{table}

\subsection{Additional Results}
\label{app:clip-results}

\begin{figure*}[t]
\centering
\begin{subfigure}{\textwidth}
\includegraphics[width=\linewidth]{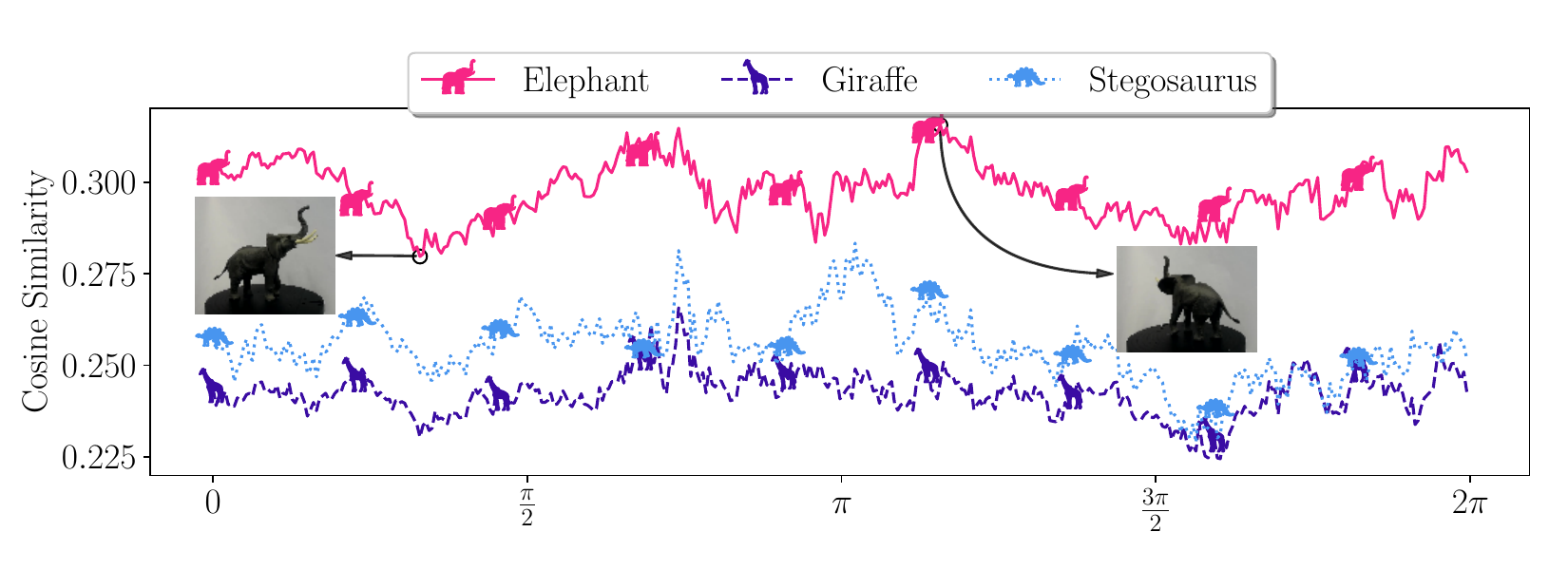}
\caption{Average cosine similarities in CLIP \cite{radford2021learning} latent space for images of a toy elephant with rotational interventions.}
\label{fig:clip-real-elephant}
\end{subfigure}
\begin{subfigure}{\textwidth}
\includegraphics[width=\linewidth]{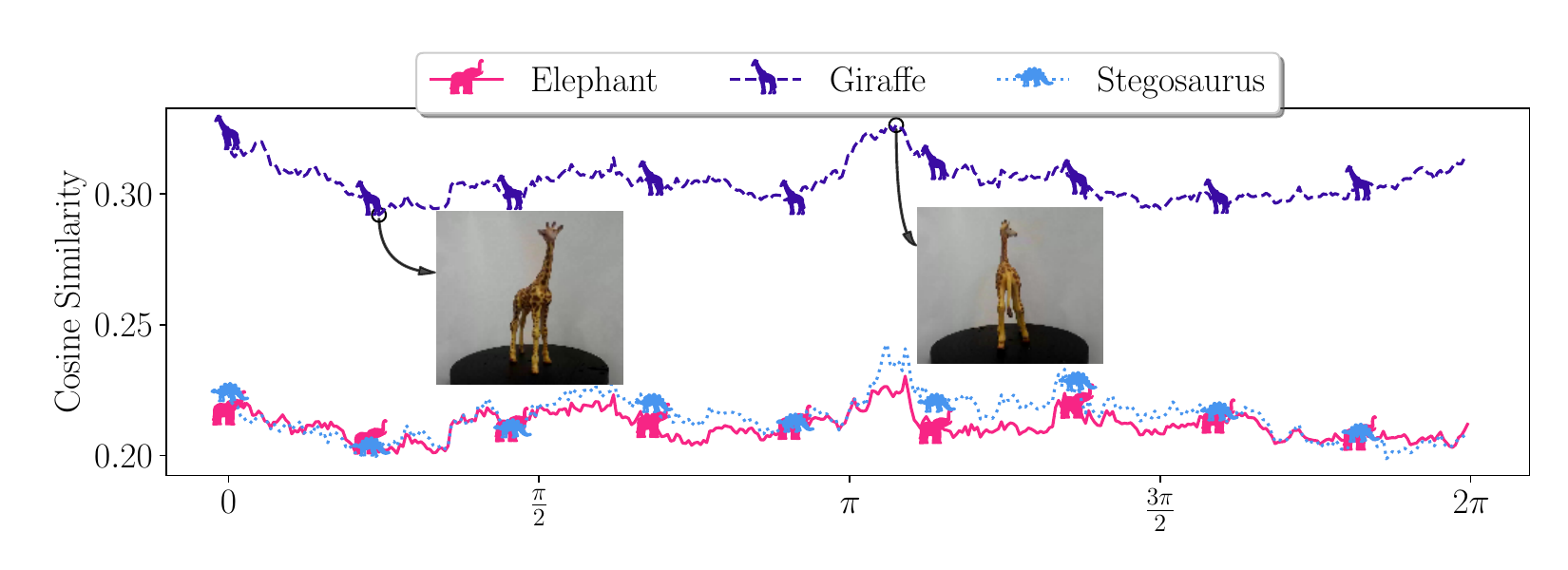}
\caption{Average cosine similarities in CLIP \cite{radford2021learning} latent space for images of a toy giraffe with rotational interventions.}
\label{fig:clip-real-giraffe}
\end{subfigure}

\caption{
Changes in CLIP \cite{radford2021learning} latent space cosine similarities for an intervention on the object position.
We showcase real-life interventions (\cref{fig:clip-real}), where we rotate around the $z$-axis using a turntable.
Here, we compare always images of the same toy model against description of all three classes.
Additionally, we visualize the minimum and maximum similarity together with the corresponding image.
}
\label{fig:clip-class}
\end{figure*}

\paragraph{Real-Life Interventions}

We present additional visualizations that complement the results shown in \cref{fig:clip-real}, focusing on the other interventional data. 
Specifically, \cref{fig:clip-real-elephant} and \cref{fig:clip-real-giraffe} display the average changes in cosine similarities for the elephant and giraffe figures, respectively, along with the minimum and maximum cosine similarity for the ground truth class.

\begin{figure*}[tb]
\centering
\begin{subfigure}{0.97\textwidth}
\includegraphics[width=\linewidth]{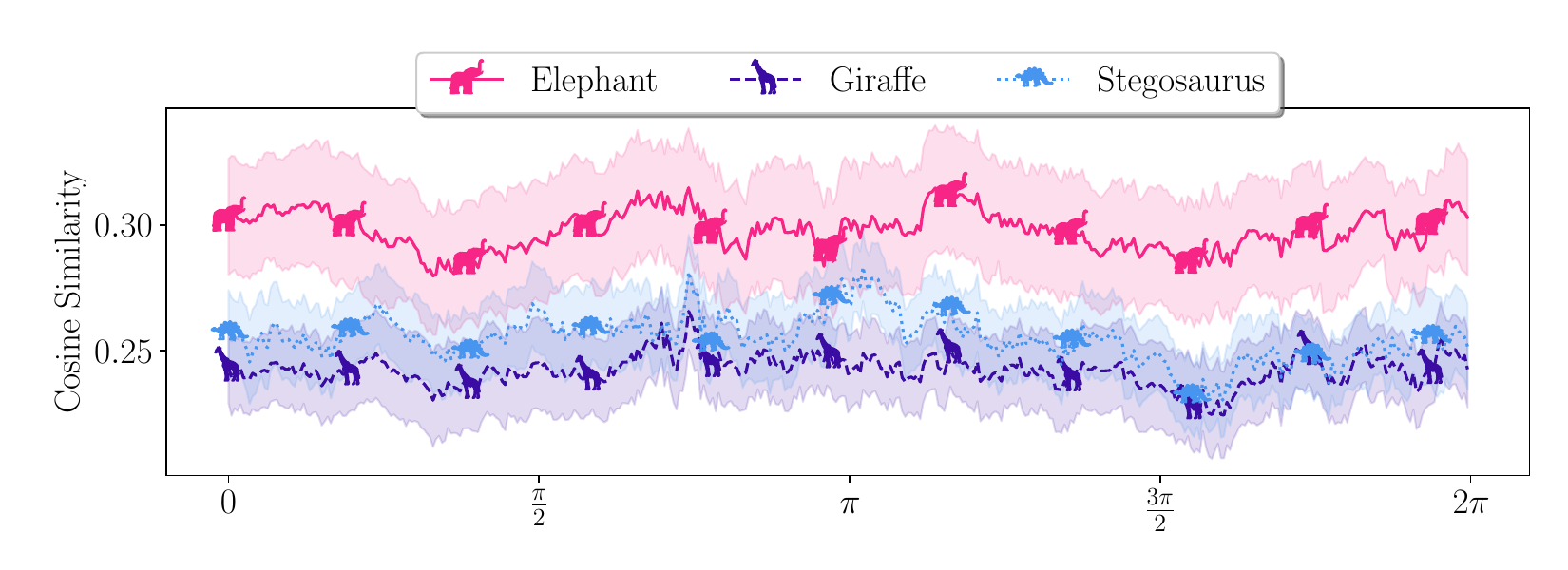}
\caption{Cosine similarities in CLIP \cite{radford2021learning} latent space for rotations of a toy elephant.}
\label{fig:clip-multi-elephant-std}
\end{subfigure}
\begin{subfigure}{0.97\textwidth}
\includegraphics[width=\linewidth]{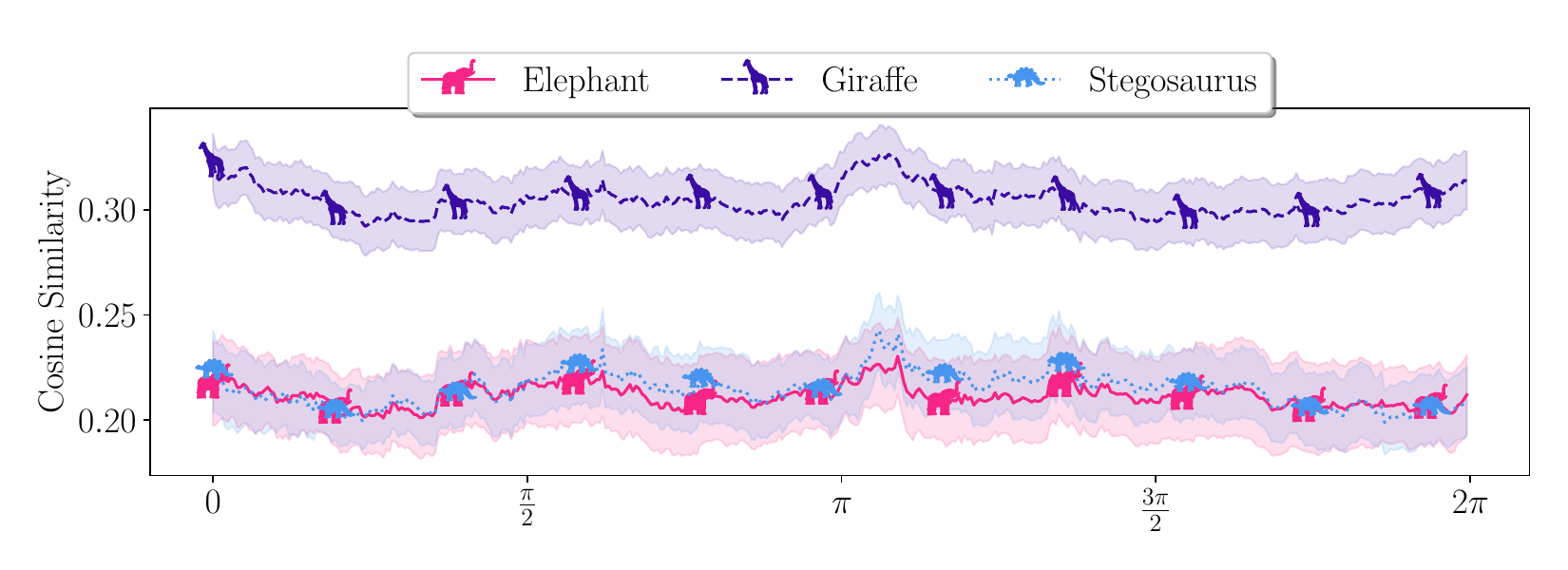}
\caption{Cosine similarities in CLIP \cite{radford2021learning} latent space for rotations of a toy giraffe.}
\label{fig:clip-multi-giraffe-std}
\end{subfigure}
\begin{subfigure}{0.97\textwidth}
\includegraphics[width=\linewidth]{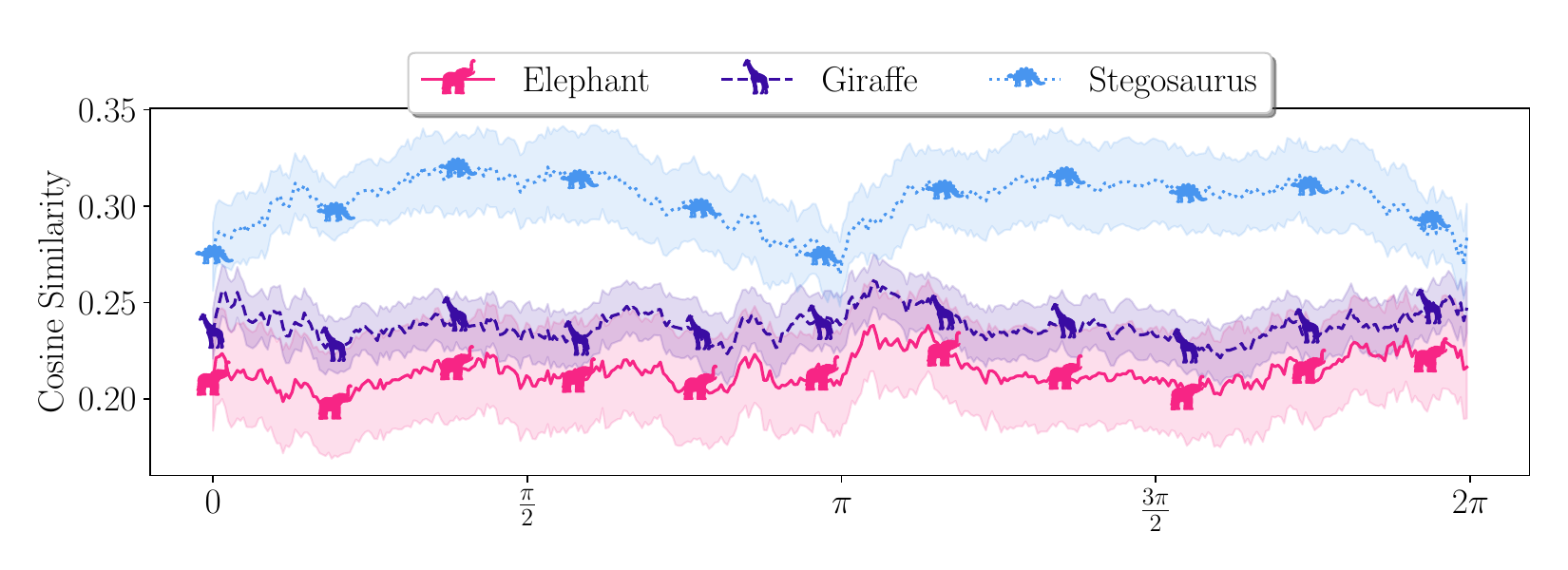}
\caption{Cosine similarities in CLIP \cite{radford2021learning} latent space for rotations of a toy stegosaurus.}
\label{fig:clip-multi-stegosaurus-std}
\end{subfigure}

\caption{
Changes in CLIP \cite{radford2021learning} latent space cosine similarities for an intervention on the object position.
We showcase real interventions, where we rotate three different toys.
We then compare against the three classes that are posed by these toys: elephant, giraffe, and stegosaurus.
The standard deviations are calculated over ten different text descriptions each (\Cref{tab:clip-texts}).
}
\label{fig:clip-real-std}
\end{figure*}

Furthermore, we provide standard deviation estimates for the interventional data in \cref{fig:clip-real-std}. 
The visualization is split according to the three real-life interventions depicted in \cref{fig:real-intervent}. 
In \cref{fig:clip-multi-elephant-std}, \cref{fig:clip-multi-giraffe-std}, and \cref{fig:clip-multi-stegosaurus-std}, we compare the visual embeddings to the three sets of phrases listed in \Cref{tab:clip-texts}. 
Particularly, \cref{fig:clip-multi-stegosaurus-std} corresponds to \cref{fig:clip-real} in the main paper, while \cref{fig:clip-multi-elephant-std} and \cref{fig:clip-multi-giraffe-std} correspond to the visualizations in \cref{fig:clip-real-elephant} and \cref{fig:clip-real-giraffe}, respectively.

As observed in the main paper, the standard deviations are consistent during the intervention. 
However, they differ between text descriptors, as evident in the comparison between elephant and giraffe descriptors in \cref{fig:clip-multi-elephant-std} and \cref{fig:clip-multi-giraffe-std}. 
Nevertheless, for specific combinations of interventional data and text phrases, we observe consistent variations. 
Moreover, consistent with the results in the main paper, the CLIP model \cite{radford2021learning} accurately predicts the shown images throughout the intervention, highlighting its robustness in our zero-shot classification setting.

In all three examples, we observe an approximately periodic behavior, which is a consequence of our experimental setup. 
However, the exact behavior varies significantly between the different toy figures. 
While the stegosaurus model exhibits a maximum similarity during a sideways orientation, this is not the case for the other figures. 
In contrast, the elephant and giraffe models achieve maximum similarity shortly after a rotation angle of $\pi$.
Here, the giraffe's maximum corresponds to a sideways position of the head, which we believe is a crucial feature. 
Notably, the elephant's trunk remains visible during the complete intervention

We calculate the corresponding \propgrad for all combinations of interventional data and text descriptions in \Cref{tab:clip-mean-scores-real}.
The results show that the images containing the giraffe model have the smallest expected property gradient magnitudes, which aligns with the visualization in \cref{fig:clip-real-giraffe} and the low standard deviations in \cref{fig:clip-multi-giraffe-std}. 
Additionally, these low scores correspond to the highest average difference between the predicted class and the rejected classes. 
We also observe that the lowest scores in \Cref{tab:clip-mean-scores-real} are concentrated along the diagonal, indicating that the rotation impacts the cosine similarities empirically the least when the similarities are high. 
In contrast, we observe higher \propgrad for the rejected classes. 
We will further investigate these observations using virtual interventions.

\paragraph{Virtual Interventions}
\begin{figure*}[t]
    \centering
    \includegraphics[width=\linewidth]{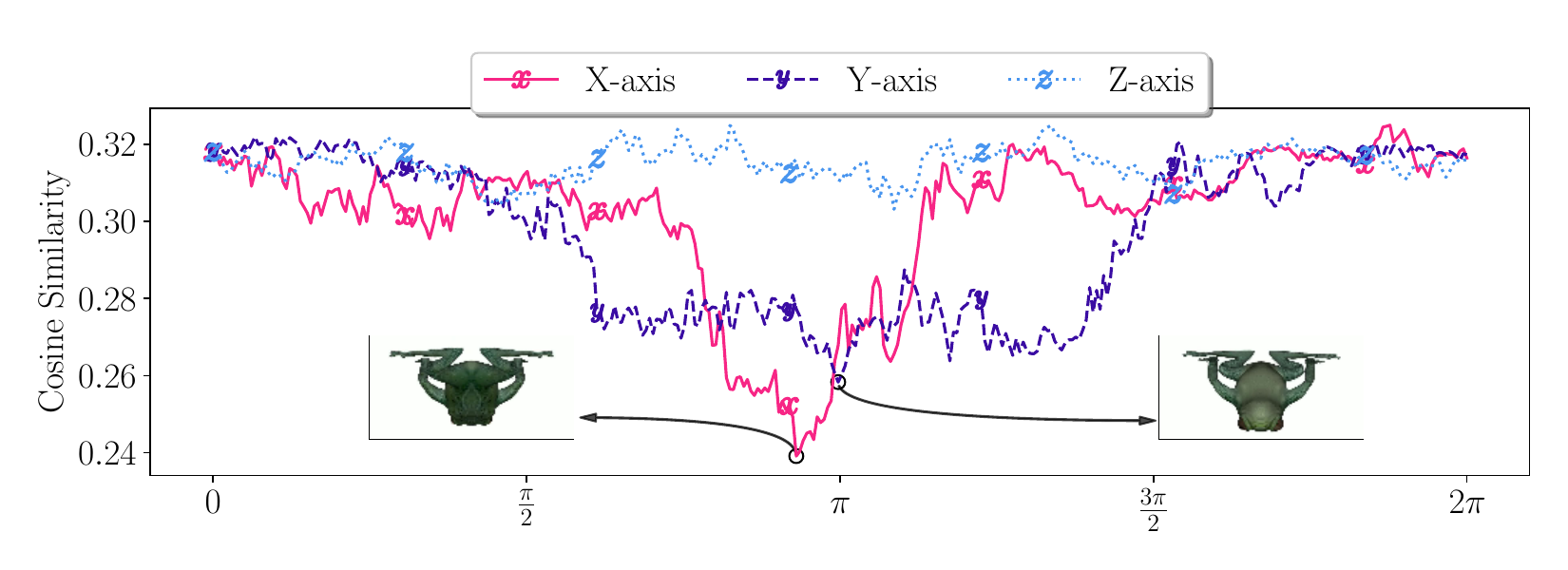}
    \caption{Average cosine similarities in CLIP \cite{radford2021learning} latent space for \textbf{virtual} interventional data. We mark the minima for the $x$ and $y$-axis rotations.}
    \label{fig:clip-virtual}
\end{figure*}

\cref{fig:clip-virtual} visualizes the influence of different rotation axes for the virtual interventions.
We find stark differences between the axes, with the most pronounced drop in similarity occurring for the rotation around the $x$-axis. 
Additionally, we note that the low point for the $y$-axis is achieved at approximately the same rotation angle.
We visualize these minima and confirm our previous hypothesis that the model locally struggles with upside-down object orientations, resulting in lower similarity.
The smaller decrease in similarity for the $y$-axis rotation may be attributed to the frontal view during rotation, which provides a more familiar perspective.
The in-plane rotation around the $z$-axis provides additional evidence that upside-down positions are problematic.
Specifically, we observe only smaller deviations in comparison.
Ultimately, our analysis reveals a consistent pattern, underscoring the model's local vulnerability to upside-down orientations.

\begin{figure*}[tb]
\centering
\begin{subfigure}{\textwidth}
\includegraphics[width=\linewidth]{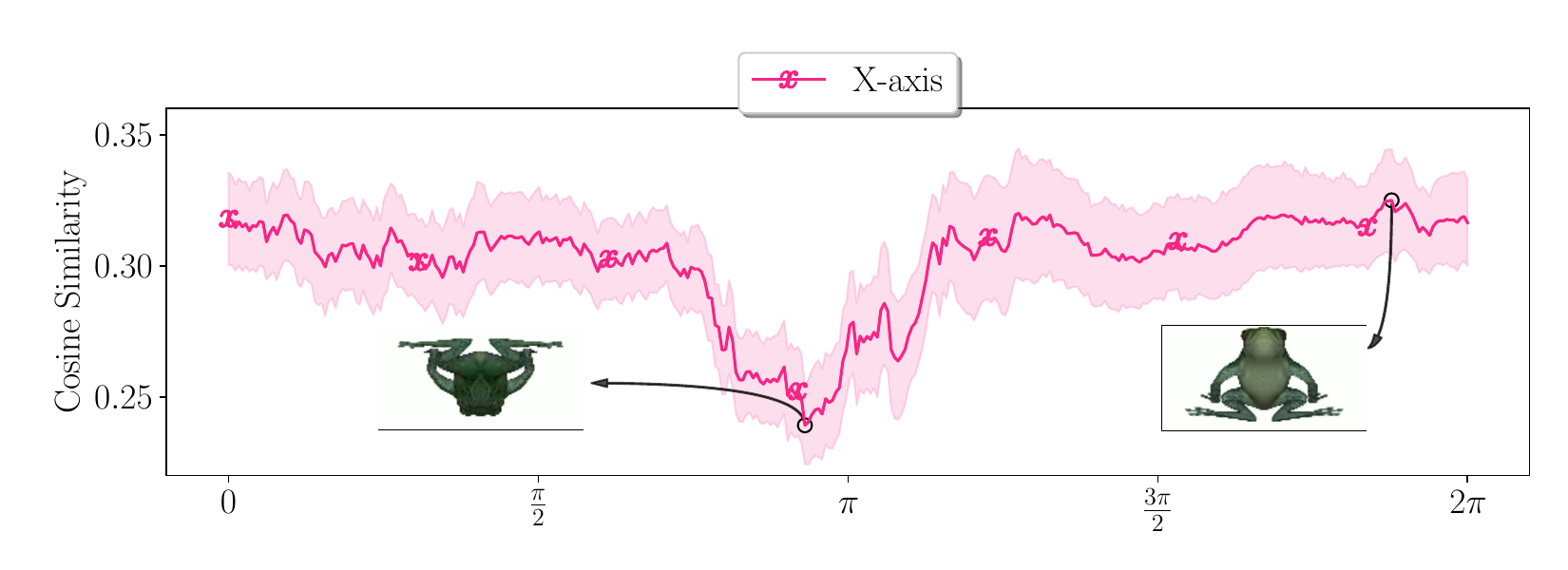}
\caption{Cosine similarities in CLIP \cite{radford2021learning} latent space for rotations around the $x$-axis.}
\label{fig:clip-multi-x}
\end{subfigure}
\begin{subfigure}{\textwidth}
\includegraphics[width=\linewidth]{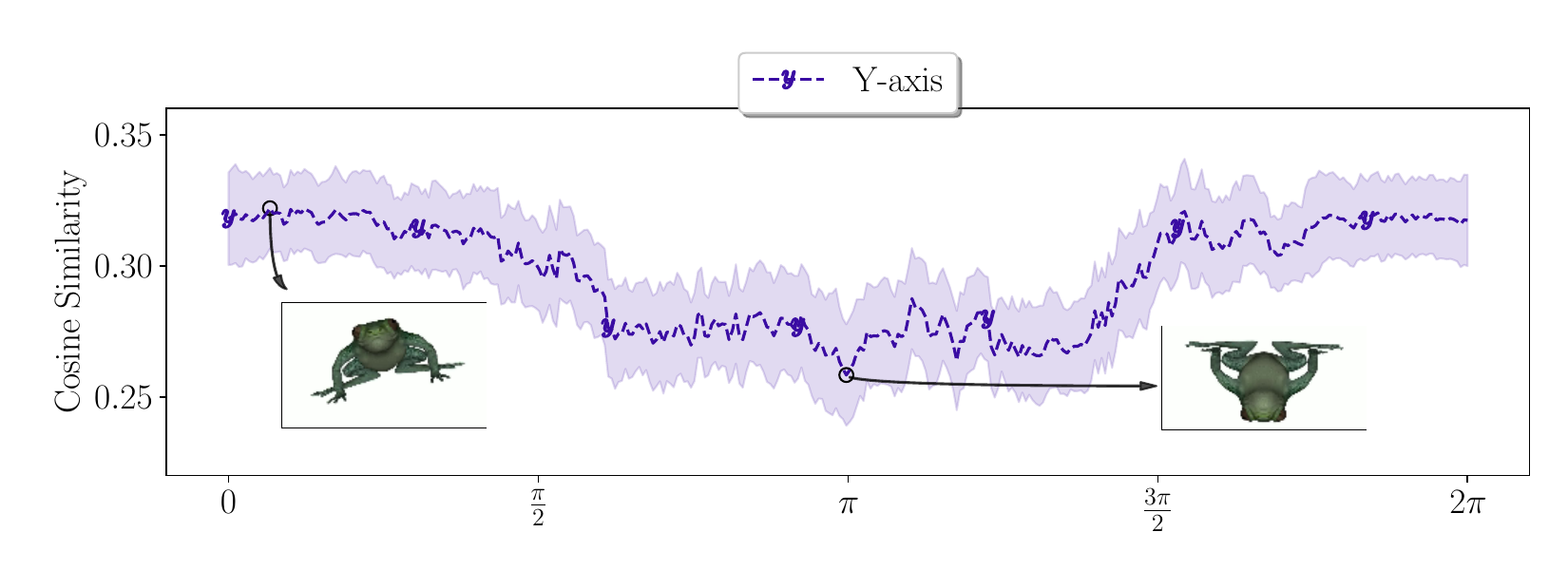}
\caption{Cosine similarities in CLIP \cite{radford2021learning} latent space for rotations around the $y$-axis.}
\label{fig:clip-multi-y}
\end{subfigure}
\begin{subfigure}{\textwidth}
\includegraphics[width=\linewidth]{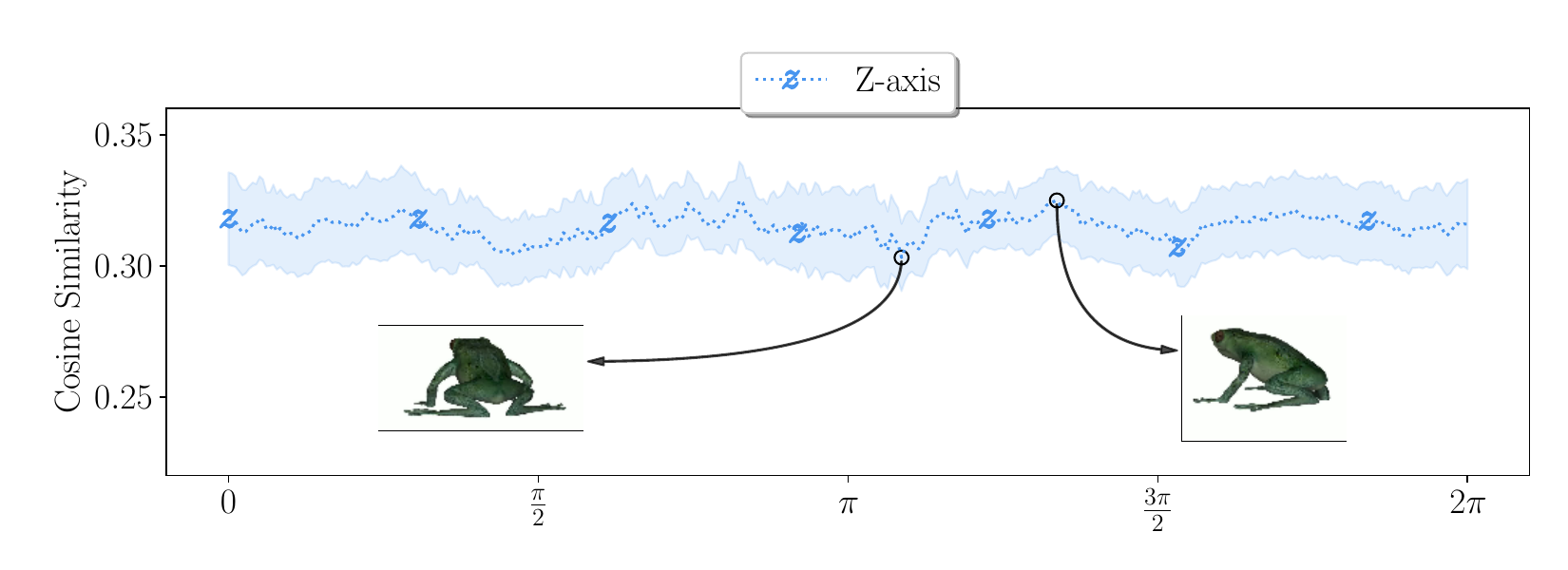}
\caption{Cosine similarities in CLIP \cite{radford2021learning} latent space for rotations around the $z$-axis.}
\label{fig:clip-multi-z}
\end{subfigure}

\caption{
Changes in CLIP \cite{radford2021learning} latent space cosine similarities for an intervention on the object position.
We showcase virtual interventions, where we rotate a frog model around all three axes.
Here, we additionally visualize the minimum, the maximum, and the standard deviation for ten text phrases used for the similarity calculations.
}
\label{fig:clip-virtual-split}
\end{figure*}

Similar to the real-life interventions, we observe consistent standard deviations throughout the complete interventions in \cref{fig:clip-virtual-split}. 
We visualize the mean behavior and the respective minima and maxima for all three interventions. 
Notably, the $x$-axis and $y$-axis rotation interventions exhibit high points for the upright position, with dips in cosine similarity observed for more uncommon upside-down orientations. 
Furthermore, we observe similar behavior to \cref{fig:clip-real} under the inplane rotation around the $z$-axis. 
Specifically, the minimum cosine similarity is achieved around $\pi$, while the maximum is observed in a sideways orientation.
In \Cref{tab:clip-mean-scores-virtual}, we summarize the approximated \propgrad scores for all three virtual rotations compared to the text embeddings in \Cref{tab:clip-texts}. 
These scores confirm our previous notion that the dependence on rotation is higher when average cosine similarities are lower. 
Specifically, we see the highest \propgrad for the $x$-axis, which also exhibits the lowest observed cosine similarity in our virtual interventional data. 
Conversely, we find a small \propgrad for the inplane rotation ($z$-axis), which aligns with the visualizations in \cref{fig:clip-virtual-split}.

Overall, our additional visualizations and \propgrad results provide further evidence to support the claims made in our main paper.
Moreover, they demonstrate that our approach can effectively handle diverse sources of interventional data.

\FloatBarrier

\clearpage